\documentclass[10pt,journal,compsoc]{IEEEtran}

\usepackage{newclude}
\usepackage{todonotes}

\usepackage[doi=false,isbn=false,url=true,eprint=false]{biblatex}
\usepackage[utf8]{inputenc} 
\usepackage[T1]{fontenc}    
\usepackage[breaklinks]{hyperref}       
\usepackage{url}            
\usepackage{booktabs}       
\usepackage{amsfonts}       
\usepackage{nicefrac}       
\usepackage{makecell}
\usepackage{microtype}      
\usepackage{lipsum}
\usepackage{graphicx}
\graphicspath{ {./img/} }
\usepackage{todonotes}
\usepackage{amsmath}
\usepackage{blindtext}
\usepackage{cuted}          
\usepackage[nohyperlinks, printonlyused]{acronym}
\usepackage{tabularx}
\usepackage{tablefootnote} 
\usepackage{subcaption} 
\usepackage{csquotes}

\AtEveryBibitem{%
  \ifentrytype{misc}{%
  }{%
    \clearfield{url}%
  }%
}

\title{Knowledge Augmented Machine Learning\\with Applications in Autonomous Driving: \\A Survey} 

\DeclareRobustCommand*{\IEEEauthorrefmark}[1]{%
  \raisebox{0pt}[0pt][0pt]{\textsuperscript{\footnotesize #1}}%
}

\author{
  \IEEEauthorblockN{%
      Julian~W\"{o}rmann\IEEEauthorrefmark{7},  
    Daniel~Bogdoll\IEEEauthorrefmark{9},        
    Christian~Brunner\IEEEauthorrefmark{6},
    Etienne~B\"{u}hrle\IEEEauthorrefmark{9},
    Han~Chen\IEEEauthorrefmark{2},
    Evaristus~Fuh~Chuo\IEEEauthorrefmark{2},
    Kostadin Cvejoski\IEEEauthorrefmark{8},
    Ludger~van~Elst\IEEEauthorrefmark{4},
    Philip~Gottschall\IEEEauthorrefmark{8},
    Stefan~Griesche\IEEEauthorrefmark{10},
    Christian~Hellert\IEEEauthorrefmark{3},
    Christian~Hesels\IEEEauthorrefmark{8},
    Sebastian~Houben\IEEEauthorrefmark{8},
    Tim~Joseph\IEEEauthorrefmark{9},
    Niklas~Keil\IEEEauthorrefmark{1},
    Johann~Kelsch\IEEEauthorrefmark{5},
    Mert Keser\IEEEauthorrefmark{3},
    Hendrik~K\"{o}nigshof\IEEEauthorrefmark{9},
    Erwin~Kraft\IEEEauthorrefmark{3},
    Leonie~Kreuser\IEEEauthorrefmark{1},
    Kevin~Krone\IEEEauthorrefmark{8},
    Tobias~Latka\IEEEauthorrefmark{6},
    Denny~Mattern\IEEEauthorrefmark{8},
    Stefan~Matthes\IEEEauthorrefmark{7},
    Franz Motzkus\IEEEauthorrefmark{3},
    Mohsin~Munir\IEEEauthorrefmark{4},
    Moritz~Nekolla\IEEEauthorrefmark{9},
    Adrian~Paschke\IEEEauthorrefmark{8},
    Stefan~Pilar~von~Pilchau\IEEEauthorrefmark{6},
    Maximilian~Alexander~Pintz\IEEEauthorrefmark{8},
    Tianming~Qiu\IEEEauthorrefmark{7},
    Faraz~Qureishi\IEEEauthorrefmark{12},
    Syed~Tahseen~Raza~Rizvi\IEEEauthorrefmark{4},
    J\"{o}rg~Reichardt\IEEEauthorrefmark{3},
    Laura~von~Rueden\IEEEauthorrefmark{8},
    Alexander~Sagel\IEEEauthorrefmark{7},
    Diogo Sasdelli\IEEEauthorrefmark{11},
    Tobias~Scholl\IEEEauthorrefmark{8},
    Gerhard Schunk\IEEEauthorrefmark{12},
    Gesina Schwalbe\IEEEauthorrefmark{3},
    Hao~Shen\IEEEauthorrefmark{7},
    Youssef Shoeb\IEEEauthorrefmark{3},
    Hendrik~Stapelbroek\IEEEauthorrefmark{2},
    Vera~Stehr\IEEEauthorrefmark{12},
    Gurucharan~Srinivas\IEEEauthorrefmark{5},
    Anh~Tuan~Tran\IEEEauthorrefmark{10},
    Abhishek~Vivekanandan\IEEEauthorrefmark{9},
    Ya~Wang\IEEEauthorrefmark{8},
    Florian~Wasserrab\IEEEauthorrefmark{1},
    Tino~Werner\IEEEauthorrefmark{5},
    Christian~Wirth\IEEEauthorrefmark{3}, and
    Stefan~Zwicklbauer\IEEEauthorrefmark{3}
  }\\\bigskip
  \IEEEauthorblockA{%
    \IEEEauthorrefmark{1}Alexander Thamm GmbH\\	
    \IEEEauthorrefmark{2}Capgemini Engineering\\
    \IEEEauthorrefmark{3}Continental AG\\	
    \IEEEauthorrefmark{4}Deutsches Forschungszentrum für Künstliche Intelligenz GmbH (DFKI)\\
    \IEEEauthorrefmark{5}Deutsches Zentrum für Luft- und Raumfahrt e.V. (DLR)\\
    \IEEEauthorrefmark{6}e:fs TechHub GmbH\\
    \IEEEauthorrefmark{7}fortiss GmbH\\
    \IEEEauthorrefmark{8}Fraunhofer-Gesellschaft zur Förderung der angewandten Forschung e.V. (FOKUS \& IAIS)\\
    \IEEEauthorrefmark{9}FZI Forschungszentrum Informatik\\
    \IEEEauthorrefmark{10}Robert Bosch GmbH\\
    \IEEEauthorrefmark{11}Universität des Saarlandes\\
    \IEEEauthorrefmark{12}Valeo Schalter und Sensoren GmbH
  }\\\vspace{0.8cm}\small{
  \begin{tabular}{p{18cm}}
    \textbf{Abstract:} The availability of representative datasets is an essential prerequisite for many successful artificial intelligence and machine learning models. However, in real life applications these models often encounter scenarios that are inadequately represented in the data used for training. There are various reasons for the absence of sufficient data, ranging from time and cost constraints to ethical considerations. As a consequence, the reliable usage of these models, especially in safety-critical applications, is still a tremendous challenge. 
    Leveraging additional, already existing sources of knowledge is key to overcome the limitations of purely data-driven approaches. Knowledge augmented machine learning approaches offer the possibility of compensating for deficiencies, errors, or ambiguities in the data, thus increasing the generalization capability of the applied models. Even more, predictions that conform with knowledge are crucial for making trustworthy and safe decisions even in underrepresented scenarios.
    This work provides an overview of existing techniques and methods in the literature that combine data-driven models with existing knowledge. The identified approaches are structured according to the categories knowledge integration, extraction and conformity. In particular, we address the application of the presented methods in the field of autonomous driving.
  \end{tabular}}
  \bigskip
  \begin{center}
      Acknowledgement: The research leading to these results is funded by the German Federal Ministry for Economic Affairs and Climate Action within the project “KI Wissen – Entwicklung von Methoden für die Einbindung von Wissen in maschinelles Lernen". The authors would like to thank the consortium for the successful cooperation.
  \end{center}
}

\newcommand{\ie}{i.e.,\ }
\newcommand{\eg}{e.g.,\ }

\begin{document}
\maketitle
\pagebreak
\onecolumn
\setcounter{tocdepth}{2}
\tableofcontents
\twocolumn
\setcounter{secnumdepth}{3}
%
%

\pagebreak
\section{Introduction}
Data-driven learning, first and foremost deep learning, has become a key paradigm in the vast majority of current \ac{ai} and \ac{ml} applications. The excellent performance of many models trained in a supervised manner can be predominantly attributed to the availability of huge amounts of labeled data. Prominent examples are image classification and object detection, sequential data processing as well as decision making. On the downside, the unprecedented performance comes at the cost of lacking interpretability and transparency leading to so called black box models that do not allow easy and straightforward verification by humans.

The application of data-driven models in safety-critical applications is therefore a major challenge. On the one hand, labeled data covering both common and critical scenarios are limited due to high acquisition costs or, not least, for ethical reasons. This makes it extremely difficult to learn robust models that can make reliable predictions even in underrepresented scenarios. On the other hand, both, developers and users postulate the requirement to be able to understand the decisions made by the deployed model. Consequently, there is a strong interest in understanding the internal information processing as well as the input-output behavior in order to identify and eliminate potential weaknesses of the models used. 

In order to tackle the aforementioned challenges, the exploitation of existing \textit{knowledge} sources in form of, e.g., basic laws of physics, logical databases of facts, common behaviour in certain scenarios, or simply counterexamples is key to evolve purely data-driven models towards robustness against perturbations, better generalization to unseen samples, and conformity to existing principles of safe and reliable behaviour. However, the utilization of knowledge raises fundamental questions. How do we represent and formalize knowledge such that it is machine readable? What kind of interfaces exist such that the knowledge component can be seamlessly integrated into the classic data-driven workflow? Does the learned model itself implicitly follow concepts that resemble existing patterns of knowledge? And finally, how do we assess and measure the impact of knowledge on the intended functional behaviour?

This survey provides a collection of existing methods and procedures from literature that facilitate the augmentation of data-driven models with knowledge, that allow for the extraction of informative concepts and patterns out of given models and that provide mechanisms to compare observed outputs and representations to existing basic assumptions and common understanding about safe, reliable and intuitive behaviour. The goal of this overview is to introduce the reader to existing approaches and methods for linking knowledge and data, paving the way to trustworthy \ac{ml} models that can be safely used in critical applications.     

Autonomous driving can be certainly considered as one of these applications, that require robust and reliable models that enable safe and comfortable driving maneuvers. In our overview, we approach knowledge augmented machine learning from this perspective, highlighting the interfaces to certain base functionalities of autonomous driving. However, we believe that this review can also provide a comprehensive overview of existing methods for many other applications as well.    

This review is structured as follows. In \autoref{ch:usecases}, we first introduce three major tasks that autonomous agents encounter during interaction with their environment, namely perception, situation interpretation and planning. \autoref{ch:knowledge_representations} reviews different perspectives to represent knowledge and to make it machine readable. Subsequently, various general approaches and techniques eligible to combine knowledge with data-driven approaches, as well as more specific methods tailored to the autonomous driving use case, are presented in \autoref{ch:integration}. Furthermore, \autoref{ch:knowledge_transfer} introduces learning paradigms in the context of knowledge transfer. 

Besides integration of knowledge, current approaches focusing on the extraction of concepts and structures are outlined in the subsequent chapters. While \autoref{ch:symbolic_extraction} summarizes methods that provide symbolic, partly natural language explanations, \autoref{ch:visual_extraction} puts emphasis on procedures that allow for visual inspection of the decision process. We conclude our survey in \autoref{ch:conformity} with an overview of techniques that consider conformity to already existing as well as newly discovered knowledge components, which eventually completes the pipeline of knowledge empowered artificial intelligence.
\section{Overview use case domains}
\label{ch:usecases}
The task of automated driving may be sub categorized into the following categories: perception, situation interpretation, planning and control \cite{hermann_driving_2008}. The foremost task in the autonomous driving is to understand and perceive the environment around the vehicle. \autoref{sec:perception} provides an introduction to the \textit{perception} module with a special focus on image-based pedestrian detection. Once the objects are detected and segmented, the second task in the autonomous driving is to understand the environment along with the road users. In order to perform safe maneuvers, the \textit{situation interpretation} is a decisive step. In this module, the goal is to answer important questions related to object's states and actions, like what an object could do next. An overview is given in section \autoref{sec:situation_interpretation}. After figuring out these situational scenarios, next task in autonomous driving is to plan the motion of ego vehicle. The \textit{planning} module described in \autoref{sec:planning} utilizes the output of the previous two modules and takes high level routing and trajectory planning decisions. 
\subsection{Perception}
\label{sec:perception}
\textit{Authors: Syed Tahseen Raza Rizvi, Mohsin Munir, Ludger van Elst}

\smallskip
\noindent
\subsubsection{Perception in the AD Stack}
Perception plays a crucial role in attaining the goal of autonomous driving. An ego-vehicle is generally equipped with a variety of sensors including cameras, lidar and radar. These sensors serve as the senses of an ego-vehicle and therefore enable the capability of perceiving the environment around the ego-vehicle in different spectrums. Object detection, and in particular pedestrian detection, has significant importance in the perception spectra as it serves as a critical piece of information for the downstream tasks associated with the autonomous driving pipeline. 

\subsubsection{Task Formulation}
Autonomous driving systems highly rely on object detection models to identify all the traffic participants. Pedestrians are usually the most common and abundantly found traffic participant. Therefore, the detection of a pedestrian is more prominent and crucial for the perception of an autonomous driving system.

Pedestrian detection deals with the identification of pedestrians in the environment around an ego-vehicle. There exist approaches in the literature which perform pedestrian detection only using lidar sensors~\cite{lin_pedestrian_2018}. However, such approaches are usually not popular in the community due to fact that the features obtained from camera images are significantly richer as compared to the ones obtained from lidar or radar. On the other hand,~\cite{furst_lrpd_2020} uses lidar to incorporate depth information into the image data for the pedestrian detection task. Therefore, the approaches to perform pedestrian detection mainly using camera images are generally widely adopted. The images from the mounted cameras serve as an input from which individual pedestrians are identified and are enclosed in a bounding box. A variety of solutions have been proposed to effectively identify individual pedestrians in the surrounding environment. 

The neural network based object detection solutions can be divided into two main categories: One-stage and Two-stage approaches. One-stage approaches are generally based on a fully convolutional architecture and consider the object detection problem as a simple regression problem~\cite{soviany_optimizing_2018}. For a given input image, the One-stage detectors learn class probabilities and the coordinates of a bounding box encompassing an object. On the other hand, Two-stage approaches are more sophisticated where each stage specializes in a sub-task which eventually contributes to the final output of the system. The first stage is responsible for identifying the region of interest and the second stage is responsible for the object classification and bounding box regression. Both types of approaches have certain pros and cons. Most notably, Two-stage approaches yield better detection accuracy than One-stage approaches as they have specialized stages where the output of the second stage is built on top of the output of the first stage. However, One-stage approaches are much faster than Two-stage approaches as they do not have an additional stage with supplementary computational overhead. 

\ac{ssd}~\cite{liu_ssd_2016}, You Only Look Once (YOLO)~\cite{redmon_yolo9000_2017,redmon_you_2016,redmon_yolov3_2018}, {RetinaNet}~\cite{lin_focal_2017} and Fully Convolutional One-Stage object detector~\cite{tian_fcos_2019} are the most prominent One-stage object detectors. Generally, these approaches divide the image into a grid followed by predicting the probability of a class object in each grid box along with its bounding box coordinates. However, some of these approaches are slightly different as they employ a unique focal loss or pixel-wise classification to achieve a higher detection accuracy in real-time. On the other hand, {Fast \ac{r-cnn}}~\cite{girshick_fast_2015}, {Faster \ac{r-cnn}}~\cite{ren_faster_2016}, {Mask \ac{r-cnn}}~\cite{he_mask_2017}, MimicDet~\cite{lu_mimicdet_2020} are the most common examples of a Two-stage object detector. Generally the first stage in these Two-stage object detection model consists of a \ac{rpn}, where in the second stage the candidate region proposals are classified based on the feature maps. Approaches like Mask \ac{r-cnn} have a mask branch which is a small \ac{fcn}~\cite{dai_r-fcn_2016} applied to each \ac{roi}, predicting a pixel-wise segmentation mask. Additionally, \ac{fpn}~\cite{lin_feature_2017} is generally used in combination with \ac{rpn} and Faster \ac{r-cnn} to make bounding box proposal more robust especially for small objects. Recently, Khan et al. \cite{khan2022f2dnet} proposed a two-stage pedestrian detection architecture that eliminates redundancy of current two-stage detectors by replacing the region proposal network with our focal detection network and bounding box head with our fast suppression head. Furthermore,  their method has significantly lesser inference time compared to the current state-of-the-art methods.

\bigskip Pedestrian detection is applied in various vision-based applications ranging from surveillance to autonomous driving. Despite their good performance, it is still unknown how the detection performs on unseen data. Hasan et al. \cite{hasan_generalizable_2021} presented a study in quest of generalization capabilities of pedestrian detectors. In their cross-dataset evaluation, they have tested several backbones with their baseline detector (Cascade \ac{r-cnn}) \cite{cai_cascade_2019} on famous autonomous driving datasets including Caltech~\cite{dollar_pedestrian_2009}, CityPersons~\cite{zhang_citypersons_2017}, ECP \cite{braun_eurocity_2019}, CrowdHuman \cite{shao_crowdhuman_2018}, and Wider Pedestrian \cite{noauthor_codalab_nodate}. Cross-dataset evaluation is an effective way of evaluating a method on unseen data and checking its generalization capability, otherwise, a method may overfit on a single dataset. The analysis presented in the paper is very interesting. The authors have demonstrated that the existing pedestrian detection methods perform poorly when compared with general object detection methods given larger and diverse datasets. A carefully trained state-of-the-art general-purpose object detector can outperform pedestrian-specific detection methods. The trick lies in the training pipeline and the dataset. In this study, the authors used large datasets that contain more persons per image. These general purpose datasets, generally collected by crawling the web and through
surveillance cameras, are likely to have more human poses, appearances, and occlusion cases as compared to pedestrian-specific datasets. It is also shown in this study that by progressively fine-tuning the models from largest (general purpose) to smallest (close to target domain), performance can be improved. The generalization ability of pedestrian detectors has been compromised due to the lack of diversity and density of the pedestrian benchmarks. However, benchmarks such as WiderPerson \cite{zhang_widerperson_2019}, Wider Pedestrian \cite{noauthor_codalab_nodate}, and CrowdHuman \cite{shao_crowdhuman_2018} provide much higher diversity and density.

Pedestrian detection has improved a lot in recent years, however, it is still challenging to detect occluded pedestrians. The pedestrian appearance varies in different scenarios and depends on a wide range of occlusion patterns. To address this issue, Zhang et al. \cite{zhang_occluded_2018} proposed an architecture for pedestrian detection based on the Faster \ac{r-cnn}. In contrast to ensemble models for most frequent occlusion patterns, the authors leverage different attention mechanisms to guide the detector in paying more attention to the visible body parts. The authors proposed to employ channel-wise attention in a convolution network that allows the network to learn more representative features for different occluded body parts in one model. The observation that many \ac{cnn} channels in a pedestrian \ac{cnn} are localizable, strongly motivates them to perform re-weighting of channel features to guide the detector to pay more attention to the visible body parts. In order to generate the attention vector, different realizations of attention networks are examined. The attention vector is trained end-to-end for all of the attention networks either through self-attention or guided by some additional external information like convolution features, visible bounding boxes, or part detection heatmaps. Eventually, the features are passed to the classification network for category prediction and bounding box regression. The experimental results are shown on the CityPersons~\cite{zhang_citypersons_2017}, Caltech~\cite{dollar_pedestrian_2009}, and ETH~\cite{ess_depth_2007} datasets. The results show improvements over the baseline Faster \ac{r-cnn} detector. Another crucial challenge of pedestrian detection, which is not widely discussed, is to detect pedestrians even with diversities in appearance. Most of the current detectors learn these diverse appearance features individually, but the training dataset might not comprise of good number of viewpoints or dressing diversities. To address this issue, Lin et al. \cite{lin2022pedestrian} introduced a pedestrian detection based on contrastive learning. The proposed method guides feature learning in such a way that the semantic distance between pedestrians having different appearances is minimized and the distance between the pedestrians and the background is maximized.  

Vulnerable road user detection is another major challenge in pedestrian detection. The safety of road users is and should be the utmost priority in the domain of autonomous driving. In addition to detect occluded pedestrians, another key challenge is to detect pedestrians at long range. When a pedestrian is detected at long range, it increases the security of the pedestrian and driver at the same time, also, it leads to a comfortable driving experience. Fürst et al. \cite{furst_lrpd_2020} introduced an approach that targets long range 3D pedestrians detection. Their approach leverages the density of \ac{rgb} images and precision of lidar. The symmetrical fusion of \ac{rgb} and lidar helps them outperform current state-of-the-art for long range 3D pedestrian detection.

\subsubsection{Goals and Requirements}
Perception plays a pivotal role in autonomous driving. It enables the ego vehicle to analyze and understand the traffic scene and surrounding circumstances. Detection of traffic participants, i.e., pedestrians, vehicles, cyclists, etc. serve as the core of perception involved in autonomous driving. Additionally, traffic circumstances like road, weather, and light conditions are also important factors in a traffic scenario. For instance, rainy weather results in a wet road which consequently has a direct impact on the decisions like breaking distance, because the braking distance increases in wet conditions as compared to normal drys. Therefore, traffic participants and their surrounding circumstances collectively provide a basis for planning and executing decisions taken by an ego vehicle. The significance of the perception can be understood by the fact that it directly contributes towards use cases like collision avoidance, trajectory planning, etc.

With the rise of deep learning for solving a universe of different tasks, object detection has also benefited from One- and Two-stage deep learning-based models to achieve higher detection performance. The effectiveness of an object detection approaches heavily relies on the efficacy of the trained object detection model. In other words, it can be said as, provided an effective object detection model, the quality of perception can be ensured. In order to train an effective object detection model, it requires a large amount of high-quality data. For this purpose, several real-life public datasets are available, i.e., Caltech~\cite{dollar_pedestrian_2009}, CityPersons~\cite{zhang_citypersons_2017}, ECP~\cite{braun_eurocity_2019}, etc. However, certain scenarios are possibly scarce or outright not feasible in such pedestrian detection datasets. For example, it is infeasible to find a dataset that contains a traffic scenario where the ego vehicle is about to collide with another traffic participant. Such a scenario can be helpful to evaluate the performance of an object detection model to detect and evade collision in such a hazardous environment. For this purpose, datasets with simulated custom scenarios can be generated to fill this gap in real-life datasets. Ultimately, a combination of real and simulated data is the key thus enabling the object detection model to effectively perform under several unseen or rarely occurring traffic scenarios.

\subsubsection{Necessity of Knowledge Integration}
Computer vision methods and in general \ac{ml} methods have significantly improved over the last years. Different methods are able to accurately interpret a situation presented in an image or video. Even with such advancements, there are scenarios where \ac{ml} methods react differently as humans. The main reason of this gap is the absence of the background knowledge from the learned model. The \ac{ml} methods only account for patterns present in the training data, whereas humans have implicit knowledge that could help them to interpret a critical situation more robustly. In the context of autonomous driving, and in general too, it is not possible to train a model for every possible scenario that could happen on road. To provide a safer environment for pedestrians and autonomous vehicles, it is important to incorporate knowledge in the module that is responsible for taking important decisions. 
\subsection{Situation Interpretation}
\label{sec:situation_interpretation}
\textit{Authors: Daniel Bogdoll, Abhishek Vivekanandan, Faraz Qureishi, Gerhard Schunk}

\vspace{-3pt}
\noindent
\subsubsection{Situation Interpretation in the AD Stack}
Situation interpretation is typically a follow-up module of the perception stage as shown in \autoref{sec:perception}. Accordingly, this module is aware of objects, their states, and classifications within the surrounding environment. Its main objective is to interpret the situation, which includes questions such as “What is an object doing next?”, “Is there an implicit meaning of an object's action?” or “Is a rule exception applicable right now?”.

\subsubsection{Task Formulation} 
Automated driving relies on accurate perception of the environment. We follow the concept of Gerwien et al.~\cite{gerwien_towards_2021}, who describe \emph{situation interpretation} as a module which provides a “situation-aware environment model”, that expands an environment model, which is typically the results of the perception stage, by \emph{situation recognition} and \emph{situation prediction}. They classify these three modules as \ac{sa} levels 1-3. The output of the perception layer can be represented in various forms, for instance with object lists or probabilistic maps. Independent of the structure, the output is critical for the functioning of subsequent \ac{ad} layers, which are tasked with situation interpretation, path planning – as shown in \autoref{sec:planning} – and vehicle control.
 
Nevertheless, sometimes raw data in addition to the outputs of the perception layer is relevant to detect intentions or meanings which are typically not addressed by perception systems. Two examples are direction of view~\cite{hasan_forecasting_2019} and hand gestures~\cite{weaver_self-driving_2020}.

Situation interpretation works in tandem with perception, planning and control. A typical example of situation interpretation may involve cut in scenarios during automated driving using adaptive cruise control~\cite{perdomo_lopez_scenario_2017}. In a cut in scenario, the situation interpretation system shall be able to detect if a collision is imminent (using perception and planning output) and employ mitigation measures (braking in this case) in due time, ensuring the safety of the ego vehicle and its occupants.

In the aforementioned example, the collision detection and avoidance can be designed by using vehicle motion models and traffic rules. In complex situations, however, the task of situation interpretation may not be accomplished by only using a predefined set of rules. Especially for urban scenarios where the number of interactions between the ego vehicle and the objects in the scene are significantly higher. Additionally, there might be situations where a particular rule needs to be violated in order to ensure safety of human life.
 
\subsubsection{Goals and Requirements}  
To be consistent with the previously defined \ac{sa} levels, level 2 takes in raw data and adds semantic meaning to it in the form of semantic data models. Many works, especially~\cite{gerwien_towards_2021}, have defined the operational context in regard to adding more semantic structure to identify situations of interest.

As with \ac{sa} level 3 defined by~\cite{gerwien_towards_2021}, motion prediction forms the abstract layer for situation understanding, which comprises different actors in the ego space. It plays a crucial role in determining safety critical applications for the autonomous driving stack by providing the service of estimating the future positions of an object. For instance, when driving in a highway scenario, assuming that a lead vehicle suddenly merges or cuts-in to the ego lane; the primary goal of this layer is to mitigate the collision by anticipating the intention of the lead vehicle(s). The crash avoidance maneuver should have safety properties such that the maneuver itself should not cause an additional collision, e.g., while hard braking could prevent the crash it could lead to a rear ended collision with other vehicles. This requires not only a prediction module but also a system that checks for the validity of the planned decision based on dynamic safety reasoning methodologies which could influence the \ac{ttc}, such as including weather constraints.

Most of the existing behavior prediction approaches perform simultaneous tracking and forecasting with the use of Kalman Filters or in the form of rule based approaches, as can be seen from the previous works \cite{lefevre_survey_2014}. Although variants of Kalman filters are good for short term predictions, their performance degrades for long term motion problems as they fail to make use of the situation or environmental knowledge~\cite{cosgun_towards_2017} which could be obtained via vectored maps. As a result, prediction modules should make use of domain knowledge to forecast reliable predictions~\cite{boulton_motion_2021}.

In a typical \ac{ad} stack, motion prediction is a separate module which does prediction based on the outputs from the previous perception layer. For example, the object detection outputs bounding box coordinates of an object along with the probability score of a class it belongs to such as truck, car, or construction cone. When this is used as an input to the motion prediction, a failure to propagate uncertainty happens due to the softmax outputs~\cite{gal_uncertainty_2016}. To alleviate those shortcomings, end-to-end networks, which take raw inputs such as lidar point clouds and camera fusion to produce motion predictions directly~\cite{wu_motionnet_2020, djuric_uncertainty-aware_2020} should be considered. Additionally, knowledge about one's own path planning can be integrated into the prediction component~\cite{aurora_cvpr_2021}.
 
\subsubsection{Necessity of Knowledge Integration}
Vehicles equipped with a level 4 or 5 driving automation system are expected to master a wide variety of situations within their \ac{odd}~\cite{sae_j3016c_2021}. Since many situations do not occur frequently in real life, \ac{ml} based systems are struggling to extrapolate from their trained domain. Therefore, hybrid approaches that integrate rule- and knowledge based algorithms and insights into \ac{ml} systems have the potential to combine the best of two worlds – great general performance and improved handling of rare situations, such as corner cases.

\subsection{Planning}
\label{sec:planning}
\textit{Authors: Etienne B\"{u}hrle, Hendrik K\"{o}nigshof, Abhishek Vivekanandan, Moritz Nekolla}

\smallskip
\noindent
\subsubsection{Motion Planning in the AD Stack}
The planning module uses the outputs of the perception and prediction modules to plan a trajectory for the vehicle, which is subsequently handed down to the vehicle controls to be executed. This plan considers high-level routing decisions, and follows the rules of the road as well as basic principles of safe and comfortable driving.

A wide range of methods has been developed to tackle the trajectory tracking control problem, and we refer to \cite{paden_survey_2016} for an overview. However, the motion planning problem, especially in highly complex and dynamic environments like road traffic, remains largely unsolved and constitutes an area of ongoing research.

\subsubsection{Task Formulation}
Formally, the solution to the trajectory planning problem is a function that assigns every point in time a position in configuration space (typically, planar coordinates and heading). Classical approaches include variational methods (which represent the path as a function of continuously adjustable parameters), graph-search methods (which discretize the configuration space), and incremental search methods (which improve upon graph-search methods by using iterative refinement procedures). An excellent overview is given in \cite{paden_survey_2016}.

The mentioned approaches are usually modular and interpretable. However, as hand-engineered solutions to difficult problems, they tend to be brittle and require extensive manual fine-tuning. Additionally, isolated changes to parts of the system might reduce or break the overall system performance, requiring careful re-tuning \cite{zeng_end--end_2021}.

These drawbacks motivate the use of deep learning based approaches, which have proven more robust to variations and can be trained in an end-to-end fashion. The current applications of deep learning to autonomous driving can roughly be classified into two groups. Full end-to-end approaches that map raw sensory input directly to vehicle commands (steering, acceleration), and methods that produce or work on intermediate representations. An overview can be found in \cite{tampuu_survey_2020}.
    
\subsubsection{Goals and Requirements}
The motion planning system is in charge of ensuring behavioral safety of the self-driving vehicle \cite{nhtsa_automated_2017, nhtsa_federal_2016}. This includes taking the correct behavior and driving decisions, based on the knowledge of traffic rules and the behavior of other traffic participants, as well as the ability to safely navigate expected and unexpected scenarios.

The \ac{dot} has recommended that Level 3, Level 4, and Level 5 self-driving vehicles should be able to demonstrate at least 28 core competencies adapted from research by \ac{path} at the Institute of Transportation Studies at University of California, Berkeley. These basic behavioral competencies include, amongst others, keeping the vehicle in lane, obeying traffic laws, following road etiquette, responding to other vehicles, and responding to hazards \cite{nhtsa_federal_2016}.

While the majority of these behavioral competencies cover normal driving, i.e., regularly encountered situations, a self-driving vehicle is also responsible for \ac{oedr}, which includes detecting unusual circumstances (emergency vehicles, work zones, ...) as well as planning an appropriate reaction, which typically takes place in the behavior and planning components. Above all, the planning system is responsible for crash avoidance, and should be able to handle control loss, crossing-path crashes, lane changes/merges, head-on/opposite-direction travel, rear-end, road departure, and low speed situations (backing, parking). At any time, the system should be able to execute a fallback action that brings the vehicle to a minimal risk condition. According to \cite{nhtsa_automated_2017}, "a minimal risk condition will vary according to the type and extent of a given failure, but may include automatically bringing the vehicle to a safe stop, preferably outside of an active lane of traffic."

Finally, the motion planner not only interacts with other traffic participants, but also to a great extent with its passengers. In particular, it must be able to communicate proper function, malfunction, as well as an eventual takeover request to a human driver, who must be able to take over in time.

\subsubsection{Necessity of Knowledge Integration}
Level 5 self-driving vehicles are expected to function in a wide variety of operational design domains (we refer to \cite{bsi_operational_2020} for a taxonomy). While the basic principles of safe and comfortable driving remain unchanged, the concrete implementations at the level of traffic laws, customary behavior, and scene structure might be subject to change. We argue that the inclusion of knowledge into a motion planning system will make it easier to handle these situations by increasing traceability (e.g., in the case of crash reconstructions) and reliability. Furthermore, a transparent decision process based on a common understanding between humans and machines will increase interpretability and trust. Finally, we expect the emergence of alternatives to extensive simulation testing, which is at the core of present validation concepts \cite{uber_atg_uber_2020, waymo_llc_waymo_2020, lyft_inc_lyft_2020}.

Emphasizing the advantages of Knowledge Integration, \cite{chen_socially_2017} demonstrates many of the aspects mentioned above. Fan Chen et al. integrate rules, in the form of social norms, by extending the agents reward function, e.g., passing objects with a minimum distance. Violating these rules results in a reward penalty. According to their results, agents with such restrictions exhibit behavior more similar to a human level. Therefore, when integrating knowledge into the machine learning pipeline, models become more interpretable and confidential not solely for experts but for ordinary people since these constraints occur in everyday life.
Furthermore, their extension of the agent´s knowledge reduces learning effort which accelerates training and enables them to outperform their benchmark algorithm in most cases.
Despite those promising benefits, integrating knowledge typically narrows down the broad variety of possible solutions while consuming human work force for hand engineering. This shrinks the original, holistic approach of machine learning. Therefore, the trade-off between knowledge integration and self-learning needs to be chosen carefully \cite{chen_socially_2017}.

\section{Knowledge Representations}
\label{ch:knowledge_representations}
The symbolic and the sub-symbolic methods represent two ends of the \ac{ai} spectrum. The former is more driven by the knowledge and the latter by the data. A plethora of ongoing research can be found in the literature to develop hybrid-\ac{ai} systems which exploit the strengths of one another.  However, there still exists a core challenge in representation of knowledge used in symbolic space  to integrate or augment within the data-driven sub-symbolic/statistical world.  An overview of formalism and languages for representing symbolic knowledge which exists in the form of facts, rules and structured information is reviewed  in \autoref{sec:symbolic_representation_and_knowledge_crafting}.  Furthermore, in \autoref{sec:knowledge_representation_learning} a survey on knowledge embedding is presented, which focuses on transforming prior knowledge from the symbolic space to a real-vector space, i.e., embeddings. These embeddings can be leveraged to improve the sub-symbolic methods (\ac{nn}, \ac{dl}) for effective training, inference and improved reasoning. In addition to it, methods and approaches dealing with injection of hard and soft rules together with embeddings are discussed in \autoref{subsec:Knowledge Graph Embeddings with Rule Injection}.
Each of the sections in this chapter dealing with different mechanisms in representing knowledge is concluded with an outlook that is more tailored to the field of autonomous driving. Mapping perceived information to semantic concepts and reasoning using symbolic models provides improved understanding of driving situation. Furthermore, formalized traffic rules and legal concepts are used to derive possible driving actions conditioned on their legal consequences  \autoref{subsec:application_symbolic_representation}.

\subsection{Symbolic Representations and Knowledge Crafting} \label{sec:symbolic_representation_and_knowledge_crafting}
\textit{Authors: Denny Mattern, Diogo Sasdelli, Tobias Scholl}

\smallskip
\noindent
In contrast to numerical representations (e.g., vector embeddings), which focus on quantitative aspects, logic formalisms  use symbols to represent \emph{things in a logical sense} -- which include physical things (cars, motorcycles, traffic signs), people (pedestrians, driver, police), abstract concepts (overtake, brake, slow down) and non-physical things (website, blog, god) --, as well as propositions expressing their properties and relations obtaining among them. 
Symbolic knowledge representations comprise all kinds of logical formalisms, 
as well as structural knowledge representing entities with their attributes, class hierarchies and relations.

\subsubsection{Logic Formalism}
Logic formalisms are used to express knowledge (mostly facts and rules)
through formal logical expressions.
Different logic formalisms (or \emph{logic systems}) may differ in their complexity, which has consequences for their overall expressivity and for their decidability. 
Choosing an adequate formalism depends on the concrete problem one wishes to model.
Among the most widely used formalisms, propositional logic has the simplest structure. 
It provides a set of symbols for representing individual propositions and a set of operators that can be applied to these propositions in order to generate new propositions.
In classic, two-valued propositional logic, a Boolean value-assignment assigns to each proposition one of two values, e.g., 0 or 1, \emph{T} or \emph{F} or \texttt{true} or \texttt{false}.

For example, the idea that a car does not cause an accident if it is in good condition and is driven carefully can be represented by the expression 
\begin{equation}
    (P\land Q)\to R
\end{equation}
where $P$ is taken to mean \emph{The car drives carefully}; $Q$ to mean \emph{The car is in good condition}, and $R$ to mean \emph{The car does not cause an accident}.

As the name suggests, propositional logic can only be adequately employed to represent \emph{propositions}, i.e., apophantic linguistic utterances. In order to model logical structures concerning not only propositions, but also objects, their properties, and their relations, a more complex logic formalism is required, e.g., first-order logic (FOL). FOL is a kind of predicate logic that extends propositional logic by introducing symbols used to represent functions, constants, variables, predicates and quantifiers (e.g., $\forall$, $\exists$). While FOL is more expressive than propositional logic, it is not \emph{decidable}, i.e., it is not possible to design an algorithm that is able to decide the semantic status of every single FOL-proposition.

For example, the idea that cars are destructible objects, i.e., that they possess the property of being destructible, can be formalised as 
\begin{equation}
    \forall x (Car(x) \rightarrow Destructible(x))
\end{equation}
where $Destructible(x)$ is taken to mean \emph{x is destructible} and $Car(x)$ to mean \emph{x is a car}.

Assuming the validity of the sentence above, it is then possible to infer that a specific car, e.g., \emph{Model T} is destructible, i.e., it is possible to infer the sentence
\begin{equation}
       (Car(Model\;T) \rightarrow Destructible(Model\;T))
\end{equation}
Overall, it is important to notice that the truth-value of FOL-expressions will depend on how their variables are interpreted with respect to some given set of objects, over which the variables range, i.e., to some given \emph{domain}. 

Although predicate logic is more expressive than propositional logic, both share the property of being extensional, i.e., the truth-value of any complex expression depends solely on the truth-values of the expressions it is composed of, and, in the case of systems of predicate logic (e.g., FOL) the definition of any property is reduced to the set of objects containing this property. Hence, these formalisms are unable to adequately represent the distinction between sentences that are true under the same conditions (and of properties that, although distinct, obtain for the same set of objects). For example, the sentences:
\begin{enumerate}
    \item It is sunny and cold
    \item It is sunny, but cold
\end{enumerate}
have slightly different meanings: the word \emph{but} in the second sentence indicates an opposition between it being sunny and it being cold. Notwithstanding, both sentences share the same \emph{extensional meaning}: they are true under the same condition, i.e., when it is both cold and sunny. The semantic difference between these sentences concerns their so-called \emph{intensional meaning}, which cannot be adequately grasped by the formalisms discussed above. 

Systems of so-called \emph{intensional logic} (cf., e.g., \cite{Kutschera1976}) try to model precisely these semantic aspects that do not depend solely on the extension of a given expression (i.e., the intensional semantics). The most well-known examples are systems of \emph{alethic} logic, which try to model the concepts of \emph{possibility} and \emph{necessity}. These concepts are intensional because the possibility (or the necessity) of something depends on more than on whether this something is true or not: while truth does imply possibility (and falseness excludes necessity), falsehood does not exclude impossibility (and truth does not imply necessity).

Systems of intensional logic are usually built on the basis of so-called \emph{modal logic} (ML) (cf., e.g., \cite{Hughescresswell1996}). A modality can be defined as a row of zero or more uninterrupted (e.g., by a parenthesis) monadic operators which cannot be \emph{reduced} to a shorter one, i.e., which is not equivalent to a shorter row. E.g., classic propositional logic has a total of two modalities: $\neg$ and the empty modality. ML-systems introduce new modalities, which are usually defined in a way that does not depend solely on the truth-values assigned to the expressions modulated by them. Semantically, this is usually done be employing so-called \emph{possible-world-semantics}, which expand the Boolean-assignment of classic propositional logic by introducing a universe, i.e., a set of sets of formulas (or \emph{possible worlds}), to which truth-values are likewise assigned following Boolean rules. Thus, e.g., the idea of \emph{necessity} can be represented by truth in all possible worlds; the idea of possibility by truth in at least one possible world.

All of the above-mentioned basic formalisms follow the so-called \emph{bivalence principle}, i.e., they are systems of two-valued-logic. Suppressing this principle leads to so-called \emph{many-valued-logic} (MVL), which encompasses formalisms with three or more values (cf., e.g., \cite{Gottwald1989, FittingOrlowska2003}. Systems with infinitely many values are sometimes called \emph{fuzzy logic}.

An analysis of the literature in the area of logic of norms, which involves an interdisciplinary debate between philosophers, legal scholars and computer scientists, shows that, over the last decades, several different logical systems for representing (legal) norms have been proposed. Structurally, these systems are based on one of the formalisms discussed above (i.e., PL, FOL, ML, MVL). Among these, the most widely employed formalisms are based on ML (especially among philosophers, cf., e.g., \cite{gabbayetal2013, Morscher2012}) or on FOL, in particular so-called \emph{temporal logics} \cite{mainzer2023temporal} (especially among legal scholars and computer scientists, cf., e.g., \cite{maierhofer_formalization_2020, rödig1980, Satohproleg2011}). For formalisms based on MVL, cf., e.g., \cite{Menger1939, Menger1997, Storer1946, Fisher1961, Sadegh-Zadeh2012}

When built on the basis of modal logic, logic of norms is usually called \emph{deontic logic}. It introduces so-called \emph{deontic modalities}, e.g., $[OBL]$, $[PERM]$, $[FOR]$, respectively corresponding to the intuitive ideas of obligation, permission and prohibition. As monadic operators, these modalities qualify the content of respective proposition they operate on. E.g., $[OBL] p$ represents the intuitive notion that $p$ is obligatory. In many systems of deontic logic, the deontic operators satisfy the classic Aristotelian duality relations (see, e.g., \cite{gabbayetal2013}, \cite{athan_legalruleml_2015} for more details):

\begin{itemize}
    \item $[OBL]p \equiv \lnot[PERM]\lnot p$: if $p$ is obligatory, then its negation, i.e., $\lnot p$, is not permitted.
    
    \item $[FOR]p \equiv [OBL]\lnot p$: if $p$ is forbidden, then its negationm i.e., $\lnot p$, is obligatory.
    
    \item $[PERM]p \equiv \lnot [FOR]p$: if p is permitted, then $p$ is not forbidden.
\end{itemize}

\noindent Sometimes, the modality $[PERM]$, which is \emph{subaltern} to $[OBL]$ (i.e., $[OBL]p\to[PERM]p$ is valid] is called \emph{weak} or \emph{negative permission}, and another modality $[PERM']$ is introduced to represent a stronger sense of permission, which usually excludes both obligation and prohibition.

While these relations seem intuitively reasonable, they are difficult to represent in a FOL-based formalism, for if the ideas of obligation, permission and prohibition are to be modeled as properties qualifying actions (modeled as abstract objects), then it would be improper to speak of the \emph{negation} of an action, because negation, as a linguistic operation, cannot be reasonably applied to abstract objects. In other words, one cannot write $\forall x(Obl(x) \equiv \neg Obl(\neg x))$, for $\neg x$ is not syntactically well-formed. 

A promising approach involves combining modal and predicate logic, i.e., employed a formalism based on reified modal logic. However, this comes at the cost of augmented semantic complexity, leading to practical and philosophical problems (for more details, see \cite{Garsonmodallogic2023}).

While it is difficult to determine which formalism is better suited for representing (legal) norms, one can nonetheless identify certain desired properties that systems of logic of norms should ideally possess. One such property is, e.g., \emph{defeasibility}. In more technical terms, defeasibility involves suppressing (at least partially) so-called \emph{monotonicity} with respect to normative inferences. Intuitively, Defeasibility can be defined as being the property that a formalism possess when a possible conclusion is, in principle, open to revision in case more evidence to the contrary is provided \cite{athan_legalruleml_2015}. This is important when formalising (legal) norms because norms often contradict and/or override one another.

Overall, computer-readable formalisation of legal norms is an active research topic in the field of legal informatics. Literature offers multiple examples of logic formalisms for formalising legal rules and norms. Notwithstanding, there is still no consensus concerning the "best" formalism for modeling norms. In order to keep the formalised legal rules agnostic to the rules of the underlying logic formalism, an intermediate formal representation can be used. LegalRuleML (\cite{palmirani_legalruleml_2011, athan_legalruleml_2013, athan_legalruleml_2015}) aims to provide such an interchange format for legal rules, supporting deontic operators and defeasiblity among other features for formalizing legal norms. This intermediary representation can then be mapped to a specific logic in a standard format such as the TPTP~\cite{Sut17}.

As open as the question of the "best" logic formalism for norm representation is the question of a good interface for legal experts who want to represent legal norms computer understandable. A recent work proposes a dedicated editor allowing for intuitive formalization of legal texts and featuring consistency checks as well \cite{dastani_towards_2020}. Another approach proposes an agile and repetitive process \cite{bartolini_agile_2019}.

\subsubsection{Relational Knowledge}
Knowledge concerning entities, concepts, their hierarchies and properties as well as their relations to another is naturally represented by graph structures. Prominent examples for graph structured representations of structural knowledge are \emph{Taxonomies}, \emph{Ontologies} and \emph{Knowledge Graphs}.

\emph{Taxonomies} categorize entities into a hierarchy of classes and sub-classes represented as a directed acyclic graph with nodes representing the entities, classes and sub-classes, and edges representing the relations. Taxonomies categorize objects regarding one specific aspect and commonly use only one type of relation -- the "is-a" relation. E.g., a car is a vehicle, which is a machine.

An \emph{Ontology} is a formal, explicit specification of a shared conceptualization \cite{studer_knowledge_1998}. This means an \emph{Ontology} is an abstract model of explicitly defined, relevant concepts of the specific domain of discourse and their relations which is constructed in a computer understandable manner. The definitions of the meaning of the relevant concepts and relations reflect the common sense of domain experts. Exemplarily, the OpenXOntology~\cite{openxontology} is a conceptualization of traffic sceneries. It features different kinds of traffic participants, infrastructures, events, hierarchies and relations between them. The given definition of what an \emph{Ontology} actually is, implies that the development of a specific \emph{Ontology} is a process which involves different persons (e.g the knowledge engineer, the domain experts, maybe also the users) and that it takes a certain communication effort to develop a shared understanding of the concepts, the formalizations of those concepts as well as the usability of the \emph{Ontology} for the user. Hence, \emph{Ontology} building is ideally an iterative and repetitive design process for which multiple process patterns had been developed \cite{noy_ontology_2001, de_nicola_proposal_2005, garijo_best_2020}.

Concrete \emph{Ontologies} consist of classes and sub-classes which refer to domain concepts as well as the properties and relations between those, which is referred to as terminological knowledge. Additionally to the class definitions, relations and constraints for concrete instances of classes are also defined in an \emph{Ontology} and referred to as assertional knowledge. These definitions and constraints are expressed in description logic which is a decidable fragment of predicate logic, where the terminology \textit{TBox} and \textit{ABox} are often used instead of \textit{terminological knowledge} and \textit{assertional knowledge}. The logic is commonly represented in the \emph{Web Ontology Language (OWL)}, which is a computational language based on description logic that allows for formalizing complex knowledge such that it can be exploited by computer programs \cite{noauthor_owl_nodate}. An \emph{Ontology} can be interpreted as a meta-schema for domain-specific data, that not only specifies the relational structure and semantics of the data but also allows, e.g., to verify the consistency of that knowledge or to infer implicit knowledge through its strong logical foundation. \emph{Ontologies} have been developed for a wide range of domains and applications. 

In literature \emph{Knowledge Graphs} and \emph{Ontologies} had often been used as synonyms until \cite{ehrlinger_towards_2016} proposed the following definition: "A knowledge graph acquires and integrates information into an \emph{Ontology} and applies a reasoner to derive new knowledge." In a \emph{Knowledge Graph} data from heterogeneous data sources is integrated, linked, enriched with contextual information and meta-data (e.g., information about provenience or versioning) and semantically described with an \emph{Ontology}. Through their linked structure \emph{Knowledge Graphs} are prominently used in semantic search applications and recommender systems but also allow for logical reasoning when featuring a formal meta-schema in form of an \emph{Ontology}. Surveys on \emph{Knowledge Graphs} and their general applications are provided by \cite{ji_survey_2021, zou_survey_2020} and \emph{Knowledge Graphs} for recommender systems specifically by \cite{guo_survey_2020}.

\subsubsection{Applications}
\label{subsec:application_symbolic_representation}
Symbolic representations improve scene understanding
by mapping detected objects to a formal semantic representation of the current traffic scene (e.g., as a scene graph (\cite{agarwal_visual_2020, chang_scene_2021})). To integrate knowledge into machine learning algorithms, a representation of this knowledge is essential. While this knowledge is in form of embeddings, a symbolic representation allows traceability and makes it understandable for humans. 

Given a sound formalization of traffic rules and a semantic representation of the entities, actions and legal concepts in traffic scenes (analogue to the legal ontology modeling the concepts of privacy proposed by \cite{ko_pronto_2018}), we can derive the current legal state of an AD vehicle. An example where knowledge graphs are used as embeddings is \cite{oltramari_neuro-symbolic_2020}. In this case a knowledge graph is build upon a road scene ontology to recognize similar situations that are visually different. Using this technique to integrate legal knowledge and derive the legal state of different situations is a possible approach. 

Analogue to the application of symbolic representations for situation understanding we make use of formal representations of traffic rules and legal concepts as well as symbolic scene descriptions for planning tasks by ranking possible alternative trajectories and actions, e.g., according to their legal consequences.
\newpage
\subsection{Knowledge Representation Learning}
\label{sec:knowledge_representation_learning}
\textit{Author: Stefan Zwicklbauer}

\smallskip
\noindent
Complementary strength and weaknesses of data-driven and knowledge-driven \ac{ai} systems have led to a plethora of research works that focus on combining both symbolic (e.g., \acp{kg}) and statistical (e.g., \acp{nn}) methods \cite{darwiche_human-level_2018}. One promising approach is the conversion of symbolic knowledge into embeddings, i.e., dense, real-vector representations of prior knowledge, that can be naturally processed by \acp{nn}. Typical examples of symbolic knowledge are textual descriptions, graph-based definitions or propositional logical rules. The research area of \ac{krl} aims to represent prior knowledge, e.g., entities, relations or rules into embeddings that can be used to improve or solve inference or reasoning tasks (\cite{lin_knowledge_2018}, \cite{li_survey_2020}). Most existing literature narrows down the problem by defining \ac{krl} as converting prior knowledge from \acp{kg} only \cite{lin_knowledge_2018, }. Thus, our focus in this survey also lies on knowledge modeled in graph-based structures.

\subsubsection{Textual Embeddings}
With the development and advances in \ac{dl}, Natural Language Representation Learning has become a hot topic over the last couple of years. Natural Language Models, such as proposed in \cite{devlin_bert_2019}, \cite{peters_deep_2018}, \cite{brown_language_2020} are capable of directly converting natural language text, e.g., common sense text like Wikipedia articles or textual rules like road traffic regulations into embeddings that implicitly represent the syntactic or semantic features of the language \cite{qiu_pre-trained_2020}. Those embeddings are mostly used for specific downstream tasks like \ac{qa} \cite{zhu_retrieving_2021}, \ac{nmt} \cite{yang_survey_2020} or Common Sense Reasoning \cite{storks_commonsense_2019}, but probably lack power of expressiveness when it comes to representing specific rules and logic. As a consequence, most research works extract entities, relations and rules from sentences first and model them in a more expressive representation format, e.g., \acp{kg}, afterwards. In the following, we do not further elaborate literature regarding Natural Language Representation Learning but refer to the respective surveys (\cite{qiu_pre-trained_2020}, \cite{correia_attention_2021}) and assume that knowledge has already been converted to an expressive format like \acp{kg} or another logical system.

\subsubsection{Knowledge Graph Embeddings}
Many research works described how to create dense-vector representation for either homogeneous (i.e., graphs with a single type of edge) and heterogeneous (i.e., graph with multiple types of edges) graphs \cite{dai_survey_2020}. Graphs with auxiliary information (\cite{nikolentzos_matching_2017}, \cite{guo_sse_2017}) and graphs constructed from non-relational data \cite{zhang_regions_2017} are out of scope in this survey. For homogeneous graphs, the authors of \cite{perozzi_deepwalk_2014} made a significant progress in \ac{krl}. They created a node corpus by randomly walking over the graph and applied Word2Vec \cite{mikolov_distributed_2013} to generate node embeddings. The authors of \cite{zwicklbauer_doser_2016} further improved and used this approach for heterogeneous graphs. Tang et al. \cite{tang_line_2015} and especially Grover at al. \cite{grover_node2vec_2016} proposed state-of-the-art works which intelligently explore the specific and varying neighborhoods of nodes and consider the respective node order to create their embeddings. Most research works however, focus on heterogeneous graphs since they are best suited for rule and relation modeling. We first focus on pure node (entity) and edge (relation) representation learning, also called Triplet Fact-based Representation Learning Models. Hereby, we further distinguish between Translation-Based Models, Tensor Factorization-Based Models and \ac{nn}-Based Models.

Starting with Translation-Based Models, the first influential work proposed \textit{TransE} \cite{bordes_semantic_2014}, a framework to create embeddings for heterogeneous graphs. Given a triple $(h, r, t)$, with $h$ and $t$ denoting the head and tail entity and $r$ denoting the respective relation, the idea is to embed each component $h$, $r$ and $t$ into a low-dimensional space $\textbf{h}, \textbf{r}, \textbf{t}$ in a way that $\textbf{h}$ and $\textbf{r}$ translate to $\textbf{t}$: $\textbf{h} + \textbf{r} \approx \textbf{t}$. The authenticity of the respective triplet is defined via a specific scoring function, which is the distance under either $\ell_1$ or $\ell_2$ norm: 
\begin{equation}
f_r(h, t)=\lVert \textbf{h}+\textbf{r}-\textbf{t} \rVert_p
\end{equation}
with $p=1$ or $p=2$.
This objective function is minimized with a margin-based hinge ranking loss function over the training process. Since \textit{TransE} came up with several limitations, such as not being able to model one-to-many, many-to-one and many-to-many relations, various authors addressed these shortcomings by using \textit{TransE} as foundation for their works. For instance, the authors of \textit{TransH} \cite{wang_knowledge_2014} introduced relation-related projection vectors where the entities are projected onto relation-related hyperplanes. \textit{TransH} enables different embeddings based on the underlying relation. All entities and relations are still represented in the same feature space. In \textit{TransR} \cite{lin_learning_2015}, the entities $h$ and $t$ are projected from their initial entity vector space in to the relation space of the connecting relation $r$. This allows us to render entities that are similar to the head or tail entity in the entity space as distinct in the relation space. Further improvements can be found in the \textit{TransD} \cite{ji_knowledge_2015} model, which has fewer parameters and replaces matrix-vector multiplication by vector-vector multiplication for an entity-relation pair, which is more scalable and can be applied to large-scale graphs. Another problem of existing approaches is the non-consideration of crossover-interactions, bi-directional effects between entities and relations including interactions from relations to entities and interactions from entities to relations \cite{zhang_interaction_2019}. To provide an example, predicting a specific relation between two entities typically relies on the entities' relevant topic in form of their connecting entities/relations. Not all connected entities and relations belong to the topic of the relation to be found. This is modeled in \textit{CrossE} \cite{zhang_interaction_2019}, which simulates crossover interactions between entities and relations by learning an interaction matrix to generate multiple specific interaction embeddings. Another state-of-the-art approach \textit{Hake} \cite{zhang_learning_2020} is capable of modeling a) entities at a different level in the semantic hierarchy, and b) entities on the same level of the semantic hierarchy. This is achieved by mapping the entities in the polar coordinate system. Entities on a different hierarchy level are modeled with a modulus approach, whereas the phase part aims to model the entities at the same level of the semantic hierarchy.

Regarding Tensor Factorization-Based Models, \textit{RESCAL} \cite{nickel_three-way_2011} represents the foundational work for most follow-up works. \textit{RESCAL} uses a tensor representation to model the structure of \acp{kg}. More specifically, a rank-$d$ factorization is used to obtain the latent semantics: $\textbf{X}_k\approx \textbf{A}\textbf{R}_k\textbf{A}^T, for\enspace k=1,2,...,m$, with $\textbf{A}\in \mathbb{R}^{nxd}$ being a matrix that captures the latent semantic representation of entities and $\textbf{R}_k\in \mathbb{R}^{dxd}$ being a matrix that models the pairwise interactions in the $k$-th relation. Based on this principle, the scoring function is defined as $f_r(h, t) = \textbf{h}^T\textbf{M}_r\textbf{t}$, where $\textbf{h},\textbf{t}\in \mathbb{R}^d$ denote the entity embeddings and the matrix $\textbf{M}_r\in \mathbb{R}^{dxd}$ represents the pairwise interactions in the k-relation (\cite{nickel_three-way_2011}, \cite{dai_survey_2020}). The work \textit{DistMult} \cite{yang_embedding_2015} improves \textit{RESCAL} in terms of algorithmic complexity and embedding accuracy by restricting $\textbf{M}_r$ to be diagonal matrices. To overcome the problem of \textit{DistMult} that head and tail entities are symmetric for each relation symmetry, the works \textit{Complex} \cite{trouillon_complex_2016} and \textit{QuatRE} \cite{nguyen_quatre_2020} satisfy the key desiderata of relational representation learning, i.e., modeling symmetry, anti-symmetry and inversion. Both approaches leverage complex-value embeddings to support asymmetric relations. More recently proposed state-of-the-art models use special tensor factorization methods. For instance, \textit{SimplE} \cite{kazemi_simple_2018} leverages an adapted and simpler version of Canonical Polyadic Decomposition to allow head and tail entities to have embeddings that are dependent on each other, which would be impossible with the original model. Similar, \textit{TuckER} \cite{balazevic_tucker_2019} is based on the Tucker-Decomposition on a binary entity-relation-entity matrix.

Due to their success in the last decade, \ac{nn}-Based Models became also a hot topic for \ac{krl}. The first shallow \ac{nn} approaches comprise standard feed-forward networks \cite{bordes_semantic_2014} (with linear layers) and neural tensor networks \cite{socher_reasoning_2013} (with bi-linear tensor layers). Over time deeper variants such as \textit{NAM} \cite{liu_probabilistic_2016} have established to provide more flexibility when it comes to train a network towards the underlying training goal. More recently, graph neural networks \cite{zhou_graph_2018} were introduced which strive to explicitly model the peculiarities of (knowledge) graphs. In particular, graph convolutional networks for multi-relational graphs \cite{kipf_semi-supervised_2017} generalize nonvolutional neural networks to non-euclidean data and gather information from the entity’s neighborhood and all neighbors contribute equally in the information passing. Graph convolutional networks are mostly built on top of the message passing neural networks framework \cite{gilmer_neural_2017} for node aggregation. Many works are limited to create embeddings for knowledge entities only (\cite{schlichtkrull_modeling_2018}, \cite{shang_end--end_2018}), but recent approaches tried to overcome this limitation (\cite{dettmers_convolutional_2018}, \cite{vashishth_composition-based_2020}, \cite{ye_vectorized_2019}, \cite{vashishth_interacte_2019}). A neighborhood attention operation in graph attention networks \cite{velickovic_graph_2017} can enhance the representation power of graph neural networks \cite{xu_dynamically_2020}. Similar to natural language models, these approaches apply a multi-head self attention mechanism \cite{vaswani_attention_2017} to focus on specific neighbor interactions when aggregating messages (\cite{xu_dynamically_2020}, \cite{abu-el-haija_watch_2018}, \cite{ma_graph_2021}). Many authors incorporated mechanisms to improve the overall quality of entity and relation embeddings. For instance, the idea of negative sampling is to intelligently sample specific wrong samples that are needed for margin-based loss functions. Recent methods employed \acp{gan} \cite{goodfellow_generative_2014} in which the generator is trained to generate negative samples (\cite{wang_incorporating_2018}, \cite{cai_kbgan_2018}). Another work \textit{ATransN} suggested to improve existing embeddings by leveraging \acp{gan} to correctly align the embeddings with those from teacher \acp{kg} \cite{wang_adversarial_2021}. 

In this section, we mostly concentrated on methods that exclusively generated their embeddings on relational data. However, some approaches consider additional information, such es textual (entity) descriptions (e.g., \cite{fu_collaborative_2019}, \cite{wang_tackling_2019}, \cite{gupta_care_2019}), path-based information (e.g., \cite{niu_rule-guided_2020}, \cite{guo_learning_2019}) and even hierarchies (e.g., \cite{zhang_learning_2020}, \cite{zhang_knowledge_2018}) as available in ontologies.   

\vspace{-0.1cm}
\subsubsection{Knowledge Graph Embeddings with Rule Injection}
\label{subsec:Knowledge Graph Embeddings with Rule Injection}
So far, we have discussed approaches to embed knowledge that is formalized within \acp{kg}. These methods create representations that purely reflect the items' graph-based modeling (e.g., triples). In addition to this, specific rules (soft or hard rules) can be derived from \acp{kg}, which is also known as rule learning (e.g., \cite{ho_rule_2018}, \cite{zhang_iteratively_2019}, \cite{omran_scalable_2018}), or be leveraged in the embeddings learning process, also known as rule injection. In the following, we focus on the former works, how to additionally integrate pre-defined or mined rules into embeddings. The authors of \textit{RUGE} \cite{guo_knowledge_2018} presented a novel paradigm to leverage horn soft rules mined from the underlying \ac{kg} in addition to the existing triples. Their iterative training procedure improves the transfer of the knowledge contained in logic rules into the learned embeddings. The framework \textit{SLRE} \cite{guo_knowledge_2020} also presents an option to leverage horn-based soft rules with confidence scores to improve the accuracy of down-stream tasks. These rules are directly integrated as regularization terms in the training mechanism for relation embeddings. The authors of \cite{niu_rule-guided_2020} additionally enriched the horn-based rules with path information to improve the state of the art. A related work \cite{wang_logic_2019} mines inference, transitivity and anti-symmetry rules from the given \ac{kg} first and converts them into first-order logic rules in the second step. Finally, the proposed rule-enhanced embedding method can be integrated in any translation-based \ac{kg} embedding model.

Apart from rules directly mined from the underlying knowledge graph, other approaches exist that try to apply more extrinsic rules. For instance, the authors of \cite{ding_improving_2018} try to improve the embeddings' capability of modeling rules by using non-negativity and approximate entailment constraints to learn compact entity representations. The former naturally induce sparsity and embedding interpretability, and the latter can encode regularities of logical entailment between relations in their distributed representations. Other works propose to encode knowledge items into geometric regions. For instance, \cite{gutierrez-basulto_knowledge_2018} encodes relations into convex regions, which is a natural way to take into account prior knowledge about dependencies between different relations. \textit{Query2box} \cite{ren_query2box_2020} encodes entities (and queries) into hyper-rectangles, also called box embeddings to overcome the problem of point queries, i.e., a complex query represents a potentially large set of its answer entities, but it is unclear how such a set can be represented as a single point. Box Embeddings have also been used to model the hierarchical nature of ontology concepts with uncertainty \cite{li_smoothing_2019}.

Most approaches described above rely on common-sense knowledge bases like DBPedia \cite{hutchison_dbpedia_2007} or Freebase \cite{bollacker_freebase_2008} and leverage their developed embedding approaches for knowledge base link prediction or inference / reasoning tasks. However, we believe that existing models and algorithms can be similarly applied to special domain knowledge bases, e.g., knowledge bases with data for \ac{ad} \cite{wickramarachchi_evaluation_2020}.

\subsubsection{Applications}
\label{subsec:applications_knowledge_representation_learning}
The application of \acp{kg} in the \ac{ad} domain has not received too much attention at the current point of time, albeit it can be an effective way to help situation or scene understanding \cite{ulbrich_defining_2015}. For instance, the authors of \cite{halilaj_knowledge_2021} built a specific ontology to represent all core concepts that are essential to model the driving concept. The built \ac{kg} \textit{CoSi} models information about driver, vehicle, road infrastructure, driving situation and interacting traffic participants \cite{halilaj_knowledge_2021}. To classify the underlying traffic situation with a \ac{nn}, a relational graph convolutional network \cite{schlichtkrull_modeling_2018} is used to convert the underlying \ac{kg} into embeddings first. Similar, the work by Buechel et al. \cite{buechel_ontology-based_2017} presented a framework for driving scene representation and incorporated traffic regulations. Wickramarachchi et al. \cite{wickramarachchi_evaluation_2020} focused on embedding \ac{ad} data and investigated the quality of the trained embeddings given various degrees of \ac{ad} scene details in the \ac{kg}. Moreover, the authors evaluated the created embeddings on two relevant use cases, namely \textit{Scene Distinction} and \textit{Scene Similarity}.

\vspace{-0.2cm}
\section{Knowledge Integration}
\label{ch:integration}
A plethora of methods and approaches have been proposed in literature that focus on augmenting data driven models and algorithms with additional prior knowledge. Among the most prominent approaches are the modification of the training objective via customized cost functions, especially knowledge affected constraints and penalties. An overview of \textit{auxiliary losses and constraints} that take into account physical and domain knowledge in various peculiarity is presented in \autoref{sec:auxiliary_losses}. Often these approaches are accompanied by problem-specific designs of the architecture, leading to hybrid models that leverage symbolic knowledge in form of logical expressions or knowledge graphs. The merging of symbolic and sub-symbolic methods, also referred to as \textit{neural-symbolic integration} is focus in \autoref{sec:neuralsymbolic_integration}. 

Besides external input, recent methods rely on preferably internal representations in order to focus \textit{attention} on distinct features and concepts within a network itself. Key weighting and guidance approaches are discussed in \autoref{Knowledge_Integration/attention}. Last but not least, \textit{data augmentation} techniques form the backbone to integrate additional domain knowledge into the data and thus indirectly into the model. Approaches starting from data transformations to augmentations in feature space up to simulations are discussed in \autoref{Knowledge_Integration/data_augmentation}. 

In addition to these prevalent general approaches, this chapter concludes with methods and paradigms that are more tailored to the field of autonomous driving, considering multiple agents that interact with specific environments typical for the application under investigation. Especially inferring and predicting the state of an agent plays an essential role in the considered \textit{state space models} in \autoref{sec:state_space_models} and \textit{reinforcement learning} in \autoref{Knowledge_Integration/reinforcement_learning}. The involvement of positional as well as semantic information is essential part of the \textit{information fusion} approach outlined in \autoref{sec:sensor_fusion_methods}.
\subsection{Auxiliary Losses and Constraints}
\label{sec:auxiliary_losses}
\textit{Authors: Tino Werner, Maximilian Alexander Pintz, Laura von Rueden, Vera Stehr}

\smallskip
\noindent
The usual \ac{erm} principle in machine learning amounts to replacing the minimization of an intractable risk, i.e., an expected loss over a ground-truth data distribution, by the minimization of the empirical risk. A mismatch between the expected loss and its empirical approximation causes \ac{erm} to result in models that do not generalize well to unseen data.
This manifests either in overfitting, where the model represents the training data too closely and fails to capture the overall data distribution, or underfitting, where the model fails to capture the underlying structure of the data.  Regularization schemes have been proposed to mitigate the problem of overfitting. The \ac{srm} principle~\cite{guyon_structural_1991, vapnik_statistical_1998} extends the \ac{erm} principle for regularization.
\ac{srm} seeks to find models with the best tradeoff between the empirical risk and model complexity as measured by the Vapnik-Chervonenkis dimension or Rademacher complexity. In practice, this encompasses minimizing an empirical risk with an added regularization term.
This technique has successfully entered variable selection as done in the path-breaking work of \cite{tibshirani_regression_1994} who introduced the Lasso.
Regularization in general proved to be indispensable in high-dimensional regression \cite{efron_least_2004, zou_adaptive_2006, candes_enhancing_2008, yuan_model_2006, simon_sparse-group_2013}, classification ~\cite{park_l_1-regularization_2007, van_de_geer_high-dimensional_2008}, clustering~\cite{witten_framework_2010}, ranking~\cite{laporte_nonconvex_2014} and sparse covariance or precision matrix estimation~\cite{banerjee_model_2008, friedman_sparse_2008, cai_constrained_2011}.


As for knowledge-infusion into \ac{ai}, a natural strategy is to similarly use regularization terms (so-called auxiliary losses) that correspond to formalized knowledge. However, constraints may also appear in terms of hard constraints, for example, if some logic rule must not be violated so that integrating it in a soft manner via auxiliary losses would not be appropriate, as dependencies or as regularization priors. This section is structured as follows: After describing techniques that integrate physical knowledge or other domain knowledge via auxiliary losses, we review ideas to incorporate constraints into the \ac{ai} training and the \ac{ai} architecture, followed by works that propose uncertainty quantification for knowledge-infused networks. At the end, we review applications of knowledge-infused networks for perception and planning in the automotive context. 

Let us first point out the advantages of such techniques, besides the stronger adaptation of the model to the knowledge. For example, the authors in \cite{karpatne_physics-guided_2018} highlighted that the knowledge-based regularization term does generally not require labeled inputs, which enables data augmentation with unlabeled instances, saving a large amount of time and money that would be required to generate a large labeled dataset. 
The approach in~\cite{gao_phygeonet_2021} does not even require any labeled instance. Moreover, a common result is better generalizability of the model, paralleling the improved generalization ability of models trained with complexity regularization. Each improvement in explainability and interpretability of deep models is especially relevant for autonomous driving in order to increase the public acceptance of self-driving vehicles.

\subsubsection{Knowledge Integration via Auxiliary Losses}

One has to distinguish between common penalty terms that regularize the complexity of the model as outlined in the previous paragraph and knowledge-based penalty terms that integrate formalized knowledge into the model. Surveys on knowledge integration, including auxiliary loss functions, are given by~\cite{stewart_label-free_2017, deng_integrating_2020, rai_driven_2020, willard_integrating_2020, von_rueden_informed_2019, borghesi_improving_2020}. An important side effect from knowledge integration, apart from a better generalization performance, is that it increases explainability and interpretability of the model by at least partially explaining the predictions by the knowledge. This is especially true for deep models which are usually black boxes.

Regarding the success of regularization in machine learning, knowledge-based regularization terms have the potential to significantly improve machine learning models by encouraging them to respect existent knowledge without wasting computational effort for learning this knowledge again from scratch during training but more efficiently. It is important to note that knowledge integration via losses and penalties is also possible if the knowledge is not present in the data (for example, if it is related to rare cases) or if it cannot easily be derived from the data. 
Work in this direction has been done by \cite{stewart_label-free_2017}, but note that \cite{koscianski_physically_2005} already had the idea of physics-based regularization when solving inverse problems. 
The authors in \cite{stewart_label-free_2017} consider the applications of predicting a height curve when throwing an object (constraint: it is a parabola), the location of a walking person (constraint: the velocity should remain constant) and casual relationships (video game) by adding a suitable regularization term to the loss function.

\paragraph{\textbf{Physical Knowledge}}

Real-world environments are constrained by physical laws, which need to be considered for realistic modeling. Several approaches have been proposed for infusing such physical knowledge into neural networks. 
In \cite{raissi_physics_2017, raissi_multistep_2018, raissi_physics-informed_2019} physics-informed \acp{nn} are introduced to reliably solve partial differential equations (i.e., enforcing the solution to respect physical laws) like the Schrödinger equation or discrete time models like Runge-Kutta models. The authors in \cite{lu_physics-informed_2021} describe how to impose soft boundary conditions for \acp{pde} via auxiliary loss functions. The authors in \cite{shi_physics-informed_2022} use physics-informed \acp{cnn} in order to predict physical fields. The authors in \cite{hernandez_structure-preserving_2021} propose physics-guided \acp{nn} that solve \ac{pde}s while satisfying thermodynamical constraints. Further applications in differential equation and dynamics modeling are given for example in \cite{bekele_physics-informed_2020, dana_physics_2021, rao_physics-informed_2021, rao_physics-informed_2020, mao_physics-informed_2020, yin_non-invasive_2021, zhang_physics-guided_2020, wang_understanding_2021, wang_when_2022, gao_phygeonet_2021, jagtap_conservative_2020, jagtap_extended_2020, xu_physics-informed_2022}.

Several works on physics-informed \acp{nn} consider the problem of temperature modeling (\eg of lakes or sea surfaces), such as \cite{karpatne_physics-guided_2018,de_bezenac_deep_2019} or \cite{muralidhar_incorporating_2018}, who try to encourage a monotonicity constraint by an auxiliary loss term (\eg the water density increases monotonically with depth). 
The authors in \cite{jia_physics_2018} use physics-guided recurrent graph networks to model the flow and the temperature in rivers and enforce the model to respect local patterns via physics-guided regularization. 
In \cite{jia_physics-guided_2021}, an energy conservation constraint is integrated while \cite{uzun_physics_2021} apply physical regularization in fuel consumption modeling.
 
\paragraph{\textbf{Domain Knowledge}}

The authors in \cite{vo_combination_2017} incorporate domain knowledge (here: sentiment dictionary/ontology, linguistic patterns) into \ac{dl} in the context of sentiment analysis. Medical domain knowledge in terms of priors on abdominal organ sizes is integrated into \ac{dnn} models in~\cite{zhou_prior-aware_2019} for the task of segmenting organs on \ac{ct} scans.  The authors in \cite{chang_kg-gan_2019} propose knowledge-guided \acp{gan} that are trained using image data and additional textual descriptions of (potentially unseen) input images (types of flowers). They train two generators, one for generating images of seen categories and one for unseen categories, and use an auxiliary loss to transfer knowledge between the generators.
In \cite{yang_quantitative_2020} the contribution of domain knowledge is quantified by approximating the Shapley value (see also \autoref{sec:saliency_modelagnostic}) of a particular knowledge constraint.

Imposing general logical rules (equations, inequalities, orderings) on network outputs is also considered for incorporating domain knowledge in the literature. 
The authors in ~\cite{xu_semantic_2018} construct a loss function, such that the output of neural network satisfies certain first-order logic sentences upon minimization of the loss.
In \cite{fischer_dl2_2019} are more general framework is proposed that turns general first-order sentences into differentiable loss functions using max or logit operators. The training of the constrained \ac{nn} is simply done using standard optimization techniques like \ac{sgd}. 

\subsubsection{Integration of other Constraints}

Adding a knowledge-based regularization term to a loss function typically enforces constraints in a soft manner.
However, in many cases we would like to ensure that constraints are perfectly satisfied, i.e., enforce hard constraints that correspond to the limit case of auxiliary regularization terms with infinite regularization parameters. 
In the following, several approaches are introduced that aim at incorporating hard constraints.
Besides using auxiliary losses, these often employ other approaches for constraint incorporation such as a change in architecture or use different optimization schemes such as projected gradient descent or conditional gradients \cite{ravi_explicitly_2019}.
\paragraph{\textbf{Hard constraints}} 
Methods to train \ac{nn}s with hard constraints on the output layer are explored in \cite{marquez-neila_imposing_2017}, but due to the large dimensionality it is infeasible to apply standard Lagrangian techniques and even worse, if the constraints are incompatible, one will face numerical instabilities. In order to solve the linear system imposed by the \ac{kkt} conditions, they use the Krylov subspace method which iteratively solves linear equations. The required products of Jacobians and vectors are computed using the Pearlmutter trick. It is further discussed how the Krylov method can be improved to cope with ill-posed constraints, how a constrained Adam looks like and how to reduce the number of constraints during learning. The latter is achieved by randomly selecting active constraints on the unlabeled data as \ac{sgd} selects instances for labeled data. They suggest that randomly choosing them may be replaced by using the ones for which the constraint violation is largest, i.e., it is some kind of active learning approach for the constraints.
The authors in \cite{nandwani_primal_2019} show how to perform deep learning with hard constraints by converting hard label constraints to soft logic constraints over distributions. They point out that the work of~\cite{diligenti_integrating_2017} is similar to theirs but does not use a full Lagrangian for respecting the constraints (logical formulas) but modify the loss functions so that hard constraints cannot be handled. Equality constraints are formulated using the two corresponding inequality constraints. Using the Hinge loss, the constraints can be equivalently written as equality constraints (\cite{kervadec_constrained_2019} call it ReLU Lagrangian) which reduces the number of constraints and allows any constraints as long as they are differentiable. 
Training is done via subgradient descent.
The authors in \cite{fioretto_lagrangian_2020} consider inequality constraints on \ac{dnn}s and formulate it very similarly to~\cite{nandwani_primal_2019} using the Hinge loss, but they consider a primal-dual formalization as~\cite{chen_primal-dual_2013} for solving the problem. 
The authors in \cite{kervadec_constrained_2019} propose log-barrier extensions to approximate the Lagrangian optimization of constrained \acp{cnn} with a sequence of unconstrained losses with an initial feasible set of parameters.
The main idea is to first compute any feasible point of the constrained problem with an inequality constraint and to approximate the original problem with the unconstrained problem but where the inequalities enter as penalty terms with a log-barrier function which approximates the Hinge loss. They provide a continuous and twice differentiable log-barrier extension which is no longer restricted to feasible points and therefore does not require to find a feasible initial point.
The authors in \cite{pauli_neural_2022} consider the training of matrix inequality constrained (semidefinitely constrained) \acp{nn} that are used for enforcing Lipschitz continuity or stability. Training robust, i.e., Lipschitz \acp{nn} has already been considered in \cite{pauli_training_2022} who solve the Lipschitz-regularized optimization problem using an \ac{admm} scheme. In order to capture even nonlinear matrix inequality constraints, \cite{pauli_neural_2022} propose to transform the constrained problem into an unconstrained problem using log-det barrier functions.
The framework of \cite{fischer_dl2_2019} enforces besides the logic-based soft constraints, also convex hard constraints via projected gradient descent (projecting gradients back into the convex constraint region).
Besides enforcing soft boundary conditions of physics-based models, the authors in \cite{lu_physics-informed_2021} also propose a \ac{nn} architecture for encoding hard constraints.
\paragraph{\textbf{Constraint incorporation via layers}}
Other techniques address the problem which knowledge constraints to integrate when and to which extent (e.g., how the regularization parameters have to be chosen).
The authors in \cite{kursuncu_knowledge_2019} criticize that many existing approaches incorporate the knowledge before or after the learning process by feature extraction or validation and therefore propose a method how to incorporate it within the hidden layers themselves, i.e., by infusing the knowledge between the layers.
In order to decide whether knowledge should be incorporated between particular hidden layers and how the latent representation and the knowledge representation merge, they propose two loss functions. As for the knowledge representation, they build knowledge graphs.
The knowledge infusion is realized by a knowledge-infusion layer which minimizes the gap between the learned representation and the knowledge representation (called differential knowledge) using a knowledge-aware loss function, i.e., a relative entropy loss quantifying the information gain from the knowledge representation.
 Finally, a weight matrix based on the differential knowledge is learned and the \ac{ai} is trained using \ac{bp}. 

The authors in \cite{amos_optnet_2017} propose OptNet that uses special \ac{dnn} layers for solving optimization problems which encode constraints as well as complex dependencies among the hidden nodes. They concentrate on quadratic problems and the solution becomes the output of the respective layer. To enable \ac{bp} through these layers, the derivatives of the solutions (i.e., of the argmin operator) have to be computed, which is done by differentiating the \ac{kkt} conditions. They prove that the OptNet layers are subdifferentiable everywhere and that they can approximate any piecewise linear function but, however, point out that OptNet layers are costly. In \cite{de_avila_belbute-peres_end--end_2018}, a linear complementarity problem for equality- and inequality-constrained reinforcement learning (see \autoref{Knowledge_Integration/reinforcement_learning}) is formulated which, using the results from~\cite{amos_optnet_2017}, allows for gradient computation while keeping the \ac{bp} solution scheme. Their approach can be interpreted as adding a physics-based layer to the network.

There are also efforts to incorporate logic constraints directly into the network architecture.
The authors in \cite{li_augmenting_2019} consider logic rules on the activations of \acp{dnn}. To enforce such rules, the pre-activations of the network are augmented with terms that increase when given logical statements are satisfied. 
A differentiable logic layer for trajectory prediction that can incorporate symbolic priors and temporal logic formulae is proposed in \cite{li_differentiable_2020}. Since this requires much less labeled data, trajectory predictors can serve as trajectory generators. The parameter adjustment to the rules is done in the \ac{bp} step. Furthermore, the layer can check whether rules are satisfied/violated. The idea is to define a robustness function based on signal temporal logic formulae so that they are satisfied if and only if the robustness function is greater than zero. Minimum and maximum operators are smoothly approximated. Training is done by \ac{bp} where the gradients of this robustness function are used. 
In \cite{hu_harnessing_2016}, \acp{dnn} are combined with declarative first-order logic rules. This is done by enforcing the \ac{nn} to predict the outputs of a logic-rule-based teacher and updating both \acp{nn} iteratively. For a classification task, the softmax output that the student network assigns to an instance is projected onto the rule subspace where the constraints are satisfied, leading to the softmax output of the teacher network. The parameters of the student network are iteratively updated while the teacher network is trained so that it satisfies the first-order constraints by minimizing the \ac{kl}-divergence.
\paragraph{\textbf{Posterior regularization}}
The authors in \cite{mei_robust_2014} propose robust RegBayes which does not incorporate knowledge via the priors but by posterior regularization w.r.t. first-order logic rules (see  \autoref{Knowledge_Integration/symbolic_representation_and_knowledge_crafting}). The idea builds upon regularized Bayesian inference (RegBayes) from~\cite{zhu_bayesian_2014}. Robust RegBayes takes the uncertainty about the domain knowledge into account and outputs parameters that reflect the importance of each logic constraint which will be low in cases of large uncertainties about the knowledge. 
In \cite{hu_deep_2016}, the method of \cite{hu_harnessing_2016} is generalized by jointly learning both the regularized \ac{dnn} models as well as the structured knowledge. More precisely, the task is to learn the regularization parameters in the penalized objective function as well as dependency structures of the knowledge constraints. Their technique can be interpreted as regularized Bayes with generalized posterior \cite{zhu_bayesian_2014}. 
The authors in \cite{zhang_prior_2018} also propose posterior regularization (see~\cite{ganchev_posterior_2010}) for prior knowledge integration in order to handle multiple overlapping prior knowledge sources in the context of neural machine translation. They penalize the likelihood by the \ac{kl}-divergence of the resulting model and a distribution that encodes prior knowledge. 
In \cite{peng_discretely-constrained_2020}, discrete constraints and regularization priors for \acp{cnn} are proposed, leading to discrete-valued regularization terms. The optimization problem is re-formulated as \ac{alm} and solved using an \ac{admm} scheme.

\subsubsection{Uncertainty Quantification of Knowledge-based \acp{dnn}}

The authors in \cite{daw_physics-guided_2020} combine the physics-guided architectures with \ac{mc} dropout (c.f.~\autoref{sec:uncertainty}) for uncertainty quantification and show that the physics-guided \ac{nn} approach still yield black-box models and that the random dropping of weights again leads to physically inconsistent predictions. They remedy this issue by introducing physically-informed connections and physical intermediate variables which grant certain neurons a physical interpretation. 
They consider a monotonicity-preserving \ac{lstm} which extracts temporal features and predicts an intermediate physical quantity (water density) such that the monotonicity is satisfied for this quantity by hard-coding it into the architecture. Then, an \ac{mlp} combines these predictions with the inputs to get the predicted responses. The perturbations injected by MC Dropout do not destroy the consistency with the physical knowledge. 
In \cite{zhang_quantifying_2019}, a dropout variant for uncertainty estimation (both approximation and parameter uncertainty) in physics-guided \acp{nn} is suggested for the context of forward and inverse stochastic problems by invoking polynomial chaos and \ac{mc} dropout. 
The authors in \cite{yang_adversarial_2019} propose a latent-variable-based adversarial inference procedure for uncertainty quantification of physics-based \acp{nn}. In \cite{kaltenbach_incorporating_2020, kaltenbach_physics-aware_2021}, uncertainty quantification for physics-guided \acp{nn} in dynamical systems is done by a coarse-graining process which again results in a Bayesian-type approach where an evidence lower bound is maximized.

\subsubsection{Applications}

%
Knowledge integration has touched upon several perception tasks. As for object detection, \cite{luo_fast_2018} integrate prior knowledge about the size of the bounding boxes of vehicles into the model by imposing size constraints for the boxes. 
The authors in \cite{patel_weakly_2022} consider equivariance constraints in weakly supervised segmentation in order to cope with affine image transformations. As \acp{cnn} are not equivariant in general, they impose an equivariance-preserving loss and extend this technique for shared information between multiple networks. 
The authors in \cite{yi_knowledge-based_2018} propose a knowledge-based attentive \ac{rnn} (see also \autoref{Knowledge_Integration/attention}) for traffic sign detection, motivated by the fact that small objects are not yet detected reliably by \acp{dnn}. The idea is to impose a prior distribution on the location of the traffic signs that represents the domain knowledge that the driver's attention is the bias of the center and the intuitive knowledge that human's attention follows a Gaussian distribution. 
The former emerged from \cite{cornia_predicting_2018} who automatically learn priors that respect that issue, i.e., it learns the bias from eye fixations.
As for semantic segmentation, \cite{kervadec_bounding_2020} impose constraints such that each bounding box at least has to contain a foreground pixel (to prevent excessive shrinking) and no background pixel (background emptiness constraint). 
To solve the resulting problem, they employ log-barrier extensions and optimize the corresponding Lagrangian function directly via \ac{sgd} as proposed in \cite{kervadec_constrained_2019}. The authors in \cite{wang_bounding_2021} propose a bounding box tightness prior for weakly supervised image segmentation by applying a smooth maximum approximation instead of posing it directly as constraints as in \cite{kervadec_bounding_2020}.
In \cite{pathak_constrained_2015}, constrained \acp{cnn} are proposed to incorporate weak supervision into the learning procedure. 
The idea is to define linear constraints on the output layer that enforce the output being near the latent distribution from weak supervision. Their formulation covers for example bounds for the expected number of foreground and background pixel labels in a scene, suppression of a label in a scene (object is not allowed to appear) or size constraints. 
They concretize the problem using a \ac{kl}-divergence-based loss function which can be solved using \ac{sgd} on the dual. 
A related approach is presented by \cite{oktay_anatomically_2017} who impose anatomical constraints on a \ac{cnn}. The authors in \cite{wang_context-aware_2021} propose virtual adversarial training for anatomically-plausible image segmentation, i.e., they generate adversarial samples that violate the topological constraints and let the network learn to avoid such predictions. They point out that additional losses that correspond to some constraint violation may not exist or may not be differentiable. Even worse, if the constraints are complex relationships, the \ac{nn} may never violate them during training so that the constraint will always lead to a gradient of zero. They optimize a regularized cross-entropy loss where the context-aware regularizer is the maximum of a \ac{kl}-divergence, penalized by a constraint loss which encourages adversarial samples.
The authors in \cite{peng_discretely-constrained_2020} impose discrete constraints which may be lower and upper bounds for the foreground size and the regularization prior can be a measure of the similarity of the intensity or color of neighboring pixels. \cite{peng_active_2021} experimentally derive that existing \ac{al} methods work poorly for lane detection due to label noise (maybe due to occlusion or unclear lane markings) and due to the fact that the entropy criterion leads to selecting images with no or only few lanes. They propose to train a student model using the same loss as for the teacher model, regularized with a distillation loss. As for mitigating the label noise that may be the reason for large discrepancies of the teacher and the student, they train another student without knowledge distillation. They select samples where the discrepancy of the student's predictions are large but where the discrepancy of the teacher and the distilled student are low (teacher may be erroneous here) or where the latter discrepancy is large and those between the students is low (knowledge is difficult to learn). Experiments are conducted on the LLAMAS and the CULane dataset.

In \cite{marquez-neila_imposing_2017}, human pose estimation with hard symmetry constraints is considered while \cite{hu_deep_2018} impose a consistency constraint that encourages the body parts of the generated images match the respective parts in the real images. 
As for image classification, \cite{melacci_can_2020} include hierarchical domain knowledge into classification tasks, i.e., that all parts belonging to a certain vehicle or all vehicles that contain a given part are considered.
The authors in \cite{yang_quantitative_2020} incorporate symbolic knowledge in classification, i.e., they consider super-classes that provide information about the potential actual sub-classes.
Knowledge can also be integrated into tracking and trajectory prediction. 
In \cite{greer_trajectory_2021}, the Yaw loss, an auxiliary differentiable heading loss that penalized angle differences between the optimal and the predicted headings, is proposed, where the case of road intersections is also respected.
The authors in \cite{niedoba_improving_2019} propose an off-road loss for improving the movement prediction of traffic participants. This loss is the mean Euclidean distance between each predicted waypoint and the corresponding nearest feasible (drivable) point. 
In \cite{boulton_motion_2021}, this approach is extended by using a pre-trained model (according to off-road loss) and by combining it with models like CoverNet from \cite{phan-minh_covernet_2020} that respect dynamic constraints and that make multimodal probabilistic trajectory predictions or by the method from \cite{cui_deep_2020} who predict kinematically feasible trajectories using a kinematic layer. The authors in \cite{stewart_label-free_2017} enhance pedestrian tracking models by the world knowledge that the walking speed is constant. The idea in \cite{bahari_injecting_2021} is to add residuals to knowledge-driven trajectories in order to better reflect the stochastic behavior, to make it more realistic and to let the prediction effectively account for other agent's behaviors. They also consider social rules (world knowledge) concerning the movements of pedestrians. They show that their approach can also be used for multimodal prediction and combined with the kinematic layer from~\cite{cui_deep_2020}.
The authors in \cite{zhang_stinet_2020} propose STINet for joint pedestrian detection and trajectory prediction. 
The idea is to model temporal information for each pedestrian so that current and past states are predicted. 
They also model the interaction of the pedestrians with an interaction graph. 
A temporal-region proposal network is applied in order to make object proposals in terms of past and current boxes, supervised by the ground truth boxes. 
In \cite{ju_interaction-aware_2020}, the interaction-aware Kalman \ac{nn} for predicting interaction-aware trajectories is proposed.

%
As for planning, knowledge-infused models for semantic segmentation, object recognition and trajectory prediction outlined in the perception subsection can potentially be used for planning the ego-trajectory since they improve the quality of the observed and predicted states respectively. Especially approaches like STINet \cite{zhang_interaction_2019} that incorporate social interactions are candidates since the interactions with the ego-vehicle can be included.
The authors in \cite{sadat_jointly_2019} add different regularization terms corresponding to speed limits, dynamics or lane changes.
In \cite{cui_ellipse_2021}, the ellipse loss is proposed which penalizes the bounding box regression and orientation loss with an off-road loss computed by a non-drivable region mask which is added to the computed Gaussian raster. 
Position and heading of the agent enter as regularization terms in the ChauffeurNet of \cite{bansal_chauffeurnet_2018}.
The authors in \cite{casas_mp3_2021} consider penalties, for example for trajectory curvature, lateral acceleration and off-road driving.

The authors in \cite{bolderman_physicsguided_2021} propose to use physics-guided \acp{nn} for inversion-based feed-forward control applied to linear motors. Two physics-guided \acp{nn} are considered, one in which the inputs are transformed according to the feed-forward controller (i.e., physics-guided input transformation) and one in which a physics-guided layer is used where the output is transformed according to the physical model (maybe enhanced with a physics-guided input transformation). Their model is applied to tracking tasks.
\subsection{Neural-symbolic Integration}
\label{sec:neuralsymbolic_integration}
\textit{Authors: Tobias Scholl, Philip Gottschall, Christian Hesels, Gurucharan Srinivas}

\smallskip
\noindent
Machine learning and deep learning techniques (so-called sub-symbolic \ac{ai} techniques) have proven to be able to achieve great performance in pattern recognition tasks of numerous kinds: image recognition, language translation, medical diagnosis, speech recognition, recommender systems and many more. While the accuracy in performing those tasks which require dealing with large and noisy input is often on par with human abilities or even beyond that, they come with certain drawbacks: They usually offer no justification for their output, require (too) much data and computational power to be trained, are susceptible to adversarial attacks and are often criticized to generalize weakly beyond their training distribution. On the other hand, "classic" so-called symbolic \ac{ai} systems such as reasoning engines can provide output that is explainable but performs badly when it comes to handling large or noisy input.

Merging methods from the fields of symbolic and sub-symbolic \ac{ai} is the purpose of neural-symbolic integration. Its goal is to remedy the drawbacks of both approaches and combine their advantages by integrating methods of both fields. A first taxonomy for the types of those integrated systems was proposed by Henry Kautz at AAAI 2020 \cite{kautz_third_2020} and provides a quick survey on the kinds of systems in neural-symbolic integration:
\begin{itemize}
    \item Neural networks that create symbolic output from symbolic input, e.g., machine translation.
    \item Neural pattern recognition subroutines within a symbolic problem solver, e.g., the Monte-Carlo search in the core neural network of AlphaGo \cite{silver_mastering_2017}.
    \item Systems in which the neural and symbolic are plugged together and utilize the output of the other system(s), e.g., the neuro-symbolic concept learner \cite{mao_neuro-symbolic_2019} or a reinforcement agent working together with symbolic planners \cite{illanes_symbolic_2020}.
    \item Neural networks that have knowledge compiled into the network, e.g., if-then rules \cite{garcez_neural-symbolic_2008}.
    \item Symbolic logic rules embedded into a neural network that acts as a regularizer.
    \item Neural networks that are capable of symbolic reasoning such as theorem proving.
\end{itemize}

\subsubsection{Methodological Overview}
\noindent\textbf{Neural-symbolic methods for reasoning}. 
In \cite{lamb_graph_2021} multiple approaches were presented to integrate symbolic systems in \acp{gnn}.
\acp{gnn} allow for two major advantages in solving reasoning tasks. They apply an inductive bias directly through their architecture and offer permutation invariance because of their update and aggregation functions.
Permutation invariance simplifies the representation  of literals and clauses. Therefore the order of logical symbols does not impact the learning and understanding of such clauses. For example the \ac{gnn} handles the logical expression \((x_1 \lor \neg x_2 \lor x_3)\) semantically the same as the expression \((x_1 \lor x_3 \lor \neg x_2)\). \acp{gnn} enable visual scene understanding and reasoning superior to Convolutional Neural Networks as shown in \cite{santoro_simple_2017}.

Tensorization of first-order logic is another approach for solving reasoning tasks utilizing Deep Learning in combination with neural-symbolic integration. \acp{ltn} as presented in \cite{serafini_logic_2016} are able to use full first-order logic with function symbols by embedding these logic symbols into real-valued tensors. They propose a neural-symbolic formalism called Real Logic in addition to the computational model that is designed for defining logical expressions suited for tensorization in \acp{ltn}.

Real Logic is a many-valued, end-to-end differentiable first-order logic. It consists of sets of constant, functional, relational and variable symbols. Formulas build from these symbols can be partially true and therefore Real Logic includes fuzzy semantics. Constants, functions and predicates can also be of different types represented by domain symbols. The logic also includes connectives \( \diamond \in \{\neg\}, \circ \in \{\land, \lor, \rightarrow, \leftrightarrow \}\) and quantifiers \(Q \in \{\forall,  \exists\}\).
Semantically Real Logic interprets every constant, variable and term as a tensor of real values and every function and predicate as a real function or tensor operation.
Therefore Logic Tensor Networks are able to efficiently compute an approximate satisfiability by mapping logical expressions to real-valued tensors.

Moreover, \cite{badreddine_logic_2021} presents multiple related approaches that integrate logical reasoning and deep learning while being end-to-end differentiable:
\begin{itemize}
\item Logical Neural Networks \cite{riegel_logical_2020} use a logical language to define their architecture. By applying a weighted Real Logic a tree-structured neural network is built with different logical operators represented by different activation functions.
\item DeepProbLog \cite{manhaeve_deepproblog_2018} is a probabilistic logic programming language that implements a Neural Network capable of solving reasoning tasks by applying logical inference.
\end{itemize}

\noindent\textbf{Neural-symbolic architectures for context understanding.}
In \cite{oltramari_neuro-symbolic_2020} two applications for neural-symbolism are demonstrated and evaluated. The first application focuses on autonomous driving and uses Knowledge Graph Embedding Algorithms to translate Knowledge Graphs into a vector space. The Knowledge Graph is generated from the NuScenes dataset and consists of the given Scene Ontology with a formal definition of a scene and a subset of Features-of-Interests and events defined within a taxonomy. By creating the Knowledge Graph and the use of Knowledge Graph Embeddings it is possible to calculate the distances of scenes and to find similar situations that are visually different. Presented methods to create Knowledge Graph Embeddings are TransE, RESCAL and HoIE, where TransE shows the most consistent performance on the quantitative Knowledge Graph Embeddings-quality metrics.

\begin{figure}
  \includegraphics[width=\linewidth]{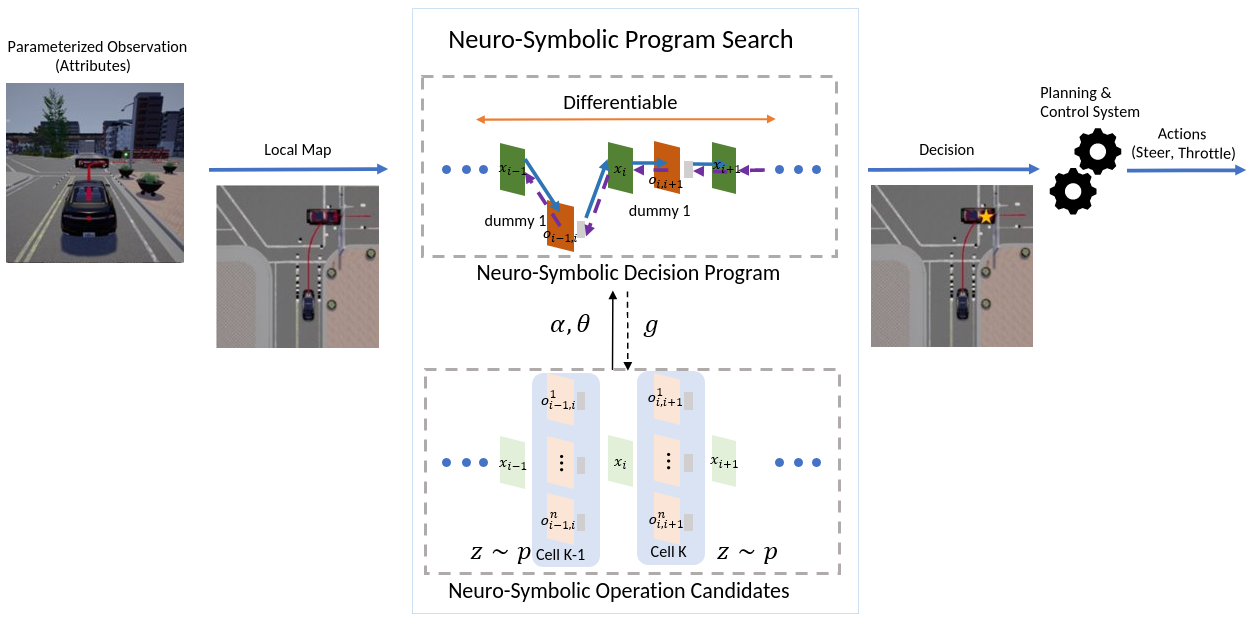}
  \caption{The \ac{nsps} searches The \ac{nsdp}, through back-propagation \textit{g} to update $\alpha$ and $\beta$. The decision policy generated by \ac{nsdp} is further used for planning and control. Semi-transparent operators shows the candidates for parent operations used in the search from \ac{nsps} \cite{sun_neuro-symbolic_2020}.}
  \label{fig:NSPS framework}
\end{figure}

The second application is "Neural Question-Answering" with knowledge integration using attention-based injection. The presented method uses knowledge from ConceptNet and ATOMIC and injects it into an Option Comparison Network by fusing the commonsense knowledge into BERT's output. It is evaluated with the CommonsenseQA dataset and the analysis suggests, that attention-based injection is preferable for knowledge injection.
\\\\
\noindent\textbf{Neural-Symbolic Program Search for Autonomous Driving Decision Module Design.}
In \cite{sun_neuro-symbolic_2020} \ac{nas} framework is proposed, which automatically synthesizes the \ac{nsdp} to improve the autonomous driving system design. \ac{nsps} synthesizes end-to-end differentiable \acp{nsp} by amalgamating neural-symbolic reasoning with representation learning. Symbolic representations of driving decisions are described with \ac{dsl} for autonomous driving. \ac{dsl} contains both basic primitives for parts with driving, along with conditional statements to enforce higher-level priors. The design of \ac{dsl} is allowed to specify all the behaviors for autonomous driving in a differentiable neural-symbolic behavior paradigm. Further \ac{nsps} is formulated as a stochastic optimization problem allowing to efficiently search for program architecture that integrates the neural-symbolic operations ensuring end-to-end learning possibilities. The \ac{nsps} in \autoref{fig:NSPS framework} is combined with \ac{gail} to learn in an end-to-end fashion to generate neural-symbolic decision programs to output specific instructions (e.g., target waypoint index, target velocity) to the motion planner and controller.
\\\\
\noindent\textbf{Integrating Prior Knowledge into Deep Learning.}
In \cite{diligenti_integrating_2017} a \ac{sbr} \cite{diligenti_semantic-based_2017} framework is used to express prior knowledge as set of \ac{fol} clauses. SBR is a statistical relational learning framework, holding the ability to learn from examples and logic rules. Partial definition of the mapping from input to output is provided through expressed \ac{fol} clauses. Statistical relational learning is employed to inject logical knowledge into  learning. It transforms logic knowledge into continuous constraints which are integrated with cost functions as a regularizer. Experimental analysis of the proposed work is focused on image classification problems. A subset of ImageNet dataset \cite{russakovsky_imagenet_2015} is used for the classification task.

\subsubsection{Applications in Perception}
\noindent\textbf{Reasoning.} 
A neural-symbolic reasoning engine could employ world knowledge or common sense knowledge to make sense of the scenery of a perception module or a combination of such modules. It could work as a regularizer (semantic loss function) during the training of a perception neural network which describes the scenery and penalizes implausible combinations of the recognized entities and their attributes or relations. Secondly, the same principle could be applicable during inference in which the perception module(s) outputs multiple possible scenery descriptions and the reasoning engine checks for contradictory elements in the scenery thus assessing the plausibility of the output.
\\\\
\noindent\textbf{Scene Consistency.}
As in \cite{oltramari_neuro-symbolic_2020}, concepts in labeled driving datasets along with domain experts can be utilized to model SceneOnotology representing spatial relational knowledge among the scene concepts. However, the ontology in \cite{oltramari_neuro-symbolic_2020} is used to generate triplets or facts exhibited in the scene. Vector representation of these facts is performed with \acp{kge}. The \ac{kge} framework allows translating the triplets to latent vector space for vertical integration of spatial knowledge with deep learning methods for downstream scene understanding task. Qualitative intrinsic evaluation of \acp{kge} for complete encoding of all knowledge facts is challenging and is subject to the different score functions utilized in \acp{kge} \cite{oltramari_neuro-symbolic_2020}. Therefore, one could use \ac{sbr} \cite{diligenti_integrating_2017} a statistical relational learning framework that can represent these facts as \ac{fol} clauses and transforms these clauses into continuous constraints. These constraints could be integrated into the cost functions as a regularizer term, allowing to find optimal parameters for the learning algorithm by restricting the solution space. The downstream deep learning task such as object detection could use this relational knowledge expressed with \ac{sbr} framework to improve the precision and accuracy by integrating the knowledge of the object's spatial-relational existence.

\subsubsection{Applications in Situation Interpretation}
\textbf{Neural-symbolic architectures for context understanding.}
Because the application of Knowledge Graph Embeddings in \cite{oltramari_neuro-symbolic_2020} is already used in the autonomous driving context, it is applicable as it is. Discovering similar situations that are visually different helps to understand certain scenarios and could be useful as an input to help classifying dangerous situations.

The attention-based injection approach can be useful for reasoning in certain situations, for example by injecting the \ac{stvo} into a \ac{qa} network to encode new sentences in BERT.
\\\\
\textbf{Behavior prediction of target vehicle.}
As presented in \cite{sun_neuro-symbolic_2020}, one could make use of \ac{dsl} for describing driving maneuvers in symbolic space. \ac{dsl} could be designed to contain \ac{stvo} logical rules along with primitive attributes (e.g, velocities, accelerations, pose, associate lane type, lane attributes, road typologies) to enforce higher-level priors for maneuvers. 

The \ac{nsps} framework proposed in \cite{sun_neuro-symbolic_2020} is a stochastic optimization problem. Joined with the mechanism of \ac{nas} it can be employed to generate decision policy for downstream motion control and planning tasks. Similarly, \acp{nsbp} are generated using the \ac{nsps} framework to reason about the target vehicles' behaviors to support cooperative planning in the scene. The synthesized \ac{nsbp} shall be an operation involving Neural-Symbolic operations (numerical operation and logical operations), rather than plain neural networks. The \ac{nsps} framework together with \ac{gail} could be used to learn \acp{nsbp} in an end-to-end fashion. 

\subsubsection{Applications in Planning}
\textbf{Reasoning.}
The creation of formalized knowledge requires a methodology capable of validating the resulting formalization. One such method is querying the formalization against test cases, e.g., check if the current formalization of the \ac{stvo} entails undesired properties such as it is possible to infer that endangering pedestrians in order to make way for an ambulance is ok. Those queries are also formalized statements and answered by a neural-symbolic reasoning engine that employs the formalized knowledge.

The formalization of legal knowledge is a prerequisite for checking compliance of an already taken or planned action for a certain traffic situation with regulations such as the \ac{stvo}. Neural-symbolic reasoners could perform such compliance checks enhancing two applications in the autonomous driving domain: Firstly, a planner could use the compliance check to assess several courses of action. Secondly, a compliance check could be employed as a regularizer during the training phase of a planner, forcing the model to prefer legally compliant solutions over non-compliant solutions.
\\\\

\subsection{Attention Mechanism}
\label{Knowledge_Integration/attention}
\textit{Author: Tianming Qiu}

\smallskip
\noindent
Human beings can focus on a specific area in fields of view or recent memories to avoid over-consuming energies. 
Inspired from the visual attention of human beings, an algorithmic attention mechanism becomes a popular concept in deep learning. 
\ac{nmt}~\cite{bahdanau_neural_2015}, a classical \ac{nlp} task, is one of the earliest successful attempts which apply attention mechanism.
Traditional \ac{nmt} approaches are based on a sequential encoder-decoder architecture which uses \ac{rnn}. 
The encoder maps source sentences word by word to hidden states and the decoder predicts target sentences. 
One of the drawbacks is that the longer the input sentence is, the more severe forgetting of previous words. 
The attention mechanism gives specific words (or tokens) more emphasis to avoid long distance forgettings.
Similar to \ac{nlp}'s attention concept, many machine learning tasks also require efficient focus on specific data or information.
Such specific focus comes from prior knowledge or experience which is very helpful for the objective task. 
Furthermore, this attentive information is usually intuitive for human understanding and it provides useful interpretability. 
For example, image captioning tasks look for heatmaps on input images which indicates where caption words refer to~\cite{xu_show_2015}.
If the attention mechanism is considered as a form of human knowledge, learning such semantic knowledge is expected to benefit networks performance.

In \ac{cv} tasks, the attention mechanisms are categorized into three different modeling approaches: \textit{spatial attention}, \textit{channel-wise attention}, and \textit{self-attention.}

\subsubsection{Spatial attention}
Spatial attention attempts to imitate how human beings are attracted by significant objects or features visually.
Technically, it emphasizes spatial areas in input images with highlighted heatmaps. 
The common spatial mechanism is written as
\begin{equation}
   \begin{split}
    \alpha_i &= f_{att}(\mathbf{v_i}), \\
    \mathbf{v_{i}^{\prime}} &= \mathbf{\alpha_i}\odot \mathbf{v_i},
\end{split} 
\end{equation}
where $\mathbf{v_i}$ represents a certain feature map of an input image, $f_{att}$ is a nonlinear mapping and $\odot$ represents Hadamard product, namely an element-wise product. 
A tiny two- or three-layer neural network is used to describe a nonlinear mapping of $f_{att}$, whose parameters are updated during training. 
The mask assigns different weights on the original feature map $\mathbf{v_i}$ by using Hadamard product so that it emphasizes information beneficial for following classification tasks and weaken less important features. 
These weighted masks on feature maps are scaled up to the original input image size and visualized by heatmaps to illustrate semantic image pixel-level attention.

The key point is to learn a nice attention function $f_{att}$ which generates a semantic attention heatmap.
Such an attention heatmap is integrated again into the neural network for improving final performances and provides semantic meaningful visualizations.
Similar to machine translation tasks, the attention on input original sentence words now switch to input image areas.
Similar works are seen in HydraPlus-Net~\cite{liu_hydraplus-net_2017}, which develops a complicated and huge neural network by duplicating Inception networks several times.
HydraPlus-Net is designed for pedestrian re-identification so it should be capable to detect detailed features on pedestrians.
All the above papers learn attention functions only by standard loss functions which only contain predicting losses for bounding box class and localization, but no predicting loss for attention heatmaps.
They design special structures but do not provide extra information for attention function training.
The only `guide' for attention learning comes from the loss functions.
Another approach to learn attention is to add an extra auxiliary loss function specifically for $f_{att}$ training~\cite{pang_mask-guided_2019, zhang_occluded_2018}.
In object detection tasks, datasets provide segmentation ground truth which is used to evaluate attention as well.
The loss function that measures overlaps between attention heatmap and ground-truth segmentation is used as a very strong guide to learn attention function~\cite{pang_mask-guided_2019}.
Another approach that leverages additional information to train attention networks is to use pre-trained attention layers from other tasks.
In a pedestrian detection task, such an additional dataset like MPII Pose Dataset~\cite{andriluka_2d_2014} which provides precise predictions of 14 human body key points demonstrates a good attention result on the primitive task~\cite{zhang_occluded_2018}.
Spatial attention is seen as a special feature representation. It learns the spatial knowledge from input images that different spatial areas have different impacts on the final outputs of neural networks.

\subsubsection{Channel-wise attention}
In computer vision tasks, channel-wise attention weighs channels of convolutional layers' outputs differently. 
Similar to the aforementioned spatial attention, channel-wise attention is still a probability mask. 
It assigns various weight values for each channel of output separately with the supervision of classification or detection outputs. 
Convolution layers are considered to be able to show the hierarchical nature of features~\cite{zeiler_visualizing_2014}.
Each convolutional kernel is assumed to represent a different feature extraction ability. 
Hence, channels of the output feature map behave differently to various image patterns.
Each channel may contain different features which might affect the final output.
Channel-wise attention was first used to aggregate information from the entire receptive field for involving more global information than local spatial information~\cite{hu_squeeze-and-excitation_2018}.
In pedestrian detection tasks,~\cite{zhang_occluded_2018} interprets \ac{cnn} channel features of a pedestrian detector visually and indicates that different channels activate response for different body parts respectively.
An attention mechanism across channels is employed to represent various body parts.
By emphasizing detected human body parts, occluded pedestrian detection results are improved.

\subsubsection{Self-attention}
Self-attention is widely used in \ac{nlp} because it is good at extracting the correlations between words.
The relationship between each word plays a significant role of text understanding.
Self-attention in \ac{cv} analyses the correlations between pixels and are formulated as a formal function of query $\mathbf{q}$, value $\mathbf{k}$ and key $\mathbf{v}$:
\begin{equation}
    \begin{split}
    \mathbb{R}^{d_k\times n_q}\times \mathbb{R}^{d_k\times n_k}\times \mathbb{R}^{d_v\times n_k} & \to  \mathbb{R}^{d_v\times n_q}, \\
    \mathbf{q,k,v} & \mapsto \text{Attention}(\mathbf{q,k,v}).
\end{split}
\end{equation}
Query $\mathbf{q}$, value $\mathbf{v}$ and key $\mathbf{k}$ concepts come from retrieval systems, where the best matched `value' should be returned according to a certain `query'.
Usually, query is first converted to keys that are connected to values.
Here query and key refer to the projected outputs of the decoder and encoder. Sometimes key and value are the same.
Attention computes for each query $\mathbf{q}$ an attention vector $\mathbf{a}_i$ by returning a weighted sum of all values, i.e.,
\begin{equation}
  \mathbf{a}_i = \sum_{j=0}^{n_k}\alpha_{i,j} \mathbf{v_j}.
\end{equation}
The weights are determined from some measurements of similarity between the queries and keys. Transformer architecture uses word relevance to improve translation performance~\cite{vaswani_attention_2017}.
Self-attention represents the image block or pixel relevance in computer vision tasks.
Apart from local features within each block, self-attention provides more global features~\cite{wang_non-local_2018, ramachandran_stand-alone_2019}.
Alternatively, each pixel in an image is seen as the query $\mathbf{q}$.
Self-attention of each query pixel is calculated on the other pixels in an image.
Compared with convolution layers, self-attention is also able to extract different levels of features at different layers.
Furthermore, due to its ability to extract global features, self-attention is able to achieve better performance than convolution in many tasks~\cite{cordonnier_relationship_2020}.
Self-attention mechanism and Transformer architecture are applied to many image-based detection tasks~\cite{carion_end--end_2020, zhu2021deformable} as well as 3D detection tasks~\cite{misra2021end}.

\subsubsection{Applications}
Attention mechanism in \ac{cv} is widely used in autonomous driving perception tasks such as pedestrian detection.
In Zhang's work~\cite{zhang_occluded_2018}, attention is integrated into the network to enhance the potential ability to find more occluded pedestrians.
Similarly, integrating attention heatmap to the existing detector backbone improves the detection results as well~\cite{pang_mask-guided_2019}.
Attention mechanism isn't used for planning directly, but it is used for interpretabilities of planning or decision making.
Works~\cite{kim_textual_2018, kim_interpretable_2017} from Berkeley Deep Drive use attention heatmap to explain why vehicle takes a certain controller behavior and textual explanations would be generated.
Attention is updated during training, meanwhile, it also affects the training results in the end. For scene understanding, it is not considered as a feasible method.
\subsection{Data Augmentation} \label{Knowledge_Integration/data_augmentation}
\textit{Authors: Stefan Matthes, Tobias Latka}

\smallskip
\noindent
Data augmentation comprises a number of techniques that increase the amount of data for little additional cost. It provides a way to integrate knowledge about how concrete changes in the input signal affect the model's target output, such as invariance to small perturbations. Training with the additional data usually improves the generalization of the model and can be especially helpful when data is scarce or imbalanced.

Which data augmentation technique can be used depends on the format of the input data (e.g., image, audio, point clouds) and the machine learning task. It is essential that the applied algorithm preserves task-relevant information. For example, color space distortions can be helpful in image-based license plate recognition (by making the model more robust to color changes), but can reduce performance in bird species classification, since color is an important distinguishing feature for many species. For some tasks, such as density estimation, it is inherently difficult to define appropriate data augmentations. On the other hand, data augmentation is even an integral part of some unsupervised models, for example in contrastive learning \cite{chen_simple_2020}.

In recent years, several surveys have been published on data augmentation \cite{wang2017effectiveness, shorten_survey_2019, wang2020survey, khosla2020enhancing, yang2022image}. \cite{wang2017effectiveness} and \cite{wang2020survey} review data augmentation methods for image recognition and face recognition, respectively, while Shorten and Khoshgoftaar \cite{shorten_survey_2019} provide a more general perspective and taxonomy of data augmentation techniques. Khosla and Saini \cite{khosla2020enhancing} focus on data warping and oversampling, and highlight how these techniques avoid overfitting. More recently, Yang et al. \cite{yang2022image} discuss data augmentation methods for common \ac{cv} tasks, including object detection, semantic segmentation and image classification based on experimental results. In this chapter, we look at data augmentation from the perspective of knowledge formalization and integration with applications in the field of autonomous driving.

\begin{figure*}
    \centering
    \includegraphics[width=0.7\linewidth]{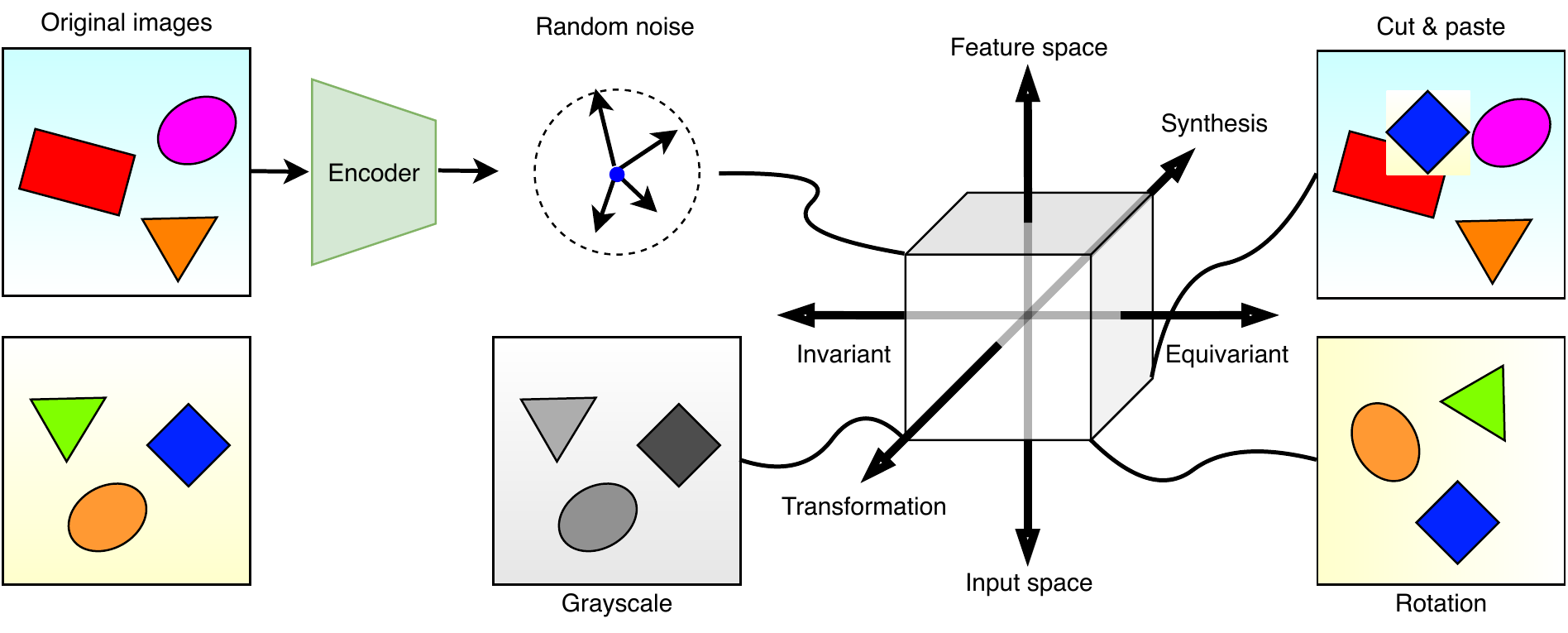}
    \caption{The dimensions of data augmentation.}\label{fig:data_aug}
\end{figure*}

Data augmentation methods can be categorized based on multiple criteria or factors (see Fig.~\ref{fig:data_aug}). First, they can leverage either invariances or equivariances in the data. The former modifies the input signal in a manner that does not affect the target, while the latter also changes the target based on certain known symmetries. Data transformation (manipulation or warping) techniques modify individual instances, whereas in data synthesis parts of two or more instances are recombined. Generative models, such as a \ac{gan} or \ac{ae}, which can be used to generate additional samples, can be seen as an extreme case of the synthesis approach. Finally, we distinguish between augmentations in data space and feature space. Some authors do not consider simulation as data augmentation, but since simulation is a useful tool for knowledge integration and plays a crucial role in autonomous driving, we will discuss it here as well.

\subsubsection{Invariance and Equivariance}

Many classical approaches using \acp{dnn} add random noise to the training data \cite{reed_regularization_1992}\cite{bishop_training_1995}, motivated by the fact that the learned function should be invariant to noise. Bishop \cite{bishop_training_1995} showed that applying small perturbations to the inputs during training leads to a smoothed target function and is equivalent to optimizing with an additional regularization term or constraining the weight updates (see also \cite{reed_regularization_1992} and \cite{niyogi_incorporating_1998}). However, it is unknown what the optimal noise distribution is.

A related technique is random erasing, for example, graying out pixels \cite{devries_improved_2017} and dropping words in text \cite{wei_eda_2019}. This is similar to dropout \cite{srivastava_dropout_2014} where instead network weights are masked with some probability in each optimization step.

For image data, the effect of many transformations is well studied. Typical image manipulations include geometric transformations, such as cropping, translations, rotations, reflections, and projections; kernel filters, e.g., sharpening and blurring; and color space transformations, such as random grayscale and color jitter \cite{shorten_survey_2019, niyogi_incorporating_1998}. While these transformations generally have no effect on the labels in image classification, in other tasks such as object detection and semantic segmentation, bounding boxes and segments must be modified equivalent to the input. Thus, how a transformation affects the target also depends on the task.

Permutations are another example of this. Sorting, for instance, is a permutation invariant task, while object tracking is permutation equivariant. However, some architectures, such as transformers~\cite{vaswani_attention_2017}, are inherently permutation equivariant and implement this type of knowledge much more efficiently. Changing the order of queries or keys affects the order of the output accordingly.

For an input target pair $(x,y) \in \mathcal{X} \times \mathcal{Y}$, we can formalize invariances and equivariances by $(x,y) \mapsto (g(x;\theta), y)$ and $(x,y) \mapsto (g(x;\theta), \tilde{g}(y;\theta))$, respectively, where $\theta$ denotes the type and strength of the applied transformations $g$, $\tilde{g}$ and is typically a random variable.

\subsubsection{Data Transformation and Synthesis}

Many important transformations for image data have already been mentioned in the previous section. Data types with other properties require different transformations. For instance, audio datasets can be enhanced using scale changes (pitch shifting and time stretching), compression, quantization, equalizing, filtering, reverberation and background noise injection \cite{mignot_analysis_2019}. Moreover, several of these elementary transformations can be combined in a myriad of ways.

A special class of data transformations are adversarial perturbations. These are slightly distorted inputs that lead to incorrect and usually overconfident predictions, but can often not be distinguished from the original by humans \cite{goodfellow_explaining_2014}. Adversarial training, i.e., feeding these examples back into the model, leads to more robust predictions \cite{goodfellow_explaining_2014}. Miyato et al. \cite{miyato_virtual_2018} extend this procedure to the semi-supervised setting by computing the adversarial examples using the model's predictions instead of ground truth labels.


In addition to modifying individual instances, new data can be synthesized by combining elements from multiple data points. One of the early approaches is SMOTE \cite{chawla_smote_2002}. It was developed for imbalanced datasets and can be used to oversample underpopulated classes by interpolating between nearest neighbors from the same class. Mixup \cite{zhang_mixup_2017} and SamplePairing \cite{inoue_data_2018} explore the same technique for image data. The former also interpolates the labels accordingly and uses soft labels, which however cannot be used in the semi-supervised setting.

Another common technique for image datasets is to cut and paste patches from different images \cite{dwibedi_cut_2017, dvornik_modeling_2018, fang_instaboost_2019, georgakis_synthesizing_2017, yun_cutmix_2019}. To avoid that the model cheats by detecting artifacts at the boundary of the inserted patches, various blending techniques and distractors (patches that do not contain any of the relevant objects) can be used \cite{dwibedi_cut_2017}. Instead of inserting objects randomly, several techniques were proposed for more realistic object placements, such as using a visual context model \cite{dvornik_modeling_2018}, depth and semantic information \cite{georgakis_synthesizing_2017}, and a heat map for appearance consistency \cite{fang_instaboost_2019}. YOLOv4 \cite{bochkovskiy_yolov4_2020}, a 2D object detector, additionally uses mosaic data augmentation that concatenates multiple images before cropping. This improves the detection of smaller objects.

The previous formulas for invariance and equivariance can be generalized to the case when new data is synthesized from multiple instances: $\{(x_i,y)\}_{1}^{n} \mapsto (g(x_1, \dots, x_n;\theta), y)$ and $\{(x_i,y_i)\}_{1}^{n} \mapsto \left(g(x_1, \dots, x_n;\theta), \tilde{g}(y_1, \dots, y_n;\theta)\right)$.

A more elaborate approach to create additional data is to first train a generative model with the given data and then sample from it. Neural Style Transfer \cite{gatys_image_2016}\cite{meshry_neural_2019} is a technique that can be used to change the appearance of an image while leaving the content unaffected. It has mainly artistic applications, but can also be used to render images with the appearance of different seasons, times of day, and different weather conditions \cite{meshry_neural_2019}. A major drawback is that these models already require large amounts of training data and may take a long time to sample.

\subsubsection{Data Augmentation in Feature Space}

The methods described so far directly modify the raw data, but it is also possible to augment data in the feature space. In the latter, the input data is fed through the first layers of the \ac{dnn} and then the intermediate representations are manipulated before being passed through the remaining layers.

New instances can be synthesized either by interpolation, extrapolation or simply by adding noise \cite{devries_dataset_2017}. Similar to Mixup \cite{zhang_mixup_2017}, Manifold Mixup \cite{verma_manifold_2019} additionally interpolates between points from different classes by also interpolating the labels accordingly, which can therefore be considered its natural extension. Alternatively, an \ac{ae} can be used to transform the modified features back into the input space \cite{devries_dataset_2017}, but unlike the other approaches, this already requires a trained decoder. These methods have the advantage of being domain agnostic. However, experiments by Wong et al. \cite{wong_understanding_2016} suggest that augmentations in the data space, when applicable, are preferable to data augmentation in feature space alone.

\subsubsection{Automatic Data Augmentation}

Towards automating the machine learning pipeline, Cubuk et al. \cite{cubuk_autoaugment_2019} applied a reinforcement learning approach to search the space of augmentations. The learned policies specify the order and strength of predefined operations, including geometric transformations, photometric transformations, kernel filters, as well as Cutout \cite{devries_improved_2017} and SamplePairing \cite{inoue_data_2018}. Since then, several variations and extensions have been developed (see \cite{shorten_survey_2019, yang2022image} for a review). In contrast to the previous approaches, Benton et al. \cite{benton_learning_2020} directly optimize distributions over augmentations with respect to the training loss.

\subsubsection{Simulation}

Simulations provide another way to generate large amounts of data at low cost, which is especially useful for data hungry models like \acp{dnn}. They enable the generation of more data for interesting situations or rarely occurring events, which is often difficult or infeasible in the real world for financial or moral reasons. Additionally, they offer the possibility to evaluate safety-critical systems in specific test scenarios.

For the development of machine learning models, simulation results are predominantly used in the natural sciences, e.g., thermodynamics, material sciences, and autonomous driving \cite{von_rueden_informed_2021}. There exists a plethora of open source and commercial simulators for the development and benchmarking of \acp{sdv} \cite{dosovitskiy_carla_2017}\cite{richter_playing_2017}\cite{bernhard_bark_2020}. A recent overview of their configuration options and available sensors can be found in \cite{rosique_systematic_2019}. Besides data generation, there are other ways to combine simulations and machine learning models. We refer the interested reader to a recent overview \cite{von_rueden_combining_2020}.

One of the biggest hurdles in transferring the trained models to the real world is the domain shift (distribution change) between simulation and reality. Even highly accurate models in simulation can perform poorly on real data if the data distributions differ too much. Therefore, the models often have to be additionally fine-tuned on real data. Apart from developing more accurate models, there are two approaches to bridge the gap, domain randomization and domain adaptation.

The idea of domain randomization is that given enough variability in the virtual environment, the model may interpret the real world as just another variation. For example, in the context of grasping experiments with a robotic arm, Tobin et al. \cite{tobin_domain_2017} demonstrated that randomizing the rendering of images improves the transferability from simulation to hardware. Interestingly, they found that designing the simulation to be as realistic as possible was less effective than varying the styles.

Domain adaptation is a type of transfer learning that leverages labeled data in one or more related source domains for prediction in a target domain. Two recent surveys with applications in computer vision can be found in \cite{csurka_comprehensive_2017} and \cite{wang_deep_2018}. The latter has a stronger focus on deep learning models.

\subsubsection{\aclp{scm} for Data Augmentation}

\acp{scm} encode knowledge about an environment \cite{pearl_seven_2019}. In that respect, they can be thought of as the data generating process. Mathematically, an \ac{scm} is nothing else than a \ac{dag} equipped with both a set of functions and a distribution on the \ac{dag}'s root vertices. While the \ac{dag}'s vertices correspond to the variables of the environment, its directed edges represent independent causal mechanisms between variables. In particular, the causal mechanisms describe how variables affect one another in a deterministic manner. Thus, every \ac{scm} naturally defines a joint distribution on its variables. Thereby, its shape is determined by the set of functions and by the distribution of the \ac{scm}'s root variables. Moreover, distribution shifts can be modeled in the \ac{scm}-framework as interventions, for instance, exchanging one function for another one.

As discussed, for instance, in \cite{scholkopf_toward_2021}, \acp{scm} are perfectly suited to generate valid and consistent samples at will in the sense that they are consistent with causal relations encoded in the \ac{scm}. In this way, \acp{scm} serve as a kind of lightweight simulator of the underlying environment leading to different interventional distributions depending on the concrete set of interventions. To be more precise, an arbitrary training set can be thought of as being composed of individual distributions. These individual distributions can either represent the distribution defined by the unmodified \ac{scm} or originate from different interventions applied to the original \ac{scm} at the time data is being generated. Hence, interventions effectively modify the environment and cover reasonable variations of the environment. In this way, sampling data from different joint and interventional distributions (as constructed from the original \ac{scm}) naturally increases the diversity of the overall training distribution and can be interpreted as some kind of data augmentation.

\subsubsection{Applications}

Data augmentation can be used in all stages of autonomous driving. Especially in stack-based architectures and end-to-end approaches that compute interpretable intermediate representations, data augmentation can be used in a variety of ways.

%
The early end-to-end approach from Bojarski et al. \cite{bojarski_end_2016} learns steering commands directly from monocular image data and emulates the \ac{sdv} at various displacements from the center of the lane and angles to the direction of the road. They extend the dataset by transforming the viewpoint of the images using two additional forward-facing side cameras and adjusting the steering angle accordingly. This results in a more robust driving model that can recover more effectively from adverse situations. Photometric transformations, kernel filters, noise injection, and various other augmentation techniques that do not affect the control commands can also be applied at this stage \cite{codevilla_end--end_2018}.

Many of the techniques used in image-based object detection were already discussed above. For 3D object detection from point clouds, it is common to randomly shift, rotate, flip and scale each ground truth bounding box and its associated points \cite{zhou_voxelnet_2018}\cite{shi_pointrcnn_2019}\cite{yan_second_2018}. Except for translations, these can also be applied to the point cloud as a whole. Yan et al. \cite{yan_second_2018} additionally synthesize new point clouds by inserting points belonging to bounding boxes from different scans. Implausible outcomes are avoided by performing collision tests.

In multimodal object detection, additional care must be taken to ensure that augmentations do not cause inconsistencies between data streams such as pasting objects at implausible locations. By performing occlusion and collision tests the cut and paste augmentation can be extended to image data for multi-modal object detection \cite{zhang_multi-modality_2020}.

Many recent approaches to object detection, semantic segmentation, and related tasks incorporate additional synthetic data from virtual environments, such as from the game Grand Theft Auto V \cite{richter_playing_2017}\cite{wu_squeezeseg_2018} and the SYNTHIA dataset \cite{ros_synthia_2016}. Integrating synthetic data generally leads to better performance, but the gains level off above a certain ratio. In addition, photorealism plays a smaller role than realistic modeling of sensor distortions and environmental distribution.

%
Several approaches advocate a bird's eye view, also known as plan view, as an intermediate representation for subsequent motion prediction, trajectory planning, and control \cite{bansal_chauffeurnet_2018}\cite{wang_monocular_2019}\cite{chen_learning_2020}. A common approach is to render detected objects and information about the environment into a multi-channel image, which is then processed by a \ac{cnn}. The bird's eye view image can be augmented with geometric transformations such as random translations and rotations \cite{chen_learning_2020}.

As yet another example of data augmentation in the context of autonomous driving, imagine, for instance, an \ac{scm} that describes the vehicle trajectories (i.e., the physical laws connecting vehicle states and actions to new states). New trajectories of the vehicle’s motion can be generated from existing ones by means of such a vehicle \ac{scm} following the subsequent steps. First, values of the external (and usually) unobserved random variables are inferred from existing trajectories (known as the abduction step) effectively reconstructing the situation the vehicle was in when the observed trajectory was recorded. Second, an intervention or sequence of interventions (actions) are applied to the vehicle \ac{scm}, while sustaining the updated distribution from the abduction step. Third, the (intervened upon) \ac{scm} predicts a new trajectory that is grounded in the observed trajectory. The procedure just described returns a so-called counterfactual trajectory that could have evolved, if another sequence of actions had been taken. In this sense, this technique transforms an observed trajectory into a counterfactual one, while complying with the rest of the \ac{scm} that was not intervened upon (data transformation). Thus, the technique allows to augment data in a way that it is still anchored in an observed trajectory, but can, at the same time, be used to generate more data, especially covering hazardous and underrepresented scenarios. Moreover, this technique was shown to be useful for explaining the causes of \acf{ml}-model decisions, as discussed, for example, in \cite{deletang_causal_2021} and thus assists in situation understanding.

%
Bansal et al. \cite{bansal_chauffeurnet_2018} increase the diversity of vehicle trajectories by adding random perturbations. The perturbations are chosen such that the vehicle is brought back to its original trajectory after a perturbation. However, too strong distortions degrade performance as the model learns bad behavior.

They also employ an augmentation technique called past motion dropout \cite{bansal_chauffeurnet_2018}. Because the model is provided with past ego-motion from expert demonstrations, it can learn to exploit cues in the motion history rather than learning the underlying causes of such behavior, such as stopping at a stop sign because it sees a deceleration. Randomly dropping the past motion forces the network to look for other signals to explain the future trajectory.

TrafficSim \cite{suo_trafficsim_2021} learns to simulate multi-agent behaviors which can be used as effective data augmentation for training better motion planner. The learned driving model is fed into other vehicles in the simulation to produce more realistic behavior.
\subsection{State Space Models}
\label{sec:state_space_models}
\textit{Author: J\"{o}rg Reichardt}

\smallskip
\noindent
Driving is an inherently sequential dynamic activity. Sensory information is acquired through a stream of observations that exhibits causal dependencies and correlations across a very wide range of time scales.

As a passive observer, our ability to understand any dynamic phenomenon depends on our ability to predict future observations $\mathbf{o}_{t..t+\Delta t}$ from past observations $\mathbf{o}_{0..t}$ \emph{in probability} \cite{shalizi_computational_2001}. We stress \emph{in probability} to highlight that possible inherent randomness of a dynamic process can only be understood and modeled in terms of distributions, not individual outcomes \cite{crutchfield_regularities_2003}. Our central object of interest is thus the distribution over future observations given past observations
\begin{equation}
p(\mathbf{o}_{t+\Delta t} | \mathbf{o}_{0..t}). 
\end{equation}
We denote all time points from start time $0$ to time $t>0$ with the indices $0..t$ and further assume $\Delta t>0$. If we are taking an active role in the dynamics of the system, then future observations will naturally also depend on our past and future actions $\mathbf{a}_{0..t+\Delta t}$ and our central object of interest becomes
\begin{equation}
p(\mathbf{o}_{t+\Delta t} | \mathbf{a}_{0..t+\Delta t}, \mathbf{o}_{0..t}).
\end{equation}
Equipped with this distribution, we are able to optimize/plan our actions to increase the chances of a desired future outcome and observation.

These two distributions are intimately related and \acp{ssm} are the most commonly applied tool for their modeling. Firmly rooted in probability theory, state space models are uniquely suited for the study of the steady stream of information that originates from traffic phenomena. They permeate all aspects of driving from perception to situation understanding and planning \cite{thrun_probabilistic_2005}.

In the context of knowledge integration, \acp{ssm} represent an \emph{algorithmic prior} \cite{jonschkowski_differentiable_2018}. They provide a scaffold of probability distributions and their corresponding conditional independence structure. Data driven methods can then be used to learn parameterizations for these probability distributions. Ideally, optimization and learning can then be achieved end to end.

We will next review the fundamental terminology and assumptions of state space models that are central to their understanding in the modeling of dynamical systems and control. With this terminology clarified, we will then focus on the peculiarities of applying \acp{ssm} to traffic and autonomous driving and review the possibilities of creating hybrid learning systems within the framework provided by \acp{ssm}. 

State space models introduce a latent dynamical random variable $\mathbf{x}_{t}$ called the system's 
\emph{state}. The system's state governs the system dynamics, but is in general not directly accessible to an observer, i.e., state information has to be inferred. The system state has two defining qualities: First, it gives rise to the system dynamics via a Markov Process:
\begin{equation}
p(\mathbf{x}_{t+\Delta t} | \mathbf{a}_{0..t+\Delta t}, \mathbf{x}_{0..t}) = p(\mathbf{x}_{t+\Delta t} | \mathbf{a}_{t..t+\Delta t}, \mathbf{x}_t),
\end{equation}
i.e., only the most recent state and future actions matter for the future evolution of the system. The relation $p(\mathbf{x}_{t+\Delta t} | \mathbf{a}_{t..t+\Delta t}, \mathbf{x}_t)$ is causal and is called the \emph{motion model} of the system. All prior knowledge about the system dynamics may enter in this model.

Second, the current state and only the current state gives rise to observations via measurements:
\begin{equation}
p(\mathbf{o}_t | \mathbf{x}_t).
\end{equation}
This relation is causal \cite{pearl_causality_2000} and is called \emph{observation model} or \emph{measurement model}. Measurements do not change the state. Measurements can only reduce our uncertainty about the state through Bayes' theorem in which the observation model plays the role of the \emph{observation likelihood}. All knowledge about the measurement process such as measurement noise or sensor transfer functions are captured in the observation model.

\begin{figure}
    \includegraphics[width=\textwidth/2]{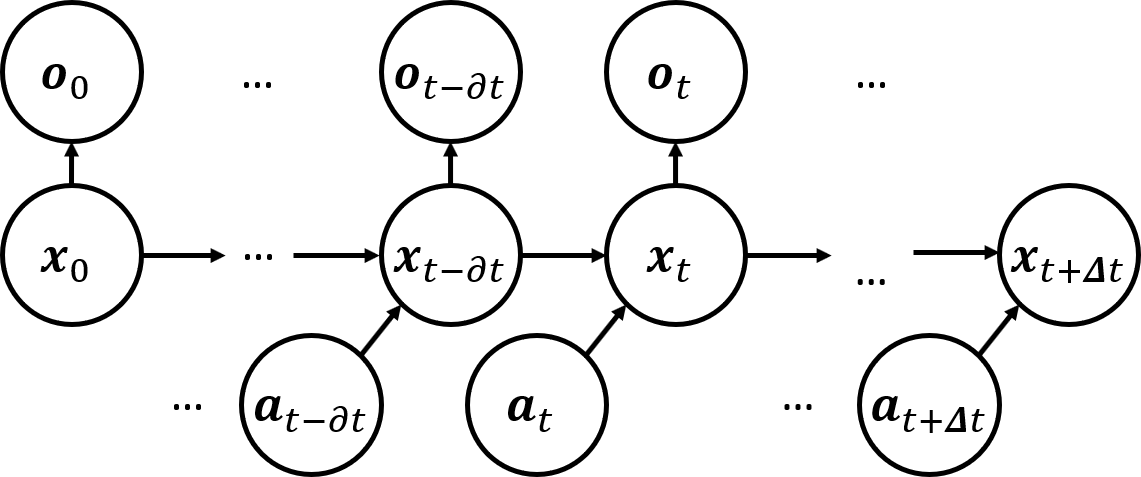}
    \caption{Conditional independence relations represented as graphical model}
    \label{fig:state_space_graphical_model}
\end{figure}

From these two qualities, it follows that the state renders past and future observations and actions conditionally independent through d-separation \cite{pearl_causality_2000}:
\begin{equation}
p(\mathbf{o}_{t+\Delta t} | \mathbf{a}_{0..t+\Delta t},\mathbf{x}_t, \mathbf{o}_{0..t}) = p(\mathbf{o}_{t+\Delta t} | \mathbf{a}_{t..t+\Delta t}, \mathbf{x}_t).
\end{equation}
This means the state provides as much information about future observations as all past observations and past actions combined. If we know the state, we can forget about all past observations and actions taken. Note that the state is not meant to provide a good reconstruction of past observations and actions - it only extracts the information from past observations that is necessary to predict future observations. The state is all we need to make optimal predictions about the future and to plan our actions \cite{shalizi_computational_2001}.
\autoref{fig:state_space_graphical_model} illustrates the conditional independence relations discussed above in the form of a graphical model.


With the state having such formidable quantities, the central object of interest for state space models becomes the \emph{posterior state density}
\begin{equation}
p(\mathbf{x}_t|\mathbf{a}_{0..t}, \mathbf{o}_{0..t}),
\end{equation}
which can be obtained through the application of Bayes' Theorem in a recursive manner from earlier state density, as given in Eq.~\eqref{eq:recursice_state_spaces},
\begin{figure*}
    \begin{equation}
    p(\mathbf{x}_{t+\Delta t} | \mathbf{a}_{0..t+\Delta t}, \mathbf{o}_{0..t + \Delta t}) = 
    \frac{p(\mathbf{o}_{t+\Delta t} | \mathbf{x}_{t + \Delta t})}{p(\mathbf{o}_{0..t+\Delta t})}\int d\mathbf{x}_t p(\mathbf{x}_{t + \Delta t} |\mathbf{a}_{t..t+\Delta t}, \mathbf{x}_t)p(\mathbf{x}_t | \mathbf{a}_{0..t}, \mathbf{o}_{0..t})
    \label{eq:recursice_state_spaces}
    \end{equation}
\end{figure*}
where we have used that only the state $\mathbf{x}_t$ can give rise to observations at time $t$ and the defining quantity of the state as the sole generator of system dynamics.

The quantity
\begin{equation}
p(\mathbf{x}_{t+\Delta t}|\mathbf{o}_{0..t}) = \int d\mathbf{x}_t p(\mathbf{x}_{t + \Delta t} | \mathbf{a}_{t..t+\Delta t}, \mathbf{x}_t)p(\mathbf{x}_t | \mathbf{a}_{0..t}, \mathbf{o}_{0..t}),
\end{equation}
is called the \emph{predictive state density} and we note the role of the motion model in this expression.

The continued update of a state estimate over time is called \emph{tracking} or \emph{Bayesian Filtering}. In the construction of the posterior state distribution the observation model plays the role of the likelihood term and the predictive state distribution that of a prior. The \emph{evidence} term $p(\mathbf{o}_{0..t+\Delta t})$ acts as a normalizing factor for the posterior state density. The Bayes Filter is a generative model for the observations.

It is this sequential process of absorption of current evidence into a latent state representation via a sound probabilistic framework that makes state space models extremely appealing from a conceptual point of view \cite{law_data_2015, thrun_probabilistic_2005}. We will next discuss under what conditions this conceptual appeal can be translated into computationally efficient algorithms and at what point approximations are introduced to obtain computational efficiency.

The first condition is that $p(\mathbf{x}_t|\mathbf{a}_{0..t}, \mathbf{o}_{0..t})$ can be normalized, i.e., we are able to evaluate $p(\mathbf{o}_{0..t})$. For all but the simplest kinds of observation spaces, this is hopeless to do exactly. One option is to resort to a sample / histogram based approach \cite{jonschkowski_end--end_2016}. Alternatively, one can make use of an \emph{assumed density} for $p(\mathbf{x}_t|\mathbf{o}_{o..t})$, i.e., via a parametric distribution for which the normalization constant can be computed from a set of sufficient statistics. The most common example of such an assumed density is the multivariate Normal. This also fixes the state representation $\mathbf{x}_t\in\mathbb{R}^n$ for which we have made no restrictions so far. The canonical example of such a \emph{state vector} is the kinematic state vector describing an object's position, velocity and acceleration \cite{thrun_probabilistic_2005}. The second condition is that we can perform the integration necessary to produce the predictive state distribution. Variational methods \cite{opper_advanced_2001} and  Monte Carlo methods are a viable option here in particular if they can be executed on highly parallel hardware \cite{jonschkowski_differentiable_2018}. If an assumption was already made on the form of the posterior state distribution, then it is natural to make the same assumption on the predictive state distribution. It then depends on the form of the motion model $p(\mathbf{x}_{t+\Delta t}|\mathbf{a}_{t..t+\Delta t}, \mathbf{x}_t)$ whether this is exact. If the motion model is also a multivariate Normal with $\mathbf{x}_{t + \Delta t}$ being a linear function of $\mathbf{x}_t$ and $\mathbf{a}_{t..t+\Delta t}$, then no approximation is introduced at all. Under assumed distributions for the state, it is further desirable that the observation model $p(\mathbf{o}_{t + \Delta t}|\mathbf{x}_{t + \Delta t})$ be conjugate to the predictive state distribution. Then condition, the posterior can be updated in closed form \cite{garcia-fernandez_poisson_2018, granstrom_poisson_2019}.

Depending on the type of approximations used, the update equations for the posterior state density are commonly referred to as \emph{Kalman Filter} for $p(\mathbf{x}_t|\mathbf{o}_{0..t})\sim\mathcal{N}(\mathbf{x}_t;\hat{\mathbf{x}}_t, \boldsymbol{\Sigma}_t)$ with linear Gaussian motion $p(\mathbf{x}_{t+\Delta t}|\mathbf{x}_t)\sim\mathcal{N}(\mathbf{x}_{t+\Delta  t};\hat{\mathbf{x}}_{t+\Delta t}=\mathbf{F}(\Delta t)\mathbf{x}_t, \mathbf{Q})$ and observation models $p(\mathbf{o}_t|\mathbf{x}_t) = \mathcal{N}(\mathbf{o}_t; \hat{\mathbf{o}}_t=\mathbf{H}\mathbf{x}_t, \mathbf{R})$ \cite{kalman_new_1960}. The co-variance matrices $\mathbf{Q}$ and $\mathbf{R}$ herein model the so-called \emph{process noise} and \emph{observation noise}, respectively. They have to account for errors both introduced by the assumption of linearity for dynamics and observation process as well as for actual noise and measurement uncertainty and can be estimated and tuned from data \cite{chen_kalman_2019, abbeel_discriminative_2005}. If these modeling assumptions are fulfilled, the Kalman filter update equations provide an exact closed form solution to the state estimation problem. In case the means of motion model and/or observation model are non-linear functions $\hat{\mathbf{x}}_{t+\Delta t} = \mathbf{F}(\mathbf{x}_t, \Delta t)$, $\mathbf{o}_t = \mathbf{H}(\mathbf{x}_t)$ one can linearize around the mean of $\mathbf{x}_t$ to obtain the \emph{Extended Kalman Filter}. The \emph{Unscented Kalman Filter} instead uses integration by Quadrature, i.e., nonlinear functions $\mathbf{F}$ and $\mathbf{H}$ are evaluated at at set of judiciously chosen \emph{sigma points} and the results are combined into a weighted average to give $\hat{\mathbf{x}}_{t+\Delta t}$ and $\hat{\boldsymbol{\Sigma}}_{t+\Delta t}$ \cite{wan_unscented_2000, haykin_kalman_2001}. Giving up on the assumption of a Gaussian state distribution, one models the state distribution as a population of sample/particles and perform prediction and update step in a Monte Carlo fashion. This approach is known as a \emph{Particle Filter} \cite{godsill_particle_2019}.

The above relations and algorithms form the basis for much of the classic model based study of dynamical systems and signal processing. We will next discuss the requirements for their application in an autonomous driving application. We will discover a wide range of opportunities for the application of data driven learning algorithms while still following the general scaffold of Bayesian data assimilation described above.

Fundamental to the specific aspects of perception, situation interpretation and planning is the substrate on which they operate: the state space. Hence, it will be discussed first.

Traffic is a \emph{multi agent} phenomenon \cite{risser_behavior_1985}. Traffic participants are very diverse - we observe fast cars, slow cars, small cars, large trucks, cable-cars, motorcycles, bicyclists, pedestrians, roller skaters, scooter drivers and - depending on country - horses, cows or moose on the road. Traffic participants are not particles, but decision making goal driven agents that are bound to move according to the laws of physics as well as by the rules of the road - though the latter being obeyed to a lesser extent in general. Goals and intentions of traffic participants are not measurable. They are either actively signaled by an agent in an explicit manner, e.g., through turn signals, or must be inferred from the agents' dynamical behavior. Traffic participants interact with each other to avoid collisions and operate in a structured dynamic environment, i.e., they follow roads and lanes or sidewalks or react to stop lights.

Hence, a state variable must be able to represent a \emph{varying number} of \emph{diverse} traffic participants at any given moment in time. If the state is to be the sole generator of system dynamics, it must include components for signaled/inferred \emph{goals and intentions} of traffic participants. It must further include components that model the \emph{environment} to the extent it is necessary to make predictions about their future movements \cite{khosla_adaptive_2003, hasberg_integrating_2008, lundquist_joint_2011, wang_simultaneous_2007}. The multi agent nature of traffic further requires that distributions over states are invariant under a permutation of objects/traffic participants \cite{zaheer_deep_2017}.

In order to cope with these requirements, two classical approaches exist. Multi-Object tracking algorithms \cite{luo_multiple_2021} abandon the representation of state as a single vector and instead use a random finite set of state vectors for individual objects \cite{vo_labeled_2014, granstrom_poisson_2019}. These representations are again fully probabilistic and even treat the number of objects in the set as a random variable. Alternatively, one can keep a fixed size state vector by rendering state information into a fixed size grid using the always available positional information of objects for positional encoding, i.e., to specify the grid cells \cite{bansal_chauffeurnet_2018, deo_convolutional_2018}. All available features are then stored in a dimension perpendicular to the grid. In essence, this amounts to saving state information in a multi-channel image and unlocks the ability to use (convolutional) neural network architectures on the state variables at the expense of a vastly increased state space dimension.

\subsubsection{Applications in Perception}

The perception stage of an autonomous driving stack enters \ac{ssm} through the generative observation model or observation likelihood $p(\mathbf{o}_t|\mathbf{x}_t)$. In principle, the observations $\mathbf{o}_t$ could be raw sensory input such as camera images or lidar points, but that is currently not computationally feasible without strong approximations \cite{ko_gp-bayesfilters_2009}. One alternative is to drop the generative nature of the observation model and learn to estimate the posterior state density $p(x_t|\mathbf{a}_{0..t}, \mathbf{o}_{0..t})$ directly, this approach is know as a discriminative Filter \cite{haarnoja_backprop_2016}. Note, however, that this approach is practical only for fixed history lengths and thus sacrifices the ability of the standard formulation to represent correlations in time of arbitrary length.

The second alternative is to pre-process raw sensor readings via detection algorithms to yield observations that correspond to object-level data \cite{yu_tracking_2016}. This corresponds to the "tracking from detection" paradigm. One can further differentiate if an object can give rise to at most one detection ("point detection algorithms") or multiple detections. If the latter is not an artifact, but a result of multiple sensor readings on an object's physical extension, so-called "extended object tracking" algorithms result \cite{granstrom_extended_2016, xia_extended_2020}. If detections are available from every available sensor modality, the observation likelihood is a natural place to perform sensor fusion. Alternatively, sensor fusion is performed prior to the application of a detection algorithm on the raw sensor level.

In correspondence to the set of objects we observe in traffic, detection algorithms will return a set of detections. Detection algorithms, however, are not perfect. There can be false detections, so called clutter, that do not arise from actual objects. There can also be objects that do not give rise to detections at the current moment due to occlusions or failures of the detection algorithm. Further, detection are not necessarily labeled, i.e., there is no known correspondence between a tracked object and its detection. From this arises the so called \emph{data association problem} of multi object tracking: the need find this very correspondence of the elements of the set of observations with the elements of the set of objects tracked. Standard Algorithms exist to solve it \cite{date_gpu-accelerated_2016, miller_optimizing_1997}. Once this correspondence is established and the posterior state estimates of the tracked objects have been updated, any errors made in updating the state vector of an object with the wrong detections cannot be recovered. In order to mitigate this problem, so-called multi-hypothesis tracking algorithms \cite{reid_algorithm_1979} maintain several plausible potential data association hypotheses until possible uncertainties in data association are resolved by additional evidence \cite{garcia-fernandez_poisson_2018}.

Two more aspects of \acp{ssm} pertain to the perception module that allow for knowledge integration. The first is the so-called \emph{birth model}, i.e., the prior distribution for the state vector for new objects $p(\mathbf{x}_0)$ that is needed for the state estimation from an object's first detection via $p(\mathbf{o}_0|\mathbf{x}_0)$. The birth model can represent the sensitivities of the sensors as well as prior knowledge about where and how objects will enter the sensor range of an autonomous vehicle. The second are the probabilities of detection $P_D(\mathbf{x})$, survival $P_S(\mathbf{x})$ and the clutter intensity $P_C(\mathbf{o})$ that can add further specify the observation model \cite{garcia-fernandez_poisson_2018}. With $P_D(\mathbf{x})$ we are able to specify that an object, though present, currently cannot be detected, e.g. due to occlusions \cite{granstrom_extended_2016}. With $P_S(\mathbf{x})$ we can express our prior knowledge about how objects leave the sensor range. For example, we can forget about oncoming traffic immediately once it has left the sensor range, while objects that have left the sensor range in the forward direction have a much higher probability to be re-encountered. Finally, with $P_C(\mathbf{x})$ we can model prior knowledge about the accuracy of the detection algorithms used.

\subsubsection{Applications in Situation Interpretation}

In the previous section, we have decidedly spoken about objects in order to address the handling of both traffic participants and elements of the environment such as roads or intersections. It is part of the appeal of \acp{ssm} that they can be used to model both the moving traffic participants as well as the static environment as seen from a moving sensor.

Situation interpretation primarily consists of two key tasks, the problem of estimation and tracking of the \emph{current} state of the system $\mathbf{x}_t$ from past observations $\mathbf{o}_{0..t}$ and actions $\mathbf{a}_{0..t}$, i.e., the state tracking and filtering problem. Specifically, this entails the mapping and localization problems, i.e., modeling the static environment from observations and referencing the ego vehicle and other traffic participants in this environment \cite{wang_simultaneous_2007}. It is important to note that the state update equations are evaluated at the rate at which new observations are available, typically $\Delta t\leq50$ms and thus the motion model is used on very short prediction horizons. Hence, one generally uses simple kinematic motion models \cite{schubert_comparison_2008}. Model uncertainty can then be adequately modeled as random noise and data can be used to tune the noise distribution \cite{chen_kalman_2019, abbeel_discriminative_2005}. In particular, in multi object tracking, one may assume independence between the motion of individual agents and neglect the interactions with both the environment and other traffic participants. This simplification results in growing errors as the time frame without new observations, e.g., due to occlusion, grows.  

\begin{figure}
    \centering
    \begin{subfigure}{\columnwidth}
        \includegraphics[width=\columnwidth]{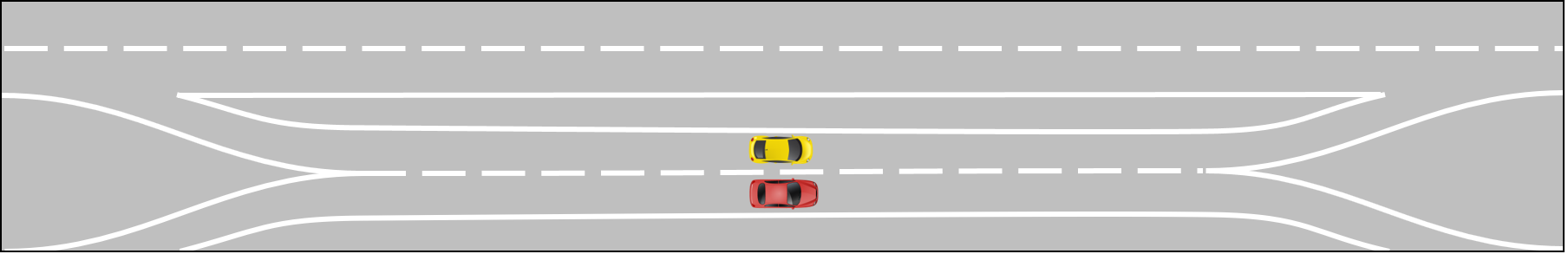}
        \caption{Two cars driving on a highway on-off ramp.}
        \label{fig:state_space_on_off_ramp1}
    \end{subfigure}
    \begin{subfigure}{\columnwidth}
        \includegraphics[width=\columnwidth]{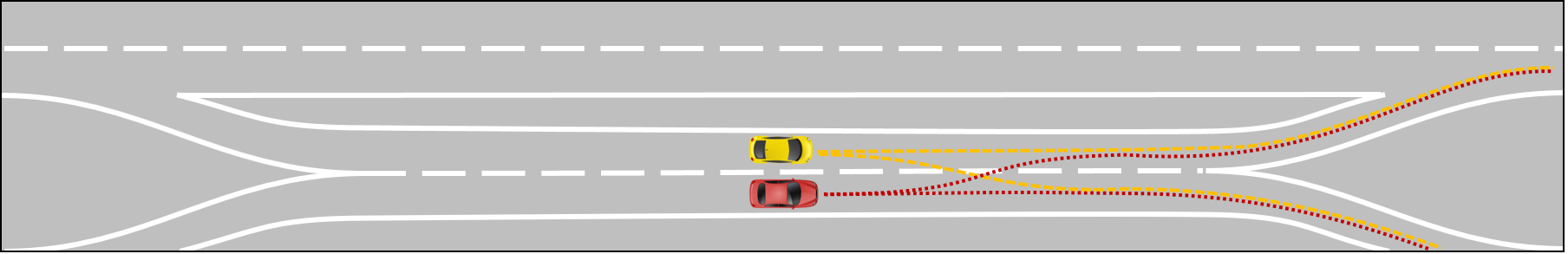}
        \caption{Four scenarios of future trajectories.}
        \label{fig:state_space_on_off_ramp2}
    \end{subfigure}
    \begin{subfigure}{\columnwidth}
        \includegraphics[width=\columnwidth]{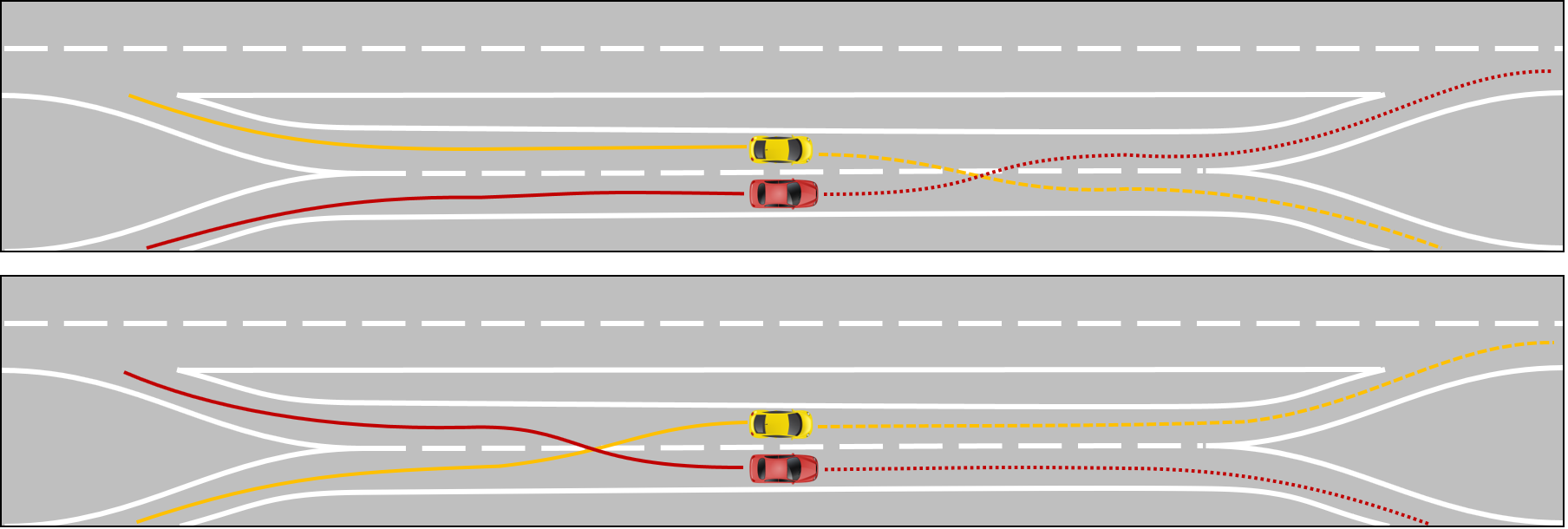}
        \caption{Driver intent based on past trajectories.}
        \label{fig:state_space_on_off_ramp3}
    \end{subfigure}
    \caption{Illustration of potentially conflicting trajectory planning of two vehicles on a highway on-off ramp.}
    \label{fig:state_space_on_off_ramp}
\end{figure}

The second task of situation interpretation is to extrapolate this estimate into the future and enable anticipatory planning which is essential for safe and comfortable driving. Now the situation becomes markedly different as we are predicting the evolution of the traffic state over time scales $\Delta t$ typical for driving maneuvers, i.e., several seconds. Consider the situation illustrated in \autoref{fig:state_space_on_off_ramp} depicting two vehicles driving on a highway on-off-ramp.

Given only the observable kinematic information for the current point in time and the environmental structures, we see that we have two possible future trajectories for each vehicle. Assuming independence of future motion for individual vehicles $i$, i.e. 
\begin{equation}
p(\mathbf{x}_{t+\Delta t} | \mathbf{x}_t) = \prod_i p(\mathbf{x}^i_{t+ \Delta t} | \mathbf{x}^i_t),
\end{equation}
we would need to consider 4 different futures for the traffic scene, three of which contain potential conflicts that a planning algorithm may have to deal with as can be seen in \autoref{fig:state_space_on_off_ramp2}.
Now consider the same situation in which two different plausible past trajectories are given as shown in \autoref{fig:state_space_on_off_ramp3}. These past trajectories provide information about the probable intent of the drivers. Now, there is only one plausible future trajectory for each vehicle and even the uncertainty with respect to the future evolution of the scene has been reduced. This underscores the necessity to model a traffic participant's intent. 

Driver intent is often modeled as an unobservable discrete state variable indicating one of several possible maneuvers classes, such as lane change left, turn right, follow road \cite{schreier_bayesian_2014, schreier_integrated_2016}. These classes must be mutually exclusive and collectively exhaustive and the concrete class has to be inferred from observations. Often, specialized motion models are associated with a maneuver class leading to so-called multiple model filters for such maneuvering targets \cite{mazor_interacting_1998}.

The complete reduction of uncertainty about future trajectories is not always possible and more than one possible option for future trajectories has to be represented. This implies that $p(\mathbf{x}_{t+\Delta t}|\mathbf{a}_{t..t+\Delta t},\mathbf{x}_t)$ should be multi modal both for the motion of individual traffic participants as well as for the entire set of traffic participants in a scene. Under the factorization assumption, this can lead to a combinatorial explosion of possible futures for the entire traffic scene including many future scenarios with conflicting trajectories. This problem can be dealt with by pruning the conflicting scenarios with corresponding computational expense \cite{wissing_interaction-aware_2018}. More desirable would be a motion model for the entire traffic scene that produces conflict free scenarios from the start.

\subsubsection{Applications in Planning}

Planning happens in state space. Given the current state of the traffic situation $\mathbf{x}_t$ including the state vector of the ego vehicle $\mathbf{x}^e_t$, the predicted trajectories of all traffic participants and dynamic aspects of the environment, the planning algorithm must find a sequence of actions that bring it closer to its destination while maintaining safety and comfort. It must do so taking into account the uncertainty in the future behavior of the traffic participants \cite{hubmann_automated_2018}. 

For this, generally model predictive algorithms are employed that optimize the expected cost of a target function $C(\mathbf{x}_{t}, \mathbf{a}_{t})$ over a constant planning horizon $T$ under a set of feasibility constraints \cite{neunert_fast_2016, giftthaler_family_2018}: 
\begin{equation}
\begin{aligned}
\underset{\mathbf{a}_{t..t+T}}{\mathrm{argmin}} & ~~ \int_0^{T} C(\mathbf{x}_{t+\Delta t}, \mathbf{a}_{t+\Delta t})p(\mathbf{x}_{t+\Delta t}|\mathbf{a}_{0..t}, \mathbf{o}_{0..t})d\Delta t\\
\mathrm{subject~ to} & ~~ f_i(\mathbf{x}_{t..T}, \mathbf{a}_{t..T}) \geq 0~~ \forall i.\\
\end{aligned}
\end{equation}
The feasibility constraints allow to include environmental and safety constraints. 
The optimization problem is discretized in time as a \ac{sqp} with initial condition set to the current kinematic state of the ego vehicle and continuity constraints between the individual stages of the \ac{sqp} \cite{boggs_sequential_1995}. In order to perform the optimization, the expected kinematic state of every other traffic participant has to be known at every stage of the optimization. Hence, it is required that the future trajectories of other traffic participants $p(\mathbf{x}_{t..t+\Delta t}|\mathbf{x}_t)$ can be evaluated efficiently. Ideally, this prediction model is \emph{interaction aware}, i.e., it takes the influence of the ego-vehicle's actions on the expected behavior of other traffic participants into account. 
This is a difficult problem, especially if the motion model is non-linear and thus the long term evolution is likely very susceptible to uncertainties in the initial state estimate. A possible remedy is to learn a state space representation and observation model under the constraint of a \emph{linear} motion model \cite{watter_embed_2015, karl_deep_2017}. This will shift complexity and computational expense to the observation model which may however be less critical as no temporal extrapolation is needed in the observation model.
Once an optimal sequence of controls for the ego vehicle over the entire planning horizon is found, the ego vehicle applies only the first step of this sequence and the process repeats with new observations, an updated state estimate of the traffic scene, updated predictions of future trajectories.

\subsection{Reinforcement Learning}
\label{Knowledge_Integration/reinforcement_learning}
\textit{Authors: Stefan Pilar von Pilchau, Christian~Brunner, \\ Daniel~Bogdoll, Tim~Joseph}

\begin{figure}
\includegraphics[width=\columnwidth]{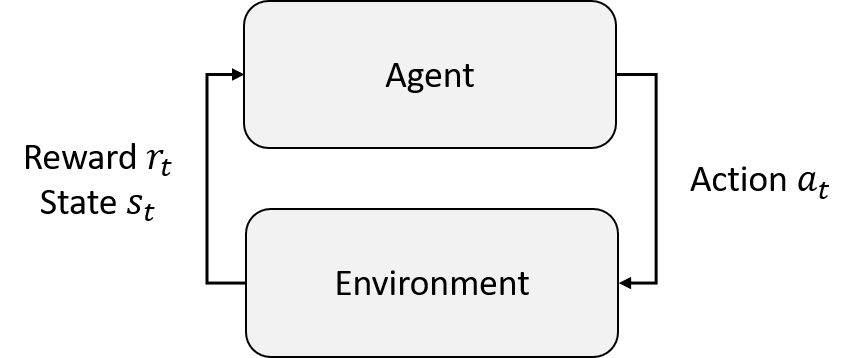}
\caption{The basic reinforcement learning setting.}
\label{fig:reinforcement_learning:rl-model}
\end{figure}

\smallskip
\noindent
\ac{rl} is a set of techniques where agents optimize their behavior given a reward signal over a period of time. A detailed introduction to the field is given in~\cite{sutton_reinforcement_2018}. Here, we stick to a brief description of the so-called \ac{rl} problem, which is depicted in \autoref{fig:reinforcement_learning:rl-model}. An \textit{agent} interacts with an \textit{environment} by executing an \textit{action} in each time step. It decides which action to choose based on the current \textit{state} and its estimated evaluation from past experiences. This mapping from the state to an action is called the \textit{policy} of the agent. Subsequently, the agent receives a \textit{reward} in each time step which reflects the notion of a local evaluation of the agents actions. But, often, this immediate reward alone is not sufficient to judge how good an action is since a larger reward will only be given after a beneficial sequence of actions, i.e., the agent faces a sequential decision making problem. For instance, the agent moves in the right direction over multiple time steps to reach a defined goal. That is why \ac{rl} algorithms commonly aim to find a policy that maximizes the expected cumulative reward instead of the immediate reward.
Over the last years, deep \ac{rl} has become the dominant form of \ac{rl}, where deep learning is used to realize an \ac{rl} agent. An introduction to these modern approaches is given in~\cite{achiam_spinning_2020}. They can be roughly divided in two categories, namely model-free and model-based algorithms. The model-based algorithms make use of an explicit model of the environment which is either given beforehand or learned from experience. Model-free algorithms on the other hand do not use such a model and always act directly in the environment. An overview regarding current approaches is given in \autoref{sec:mfrl} and \autoref{sec:mbrl}. Another common classification is the distinction between on-policy and off-policy algorithms. The former can only improve the value estimation for the policy the agent is currently carrying out. In contrast, the latter can improve the estimation of the value of the best policy independently from the actions taken by the agent. In the following, we provide an introduction to recent developments in the area of multi-agent reinforcement learning, where multiple agents operate in the same environment and might interfere (c.f.~\autoref{sec:marl}). Therefore, the agents have to take each other's actions into account. This type of reinforcement learning is especially interesting since we face a multi-agent system in the autonomous driving domain. Another current approach is the idea of inverse \ac{rl}, where one aims to learn a reward function given examples of interactions with the environments, briefly outlined in \autoref{sec:irl}. Finally, we give a brief overview regarding the recent works on the integration of knowledge into \ac{rl} algorithms (c.f.~\autoref{sec:rl-knowledge-integration}) and the current state of the art regarding \ac{rl} in the automotive domain (c.f.~\autoref{sec:rl-automotive}).

\subsubsection{Model-free Reinforcement Learning}
\label{sec:mfrl}


\ac{mfrl} algorithms learn a policy directly from real experiences in an environment without the need for a model. MFRL can be divided into methods that derive a policy from state-value estimates and methods that directly optimize a policy (policy gradient-based methods). 
Q-learning~\cite{watkins_q-learning_1992} has been one of the most popular methods based on state-(action)-value estimates. While early work in the tabular setting and with linear function approximation provided proofs on different convergence properties~\cite{tsitsiklis_analysis_1997, baird_residual_1995} and showed some first results~\cite{tesauro_temporal_1995}, more recent work showed the potential to solve complex tasks when Q-learning is combined with deep neural networks, called \ac{dqn}~\cite{mnih_human-level_2015}. Various improvements have followed~\cite{schaul_prioritized_2016, wang_dueling_2016, bellemare_distributional_2017}, including extending \ac{dqn} for partially observable Markov decision processes~\cite{hausknecht_deep_2015}, mitigating value overestimation~\cite{hasselt_deep_2016} and a combination of all previously mentioned improvements called Rainbow-DQN~\cite{hessel_rainbow_2018}. 

In contrast to Q-learning, policy gradient-based methods directly maximize the expected future sum of rewards~\cite{sutton_reinforcement_2018}. A major advantage is the ability to learn in environments with continuous action spaces, something that is not possible with standard Q-learning. An early policy gradient method is the REINFORCE algorithm~\cite{williams_simple_1992}. It uses the policy gradient in its most basic form. However, when the score function estimator is used as in REINFORCE the policy gradient is known to be of high variance and thus, learning is slow. \citeauthor{sutton_policy_2000}~\cite{sutton_policy_2000} show that the variance can be reduced when a baseline function (e.g., an advantage function) is incorporated into training. Another approach is to use \ac{dpg}~\cite{silver_deterministic_2014, lillicrap_continuous_2016} which are advantageous in environments with a high number of action dimensions. Another class of on-policy policy gradient algorithms constrains the policy change at each update step to allow for multiple updates with the same batch of data (note that the policy gradient is only valid in expectation with respect to data collected by the current policy). The most prominent members of this class are \ac{trpo}~\cite{schulman_trust_2015} and \ac{ppo}~\cite{schulman_proximal_2017}. In contrast, a recent state-of-the-art off-policy algorithm is \ac{sac}~\cite{haarnoja_soft_2018}. \ac{sac} adds an entropy bonus to the policy learning to enable better exploration and uses the re-parameterization trick~\cite{rezende_stochastic_2014, kingma_auto-encoding_2014} instead of the score function estimator which makes it more stable than methods based on \ac{dpg}. One of the major disadvantages of pure model-free reinforcement learning agents is that they often need a very high number of environment steps (e.g., a common setting for ATARI~\cite{bellemare_arcade_2013} based benchmarks is 200 millions steps) to converge to a good policy. Furthermore, the black box characteristic of most existing \ac{mfrl} algorithms makes it hard to analyze and interpret actions or future behavior. Thus, training a \ac{mfrl} in the real-world is prohibitively expensive.

\subsubsection{Model-based Reinforcement Learning}
\label{sec:mbrl}
\ac{mbrl} algorithms make use of an existing or learned model of the world to either provide imagined experiences~\cite{sutton_dyna_1991} to train a policy, to provide better gradients for policy training~\cite{heess_learning_2015} or to plan at inference time.
The two main challenges of \ac{mbrl} are to learn a model (if none exists already) and to use it effectively. 
A problem with learning a good model is model bias, i.e., that the policy will exploit regions in the model that deviate from the real environment. One approach that considers this problem is PILCO~\cite{deisenroth_pilco_2011}. The model is implemented as a Gaussian process and used to roll out imagined trajectories from which a policy can be learned with analytic gradients. \citeauthor{gal_improving_2016}~\cite{gal_improving_2016} improve PILCO with an ensemble of deep neural networks instead of Gaussian processes. Both avoid model bias by not just training the policy with a single dynamics model, but with a distribution of possible models. Similar approaches have also been used in further works in combination with TRPO~\cite{schulman_trust_2015} instead of back propagation through the transition dynamics~\cite{kurutach_model-ensemble_2018}, to learn a meta-policy~\cite{clavera_model-based_2018} or with terminal Q-functions~\cite{clavera_model-augmented_2020} for better long-term learning.
Model bias is mostly addressed when an agent acts in an environment with a low-dimensional observation space. 

For high-dimensional observations, such as images from an RGB camera, the focus shifts to higher-capacity models and efficiently predicting future rewards. \citeauthor{wahlstrom_pixels_2015}~\cite{wahlstrom_pixels_2015} learn a deep dynamics model that consists of an auto-encoder and a latent dynamics model. They then plan with this model with model predictive control. PlaNet~\cite{hafner_learning_2019} learns a sequential latent variable model to account for the stochasticity of the environment and uses \ac{mpc} to efficiently plan in latent space. \citeauthor{ke_modeling_2019}~\cite{ke_modeling_2019} uses a similar approach, but explicitly enforces correct long-term predictions through an auxiliary loss that enforces latent variables to be informative about the future.
\citeauthor{ha_recurrent_2018}~\cite{ha_recurrent_2018} train a policy purely on imagined experiences in a low-dimensional latent space that are generated with a learned world model. Dreamer~\cite{hafner_dream_2020} makes use of the differentiable model of PlaNet~\cite{hafner_learning_2019} to learn a policy by back propagating through the latent transition dynamics. Follow up works~\cite{hafner2020mastering}\cite{hafner2023mastering} improve the performance of Dreamer and demonstrate the applicability to diverse domains.

Finally, there are approaches that use an existing model, most famously AlphaGo~\cite{silver_mastering_2016} and AlphaZero~\cite{silver_mastering_2017}. Both assume that a model that can be queried efficiently for a lot of trajectories does exist. It is then rolled out with \ac{mcts} and a policy and value function are learned. However, recently \citeauthor{schrittwieser_mastering_2020}~\cite{schrittwieser_mastering_2020} have presented a variant of AlphaZero called MuZero, that uses a learned model and is able to match or outperform AlphaZero. \citeauthor{ye2021mastering}~\cite{ye2021mastering} use EfficientZero, which adapts MuZero to the setting of limited data. Three key modifications are introduced. An auxiliary loss for environment model self-supervision enforces predicted latent states to be informative about future observations. For Q-value estimation the demand to predict the exact expected reward at each timestep is loosened, predicting the cumulative discounted reward of a window of future timesteps and an off-policy correction for value estimation is implemented.

\subsubsection{Multi-agent Reinforcement Learning}
\label{sec:marl}
In \ac{marl}, the basic idea of the interaction of an agent with an environment is extended to a setting where multiple agents interact with the environment and each other at the same time. A full review of current methods and a taxonomy of the algorithms can be found in~\cite{hernandez-leal_survey_2019}. Here, we focus on some recent main achievements regarding the application of RL to multi-agent systems, as well as some extensions of standard RL algorithms to the multi-agent case.

\begin{figure}
\includegraphics[width=\columnwidth]{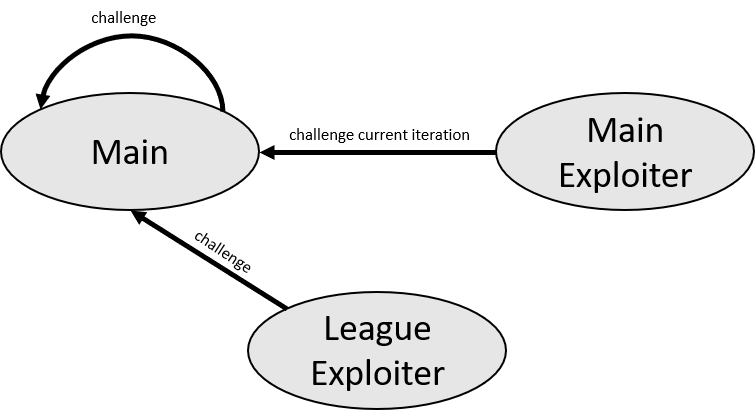}
\caption{League training as proposed in \cite{vinyals_grandmaster_2019}.}
\label{fig:reinforcement_learning:alphastar-agents}
\end{figure}

The authors in~\cite{vinyals_grandmaster_2019} present a method to train an agent capable of playing competitively with top human players in real-time strategy games. Such games can be categorized as two-player zero-sum games and a special emphasis regarding the training of the agent must be put on the fact that an agent should be robust against a variety of counter strategies. To stimulate the learning of such a behavior the authors introduced a so-called league training with the main idea to extend fictitious self-play with three types of special agents (cf.~\autoref{fig:reinforcement_learning:alphastar-agents}). The first type is named \textit{main agents} and uses prioritized fictitious play meaning it selects opponents based on the win rate against the agent. The second is the \textit{main exploiter} type which is playing against current \textit{main agents} only to find weaknesses in their behavior. The third type are the \textit{league exploiter agents} which use a similar strategy as the \textit{main agents} but cannot be targeted by the \textit{main exploiter agents}. Therefore, they have the opportunity to find strategies to exploit the entire league.

A similar challenging environment has been approached in~\cite{openai_dota_2019}. There, an agent for a \ac{moba} game has been trained which shows super human performance on a slightly less complex version of the game that, e.g., limits the number of available champions. \ac{moba} games are interesting in the context of \ac{marl} because this type of game is played five on five, meaning that there are five cooperative agents which face five competitive agents. In this work, the authors used \ac{ppo} as a basis. As in single-agent reinforcement learning, the agents start the training with a local reward only based on their own benefit from certain actions because a global reward introduced too much variance to the reward function. Later in the training, their mechanism \textit{team spirit} shifts gradually from the local reward to the global reward to encourage the agents to play as a team.
 
The authors in~\cite{ye_towards_2020} improved on the results in the \ac{moba} domain by introducing curriculum self-play learning. Here, the limitation regarding the number of champions has been weakened by training individual agents for small sets of champions and later merge them with a multi-teacher policy distillation. Furthermore, they used an off-policy variant of \ac{ppo} called Dual-clip PPO.
 
The authors in~\cite{jaderberg_human-level_2019} trained a team of agents for a capture the flag game. Each game consisted of two agents who use the same interface as humans, i.e., a \ac{rgb} image as input and produces control actions while the in-game statistics are used as reward signal. To address the special circumstances a hierarchical learning mechanism has been installed that on the one hand uses an actor-critic algorithm for the individual agents and on the other hand an evolutionary algorithm that optimizes the reward function of the agents based on the available game points. 
 
Based on \ac{mcts}, there is a multi-agent extension that has been applied to a simple grid world, where each agent has to learn to move to one of the defined goals but each tile can only be used by one agent \cite{zerbel_multiagent_2019}. The method uses \ac{mcts} with default and random policies for the rollout and combines it with difference evaluation in the reward function.

\subsubsection{Inverse Reinforcement Learning}
\label{sec:irl}
\ac{irl} is the process of learning a reward function from data-based observations. \citeauthor{arora_survey_2020}~\cite{arora_survey_2020} provide a recent survey on \ac{irl}, motivating \ac{irl} based on its potential to model the performance and preferences of others. They address two core challenges: "Finding a reward function that best explains observations is essentially ill-posed" and "computational costs of solving the problem tend to grow disproportionately with the size of the problem", which is especially relevant in the complex domain of autonomous driving, since existing methods "do not scale reasonably to beyond a few dozens of states or more than ten possible actions". They cluster existing methods based on four categories. \emph{Max margin methods} try to "maximize the margin between value of observed behavior and the hypothesis", while \emph{max entropy methods} are designed to "maximize the entropy of the distribution over behaviors". \emph{Bayesian learning methods} "learn posterior over hypothesis space using Bayes' rule" and \emph{Classification and regression methods} "learn a prediction model that imitates observed behavior". Additionally, there are many extensions to \ac{irl}, which Arora and Doshi again cluster in three categories: "Methods for incomplete and noisy observations, multiple tasks, and incomplete model parameters".

\subsubsection{Reinforcement Learning and Knowledge Integration}
\label{sec:rl-knowledge-integration}
In the following, we shortly summarize which techniques of knowledge integration have been identified.

\paragraph{Reward Shaping}
The most common form of knowledge integration is the shaping of the reward~\cite{ng_algorithms_2000, rosenfeld_leveraging_2018}. The idea is to design the reward function in a way that makes it easier for the agent to find an optimal policy, while still optimizing the original target in the limit. This can be especially useful in situations with long time horizons and sparse reward signals~\cite{openai_dota_2019}. 

\paragraph{Models}
A common way to integrate prior knowledge in an \ac{rl} algorithm is to use some sort of model of the environment. This method has defined the area of \ac{mbrl} in the first place. While the trend goes towards models that are learned by the agent during runtime, e.g., in~\cite{schrittwieser_mastering_2020}, it has been showcased that human-designed models can allow to solve very complex tasks~\cite{silver_mastering_2016} and improve the learning speed by integrating knowledge, e.g., represented by a structural causal model (cf.~\autoref{sec:causal_reasoning}) ~\cite{buesing_woulda_2018}, in the learning system.

\paragraph{Learning by Demonstration}
The idea of learning by demonstration (or apprenticeship learning) is around for some time~\cite{schaal_learning_1997}. It defines a paradigm where humans give a demonstration of a desired behavior of a learning system to speed up the learning process. One common approach to it is to use \ac{irl}~\cite{abbeel_apprenticeship_2004}. In others, there is already a reward signal available~\cite{openai_dota_2019,vinyals_grandmaster_2019}

\paragraph{Auxiliary Tasks}
A method to integrate prior knowledge into neural networks are auxiliary tasks. The main idea is to share one network over several tasks that force it to create structures that are beneficial for the main task. It has been used with actor-critique methods in a 3D labyrinth environment~\cite{jaderberg_reinforcement_2016} and a multi-agent capture the flag game~\cite{jaderberg_human-level_2019}, for instance.

\subsubsection{Applications}
\label{sec:rl-automotive}

\citeauthor{kiran_deep_2021}~\cite{kiran_deep_2021} provide a broad overview of deep reinforcement learning within the context of autonomous driving. They see many tasks where \ac{rl} could be utilized, including path planning, controller optimization and scenario-based policy learning. They provide an overview of common simulation environments and a detailed overview about the topics motion planning and inverse reinforcement learning for behavior cloning of experts. Since bridging the gap from simulation to reality is hard, they discuss many real world challenges, including validation, sample efficiency, and exploration issues.

Since \ac{rl} is often utilized to train agents end-to-end, perception tasks on an explicit level have not been within the scope of RL in the past. To tackle this issue, a recent method called \emph{Latent Deep Reinforcement Learning}~\cite{chen_interpretable_2021} utilizes the latent space to not only create control commands, but also map the sensor input, namely \ac{rgb} camera data and a \ac{bev} lidar pointcloud, to a semantic mask of the environment, including the map and surrounding objects. This way, the model does provide an interpretable environment model, which is a common output of pure perception modules. 

A major challenge in end-to-end autonomous driving using reinforcement learning is distribution shift in the simulation-to-real (sim2real) transfer. It arises when an agent trained in a simulation is deployed in the real world, degrading the driving performance. \citeauthor{so2022sim}~\cite{so2022sim} demonstrate Sim-to-Seg to cross the visual reality gap for off-road autonomous driving without using real-world data. It is accomplished by learning to translate texture randomized simulation images into segmentation and depth maps, subsequently enabling translation of real-world images. \citeauthor{10.1109/IV51971.2022.9827374}~\cite{10.1109/IV51971.2022.9827374} encode the output of a segmentation network class-wise into a latent space. Real world deployment in various environments is shown, training only the segmentation and encoder network with real world data while using the policy learned in the simulation.

\citeauthor{krasowski_safe_2020}, as part of a \emph{Safe Reinforcement Learning} framework~\cite{krasowski_safe_2020}, predict the occupancies of other traffic participants on a highway scenario. Their predictions are part of a safety layer within their \ac{rl} framework, which they utilize to only allow for safe actions during the exploration phase. Since these occupancy predictions stem from an external algorithm~\cite{koschi_spot_2017}, \ac{rl} is not utilized for situation interpretation, but situation interpretation is embedded within \ac{rl}. Further approaches on safe reinforcement learning can be found in~\cite{garcia_comprehensive_2015}.

\citeauthor{ye_survey_2021}~\cite{ye_survey_2021} provide an overview of recent methods on RL-based planning methods. They separate methods into end-to-end systems, based on sensor data as input, and motion planning modules as a follow-up module of a perception stage. The type of available actions ranges from strategic maneuvers over lane changes and trajectories to direct control. The utilized algorithms vary widely, including classical \ac{rl}, \ac{dqn}, \ac{ddpg} and \ac{a3c}.
\subsection{Deep-Learning with Prior Knowledge Maps}
\label{sec:sensor_fusion_methods}
\textit{Authors: Evaristus Fuh Chuo, Han Chen, Hendrik Stapelbroek}

\smallskip
\noindent
Object detection and recognition problems are often approached with deep-learning methods. They yet remain a great challenge in aspect of model accuracy, especially for certain circumstances, i.e., objects are occluded, too far away from the sensors or in bad light conditions. Challenges can also be found in improving data efficiency, especially when the data capacity is low. Finding a way to extract and combine information becomes important.  

\subsubsection{Semantic Segmentation }
\label{sec:sensorfusion_segemntation}
One possible way to address these issues is to incorporate prior knowledge into data-driven models.  In~\cite{ardeshir_geo-semantic_2015}, Ardeshir et al. introduced a method of combining \ac{rgb} image with information extracted from \ac{gis} system. The \ac{rgb} images are firstly segmented as the initial status. \ac{gis} databases can provide interesting semantics in 3D coordinates, which are projected onto the 2D plane of the camera. The projected geo-semantics are weighed in terms of projection accuracy before used in improving the image segmentation and projection accuracy in an iterative manner. 

In~\cite{schroeder_using_2019}, a prior fusion network is introduced, which leverages information from revisited locations during previous traversals to improve image semantic segmentation. A scene prior is defined as images frames that have a high degree of visual coherence to the current scene. In the paper, Schroeder and Alahi experimented with 3 different network architectures, two of them have access to prior knowledge, the other one is an eight-layer convolutional network that serves as the baseline model. The difference between the two prior networks lies in the integration of the scene prior, one lets the prior images pass through an encoder along with the current frame, while the other one through a decoder. The results suggest that the decoder network which enables the prior to be fused at various resolutions outperforms the others and the improving of the model performance is even more significant for dynamic objects such as pedestrians, cars, cyclists, etc. 

\subsubsection{Applications}
\label{sec:sensorfusion_object_detection}

In~\cite{yang_hdnet_2020}, Yang et al. introduced the HDNET for 3D object detection with \ac{hd} map integration. In this approach they exploit the semantic road mask available on \ac{hd} maps as prior knowledge about the scene and convert road layout information into a binary channel in discretized lidar representation. They also conduct map estimation from a single lidar revolution for where a \ac{hd} map is not directly available and apply data dropout on the semantic prior, which randomly feeds an empty road mask to the network, to get also good results when \ac{hd} maps are unavailable or noisy. Another work that aims at improving 3D object detection with \ac{hd} map integration introduced a more generalized framework that can be used with other 3D object detection algorithms~\cite{fang_mapfusion_2021}. A 2D Feature Extractor module is implemented to extract high-level features from \ac{hd} maps and the features are fused to the voxel representation of the input points. For evaluation, the framework was integrated with three major 3D object detection algorithms, SECOND~\cite{yan_second_2018}, PointPillars~\cite{lang_pointpillars_2018} and CenterPoint~\cite{yin_center-based_2020} and a general improvement of the \ac{map} on public datasets can be seen for all models. 

Instead of integrating \ac{hd} maps, another approach~\cite{chen_monocular_2016} uses ground plane as prior, as the detected objects should appear on the ground. A set of 3D sampled bounding box candidates on the ground are projected and classified in the image space.


%
%

\section{Knowledge Transfer}
\label{ch:knowledge_transfer}
In the previous chapter, knowledge integration has been predominantly addressed by means of proper modeling, \eg modifications of the architecture and cost function design. This chapter examines the task from a more algorithmic perspective, leading to learning strategies capable of transferring implicit knowledge, often captured and represented by the weights or representations of a neural network, to some target domain of interest. One of the primary goals here is to learn reliable models given only a small amount of data from the target domain. Fine-tuning of selected layers of a pre-trained network can be attributed to \textit{transfer learning} as presented in \autoref{sec_transfer_learning}. While this concept focuses on adaptation to a single target task, in \textit{continual learning} the aim is to steadily adapt to various consecutive targets, without forgetting previous tasks. An overview is given in \autoref{sec:continual_learning}. A similar idea is pursued in \textit{meta learning} where experience from multiple tasks is gathered in an episodic training process in order to improve the learning strategy with regard to upcoming tasks. \autoref{sec:meta_learning} gives an introduction to this topic. Finally, knowledge transfer is considered form the perspective of purposeful selection and re-labeling of data. Current paradigms of this \textit{active learning} strategy are outlined in \autoref{sec:active_learning}.
\subsection{Transfer Learning}
\label{sec_transfer_learning}
\textit{Authors: Florian Wasserrab, Niklas Keil}

\smallskip
\noindent
\ac{dl} has gained much momentum in recent years and is now the de facto standard approach in many applications. The use cases range from speech recognition to image processing or \ac{ad}. Despite the great success, \ac{dl} algorithms share the serious drawback that they rely on large amounts of well-annotated data to achieve their performance. The collection of such a dataset becomes very expensive and time-consuming in reality. In addition to the procurement of the datasets, the processing of such volumes of data is also a major challenge, as it requires large amounts of computing power and electricity. Transfer learning is a technique that tries to tackle these issues by reducing the amount of training data while maintaining or even improving the performance. Therefore, it has become an integral part of many deep learning models. In simple terms, this technique takes knowledge in the form of intermediate representations learned from previous tasks and applies it to novel tasks \cite{tan_survey_2018}. A great advantage of this approach is the possibility for any developer to make use of parameters optimized by data centers or supercomputers of big companies like Google (GoogLeNet) without needing resources for it themselves \cite{szegedy_going_2015}. As a contemporary example, \cite{al-huseiny_transfer_2021} implemented a lung cancer detection model based on GoogLeNet and showed that other state-of-the-art methods are outperformed on the same dataset while only requiring a fraction of training iterations.

According to a survey \cite{pan_survey_2009}, transfer learning involves the concept of a domain and a task. A domain $\mathcal{D}$ consists of a feature space $\mathcal{X}$ and a marginal probability distribution $P(X)$ over the feature space, where $X = x_1, x_2, ..., x_n \in \mathcal{X}$. In this context, $X$ is a particular dataset and $x_i$ represents the $i^{th}$ feature vector. If two domains differ, they either have different feature spaces or different marginal distributions. Based on the domain $\mathcal{D} = \{\mathcal{X}, P(X)\}$, a task is described by a label space $\mathcal{Y}$ and a predictive function $f(\cdot)$ that is also be written as conditional probability function $P(Y|X)$. The target function is not observed directly but can be learned from the data consisting of pairs $\{x_i, y_i\}$, where $x_i \in X$   and $y_i \in \mathcal{Y}$. Based on a source domain $\mathcal{D}_S$ and a corresponding source task $\mathcal{T}_S$, a target domain $\mathcal{D}_T$ and learning task $\mathcal{T}_T$, the objective of transfer learning is to improve the predictive function $f(\cdot)$ in  $\mathcal{D}_T$ with the knowledge gained from $\mathcal{D}_S$ and $\mathcal{T}_S$, where $\mathcal{T}_S \neq \mathcal{T}_T$ or $\mathcal{D}_S \neq \mathcal{D}_T$.

\subsubsection{Network-based Transfer Learning}
The authors of \cite{tan_survey_2018} classified transfer learning in the context of \ac{dl} into four different categories: instances-based, mapping-based, adversarial-based and network-based deep transfer learning. The first approach selects specific data points from the source domain and adds them to the target domain adjusted with a weight. The second method combines two domains into one data space, where instances are similar to each other. The third approach aims to find representations that apply to both the source domain and the target domain by using adversarial technology. Finally, network-based deep transfer learning takes a model which was pre-trained in the source domain and re-use it in the target domain. In the following, we focus on the latter approach, which is currently the most commonly used way to perform transfer learning. 

The intuition behind network-based transfer learning stems from the nature of neural networks. Typically, first few layers capture low-level features such as edges in an image, while higher layers focus on more complex details. Since low-level features of a pre-trained model are not domain-specific, we simply use these features for a different task and train a new model on them \cite{tan_survey_2018}. In practice, this is done by replacing the last layers of the pre-trained network with new ones. We then fix all base layers to ensure that we do not unlearn previously acquired knowledge. In this way, only the new classifier is initially trained for the intended task. Depending on how different the new use case is, the model may additionally benefit from further fine-tuning. This means instead of keeping the pre-trained parameters fixed, they are only used for initialization and included in the training process, such that they also get adjusted to the new problem. However, to ensure that the learned low-level features are still kept, the learning rate for fine-tuning is usually set rather low \cite{cui_large_2018}.
In the last years, transfer learning has been particularly popular for computer vision tasks. As one of the first implementations, \cite{oquab_learning_2014} transferred parameters taken from a classification model pre-trained on ImageNet to other visual tasks with limited data and demonstrated significant improvements over previous results. Nowadays, popular \ac{cv} models like ResNet \cite{he_deep_2016}, VGG \cite{simonyan_very_2015}, Inception \cite{szegedy_going_2015} or DenseNet \cite{huang_densely_2018} are typically pre-trained on a large image dataset to produce universal image features which serve as foundation for the specific task. For example, in \cite{studer_comprehensive_2019} the authors used pre-trained models of different domains as foundation for further fine-tuning on historical documents.

\subsubsection{Applications}
Transfer learning does not refer to a specific application, but is understood as a general approach that optimizes the learning process of complex \ac{dl} models. This often involves computer vision tasks, such as image classification \cite{oquab_learning_2014} or object detection \cite{girshick_rich_2014}, which are also applied in the field of autonomous driving. For instance, \cite{hosang_taking_2015} showed that using a pre-trained \ac{cnn} significantly improves the task of detecting pedestrians. They applied a large network trained for the ImageNet classification task and fine-tuned it for the pedestrian detection use case. In this context, they demonstrated that using another model pre-trained for a different task, i.e., scene recognition on the Places dataset \cite{zhou_learning_2014}, led to similarly performance. From this they concluded that the specific pre-training task was not crucial, since pre-training is mainly about learning basic universal features. The authors of \cite{shao_highly_2018} went a step further by using a pre-trained model from a completely different domain. In order to develop a framework for fault diagnosis of mechanical machines, they first transferred low-level features from a model pre-trained on ImageNet. Then, they converted machine sensor data to images using time-frequency imaging and performed their target task on it. Using this approach, they were able to improve accuracy and reduce the training time, even though completely different domains were combined. Of course, the idea of transfer learning is not limited to pedestrian detection or sensor data evaluation, but can be applied to various use cases in the area of AD, e.g., road lane detection \cite{zou_robust_2019}.

In many \ac{dl} scenarios, collecting data in the real world is expensive and time-consuming. For such use cases, we use simulations to acquire knowledge about the real world by pre-training our model on it. Especially in large-scale projects, such as the development of \ac{ad} technology, simulations play an essential role, as it massively reduces the training time. For example, Waymo \cite{waymore_off_2020} stated that one day of simulation is comparable to 100 years of real-world driving experience. Moreover, with simulations, we focus on the most interesting situations in AD, making the model able to handle rare edge cases more easily. For example, \cite{akhauri_improving_2021} pre-trained their model with data from CARLA \cite{dosovitskiy_carla_2017}, an open-source simulator for AD research. In addition to the CNN model, their architecture also consists of an \ac{lstm} model. This allowed them to consider temporal information, such as the speed of the vehicle, instead of focusing only on spatial features.

The planning stage of \ac{ad} deals with finding the optimal trajectory that the ego vehicle follows in certain situations. One of the most relevant examples are lane changing and overtaking scenarios where, among other parts of the environment, traffic of the current and the adjacent lanes must be considered to find potential time slots and trajectories to pass a vehicle or obstacle in front while satisfying kinematic and safety constraints \cite{dixit_trajectory_2018}. 
Most of the recently proposed ways for controlling an \ac{ad} vehicle use methods like potential fields, cell decomposition, optimal control or model predictive control \cite{palatti_planning_2021}. As these strategies are real-time optimization problems, i.e., they do not fall in the category of deep learning, the above mentioned network-based deep transfer learning is not applicable here.
However, besides them, there are also learning based methods for controlling \ac{ad} vehicles that incorporate previously learned knowledge. These are mainly (deep) reinforcement learning or (semi-)supervised learning approaches for trajectory planning for which literature shows that transfer learning could indeed be utilized.

As an example, \cite{saha_real-time_2019} used transfer learning for real-time robot path planning, which is similar to the planning of AD vehicles. The robot recorded sequences of actions (called options) while passing different obstacles which were reused afterwards to learn suitable, collision-free trajectories in more complex situations in the future. The options were stored in a library $\mathcal{L}$ and depending on the fitness score or Jacard Index of the currently perceived obstacle and those of $\mathcal{L}$, a suitable option of $\mathcal{L}$ is retrieved. Subsequently, it is transformed to best align with the new obstacle and finally applied to the current environment. Also for \ac{uav} as provider for emergency communication services, the trajectory design problem was modeled as a deep reinforcement learning process with transfer learning to benefit from previously learned knowledge \cite{zhang_trajectory_2020}. In case the base station of a communication system fails (for instance, due to a natural catastrophe), an \ac{uav} then served as aerial communication station that needs to adjust its position to dramatic changes of geography, surviving user distribution and demand.
Even though the domain of both applications is different, such approaches also serve as basis for the development of \ac{ad} technology.
In the area of semi-supervised learning, \cite{lamm_vehicle_2020} showed that scaling up the pre-training by enabling the usage of unlabeled data considerably outperformed supervised models on state-of-the-art benchmarks. Due to the significantly increased number of images in pre-training, the models benefit even more from transfer learning effects. 

Summarizing, for autonomous driving, one of the major present challenges remains to close the simulation-reality gap, where transfer learning plays an important role \cite{kiran_deep_2021}. As it is the case for nearly all AI problems that rely on data, transfer learning definitely helped solving many AD use cases. However, it should be mentioned that transfer learning can only be applied to trajectory planning, if the vehicle controlling strategy actually learns from data.

\subsection{Continual Learning}
\label{sec:continual_learning}
\textit{Authors: Christian Wirth, Stefan Pilar von Pilchau}

\smallskip
\noindent
Continual learning aims at learning a model for multiple tasks sequentially without performance degradation on preceding tasks. It is closely related to transfer learning, as it is also applicable to reuse already trained models as a starting point for learning a new task. However, in contrast to transfer learning, continual learning aims to keep the important parts of the learned model intact. Therefore, continual learning does not use past task for (only) improving the results for new tasks, but for creating a unified model that for solving all tasks. Technically speaking, continual learning does not allow arbitrary changes to the model, but only such changes that will not introduce a degradation of the performance on past tasks. Therefore, the most important aspect is to identify which parts of the model are relevant for an observed outcome. Most of the continual learning methods work with the assumption, that it is relevant to determine the relevance of models parameters in regard to the prediction. For instance, \ac{si} \cite{zenke_continual_2017} determines the importance of a parameter by its contribution to the loss. An alternative approach is approximating how much a weight can be changed while not reducing the performance on already observed tasks \cite{kirkpatrick_overcoming_2017}. For these approaches, the scalar parameters are replaced by parametric probability distribution, e.g. a Gaussian, by directly learning the mean and variance of these distributions during training (see \autoref{sec:uncertainty_general_methods}). It is then assumed, that it is possible to limit the acceptable parameter changes in terms of the Fisher matrix or the \acf{kl} divergence. Another, recent family of methods are the gradient based algorithms. They alleviate catastrophic forgetting during the training by directly manipulating the gradient for the weight update. 

In this chapter, we are focusing on Bayesian and gradient based approaches. Bayesian approaches are interesting, due to the desired properties of Bayesian Neural Networks, as explained in \autoref{sec:uncertainty}. The gradient-based approaches have shown good results in the past years and especially they promise to allow for positive backward transfer, meaning the accuracy on past tasks can be improved while training new tasks.

Known types of Continual Learning are \cite{hsu_re-evaluating_2018}:

\begin{itemize}
\item Incremental domain learning: The input and the output space are assumed to be identical but the data distribution differs.

\item Incremental class learning: The input space is identical, but the output space differs. Each task covers a subset of classes from the same multi-class set. The data distribution may also differ.

\item Incremental task learning: The input space is identical, but the output space differs. Each task covers a distinct set of different prediction targets. The data distribution may also differ.
\end{itemize}

\subsubsection{Bayesian Continual Learning}
Nearly all Bayesian Continual Learning approaches follow the same principle:

\begin{enumerate}
\item Learn parametric weight distributions on task N
\item Use (parts of) the weight distributions as a preconditioner
\item Add a regularization term for training on task N+1, that penalizes the distance to the preconditioner
\end{enumerate}
Some approaches \cite{titsias_functional_2020,pan_continual_2020} do not operate in the weight space, but in the function space. However, the differences revolve more around the structure of the (neural network) model and that will be covered in \autoref{sec:cl_function_space}. We categorize Bayesian Continual Learning algorithms by their following properties:

\begin{itemize}
\item Approximation: Which approximation of the weight distribution is used?
\item Preconditioner: How is the approximation captured?
\item Regularizer: How are the parametric distributions regularized against the preconditioner?
\item Update Mode: Which information from seen tasks has to be preserved?
\end{itemize}

Furthermore, some approaches do not work on a purely incremental scheme, where explicit information over all history tasks is captured in the last model. These methods merge the task specific models \cite{kirkpatrick_overcoming_2017,ebrahimi_uncertainty-guided_2020}, requiring an additional computation step. On the other side, some approaches work fully online, meaning it is possible to discard information from all but the latest model. Additionally, some approaches use a replay buffer \cite{titsias_functional_2020,pan_continual_2020,nguyen_variational_2018}, meaning parts of the training data are preserved and reused. For an method overview, see \autoref{tbl:Bayes_CL}.

\begin{table*}[t]
\centering
\caption{Overview of Bayesian Continual Learning Methods}
\begin{tabular}{|l|l|l|l|l|l|l|l|} 
\hline
\textbf{Method} & \textbf{Year} & \textbf{Space} & \textbf{Preconditioner} & \textbf{Regularizer} & \textbf{Approximation} & \textbf{Replay Buffer (Selector)} & \textbf{Mode}  \\ 
\hline
EWC \cite{kirkpatrick_overcoming_2017}                                                                          & 2017                                                    & Weight                                                   & Fisher Matrix                                                     & L2                                                             & \ac{mf} Laplace                                               & -                                                                           & Incremental                                              \\ 
\hline
EWC \cite{ritter_online_2018}                                                                          & 2018                                                    & Weight                                                   & Fisher Matrix                                                     & L2                                                             & KF Laplace                                       & -                                                                           & Incremental                                              \\ 
\hline
IMM \cite{lee_overcoming_2017}                                                                         & 2017                                                    & Weight                                                   & Fisher Matrix                                                     & L2, Custom                                        & \ac{mf} Laplace                                            & -                                                                           & Incremental                                              \\ 
\hline
VCL \cite{nguyen_variational_2018,swaroop_improving_2019}                                                                      & 2018/19                                              & Weight                                                   & Mean \& Scale                                                    & KL                                                             & MF VI                                 & (K-means)                                                          & Online                                                   \\ 
\hline
GVCL \cite{loo_generalized_2021}                                                                        & 2020                                                    & Weight                                                   & Mean \& Scale                                                    & Tempered KL                                                    & MF VI                                 & -                                                                           & Incremental                                              \\ 
\hline
UCL \cite{pan_continual_2020}                                                                          & 2020                                                    & Weight                                                   & Variance                                                          & LR                                                             & MF VI                                 & -                                                                           & Online                                                   \\ 
\hline
CBLN \cite{li_continual_2019}                                                                        & 2020                                                    & Weight                                                   & Mean \& Scale                                                    & (Expectation)                                              & MF VI                                 & (GMM)                                                                   & Incremental                                              \\ 
\hline
BGD \cite{zeno_task_2018}                                                                         & 2018                                                    & Weight                                                   & Mean \& Scale                                                    & KL                                                             & MF Approx.                            & -                                                                           & Online                                                   \\ 
\hline
FRCL \cite{titsias_functional_2020}                                                                         & 2020                                                    & Projected Fn.                                       & Cholesky                                         & KL                                                             & FC Approx./VI              & Reconstruction Error                                          & Incremental                                              \\ 
\hline
FROMP \cite{pan_continual_2020}                                                                        & 2020                                                    & Full Fn.                                            & Mean \& Scale                                                    & Expectation                                                    & MF Laplace                                               & Maximal Uncertainty                                                         & Incremental                                              \\
\hline
\end{tabular}
\label{tbl:Bayes_CL}
\end{table*}

\subsubsection{Gradient-based Methods for Continual Learning}
The gradient-based approaches are a recent development in the continual learning domain. The main idea is to:
\begin{enumerate}
\item Save a small set of samples from previous tasks in a memory (similar to a replay buffer)
\item Constraint the gradient calculated for the samples of the current task by the gradient wrt. the replay buffer samples.
\end{enumerate}

This avoids that the results on the previous tasks, which are represented by the replay buffer, degenerates since the loss on samples does not increase.

\begin{table*}[t]
\centering
\caption{Overview of Gradient-based Continual Learning Methods}
\begin{tabular}{|l|l|l|l|l|l|} 
\hline
\textbf{Method} & \textbf{Year} & \textbf{Space}     & \textbf{Constraint}                                               & \textbf{Replay Buffer (Selector)}    & \textbf{Update Mode}  \\ 
\hline
GEM \cite{lopez-paz_gradient_2017}             & 2017          & Gradient           & Gradient angle                  & Random      & Incremental           \\ 
\hline
A-GEM \cite{chaudhry_efficient_2019}           & 2019          & Gradient           & Gradient angle  & Random      & Incremental           \\ 
\hline
MER \cite{riemer_learning_2019}             & 2019          & Loss               & Regularize on gradient angle                            & Random & Online           \\ 
\hline
MEGA \cite{guo_improved_2020}            & 2020          & Gradient           & Adaptive loss constraint                & Random      & Incremental           \\ 
\hline
Adam-NSCL \cite{wang_training_2021}       & 2021          & Projected Gradient & Project update in null space                                      & -                                    & Incremental           \\ 
\hline
GPM \cite{saha_gradient_2021}             & 2021          & Projected Gradient & Orthogonal task updates                                   & -                                    & Incremental           \\
\hline
\end{tabular}
\label{tbl:Gradient_CL}
\end{table*}

Such a method has first been proposed in \cite{lopez-paz_gradient_2017} and is called \ac{gem}. There, the main idea is to keep the angle between gradient of the samples from the memory and the gradient of the currently new sample below 90 degree to avoid the adaption of the weights in a direction that will likely make the accuracy on a previous task worse. The authors formulate an optimization problem that is solved during the training phase and finds the gradient that does not violate the constrains regarding the memory and is closest to the original gradient of the current samples.

Since the runtime to solve the optimization problem is considerably long \ac{agem} \cite{chaudhry_efficient_2019} has been introduced. It provides similarly good results with a lower runtime by using only a randomly sampled subset instead of all samples of the memory to constrain the gradient. Furthermore, they make use of task descriptors that allow for improvements regarding zero-shot learning techniques.

In \cite{guo_improved_2020}, a unified view on the approaches GEM and A-GEM has been found and two new methods based on a Mixed Stochastic Gradient (MEGA) have been introduced. The generalization uses a reformulation of the problem to reveal the fact that it is possible to weight between the current and past tasks via a parameter. Based on this knowledge the new methods utilize information from the loss to balance between these tasks automatically by setting the parameters appropriately, i.e., they put emphasis on the current task if the loss is high and emphasis on the past tasks if the loss is low.

Recently, \ac{gpm} has been introduced which uses gradient subspaces that are only calculated once after each task with a singular value decomposition and a basis is stored \cite{saha_gradient_2021}. For the current task, the gradient is then projected in a space that is orthogonal to this projected subspace. A property of this algorithm is that it does not require to hold samples from past tasks in the memory but only the basis of the subspace. This method reduces the required memory and improves privacy. In \cite{wang_training_2021}, another idea based on singular value decomposition called Adam-NSCL has been introduced. Here, the authors calculate the feature covariance of past tasks and approximate the null space of it. Subsequently, they calculate a gradient that lies in this null space. Similar to \ac{gpm} the samples for each tasks do not have to be stored after one task has been learned. See \autoref{tbl:Gradient_CL} for an overview.

\subsubsection{Update Mode}
Most continual learning algorithms work in an incremental update mode. That is, each new model will be suitable for predicting all already-seen tasks and are regularized against all preconditioners or replay buffer samples from all those older tasks. Hence, these algorithms require to store the preconditioner and replay buffer for all tasks, which is a memory overhead. Different to that, online approaches use only the solution from the last task. A special case is the Progress \& Compress algorithm \cite{schwarz_progress_2018}, which works by training each task independently (possible initialized by the last solution), but computes a joined model by regularizing against the preconditioners in an online fashion. 

\subsubsection{Replay Buffer}
Some algorithms reuse data from old tasks as a replay buffer. This means, a subset of the training data is used as "keypoints", where it is assumed that retraining (or constraining) with these will ensure the performance on already seen task. Therefore, all data needs to be stored, but only small parts are used for continual trainings. The methods for selecting datapoints for replay try to obtain elements, that are interesting, either in terms of representativeness by using cluster centers \cite{nguyen_variational_2018,swaroop_improving_2019}, or by determining their importance for the prediction quality, in terms of a (model) reconstruction error \cite{titsias_functional_2020} or uncertainty \cite{pan_continual_2020}.  Gradient-based continuous learning methods usually simply use a randomly sampled subset of all training data \cite{lopez-paz_gradient_2017,chaudhry_efficient_2019}.

\subsubsection{Parametric Approximation}
Bayesian Continual Learning methods usually require to approximate the posterior parameter distribution of the model. As explained in in the Subspace and Parametric Approximation parts of \autoref{sec:uncertainty_general_methods}, the approximations differ in expressiveness and runtime requirements. In the following, we will only present a short recap. Please refer to the according chapter for more details.

The most common approximation is the \acf{mf} approximation  which only covers the mean and variance of each parameter distribution independently. A \ac{fc} approximation also captures the layer-wise covariances, whereas the \acf{kf} is an approximation of the full covariance matrix. For learning the parametric distributions, usually \acf{vi} (c.f.~\autoref{sec:sample_based_methods}) is used, which learns the mean and (co-) variance by gradient decent. The Laplace approximation can be computed after learning a deterministic model, but is computationally more costly and does not scale well. Only few methods use method specific approaches.

\subsubsection{Preconditioner}
The preconditioner determines the form in which the information captured by the parametric approximation is stored. This implicitly defines the regularization basis for learning additional tasks, as the regularizer works by penalizing the distance between the preconditioner and the current model in the same space. Therefore, the regularizer employs a given distance function.

A common preconditioner is the Fisher matrix \cite{kirkpatrick_overcoming_2017,schwarz_progress_2018,ritter_online_2018,lee_overcoming_2017}, which is an approximation of the second order derivative of the loss near a minimum \cite{pascanu_revisiting_2014}. Gradients are sufficient for its computation, but the method induces some computational overhead. However, this preconditioner is usually paired with a Laplace approximation of the posterior distribution, which requires an approximation of the Hessian, which is computationally substantially more costly. An alternative is directly using the mean and variance of posterior weight distribution, which can be learned "online" with variational inference. However, in this case common L2 regularizers are ususally not applicable, as described in the regularizer section. A substantial limitation of nearly all used preconditioners is the disregard of the covariance, with the exception of FRCL \cite{titsias_functional_2020}, which uses the lower triangular cholesky matrix. However, this limitation is not induced due to issues with the preconditioner or the reguarlization, but by the scalablility of the parametric approximation.

\subsubsection{Regularizer}
As explained before, the basic idea is to ensure that parameter values relevant for already observed tasks stay the same. Therefore, it is important to consider which information is contained in the used preconditioner. The Fisher information matrix is a measure of the information contained in a parameter, given a loss function (and examples). Hence, large changes of the values in the Fisher matrix also mean large changes in the information contained within a parameter and therefore it is sufficient to place an l2 regularization on these values. An alternative is the \acf{kl} divergence over the posterior distribution. If the variance of the preconditioner is small, relevant changes to the mean will result in large \ac{kl} values. However, when changing the mean for values with large variance, the \ac{kl} loss will stay small. Hence, parameters where it was possible to determine precise values (low variance) stay the same.

\subsubsection{Constraints}
Regarding the gradient-based methods, we see different possibilities to formulate concrete constraints that are used to avoid an update that makes the network perform worse. Early algorithms \cite{lopez-paz_gradient_2017,chaudhry_efficient_2019} simply limit the actually used gradient to one that has a smaller angle than 90 degrees to the gradient from the old tasks, or regularize based on the difference \cite{riemer_learning_2019}. Modern approaches \cite{wang_training_2021,saha_gradient_2021}, project the gradients into a subspace that fulfills certain conditions, like orthogonality to previous task(s).

\vspace{-3pt}
\subsubsection{Function Space Regularization}
\label{sec:cl_function_space}
Most continual learning approaches regularize in the weight space, assuming this is sufficient for preserving performance on old tasks. However, function space regularization \cite{titsias_functional_2020,swaroop_improving_2019} tries to preserve the function (output) values for old tasks. This is computational infeasible to do directly, as it would require storing (and reusing) all training data. Nevertheless, approximations are possible with the use of kernel neural networks, as explained in \autoref{sec:deterministic_inference}, since these approaches capture the full output distribution in the model parameters of the kernel. However, using a full kernel matrix is usually computationally too expensive and therefore variational distributions at inducing points are often used. With knowledge over the kernel matrix and possibly the variational distribution from past tasks, it is possible to compute the difference of the output distribution between the model of the new and the old model(s) at a given set of points. This will require to store (a subset) of the training data from past tasks, as described in the replay buffer section. 

\vspace{-3pt}
\subsubsection{Applications}
Most aforementioned applications of continual learning focus on image classification tasks, like rotated/split MNIST, belonging to the domain of incremental class learning (c.f.~\autoref{sec:continual_learning}). With regard to \ac{ad}, this bears relation to adding new classes to object detection tasks. Hence, continual learning allows to increase the number of available classes later on.

Furthermore, the incremental domain learning setting is of interested in the area of AD. It allows for incremental learning, in the sense that it is possible to improve already learned predictors with newly available data, without training from scratch. Although, with regard to knowledge integration, we use incremental domain learning for integrating multiple knowledge sources into a joined model. The knowledge sources can relate to physical knowledge, expert knowledge, rules of the road or observed examples. As example, it is possible to train a model that predicts all locations, where it is physical possible that a vehicle can occur or drive to and then use continuous learning approaches for learning with examples where vehicles have been observed. The joined model can then correctly predict all viable positions (with low probability) for vehicles, while still emphasizing positions where this is most likely, based on the observed examples.
\subsection{Meta Learning}
\label{sec:meta_learning}
\textit{Authors: Julian W\"{o}rmann, Alexander Sagel, Hao Shen}

\smallskip
\noindent
One factor of intelligent behavior is the capability to effortlessly transfer skill and knowledge from one task to a related one. While humans generally exhibit this ability, e.g., humans can derive concepts about certain object classes from only a few representative examples, data-driven AI approaches usually require a vast amount of samples in order to solve the intended task. Meta-learning, also referred to as learning-to-learn \cite{thrun_learning_1998}, addresses this shortcoming via modifications of the learning strategy. The core element here is the consideration of tasks rather than instances of data. Given a distribution of tasks, in meta-learning we try to gather knowledge about an optimal learning strategy, such that an incoming task from the same distribution can be learned effortlessly as possible. In this context, learning a classifier from only a few given examples, or predicting the trajectory of a robot in a specific environment can be considered a task. Related tasks include instances of a new class, or án modified environment in the path planning setting. Eventually, the meta-learning approach allows the learning algorithm to re-use some structure that is shared across different tasks in order to improve its performance on a similar task, which can hardly be solved by learning entirely from scratch. 

Guiding the training phase via meta-learning can thus be considered as implicit integration of knowledge into data-driven models. The preparation and organization of the data, as well as the inclusion of an additional meta-objective is key in order to lift existing architectures to knowledgeable systems.

\subsubsection{Concepts and Algorithms}
Hosepdales et al. \cite{hospedales_meta_2022} provide a comprehensive overview of the subject and associated application area.  Ongoing from the task oriented perspective introduced above, the optimization outlined in~ \cite{hospedales_meta_2022} enables to formalize the concept as an explicit bi-level optimization problem. In general, two intertwined objectives are pursued. Base learning aims at optimizing a model that is capable in solving a base problem, e.g., achieving high classification accuracy. In the associated meta-learning stage, the meta algorithm updates the base algorithm, i.e., the algorithm used to learn the base model. Through updating the learning strategy of the base algorithm, the base model improves an outer or meta-criterion, e.g., fast learning due to effective initialization. 
The progress is generally measured by means of the outcome of learning episodes. An episode represents the current model as well as its performance on instances from the training tasks. After updating the base algorithm, the performance of the resulting model is assessed again. Via iterating this scheme, the training algorithm itself 'learns how to learn' such that, e.g., generalization to new classes can be achieved from only a few samples.

One of the most popular realizations of the meta-learning paradigm is presented by Finn et al. \cite{finn_model-agnostic_2017}, called \ac{maml}. Their framework focuses on acquiring meta-knowledge about how to learn the parameters of a model, e.g., a neural network, such that the learned model can be fine-tuned to a new task in only a few gradient steps with a small amount of training data.
To put things more formally, denoting $T_i$ as a task sampled from a distribution of tasks $p(T)$, $f_{\theta}$ the sought model parameterized by $\theta$ and $L_{T}(\cdot)$ the loss according to task $T$, the base objective is to update the parameters, such that the loss on task $T_i$ is improved. Applying gradient descent, this step reads
\begin{equation}
    \theta^{\star} = \theta - \alpha \nabla L_{T_i}(f_{\theta}),
    \label{eq:meta_inner}
\end{equation}
with $\alpha$ denoting the step size.
The outer meta-objective consists in finding the optimal parameters $\theta$ across several tasks drawn from $p(T)$, i.e.
\begin{equation}
    \underset{\theta}{\operatorname{min}} \sum_{T_i \sim p(T)} L_{T_i} (f_{\theta - \alpha \nabla L_{T_i}(f_{\theta})} ).    
    \label{eq:meta_outer}
\end{equation}
Via optimizing the outer criterion~\eqref{eq:meta_outer}, the parameters $\theta$ are forced to a representation where, according to~\eqref{eq:meta_inner}, only one gradient step improves model performance regarding the individual tasks $T_i$. In this way, a suitable initialization of the weights is provided that allows for fast adaptation. 

The gradient based \ac{maml} framework has been applied in various domains, e.g., fast adaption to environmental changes in a trajectory estimation problem tackled via reinforcement learning \cite{clavera_learning_2019}, or acquiring meta-knowledge from clustered data in an unsupervised learning setting \cite{hsu_unsupervised_2019}. Furthermore, extensions exists with regard to Bayesian Meta-Learning \cite{yoon_bayesian_2018}, which aims at rapid estimation of an approximate posterior that allows to infer uncertainty, as well as Probabilistic Meta-Learning \cite{finn_probabilistic_2018}, that combines \ac{maml} with structured variational inference in order to enable simple and effective sampling of models for new tasks.

Meta-learning or meta-criterion design can also be used to learn other design parameters of the optimization process, e.g., an optimal step size \cite{antoniou_how_2019,li_meta-sgd_2017}, the weighting of regularizers \cite{franceschi_bilevel_2018} or a suitable update direction \cite{rajeswaran_meta-learning_2019}, leading to an approach that is agnostic of the choice of base optimizer. 

The approaches mentioned so far gain their adaptation power primarily through knowledge augmented optimization strategies. Model-centric, black-box or \textit{amortized} approaches represent another important line of research in meta-learning. Unlike iterates of optimization, the knowledge that allows adaptation to a new task is infused into the model via a simple feedforward-pass of an associated embedding network \cite{rusu2018metalearning, mishra2018asimple, gordon2018metalearning}. 
This way, the parameters $\theta$ (in this context, $\theta$ often comprises only a part of the network weights, e.g., one layer) can be adjusted directly without any task-specific optimization required, i.e., $\theta~=~g(S)$, where $S$ denotes the input data, and the embedding function $g(\cdot)$ is approximated by means of a neural network architecture \cite{snell_prototypical_2017, vinyals_matching_2016}. The embedding approach is also closely related to metric learning, where the high-dimensional input $S$ is mapped to a low-dimensional, even scalar representation $\eta$ such that the similarity between two instances can be easily compared. Another interesting field of application is to meta-learn the loss used in the base algorithm by means of a simple neural network \cite{grabocka_learning_2020}. In this way, non-differential objectives like, e.g., Area under the ROC curve, or F1 measure, can be approximated and thus, the base model can be tuned towards certain metrics. 
Meta-knowledge can further be gathered about architectures, where for instance \cite{liu_darts_2019} propose a gradient-descent based method that efficiently searches for suitable convolutional or recurrent network architectures.
Other approaches in turn focus on curriculum learning \cite{jiang_mentornet_2018}, which aims at finding a sample strategy such that the meta-learned models require less resources (training data or iterations) and generalize better to unseen data.

Many meta-learning approaches, however, require the meta-objective to be differentiable. The authors in \cite{hospedales_meta_2022} denote another dimension of meta-learning as the choice of optimization strategy. Besides the presented gradient descent scheme, especially reinforcement learning or closely related evolutionary algorithms are appropriate methods. The design of the overall goal pursued in meta-learning, e.g., generalization across many tasks vs. efficient optimization in single-task scenarios can be considered another axis.

In the remainder of this section, applications of meta-learning to perception, situation interpretation and planning are discussed, while the range of applications goes even beyond, covering, e.g., architecture search, transfer from simulation to real environment, or language processing. 

\subsubsection{Applications}
One of the main application domains of meta-learning is the problem of few-shot learning, which is of high practical relevance in many real world computer vision settings. The main goal of few-shot learning is to learn reliable models from only limited amount of data such that the model generalizes well to unseen or underrepresented instances. Meta-learning is often used to acquire class-agnostic features that contain information shared by many classes. Adaptation is then achieved via modulating these features based on class-specific information. Among the few-shot learning approaches, one-shot or few-shot classification \cite{vinyals_matching_2016, koch_siamese_2015} aim to learn a class from only one or a few labeled instances. Few-shot meta-learning usually involves a set of $N$ support and query pairs that constitute a base set, i.e., $\mathcal{B}= \{ (\mathbf{S}, \mathbf{Q})_i \}_{i=1}^{N}$, where $\mathbf{S}$ and $\mathbf{Q}$ represent a set of images, respectively. Each of the $N$ pairs can be considered a task, where meta-training consists in updating the model given $\mathbf{S}_i$, while the performance on the task is measured by means of $\mathbf{Q}_i$. While updating the model after encountering several tasks, the learning algorithm gathers knowledge about how to learn such that during meta-testing, the same learning algorithm can be used to train a model on the support set of a new, unseen pair $(\mathbf{S}_{test},\mathbf{Q}_{test})$. The knowledge about how to learn allows the algorithm to learn a model that generalizes to the samples in $\mathbf{Q}_{test}$, even when the number of samples in $\mathbf{S}_{test}$ is small.

As another branch, few-shot \textit{detection} has emerged recently, focusing on recognition and localization of multiple objects in images from only a few annotated samples \cite{wang_meta-learning_2019, fan_few-shot_2020}. Rather than missing class labels, few-shot detection handles cases where only a few bounding box annotations are available. Analogously to the standard object detectors outlined in \autoref{sec:perception}, also their meta-counterparts can be divided into one-stage and two-stage approaches. Kang et. al propose Meta-YOLO \cite{kang_few-shot_2019} that combines feature map learning with a reweighting module. In the first stage, generalizable meta features are learned from base classes, that allow for simple adaptation to novel classes. The subsequent reweighting scheme transfers class specific feature information from few support images to the meta features of the query image via modulating the feature coefficients. Together with the detection prediction module that outputs class labels and bounding boxes, the whole detector can be trained end-to-end. Another approach is presented in \cite{perez-rua_incremental_2020}, that augments the one-stage CentreNet object detection model with a meta-learned class-specific code generator in order to achieve incremental few-shot detection.
A meta variant of the two-stage R-CNN based object detector can be found in \cite{yan_meta_2019}. Given the features generated by the original Faster/Masked R-CNN implementation, meta learning is applied to \ac{roi} features where background and potential objects are already separated. Similar to \cite{kang_few-shot_2019}, feature reweighting in form of class specific channel-wise attention vectors is performed on the ROI features. 

Object detection usually requires the definition of anchor points or identification of auspicious regions (via Region Proposal Networks). In order to allow for an end-to-end learning without the need for an intermediate region proposal step, Zhang et al. \cite{zhang_meta-detr_2021} recently introduced Meta-DETR, which combines meta learning for few-shot object detection with the transformer structure presented in \cite{carion_end--end_2020}. 

In order to increase the generalization capabilities of data driven classification or detection models, data augmentation has been used as pre-processing step (see also \autoref{Knowledge_Integration/data_augmentation}). Meta-learning has also been used to automate the augmentation process such that models can efficiently adapt to given tasks or datasets. One of the most famous approaches has been introduced by Cubuk et al. \cite{cubuk_autoaugment_2019}. The overall goal consists in searching for an augmentation policy that maximizes accuracy in the respective downstream task. For this purpose, different augmentation operations, e.g., translation, or rotation with varying parameters are applied to the images. The corresponding search algorithm is implemented as a reinforcement learning problem. The authors in \cite{li_differentiable_2020} propose to relax the costly policy selection to a differentiable optimization problem with the aim to reduce the computationally complexity on the one hand, and to allow for a joint optimization of the network weights and the augmentation parameters on the other hand.


The concept of meta-learning offers useful advantages also in the context of \ac{rl}, which has been applied for motion planning in dynamic environments. Especially, meta-knowledge about, e.g., goals, rewards or environments can be used in order to reduce the training time in terms of sample efficient exploration strategies \cite{hospedales_meta_2022}. 

Gupta et al. \cite{gupta_meta-reinforcement_2018} propose an exploration strategy based on the \ac{maml} approach, that is informed by prior knowledge due to a meta-learned policy and latent space. Quick adaptation of the \ac{rl} agent to new real world tasks or environments via meta-learning is also investigated in \cite{clavera_learning_2019}. \ac{rl} combined with a meta-learned self-supervised interaction loss is proposed in \cite{wortsman_learning_2019}. By aligning the gradients of a supervised and self-supervised loss in the training phase, during inference the agent can use the gathered meta-knowledge about the imitated supervised loss for self-adaptive visual navigation.   

Applications to trajectory planning scenarios are presented in \cite{ye_meta_2020}, where the authors combine meta learning with \ac{rl} to develop a strategy for decision making for lane change maneuvers on highways. Again, a variant of \ac{maml} is used to improve adaptation speed for new scenarios, e.g., traffic density, road geometries, or traffic environments.
\subsection{Active Learning}
\label{sec:active_learning}
\textit{Author: Tino Werner}

\smallskip
\noindent
\ac{al} is an iterative learning strategy that incorporates human/expert feedback. The idea is to query the labels of some yet unlabeled instances resp. to create new labeled instances and to update the model taking the new information into account, alternatingly. The advantage of \ac{al} when querying labels is the reduction of the annotation cost while clever sampling of new data points enables to train the model with less data than with uninformed sampling. Due to these benefits and the opportunity to better estimate the model uncertainty (see also \autoref{sec:uncertainty}), \ac{al} is a promising strategy for integrating knowledge into the \ac{ai} training process.

Querying labels requires the existence of a so-called unlabeled pool and is therefore known as pool-based sampling \cite{lewis_sequential_1994}, so one selects an unlabeled instance or a batch of unlabeled instances and asks for its label. Another strategy is to generate new samples which is called selective sampling if such samples can be selected from an existing pool or stream of data so the learner can still decide whether to ask for the label or not without having wasted significant computational time. The latter method is also called membership query synthesis if artificial new instances are generated according to criteria listed below or by just combining valid features from an abstract feature space (\cite{cohn_improving_1994, angluin_queries_2001, settles_active_2009}, see also for example \cite{dasgupta_general_2007, dasgupta_hierarchical_2008}). The key question is which of the unlabeled samples have to be considered for annotation resp. how new samples can be generated since, evidently, random sampling is not efficient and leads to high annotation costs. There are different sampling paradigms like uncertainty/margin-based sampling, see \cite{lewis_sequential_1994,balcan_margin_2007}, i.e., sampling from regions where the model is uncertain which for classification model means to sample from near the decision boundaries or query-by-commitee \cite{seung_query_1992} where one invokes different models and samples from regions where the models disagree most. Another paradigm is density-based sampling, i.e., sampling from representative regions, emerging from the fact that uncertainty sampling is prone to select outliers. Other paradigms consider the informativity/impact of the samples which is quantified by criteria like expected model change, expected error reduction or variance reduction. Furthermore, active learning is not limited to querying the label of one single unlabeled instance per iteration but can also consider whole batches of unlabeled instances. This has been studied for example in \cite{chakraborty_active_2015, jain_active_2016, kading_active_2016, xu_understanding_2019} who aim at finding the batch of the best instances according to the given paradigm but by simultaneously being aware to minimize the redundancy of the single instances in the batch. The number of samples in a batch does not have to be uniform over the iterations and may have to be appropriately chosen, see, e.g., \cite{paul_non-uniform_2017}. \ac{al} can be used for knowledge integration in \ac{ai} training by augmenting the training data (see also \autoref{Knowledge_Integration/data_augmentation}) with instances where high losses are suffered, where the model is uncertain or where an expected model change or variance reduction is possible in supervised tasks like object recognition, semantic segmentation, person re-identification or tracking. Furthermore, \ac{al} can even enter unsupervised tasks by considering for example the model’s entropy. One may invoke \ac{al} in knowledge extraction or conformity checking tasks by sampling from regions where the model output diverges from existing knowledge. As for testing a particular trained model, active testing may be applied in order to reduce the annotation costs for the test data.

See \cite{settles_active_2009} for an excellent overview of sampling paradigms for \ac{al}. See \cite{ren_survey_2021} for a recent overview of deep \ac{al} which is a combination of deep learning and \ac{al} in order to keep the high predictive quality of deep neural networks, including feature abstraction while reducing the labeling costs.

\subsubsection{Different paradigms for active learning}

An important result that connects the paradigm of \ac{al} with the available sampling budget is given in \cite{hacohen_active_2022} who theoretically derive that the best \ac{al} strategy crucially depends on the available labeling budget. More precisely, for a high budget, one should query uncertain samples while for a low budget, one should focus on "typical" points where they define the typicality as the inverse of the average squared distance to the nearest neighbors. In order to encourage diversity, they first cluster the points and then sample typical points from clusters which not yet contain labeled points.

\paragraph{\textbf{Uncertainty sampling}}
For uncertainty sampling, a lot of different criteria have been suggested in literature. The authors in \cite{brust_active_2018} focus on the 1-vs-2-score, i.e., the difference of the class-probabilities for the best and second best class for a given sample, in the context of deep object detection, based on the fact that a small difference between the probability of the most resp. second most probable class indicates uncertainty, more precisely, the sample is located near the intersection of the decision boundaries. The authors in \cite{ducoffe_adversarial_2018} propose DeepFool active learning (DFAL) where the selection of unlabeled data is performed by choosing those which are closest to their adversarial attacks (generated by DeepFool, see \cite{moosavi-dezfooli_deepfool_2016}) whose corresponding perturbation magnitudes can be seen as approximating the distance to the nearest decision boundary. A Bayesian approach is given by Bayesian active semi-supervised learning for Deep \ac{cnn}s \cite{rottmann_deep_2018} that iteratively reduces the classification entropy on the unlabeled data which can be combined with \ac{mc} Dropout \cite{srivastava_dropout_2014, gal_dropout_2016}. Other Bayesian approaches include \ac{gaal} \cite{zhu_generative_2017} where training instances are iteratively generated by a \ac{gan} and task-aware VAAL (variational adversarial active learning), \cite{kim_task-aware_2021} who combine the VAAL strategy from \cite{sinha_variational_2019} with the loss prediction technique of \cite{yoo_learning_2019} which predicts the target loss for the yet unlabeled instances. Ensemble methods for uncertainty estimation have been considered in \cite{beluch_power_2018} and \cite{kuo_cost-sensitive_2018}. In \cite{kao_localization-aware_2018}, specific uncertainty measures for object detection are proposed, namely localization tightness (how tight the bounding boxes are around the object) and localization stability (how stable are the boxes in the presence of input perturbation). Imbalanced data has been considered in \cite{qu_deep_2020} who propose to compute a weighted uncertainty to take the imbalance into account. The authors in \cite{hospedales_finding_2011} take a contrary position to the paradigm to sample from the vicinity of the decision boundaries by performing lowest likelihood sampling where the likelihood is approximated by a \ac{gmm} in order to discover yet unseen classes. In \cite{desai_adaptive_2019}, an adaptive supervision framework for \ac{al} for object detection is proposed. The idea is to avoid querying strong labels, here bounding boxes, but weak labels in the sense of weak supervision in order to optimize the model. Concretely, they use center-clicking from \cite{papadopoulos_training_2017} and derive an algorithm that decides whether one has to switch to querying strong labels. \cite{hemmer_deal_2022} propose DEAL (deep evidential \ac{al}) for image classification by replacing the softmax output of a \ac{cnn} by a Dirichlet distribution that allows for uncertainty quantification. \cite{bengar_when_2021} consider the label-dispersion, a measure to quantify how much the label assigned to a specific instance changes during NN training, i.e., over different epochs. A high label-dispersion indicates that there is a large uncertainty. Experiments show a considerable benefit of this method. The method from \cite{aggarwal_optimizing_2022} can be interpreted as a narrower version of label-dispersion. They introduce the alamp uncertainty measure that is given by the difference of the 1-vs-2-scores for a given sample on the current and the previous iteration, divided by the sum of these two quantities. 

\paragraph{\textbf{Diversity/informativity sampling}}
Diversity of the samples is encouraged using clustering \cite{nguyen_active_2004, chen_robust_2018} and sampling representative points from each cluster. Methods based on the KL-divergence or the cross-entropy have been suggested in \cite{xu_person_2020, ebrahimi_minimax_2020, chakraborty_active_2015, paul_non-uniform_2017}. Adversarial active learning has been considered in \cite{zhang_state-relabeling_2020} who introduce state-relabeling adversarial active learning (SRAAL) where a state discriminator identifies the most informative samples which are those being dissimilar from the already labeled ones. Further adversarial strategies are VAAL, based on a discriminator which considers the difference of the unlabeled instances to the latent representation of the labeled instances in \cite{sinha_variational_2019} and adversarial active learning for \ac{lstm}s \cite{li_adversarial_2021} where again the similarity of labeled and unlabeled samples is considered. In \cite{tran_bayesian_2019}, the most informative sample is detected using a \ac{vae}. The authors in \cite{wang_deep_2018} propose true positive sampling in the context of person re-identification with multi-view data and argue that true positives are informative. Since true positives from the same camera view are often similar, so that a view-specific bias would arise, they compute the distance of two tracklets for an adaptive resampling strategy. Density-based approaches are given in \cite{li_adaptive_2013}, based on an information density measure related to the conditional entropy, and in \cite{yang_suggestive_2017} where uncertain samples that are similar to most of the training samples are selected. The authors in \cite{wang_cost-effective_2016} consider informativity based on either entropy, least confidence or margin and iteratively select the $K$ most informative unlabeled samples. Their CEAL algorithm checks for which of these instances the current model makes a high-confidence prediction and adds these pseudo-labeled instances into the labeled set so that the model can be updated. \cite{elenter_lagrangian_2022} state the problem of selecting informative points to query as a constrained optimization problem and derive that the Lagrangian dual arising by optimizing some loss $L$ subject to some loss $L'$ being bounded by some threshold on the whole training set allows for using the dual variables as an informativity measure. \cite{haussmann_scalable_2020} experimentally compare different informativity scores and techniques to ensure the diversity of the selected samples. \cite{gentile_achieving_2022} derive minimax rates for the excess risk in batch AL. They propose to optimize a tradeoff between informativity and diversity in pool-based \ac{al}. More precisely, for noisy linear models, their algorithm first considers all points where the current model is not confident and queries the labels for a batch of points whose Mahalanobis distance from the already queried points exceeds an iteratively refined threshold, allowing for varying batch sizes among the iterations. They extend their method to generalized linear models and detail out the logistic case. As for the noise, they point out that by re-training the model at most logarithmically often w.r.t. the size of the available pool, their algorithm adapts to the noise level provided that a low-noise condition is satisfied.

\paragraph{\textbf{Variance reduction/expected model change/expected model improvement}}
An approach based on kernel mean matching where two conditional distributions are matched without requiring to estimate the densities has been introduced in \cite{wang_active_2014} for transfer learning (match the conditional distributions of the responses in the training resp. the test domain) with the goal to reduce the predictive variance. Variance reduction has also been the goal in \cite{beygelzimer_importance_2009, rotskoff_learning_2020, pydi_active_2019} who combine \ac{al} with importance sampling where one samples from a so-called proposal distribution in order to reduce the variance compared to sampling from the original distribution. Several works aim at reducing the misclassification risk, either by approximating it with the Shannon entropy in person re-identification tasks \cite{das_active_2015}, by finding wrong detections by exploiting the temporal coherence in object detection in videos \cite{bengar_temporal_2019}, via decomposing the expected loss \cite{sener_active_2017}, via Bagging to reduce the variance of the error estimate \cite{roy_toward_2001} or by approximating the expected error/performance by Bayesian approaches \cite{kottke_multi-class_2016, kottke_toward_2021, sahli_costabal_physics-informed_2020}. Analogously, \cite{hospedales_unifying_2012} and \cite{wu_cost-sensitive_2019} consider the expected accuracy/utility improvement. The authors in \cite{xu_understanding_2019} approximate the expected improvement by means of influence curves which assign a real number to each point of the input space, quantifying its infinitesimal impact on the estimator. The authors in \cite{houlsby_bayesian_2011} consider entropy reduction by maximizing the decrease in the expected posterior entropy in a Bayesian approach. Their objective is termed as BALD objective while \cite{gal_deep_2017} also consider a Bayesian setting and aim at optimizing a generic acquisition function. In \cite{wu_cost-sensitive_2019} and \cite{wang_cost-sensitive_2019}, one aims at minimizing the total annotation cost by dividing the data into clusters and by determining the optimal number of queried labels per cluster, leading to the algorithms CADU resp. CATS. The expected model change is considered as criterion in \cite{he_nearest-neighbor-based_2007} (change in the modelled local density around each sample). The authors in \cite{freytag_labeling_2013} estimate the expected changes in a Gaussian process model and \cite{freytag_selecting_2014} generalize the works of \cite{freytag_labeling_2013} and \cite{vezhnevets_active_2012} who consider Gaussian process regression and provide the EMOC algorithm. This EMOC algorithm also has been applied in \cite{brust_active_2020, kading_active_2018, kading_active_2016}. Expected model change can even enter as \ac{al} strategy in reinforcement learning where the sensitivity of the optimal policy towards changes in the rewards resp. transitions are considered \cite{epshteyn_active_2008}. In a tracking and video annotation context, \cite{vondrick_video_2011} interpolate the trajectory and approximate the expected trajectory label change when querying the annotations for selected frames.

\paragraph{\textbf{Combined approaches}}
Several strategies explicitly combine different sampling paradigms. In \cite{siddiqui_viewal_2020}, superpixels of images are selected in segmentation tasks in an uncertainty- and diversity-based manner. Uncertainty and diversity are also simultaneously considered in \cite{elhamifar_convex_2013}, based on the entropy resp. the ratio of the minimal distance to another unlabeled sample, see also \cite{yang_multi-class_2015}. The authors in \cite{wang_learning_2021} address person re-identification with weak labels, i.e., which only indicate whether a particular person appears in a video but not in which frame, and propose a combined criterion based on the cross-entropy and a penalty term that discourages sampling highly correlated videos. The authors in \cite{wang_multi-class_2018} concentrate on informativity and diversity while \cite{jain_active_2016} aim at selecting influential, uncertain and diverse instances. The authors in \cite{yuan_active_2021} apply \ac{al} in deep visual tracking using \ac{cnn}s by considering representativeness and diversity of the queried samples. Diversity is encouraged by a cooperation-based selection method using multiple frames in order to eliminate the background so that it does not affect the selection, while for the representativeness, a nearest neighbor discrimination method is applied in order to detect low-quality samples.

\paragraph{\textbf{Other applications and paradigms of active learning}}
Other approaches include concentrating on strong negatives in person re-identification tasks \cite{wang_human---loop_2016}, learning a selection strategy for \ac{al} using \ac{rl} \cite{fang_learning_2017, liu_deep_2019, woodward_active_2017}, \ac{al} for ranking \cite{long_active_2014} or Kriging (Gaussian process regression) \cite{yang_active_2015} or active \ac{svm} \cite{tong_support_2001}. The authors in \cite{krueger_active_2020} propose active reinforcement learning where one has to pay for observing the rewards. \cite{konyushova_active_2021} consider active offline policy selection where a set of candidate policies can only be evaluated on a fixed total budget of rollouts, which is solved by Bayesian optimization. 
\ac{al} is not limited to select single instances resp. batches of instances. For example, \cite{siddiqui_viewal_2020} sample superpixels instead of whole images and \cite{li_uncertainty_2020} experimentally compute the uncertainties for the estimation in active learning approaches in image segmentation and derive that such approaches where regions instead of whole images are labeled are much more effective and significantly reduce the labeling effort. The authors in \cite{wang_learning_2021} consider the case of weak labels for videos which do not provide the frame indices in which a particular person appears but only the binary response whether this person appears in the whole video and \cite{vondrick_video_2011} propose to query only the annotations in selected frames of videos. \cite{gao_unsupervised_2021} propose unsupervised clustering \ac{al} (UCAL) for person re-identification. After having applied a clustering algorithm like DBSCAN, they consider centroid pairs of person IDs that are erroneously grouped into the same resp. different clusters. These clusters are re-organized via representative pairs, i.e., splitting clusters is done via an independence-compactness tradeoff while for merging, one decides which pairs to label via the similarity of adjacent centroids. \cite{segal_just_2022} propose fine-grained AL for partially labeled scenes, e.g., for predicting future positions by autonomous vehicles. They criticize that there exist \ac{al} approaches for object detection but none for prediction and perception. Sensor observations are represented by HD maps and the labels by bounding boxes. They point out that considering a fixed batch size in \ac{al} per iteration assumes that each sample can be labeled for the same cost which is not true. They introduce a cost function so that the different efforts for manually labeling sparse or crowded scenes are reflected and fix a cost bound per iteration. They choose the next scenes to label by the ratio of the informativity score (e.g., the detection or prediction entropy) and the cost. The costs are unknown but can be approximated by the number of detections after non-maximum suppression \cite{felzenszwalb_object_2009}. They generalize their method to partially labeled scenes where they consider a fine grid of the scene, allowing for a fine-grained selection, again by the ratio of the informativity and the cost. \cite{bengar_reducing_2021} study the combination of self-supervised learning and \ac{al} and derive than it is only beneficial for high labeling budgets, marginally improving the cost for self-supervision and considerably for \ac{al}. The two paradigms themselves are also compared with the result that the former is way more efficient in reducing the labeling cost than \ac{al}. 

\paragraph{\textbf{Limitations}}
Despite the successes of \ac{al}, there are several limitations. The authors in \cite{mittal_parting_2019} experimentally show that \ac{al} without data augmentation for semantic segmentation and image classification may fail (i.e., they are worse than random sampling) if the labeling budget is small and postulate that this may result from biasing the distribution of the annotated samples. Especially, their findings encourage to combine semi-supervised learning with \ac{al}. Another study has been executed in \cite{munjal_towards_2020} who derive that the question whether \ac{al} in a \ac{dl} context provides an improvement heavily depends on the hyperparameters, the paradigms, the datasets etc.. The authors in \cite{lowell_practical_2018} derive that it is difficult to assess the quality of \ac{al} since this would normally require a large number of queried samples. Additionally, the first queried samples may improve the model, but later queried samples may decrease its performance. Moreover, the training set and the model are coupled due to \ac{al} which decreases the generalization ability across models and tasks. The authors in \cite{jain_active_2016} point out that human annotations in semantic segmentation are often too expensive while automatic segmentations are not sufficiently accurate. The sampling bias is studied in \cite{prabhu_sampling_2019} who derive that methods based on the posterior entropy are robust to sampling biases while \cite{imberg_optimal_2020} show that sampling probabilities proportional to the uncertainty often lead to suboptimal proposals and can perform worse than random sampling. They propose optimal sampling schemes where optimality is meant either w.r.t. the variance of the estimated loss, the mean squared error of the prediction or the expected total loss. The authors in \cite{farquhar_statistical_2021} study the sampling bias and propose corrective weights which leads to estimators which are unbiased and convergent as long as the acquisition proposal puts non-zero mass on all training samples. \cite{bengar_class-balanced_2022} point out that AL may become problematic in the presence of class imbalance due to \ac{al} may be biased towards particular regions, see also \cite{beygelzimer_importance_2009}, \cite{farquhar_statistical_2021}. \cite{bengar_class-balanced_2022} correct this bias, due to the true labels not being available, by treating the predicted labels (more precisely, the softmax output) as proxy. Having this softmax matrix $P$ and $z \in \{0,1\}^n$ which represents the selection of samples in \ac{al}, the goal is to minimize $||\Omega(c)-P^Tz||_1$ for the number $\Omega(c)$ of required samples of class $c$. This leads to the class balanced \ac{al} (CBAL) algorithm which minimizes the entropy regularized with the norm above. They also show how to find a tradeoff between representativeness and class balance. \cite{aggarwal_minority_2021} propose an \ac{al} strategy that is more focused on the minority class in imbalanced data by assigning more budget to them. They experimentally derive that transfer learning an \ac{al} can be combined by using the features extracted from a pre-trained model on a generic dataset. \cite{kwak_trustal_2022} experimentally derive that \ac{al} indeed does not imply monotonically increasing accuracy but that learned knowledge can be forgotten during the \ac{al} iterations. They propose to use knowledge distillation in order to mitigate this but point out that the current model may not be the best teacher due to this issue. They select the teacher by a tradeoff between accuracy and consistency (i.e., making correct predictions over subsequent \ac{al} iterations).  The authors in \cite{chandra_initial_2021} study how to select the initial pool in pool-based \ac{al}. The initial pool usually contains around 1-10\% of the dataset. The goal is to cope with the disadvantage that deep \ac{al} seldomly outperforms random sampling \ac{al} in the presence of changes in the data (class imbalance) or training procedures (regularization, data augmentation). They propose two strategies. The first is to train a self-supervised model on the raw data and to include the samples on which this model suffers a high loss into the initial pool. The other strategy is cluster sampling, i.e., the data are divided into clusters and one samples from each cluster with equal weight in order to encourage diversity and that the initial pool represents the dataset well. A related semi-supervised strategy is given in \cite{simeoni_rethinking_2021} who point out that semi-supervised learning concentrates on the most certain samples while \ac{al} concentrates on the most uncertain ones. However, \cite{gao_consistency-based_2020} criticize that the method of \cite{simeoni_rethinking_2021} is independent from the model training and propose to unify selection and model updates. They point out that uncertainty sampling is often not appropriate for deep learning \ac{al} since \ac{dnn}s tend to be overconfident. They found out that their consistency-based approach outperforms the entropy-based approach and also works sophisticatedly in terms of diversity of the queried samples. A severe issue of their method (including related \ac{al} methods) is the cold start failure which means that the initial model trained on the initial pool has a poor performance and leads to significantly wrong decision
boundaries, so an entropy-based method would query from near this wrong boundary. Their semi-supervised strategy also suffers from cold start failures. They point out that if one had a
validation set, one could track the performance of the model to avoid cold start failures, however, this is often not the case in practice. For the CE-loss, they propose an upper and a lower bound which enable a tracking during training.
As for a validation and a test set, \cite{kossen_active_2021} introduce active testing, i.e., sample-efficient model evaluation where test data is labeled in an active manner by maximizing the accuracy of the risk estimate. They point out that (naturally) selecting most uncertain points to label overestimates the risk which is even worse for overconfident models. To remedy this, they apply the IS strategy from \cite{farquhar_statistical_2021} since a sampling bias is much more harmful for testing than for training.

\vspace{-5pt}
\subsubsection{Application}
We point out that \ac{al} is not tailored to a specific task but serves as a paradigm for sophisticated iterative (re-)training of the respective \ac{ai}. Therefore, \ac{al} can be used to improve a particular learning algorithm for perception or planning models, but as already outlined above, the concrete sampling or selection paradigms of \ac{al} differ.

As for existing work for perception, \ac{al} has been applied to object detection/segmentation in \cite{brust_active_2018} who invoke the PASCAL VOC 2012 dataset. The authors in \cite{desai_adaptive_2019} additionally run experiments on the PASCAL VOC 2007 dataset and point out that it is crucial how the threshold in their strategy is chosen since switching too quickly to strong labels causes much higher annotation costs than staying in the weak labeling case which however is prone to produce noisy labels. The authors in ~\cite{kao_localization-aware_2018} consider the PASCAL VOC 2007, the PASCAL VOC 2012 and the MS COCO dataset and derive that their localization tightness metric improves the model significantly provided that the ground truth is known, otherwise the improvement is rather low. Their localization stability metric improves the model but is more expensive to compute than other metrics. The authors in \cite{li_uncertainty_2020} apply their region sampling method to the GlaS Challenge Contest and the 2016 International Skin Imaging Collaboration data and find out that sampling regions instead of whole images can indeed lead to a better calibrated model and significantly reduce the annotation cost. Evidently, all these approaches and other ones for classification or segmentation tasks, even if not primarily designed for automotive use cases, can directly be applied for automotive data. As for person~(re-)identification which is of course directly related to the automotive setting, \cite{wang_deep_2018} apply their method to the PRID, the MARS and the Duke-video dataset. On all datasets, a significant improvement is observed while keeping the annotation ratio below 3\%. The authors in \cite{das_active_2015} consider the iLIDS-VID dataset and \cite{bengar_temporal_2019} consider a synthetic dataset as well as the ImageNet-VID dataset where they only require to label 10\% of the latter dataset. The authors in \cite{vondrick_video_2011} make experiments on the VIRAT data and report that they are able to reduce the annotation costs by 90\% compared to a labeling strategy where a fixed rate of instances is to be labeled. In \cite{wang_human---loop_2016}, experiments are executed on the CUHK03, the Market-1501 and the VIPeR dataset where also the human-out-of-the-loop (HOL) case is considered which means that the number of human annotations is limited, so human feedback becomes eventually unavailable. They show that their \ac{al} strategy even leads to competitive performance in the HOL setting. The authors in \cite{wang_learning_2021} consider the two weakly labeled datasets WL-MARS and WL-DukeV and even study the setting of missing annotations where for example a person that is visible in a few frames is missed by the annotator so that even weak labels are not available. They show that even in the latter setting, their method can improve the performance of the model. \cite{yuan_active_2021} apply their \ac{al} method on seven benchmark datasets for tracking and derive that it achieves a comparable performance to methods that require huge labeled training data. \cite{feng_deep_2019} apply \ac{al} with the uncertainty criterion (measured using MC dropout or ensembles) for lidar 3D object detection. They conduct experiments on the lidar depth and intensity maps from the KITTI dataset and derive superior performance in terms of required labeling effort in comparison to a training method where samples are queried randomly.


Active learning also successfully entered trajectory planning and \ac{mpc}, see, e.g., \cite{mesbah_stochastic_2018} where the uncertainty is actively learned and \cite{zhang_active_2021} where informative trajectories for \ac{mpc} are generated. In stochastic \ac{mpc}, \cite{heirung_stochastic_2017, heirung_model_2018, heirung_model_2019, arcari_dual_2020} apply \ac{al} for learning structural uncertainties of the model in order to enable better model discrimination. In \cite{buisson-fenet_actively_2020, capone_localized_2020, cowen-rivers_samba_2022, zimmer_safe_2018}, \ac{al} for Gaussian process dynamics, where either uncertainty minimization or informativity maximization is performed, is considered. Model-based active exploration in \ac{mpc}s has been suggested in \cite{shyam_model-based_2019, schultheis_receding_2020}. \cite{hu_active_2022} propose active uncertainty learning for motion planning with stochastic \ac{mpc}.

\section{Knowledge Extraction - Symbolic Explanations}
\label{ch:symbolic_extraction}

The explanation and interpretation of the functioning of data driven models is an essential prerequisite on the way to trustworthy systems. Special focus is put on an understandable representation and formalization of the behavior learned by a neural network. The extraction and consideration of learned decision patterns and concepts not only motivate a final validation of the function but are already equally useful in the development of methods for use in safety-critical applications.

Symbolic interpretations play an important role and have been extensively investigated in recent years. In order to inspect the functionality of a neural net in a more formal way, \textit{rule learning} has established itself as one of the most prominent methods in this context. The approaches introduced in \autoref{sec:rule_learning} range from methods that focus on the internal structure and interplay of neurons to methods that try to model the input-output relations in a human interpretable way. The latter is also connected to rule extraction and pattern mining with respect to the input data as discussed in \autoref{sec:structured_output_prediction} focusing on \textit{structured output prediction}. 

Natural language, as another form of symbolic representations beyond rules, is considered in the two remaining sections. While knowledge extraction in the context of regulations and norms from \textit{legal domain} is the focus in \autoref{sec:nlp_for_legal_domain}, natural language as accompanying explanation to visual stimuli is the goal in \textit{visual question answering}, presented in \autoref{sec:visual_question_answering}. 

\subsection{Rule Extraction and Rule Learning}
\label{sec:rule_learning}
\textit{Authors: Adrian Paschke, Ya Wang, Kevin Krone, Etienne B\"{u}hrle, Hendrik K\"{o}nigshof}

\smallskip
\noindent
This section looks at algorithms for rule extraction and rule learning. A distinction is made between extraction approaches that use the internal structure of the trained models (e.g. decompositional rule extraction) and approaches that use the models and extracted concepts as a whole (pedagogical and eclectic rule learning) to derive new (symbolic) knowledge, and inductive learning approaches (e.g. Inductive Logic Programming), which learn rules from training data. The section first considers the extraction approaches, then approaches for inductive logic programming (inductive learning of symbolic rules), and last applies rule learning in the context of planning.
\subsubsection{Rule extraction from trained neural netwoks}
\begin{figure}[t]
    \centering
        \includegraphics[width=0.4\textwidth]{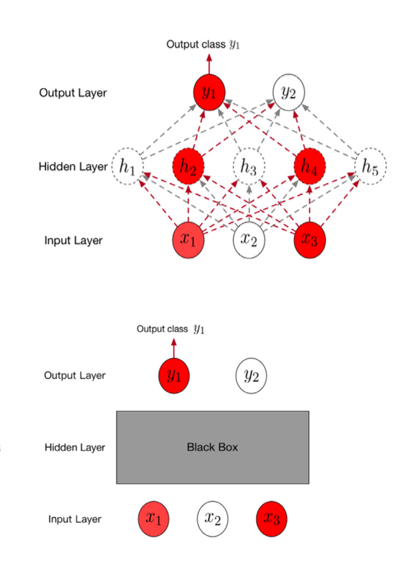}
    \caption{Illustration of the functionality of decompositional (top) and pedagogical (bottom) approaches \cite{he_extract_2020}. (red color stands for activated state and transparent color for deactivated state) }
    \label{fig:Compare_deco_pad}
\end{figure}

In recent years \acp{dnn} have achieved remarkable performance. However, their extremely complex internal structures make them incomprehensible as a black box, which is not acceptable in some safety crucial application domains, such as medical diagnosis, industrial process control and fully autonomous driving \cite{he_extract_2020}. Many approaches have been proposed to interpret and explain the \acp{dnn} in a human acceptable way. One feasible approach is to extract understandable symbolic rules from neural networks. The study on rule extraction dates back to early 90s, which originally aimed to synthesize knowledge for knowledge-based systems and discover the full potential of neural networks \cite{andrews_survey_1995}. It is basically recognized that acquired knowledge during training is encoded as the network parameters, activation and architecture design. Knowledge could be extracted in form of If-Then rules, M of N rules \cite{towell_extracting_1993}, decision tree, decision table, etc. \cite{sethi_extended_2012}.  Based on whether extracted rules reveal the internal structure of neural networks, Craven and Shavlik \cite{craven_using_1994} proposed a classification of rule extraction into two basic categories, namely the decompositional and pedagogical approaches. The former \cite{benitez_are_1997, limin_fu_rule_1994,  tsukimoto_extracting_2000, sato_rule_2001} extracts the rules by analyzing the activation and parameters of the layers, while the latter \cite{craven_extracting_1995,schmitz_ann-dt_1999, sethi_kdruleex_2012, thrun_extracting_1994, thrun_extracting_1993, taha_symbolic_1999, zhi-hua_zhou_statistics_2000} extracts rules by mapping the input-output relationships as closely as the neural networks understand without considering internal structure. As illustrated in \autoref{fig:Compare_deco_pad} (top), the decompositional approach  extracts the rule $IF \enspace \sigma({x_1}) \wedge \sigma({\neg{x_2}}) \wedge \sigma({x_3}), THEN \enspace h_2, h_4$ from the first layer and the rule $IF \enspace \sigma(\neg{h_1}) \wedge \sigma({{h_2}}) \wedge \sigma{(\neg{h_3})} \wedge \sigma({{{h_4}}}) \wedge \sigma{(\neg{h_5})} , THEN \enspace y_1$  from the second layer. As we aggregate them and eliminate $h_2, h_4$, the rule mapping between input layer and output layer is extracted as $IF \enspace \sigma({x_1}) \wedge \sigma({\neg{x_2}}) \wedge \sigma({x_3}), THEN \enspace y_1$. In contrast, the pedagogical approach (\autoref{fig:Compare_deco_pad}, bottom) treats the whole network as a black box and learns the rule mapping directly by observing the inputs and outputs.
Besides that, a third class could be the eclectic approach \cite{towell_extracting_1993, setiono_fernn_2000}, which is a combination of both decompositional and pedagogical approaches. Since our goal is to understand \acp{nn} by analyzing their internal structure, we will only focus on the decompositional approach in this subsection. 

Though Neural Networks have been widely studied and applied, there is still a lack of research on rule extraction from \acp{dnn} . Most proposed algorithms target shallow neural networks \cite{hailesilassie_rule_2016, calders_deepred_2016}, except for the work by Zarlenga et al. \cite{zarlenga_efficient_2021}. The reasons can be attributed to two aspects. First, the number of extracted rules is proportional to the depth of the network. \cite{calders_deepred_2016} applied their algorithm to a four-layer \ac{nn} with structure of \textit{784-10-5-2} and extracted over $10^5$ rule terms and consumed large amount of computation time and memory space. Nowadays, even the first version of \acp{dnn} AlexNet consists of 8 layers, state-of-the art \acp{nn} are often designed over 100 layers which makes decompositional rule extraction intractable without pruning.  Secondly, the interpretability of the rules will be lost if the number of rules is too large. Many End-to-End \acp{nn} are applied directly on perceptional level which take the raw sensory data as the input. Although Guido Bologna’s \cite{bologna_simple_2019} algorithm extracts rules from a simple convolutional neural network for image classification task, the relationship between individual input pixel and the output class does not explain the decision process very well. After taking into account these considerations, we argue that the use of the decompositional approach on the top layers of deep neural networks \acp{dnn} will result in more interpretable extracted rules. In order to enhance the readability and comprehensibility of the extracted rules, it would be beneficial to develop a rule visualization tool that enables users to explore and inspect them. This tool would provide a more intuitive and user-friendly interface for understanding the relationships between interpretable inputs variables and the prediction. 

\begin{figure}[t]
    \centering
        \includegraphics[scale = 0.7]{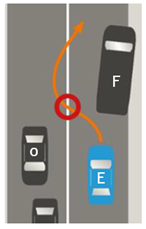}
    \caption{A controlled rule exception scenario}
    \label{fig:controlled_RuleEx}
\end{figure}

The decompositional approach helps to understand the classification process of \ac{dnn}. In the context of autonomous driving, the reason for identification of a rule exception can be extracted as human understandable knowledge. We consider the following specific driving scene as shown in \autoref{fig:controlled_RuleEx}:  an ego-vehicle E (blue marked) drives along a lane which is partially blocked by an obstacle F (black marked) and limited to the right by a solid line. A rule exception of crossing solid line is allowed if it is not foreseeable when the obstacle F will disappear. Before path planning and legal rule integration, the class of the situation should be correctly predicted and explained. Given a well-trained \ac{nn} and a set of examples as the input, the goal of the rule extraction is to find rule sets $R^v_{i\rightarrow o}$ that describe the relationship between the input variables $i$ and the output variables $o$ for each class $v$ as closely as possible. In this section, we will introduce two approaches for extracting rules that are potentially applicable to this problem. 

One approach to extracting rules from a trained neural network is to generate a set of decision trees based on the training samples and their labels. These decision trees can then be converted into IF-THEN rules. This method was first introduced by Sato et al. in their CRED \cite{sato_rule_2001} algorithm, which was designed for three-layered networks. CRED uses decision trees to learn the relationship between the hidden layer and the output layer by analyzing the activation of the units, and then builds new decision trees to learn the mapping between the activations and the input layer. Finally, the rules are merged to describe the overall behavior of the neural network. Zilke et al. extended this approach to networks with more than two layers in their DeepRED \cite{calders_deepred_2016} algorithm, making it the first algorithm capable of handling deep networks, despite its large computational complexity and time-consuming nature. A more advanced algorithm called Eclaire \cite{zarlenga_efficient_2021} was proposed by Zarlenga et al. in 2021. Compared to DeepRED, Eclaire has significant advantages in terms of efficiency of rule extraction and less rule complexity. Eclaire focuses only on the mappings between each hidden layer and the output layers, thus utilizing the distinct latent representation of the data and extracting rules in parallel. This approach avoids waiting for computational results from previous layers, resulting in significantly accelerated extraction processes and more interpretable rules. Although the rules generated by Eclaire do not reveal the topology of the network, the extraction process itself has been significantly improved, resulting in faster and more interpretable rules.  As mentioned before, when applying rule extraction, the input layer must not be the example itself but the features of intermediated layers in the \ac{nn}. For classification of rule exception scenarios problem, the input can be the result of different detectors and sensor data, and the output can be the class of one or multiple defined use cases.
\begin{figure}[t]
    \centering
        \includegraphics[width=0.4\textwidth]{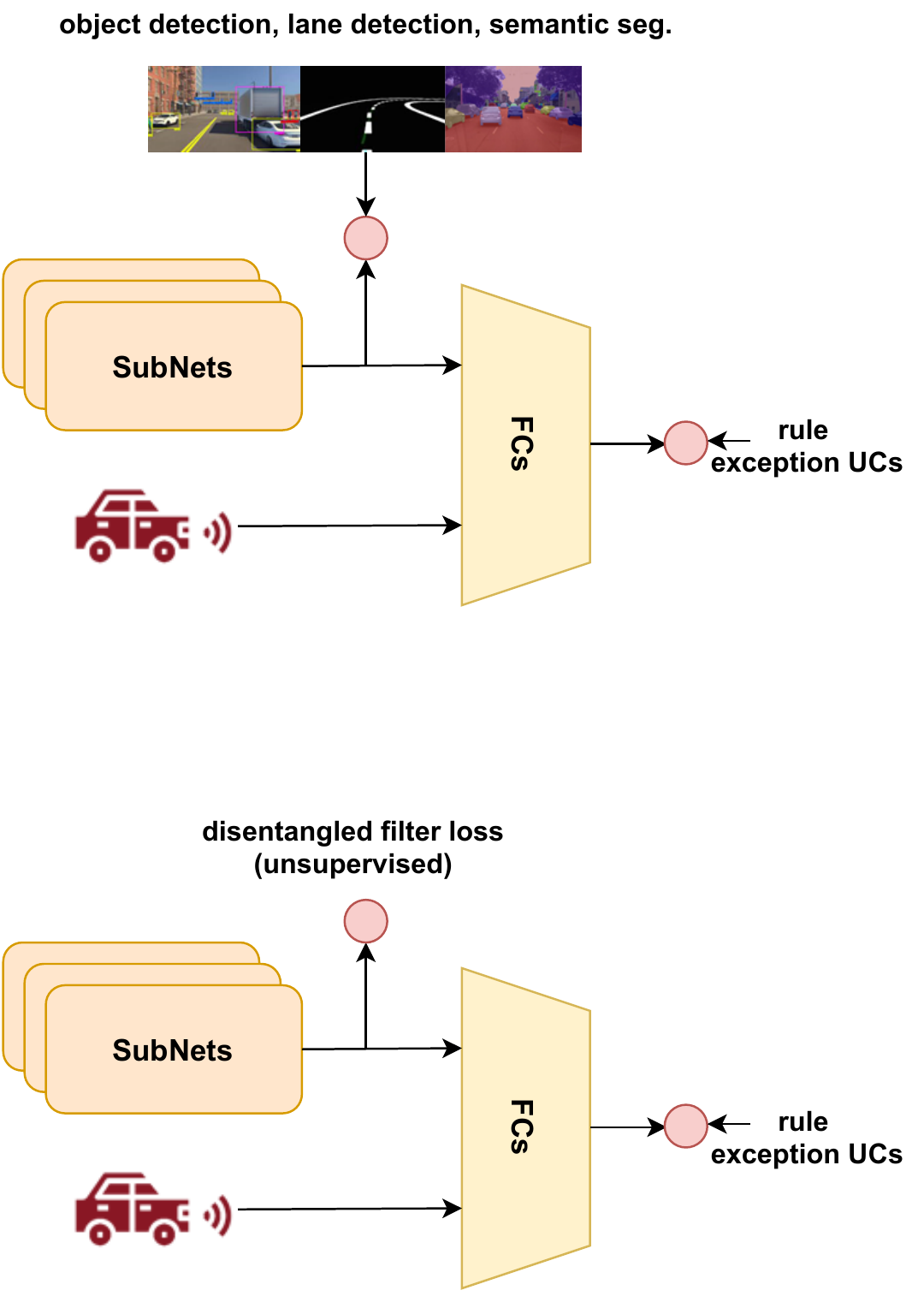}
    \caption{Comparison of two different approaches for rule extraction from FCs in neural network models (top: DeepRED, bottom: ICDT)}
    \label{fig:DeepRed_ICDT.pdf}
\end{figure}

\begin{figure}[t]
    \centering
        \includegraphics[width=0.4\textwidth]{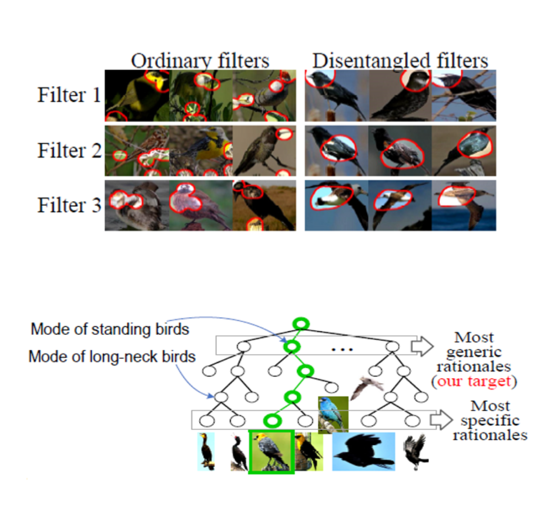}
    \caption{top: Comparisons between ordinary CNN feature maps and
disentangled feature maps; bottom: Decision tree that encodes all potential decision modes of the CNN in a general-to-specific manner \cite{zhang_interpreting_2019}}
    \label{fig:ICDT.pdf}
\end{figure}

As shown in \autoref{fig:DeepRed_ICDT.pdf} (top), the FCs (fully connected layers) are built on the top of one or multiple subNets (e.g., \ac{cnn} models). The supervisory signals for different tasks, such as for object detection, lane detection assign semantic meanings to the input of FCs. When applying DeepRED to the FCs, the following rule set is expected to be extracted:
\vspace{0.5cm}

\[IF \enspace isSolid(Line)= True, \] 
\[AND \enspace Speed(VehicleF)=0, \]
\[AND  \enspace LocX(Line)-LocX(VehicleF)>3, \]
\[AND \enspace LocY(VehicleO)-LocY(VehicleE) < 0,\]
\[THEN \enspace class Rule exception 1.\]
In other words, the rule set says a rule exception case is identified, when a left ego-line is solid, an upfront vehicle is stopping and it is more than 3 meter right from the solid line and all oncoming vehicles are already behind the ego-vehicle. In this case, an overtaking of the upfront vehicle is allowed. These rationales explain the decision very well, however, we still need ground truth data to supervise different detection tasks.  To overcome this problem, an approach proposed by Zhang et al. \cite{zhang_interpreting_2019} is potentially transformable to this task. Instead of using additional supervisory signal, the author designed specific filter loss \cite{zhang_interpretable_2018} (\autoref{fig:DeepRed_ICDT.pdf}, bottom) that disentangles the feature representations in the top layers. It is commonly recognized that, an ordinary filter in the top conv-layers usually represent the pattern that is the mixture of parts and textures. Using such disentangled filters we make each filter represent a specific object part (\autoref{fig:ICDT.pdf}, top). Different from the first approach, the author designed a clustering algorithm on the FCs to derive a decision tree to explain the decision process (\autoref{fig:ICDT.pdf}, bottom). 
Despite its advantages of no need for additional supervisory signal for object parts, it requires adaption of the \ac{nn} structure and finetuning of FCs. In the case of lane detection, disentangled filters may not work, since the texture of the lane lines plays key role in distinguishing. Another shortcoming as claimed by the author is that it is not compatible with the Nets that have skip connections, such as ResNet. Though the first approach requires additional annotations for different tasks, the symbolic rule sets extracted from fully connected layers are more expressive and easier to understand.
\subsubsection{Inductive Logic Programming}
While modern \ac{ml} approaches based on deep learning have been
extremely successful in recent years, they still often fail to
generalize on seemingly simple tasks. For example, given enough training
data a standard deep learning model learns the less-than relation on
handwritten numbers, but it will fail to generalize it to previously
unseen pairs of digits. \ac{ilp} is a set of
techniques for learning logic programs from given examples. Instead of
relying on statistical mechanisms, \ac{ilp} is based on logical inference
while incorporating ideas from automated reasoning and knowledge
representation. This way \ac{ilp} programs are often able to generalize from very small datasets. Another key advantage of \ac{ilp} is the fact that its solutions are comprehensible and verifiable.

While different \ac{ilp} frameworks employ different strategies to solve
a given task, most of them agree on the following definition of an \ac{ilp}
problem. An \textbf{\ac{ilp} problem} $(\mathcal{B},E^{+},E^{-})$ consists of
the following data:
    \begin{enumerate}
      \item $\mathcal{B}$ is a set of background assumptions,
      \item $E^{+}$ is a set of positive examples,
      \item $E^{-}$ is a set of negative examples.
    \end{enumerate}
    The goal of \ac{ilp} is to construct a logic program (a hypothesis) that,
together with given background assumptions, entails the positive
examples and rejects the negative examples.\\[1em]

Over the last decades many significant improvements to \ac{ilp} have been
made. Today, it is possible for \ac{ilp} systems to learn recursive programs
and to invent auxiliary predicates from background assumptions
\cite{cropper_inductive_2020}. However, there are a few weaknesses most
\ac{ilp} frameworks have in common. Since \ac{ilp} represents data as logic
programs this requires the user to choose   from a range of available
languages for this task. For example, Prolog is a Turing-complete logic
programming language that is commonly used in ILP. Datalog is a subset
of Prolog, which trades Turing-completeness for efficiency and decidability.
    Similarly to feature selection in classical ML approaches, choosing
the appropriate background assumptions is crucial for the success of \ac{ilp}.

One of the main disadvantages of traditional \ac{ilp} frameworks is
their inability to handle mislabeled data. Recently, there have been
attempts to combine the strengths of deep neural networks with the
benefits of \ac{ilp} by replacing absolute logical reasoning with continuous
values reflecting the confidence of the conclusion. Richard Evans and
Edward Grefenstette \cite{evans_learning_2018} presented a
reimplementation of \ac{ilp} in an end-to-end differentiable architecture
called $\partial$\ac{ilp}. According to the authors $\partial$\ac{ilp} is robust
to mislabeled data (up to $20\%$), while also being able to handle
ambiguous and fuzzy data. However, the search space needs to be
constrained via handcrafted rule templates and all logic programs are
restricted to definite Datalog clauses, disallowing function symbols.
Furthermore, $\partial$\ac{ilp} only allows for two atoms per rule and
predicates of arity of at most two. Shindo et al.
\cite{shindo_differentiable_2021} improve on $\partial$\ac{ilp} by introducing
several new algorithms that deal with more complex programs
including function symbols. For both frameworks scalability remains an
issue. Payani et al. proposed \cite{payani_inductive_2019} a differentiable version of forward chaining using neural network, in which first-order rules are learned without prior specification of a rule template, while maintaining desirable \ac{ilp} features such as being able to learn recursive predicates as well as predicate invention.  Similarly, \ac{nlm} proposed by Matthieu Zimmer et al. \cite{zimmer_differentiable_2021} also avoids the use of rule templates. It combines reinforcement learning and \ac{ilp} in a differentiable framework, while retaining full interpretable solutions. However, only a restricted space of first-order programs is learned and predicate invention is only possible at the loss of explainability of the solutions. Other works such as Yang et al. \cite{yang_learn_2020, yang_differentiable_2017} consider learning rules for a target predicate as a process of searching for relational paths from the subject to the object in a query. These approaches represent predicates as binary tensors over the domain of constants and perform reasoning through chains of tensor products that emulate clauses. By utilizing this technique, these approaches can efficiently learn interpretable and accurate rules from large, complex datasets.
\subsubsection{Applications}
In the context of autonomous driving, extracted symbolic rules reveals the internal functionality of the neural networks, which can explain the result of a scene classification task by identifying key features and their importance. Furthermore, \ac{ilp} can be employed to derive unwritten legal or behavioral rules from a given formalization of traffic rules. This extracted knowledge can afterwards be incorporated into the body of legal or behavioral knowledge. 
\subsubsection{Rule Learning for Planning}
The goal of rule learning is to infer the constraints that govern expert behavior based on observed demonstrations. A difficulty arises from the fact that the rule inference problem is ill-defined -- the rules are not unique. Furthermore, human expert demonstrations are noisy, i.e., suboptimal, and might even occasionally violate rules.

The following sections highlight methods to extract rules from expert data, with applications to traffic rule extraction. The extracted rules can be used as knowledge bases for subsequent algorithms. The presented methods leverage an explicit formalization of the rules, which, in the case of traffic rules, is still an area of ongoing research \cite{esterle_formalizing_2020}.

\paragraph{Parameter Learning}
The task of rule learning can be framed as learning an objective function which includes penalty terms for rule violations. In fact, Ng and Russell argue that "the reward function is the most succinct, robust, and transferable definition of the task" \cite{ng_algorithms_2000}. Under this interpretation, methods from behavioral cloning and inverse reinforcement learning can be used.

\textbf{Behavioral Cloning.} Behavioral Cloning aims at replicating the observed expert behavior in a supervised learning setting. In the context of autonomous driving, the method was notably used by Pomerleau in his seminal work \cite{pomerleau_efficient_1991}. A large body of work has used this approach to learn end-to-end driving policies, which, given sensory information, reproduce the observed system inputs \cite{bansal_chauffeurnet_2018, salzmann_trajectron_2020, ivanovic_mats_2020}. More recently, \acf{gail} has been shown to require less training data, while yielding more robust policies \cite{ho_generative_2016}.

Behavioral Cloning can be used to tune parameteres of explicitly stated rules. In \cite{zeng_dsdnet_2020, liu_deep_2021}, a \ac{mrf} is used to jointly reason over actor trajectories. The \ac{mrf} encodes rules as potential functions that penalize violating settings of nodes, which represent the trajectory choices of the actors present in the scene. To find the combination of trajectory choices with the lowest potential, Loopy Belief Propagation is used, which is a differentiable message-passing procedure. Thus, the weights and parameters of the rules can be adjusted by backpropagation of a loss function. The authors use a standard classification loss with expert data as the ground truth.

In \cite{li_differentiable_2020}, a \ac{dll} is presented that can be used as an additional layer on top of any network, and takes a set of \ac{stl} rules as inputs. Temporal Logics are a popular way to express constraints on trajectories, and have been used to express traffic rules \cite{maierhofer_formalization_2020, esterle_formalizing_2020, arechiga_specifying_2019, cho_learning-based_2018, cho_deep_2019}. The \ac{stl} semantics have the added advantage of being differentiable. The \ac{dll} can thus act as a corrector in the forward pass, increasing rule conformity. In the backward pass, gradients can be propagated to both the underlying network and the \ac{stl} rule parameters.

\textbf{Inverse Reinforcement Learning.} The goal of \acf{irl} is to estimate the objective function governing expert behavior. It was first posed by Kalman \cite{kalman_when_1964} in the context of Inverse Optimal Control, and later by Ng and Russell \cite{ng_algorithms_2000}. Typically, the structure of the objective function is assumed to be known, reducing the problem to a parameter estimation problem. A recent overview is given in \cite{arora_survey_2020}.

\textbf{Maximum Margin \ac{irl}}. The \ac{irl} problem is ill-posed, because multiple objective functions can explain a given expert behavior, including the "always zero" reward. Maximum Margin \ac{irl} overcomes this difficulty by requiring that expert trajectories be optimal by some margin \cite{arora_survey_2020}. This requirement can be encoded as a loss function.

In \cite{zeng_end--end_2021, wei_perceive_2020}, the max-margin loss is used to train a numerical scene-dependent planning cost volume end-to-end. In \cite{sadat_jointly_2019}, it is used to learn the parameters of an objective function that is subsequently used for behavior (coarse-scale) and trajectory (fine-scale) planning. In \cite{vedaldi_perceive_2020}, it is used to learn the weights of an objective function that uses a semantic occupancy grid map to score trajectory candidates. A similar idea is presented in \cite{casas_mp3_2021}, which uses the max-margin loss to learn penalties for a small set of illegal behaviors, including off-road and opposite-direction driving, and running traffic lights.

\textbf{Maximum Entropy \ac{irl}.} Maximum Margin \ac{irl} suffers from a problem known as label bias, which arises from the fact that the algorithm has to commit to what it deems to be the optimal objective function with a margin. Maximum Entropy \ac{irl} overcomes this by maximizing the entropy of the trajectory distribution \cite{ziebart_maximum_2008}. It can be shown that the distributions with highest entropy belong to the exponential family \cite{arora_survey_2020}. The principle can be extended to deep architectures \cite{wulfmeier_maximum_2016}.

\paragraph{Structure Learning}
The previous section presented methods for parameter learning, where the learning problem was framed as an optimization problem over a fixed class of functions with tunable parameters (e.g., the weights of a neural network). However, the structure of rules is often hard to express succinctly and needs to be discovered, requiring a joint optimization over the (typically) continuous parameter space and the (typically) discrete rule space. The following paragraphs highlight methods that bridge this gap.

\textbf{Maximum-Likelihood Constraint Learning.} The maximum entropy principle yields a distribution over trajectories. This can be used to find constraints with maximum likelihood, given an objective function and system model. In \cite{scobee_maximum_2020}, the case of of an \ac{mdp} with markovian constraints is studied. The authors present an algorithm that iteratively maximizes the probability mass of forbidden trajectories, while keeping observed trajectories legal. To prevent overfitting, the process is stopped once the decrease of the \ac{kl} divergence between the observed and inferred trajectory distributions crosses a lower threshold. A similar method is presented in \cite{vazquez-chanlatte_learning_2018}, which uses a surrogate likelihood criterion and leverages inclusion relations to efficiently search the space of constraint candidates.

\textbf{Neural Architecture Search.} Neural Architecture Search attempts to find the optimal structure of network layers for a given task. This problem has many similarities with rule structure learning. A multitude of methods, including Reinforcement Learning, Evolutionary Algorithms, and Gradient Descent can be applied, which are surveied in \cite{elsken_neural_2019}.

A related field is Neural Program Synthesis, which attempts to infer programs based on input-output examples. Successful approaches include the use of sequence-to-sequence models and beam search to generate valid programs in domain-specific languages \cite{parisotto_neuro-symbolic_2016, devlin_robustfill_2017}. An overview of Neural Program Synthesis is given in \cite{kant_recent_2018}.

In \cite{sun_neuro-symbolic_2020}, a gradient descent-based method from neural architecture search is used to learn a symbolic decision program for autonomous vehicles in a domain-specific language. To this end, the space of programs is expressed as a supergraph, whose edge weights define the inferred computation graph. The authors use the \ac{gail} objective to learn the computation graph from expert demonstrations.

In \cite{qu_rnnlogic_2020}, a recurrent neural network is used to generate rule candidates for knowledge graph completion. A survey on knowledge graphs is given in \cite{ji_survey_2021}.
\subsection{Structured Output Prediction}
\label{sec:structured_output_prediction}
\textit{Author: Christian Hesels}

\smallskip
\noindent
Pattern Mining is used to discover unexpected and useful patterns in a database. The resulting patterns are human interpretable and thus useful for data understanding and decision making. Pattern Mining first was introduced to create association rules for market basket analysis \cite{agrawal_mining_1993}. The subsequent algorithms built up on this idea and are generally summarized as Frequent Pattern Mining. Algorithms like association rules only consider the occurence of itemsets and they are applied to non sequential databases. This is called Frequent Itemset Mining. Another approach in Frequent Pattern Mining is Sequential Pattern Mining, where not only the amount of occurences is taken into account, but also the order determined by a timestamp in a given database. A slightly different approach but yet very close to the mentioned ones is Time Series Pattern Mining. While Frequent Itemset Mining and Sequential Pattern Mining receive a symbolic representation as input data, Time Series Pattern Mining is used to find patterns in high dimensional data \cite{fu_review_2011}. Generally, sorted episodes or sequences are defined as serial, not sorted ones are called parallel and if both is possible epsiodes are named general.
\subsubsection{Pattern Mining}
Frequent Itemset Mining algorithms like Association Rule Mining searches for if.. then.. statements for Items. Which in the context of Basket Case Analysis means, if a customer buys one item then he buys another one with a certain confidence. This rules are not limited to single cardinality, but also work for more combinations. To find these rules, Association Rule Mining calculates three metrics: Support, Confidence and Lift. The support is calculated from the item count $N$ and the frequency of two items $freq(A,B)$ occuring together. Confidence gives information about how often the items $A$ and $B$ occur together. Lift is calculated from the support of $A$, $B$ and the global support.
\newline
Apriori is an algorithm to optimize this principle. Because with this Association Rule Mining approach thousands of rules would be generated without pruning the input, it takes a lot of space and iterations to mine those rules. Apriori is based on the concept, that a subset of a frequent itemset must also be a frequent itemset. Therefore when a subset does not fullfill a given minimum support every itemset in this subset might as well be removed from the dataset \cite{agrawal_fast_1994}.
\newline

For Sequential Pattern Mining the definitions are the following: an event is defined by a set of items $i$ with the length $k$, i.e., $(i_1, i_2, ..., i_k)$, often in combination with a timestamp. A sequence is a set of events and a sequence with $k$-items is called a $k$-sequence. The events are taken from a database $D$. The support (or frequency) is given by $\sigma(\alpha, D)$. When a sequence pops up more often than the user defined minimum support value, it is considered a frequent sequence. Often a user defined Time Window is given, in which range the patterns need to appear in. Otherwise frequent patterns from huge sequences would be mined, that may not have any correlation.
\newline

Time Series Mining has two different approaches to generate a suitable data representation. One is to reduce the dimension of the input data with methods like basic sampling or critical point model \cite{bao_intelligent_2008}. Yet the goal remains to reduce the dimensions and calculate the distances between multiple time series to cluster them. The second approach is to discretize the data to receive a symbolic representation. On this symbolic representation, algorithms from Sequential Pattern Mining are applied. Measuring the similarity of time series is done in two ways. Whole sequence matching and subsequence matching. To calculate the distance between the time series the nearest pairs need to be saved and the distance measurement has to be done with all possible offsets \cite{fu_review_2011}. One way to do it for whole sequence matching is to minimize the dimension as mentionend before (for example with DFT coefficients \cite{agrawal_efficient_1993}) and calculate the Euclidean distance on it. In subsequence matching, a query is given and the goal is to find it in the longer time series. The subsequence therefore needs to be compared to every offset in the longer time series. There are a lot of ways to do this like DualMatch \cite{yang-sae_moon_duality-based_2001} or GeneralMatch \cite{moon_general_2002}. The final pattern mining is mostly done with clustering. An initial cluster center is defined (random or from a sequence). As a parameter the amount of clusters is given (static or variable). Afterwards the distance of each input datapoint to the cluster is calculated and the closest cluster wins and updates the cluster center. The algorithm either converges or finishes when the maximum iteration number is reached \cite{fu_review_2011}.

\subsubsection{Applications in Situation Interpretation}
Because the input is a high level symbolic representation, Time Series Pattern Mining is not suitable for this approach (except sensor data is used, in which case a suitable sampling method is needed to reduce the complexity). Assuming that in traffic a sequence is of importance, for example stopping the car before a pedestrian is in front of the car, Sequential Pattern Mining is reasonable. State of the Art algorithms like CM-Spade are used to extract the patterns \cite{fournier-viger_fast_2014}. These extracted Patterns (complex events) are used to train a transparent classifier for action/event prediction. For this approach the training data is created in a way where a certain time window is given and every time the events to predict occur, it is defined as a label and the previous events (depending on the time window) as training data.
\newpage
\subsection{Natural Language Processing for Legal Domain}
\label{sec:nlp_for_legal_domain}
\textit{Authors: Stefan Griesche, Anh Tuan Tran}

\smallskip
\noindent
Previous sections about knowledge representation learning and knowledge crafting highlighted knowledge graphs and ontologies as a feasible approach to represent knowledge. These concepts rely on annotated data. For smaller data sets and proof-of-concepts this can be achieved by a human annotator or crafting by hand. However, legal knowledge in form of norms, traffic regulations and laws sum up to a large amount of unstructured data in different languages. \ac{nlu} and \acf{nlp} provide  methods to tackle this large amount of data. However, for the legal domain additional challenges have to be considered. As mentioned by \cite{prakken_problem_2017}, computational models which deal with legal texts and traffic regulations have to take exceptions, rule conflicts, open texture and vagueness, rule change, and the need for commonsense knowledge into account. Therefore, existing \ac{nlp} concepts and models have to be tailored or trained to the domain and domain specific solutions have to be developed \cite{robaldo_introduction_2019}. 

\subsubsection{\ac{nlp} tasks}
The adaptation to the legal domain addresses already different \ac{nlp} tasks. This includes legal entity recognition \cite{leitner_dataset_2020}, text similarity \cite{minocha_legal_2018}, \cite{nanda_unsupervised_2019}, legal question answering \cite{zhong_iteratively_2020}, legal summarization \cite{kanapala_text_2019} and pretrained legal language models  \cite{chalkidis_legal-bert_2020}. Zhong et al. \cite{zhong_how_2020} provides a good summary on the benefits of AI in understanding legal text. An additional task that is particularly well-suited to legal \ac{nlp} is argument mining \cite{lippi_argumentation_2016}, in which the arguments are extracted from unstructured documents in a hierarchical form: in the lower level exists linguistic form such as argumentative sentences, argumentation boundary, etc. while in the higher level, the arguments are formed from argumentative utterances and evidence. Further tasks in legal \ac{nlp} are also of recent research interest \cite{noauthor_competition_nodate}, including legal case retrieval, legal case entailment, judicial prediction. Sleimi et al. \cite{sleimi_query_2019} used \ac{nlp} to query legal requirements and support the legal requirement handling.

\subsubsection{Datasets and legal language models}
There is a increasing proliferation of new corpora specific to legal domains such as \cite{poudyal_echr_2020}, \cite{leitner_dataset_2020}, \cite{hendrycks_cuad_2021}, \cite{urchs_design_2021}. Most of these corpora are designed to facilitate a specific task. Recently, there are attempts to provide the data for multiple task training (and with it, a pre-trained model that can be customized and fine-tuned to further downstream tasks), for example CaseHold \cite{zheng_when_2021}, Edgar \cite{borchmann_contract_2020}. Most pre-trained models rely on transformer architectures and provide a lightweight variant, fine-tuning from larger models such as BERT or GPT, namely Legal-BERT \cite{chalkidis_legal-bert_2020} and  Legal-GPT \cite{borchmann_contract_2020}. The Natural Legal Language Processing (NLLP) community \cite{noauthor_nllp_nodate} provides a good amount of references to these corpora. 

\subsubsection{Software}
There is an increasing number of open-source and proprietary software system for legal \ac{nlp}. ICLR\&D (Incorporated Council of Law Reporting for England and Wales) has open-sourced Blackstone \cite{noauthor_blackstone_nodate}, which is a legal \ac{nlp} pipeline built on top of Spacy \cite{noauthor_explosionspacy_2021}. Blackstone \ac{nlp} models consider the nature of legal texts. The named-entity recognizer is for instance able to detect legal concepts such as provisions, citations, the text categorizer annotates axioms, issues, conclusions. LexNLP \cite{noauthor_lexnlp_nodate} offers segmentation, tokenization, pre-trained word embedding, classifiers, fact extraction specific for legal texts. FlairNLP also provides easy to access API for working with deep learning models via HuggingFace and PyTorch. QuantLaw \cite{noauthor_quantlaw_nodate} has recently open-sourced their toolkits for analyzing and detecting legal references from text.

\subsubsection{Research projects}
There have been two European projects in the past years of interest in the legal domain which considered \ac{nlp} techniques and made an essential contribution to the state-of-the-art. The Mirel project \cite{noauthor_mirel_nodate} developed tools for mining and reasoning with legal text. The project aimed to close the gap between the community working on legal ontologies and \ac{nlp} parsers and the community working on reasoning methods and formal logic. The project provided first legal ontologies \cite{batsakis_d31_2017}, \cite{di_caro_d22_2017} and used \ac{nlp} techniques for legal reasoning \cite{dragoni_combining_2016}. The lynx project \cite{noauthor_lynx_nodate} developed  a legal knowledge graph for information retrieval services. Moreover, the lynx project developed an annotation service, published legal datasets and trained language models to the legal domain \cite{leitner_dataset_2020}.

\subsubsection{Applications}
The state-of-the-art shows an increasing activity of research on \ac{nlp} in the legal domain. For automated driving, traffic rules and parameters written in traffic regulation and norms are of utmost concern. However, to the authors' best knowledge, \ac{nlp} for traffic regulations and norms seems currently largely unexplored. Even though there are first \ac{nlp} applications which focus on analyzing requirements and technical documentation (test specifications) in software and system development \cite{schraps_knowledge_2016}, \cite{ferrari_nlp_2019}, the potential of \ac{nlp} is not fully exploited. \ac{nlp} can play an essential role to analyze legal requirements and find similarities and conflicts among traffic regulations and norms of different countries. Furthermore, \ac{nlp} can deliver the metadata to enrich requirements with different sources such as legal commentaries to interpret fuzzy traffic rules. This can be further processed in knowledge graphs or reasoning engines and support the software development process to design for instance legal conformity planner. \ac{nlp} methods also have the potential to support editors such as \cite{dastani_towards_2020} to semi-automate the process of formalizing traffic rules \cite{esterle_formalizing_2020}. Ontologies for traffic regulations introduced by \cite{bou_ontology-style_2020} or \cite{buechel_ontology-based_2017} provide first formal descriptions. Using this as an input, \ac{nlp} methods can be seen as a part of a pre processing to accelerate / to automate the process of formalization legal statements into logical statements by supporting the understanding of legal texts through a semantic annotation. These logical statements  can be further processed and integrated into hybrid \ac{ai} models for situation understanding or planning as described in \autoref{ch:integration}.
\vspace{30pt}

\subsection{Question Answering}
\label{sec:visual_question_answering}
\textit{Author: Tianming Qiu}

\smallskip
\noindent
\acf{qa} is a \ac{nlp} task of answering questions based on a given certain question and the corresponding context.
As a popular task towards general artificial intelligence, the algorithm is designed to understand the context and grasp some human knowledge.
A step further, \ac{vqa} extends question answering by combining it with computer vision tasks.
The \ac{vqa} task aims at answering a given question based on images.
Hence, \ac{vqa} approaches are required to possess not only language understanding but also image reasoning ability.
Answers predicted by a \ac{vqa} model can be seen as knowledge in visual scenes.
The more questions can be answered, the more knowledge the model extracts from images.

\subsubsection{\ac{vqa}: a combination of \ac{nlp} and \ac{cv}}
Majority \ac{vqa} tasks are trained using supervised learning: an image and a language feature extractors are applied to input containing image-question pairs.
Based on certain feature fusion procedures, the succeeding step is to forward the image-language features via another network module to get the final output answer.
The whole model is trained based on the loss of output answers and ground-truth labels.
Therefore, mainstream \ac{vqa} algorithms investigate different modifications mainly on the following three perspectives: 
(i) Image or language feature extraction;
(ii) Fusion of image and language features;
(iii) Network model for final answer. prediction. 

Since \ac{vqa} tasks are extended from \acf{qa} tasks, a language model like \ac{rnn} or \ac{lstm} model is often used as both the language feature extractor and the final language output generator.
Based on the intrinsic connection with \ac{nlp}, it is common to see many techniques developed by \ac{nlp} models are also applied in \ac{vqa} models.
Similarly, regarding the image processing part, various \ac{cnn}-based backbones are widely used.
As an intersection of \ac{nlp} and \ac{cv} techniques,
A key issue in the \ac{vqa} problem is how to fuse image features together with language representation.
Two-modal feature fusion approaches are summarized in a comprehensive survey work~\cite{wu_visual_2017}.
A basic method is the joint embedding approach which maps image features and language features to a shared space with same dimension.
Other approaches apply the attention mechanism to find the most relevant part between two modalities for final answers.

Since Transformer~\cite{vaswani_attention_2017} has been widely used in \ac{nlp} and \ac{cv} communities, the \ac{vqa} community is also looking for breakthroughs with the help of this powerful framework.
The original Transformer~\cite{vaswani_attention_2017} is a sequence-to-sequence \ac{nlp} model which uses a self-attention mechanism.
For a input sentence, the Transformer encoder calculates the relevance among tokens in this input sentence via the inner product.
Such a relevance matrix is called a self-attention matrix.
In the \ac{vqa} task, the Transformer structure replaces the traditional \ac{nlp} model such as \ac{rnn} or \ac{lstm}.
And the self-attention between language tokens and image feature units provides a effective fusion mode due to its semantic robustness of its inner product.
A unified Transformer model is used for a multi-task learning which includes \ac{vqa} task~\cite{hu_unit_2021}.
Another large scale model from OpenAI called Contrastive Language-Image Pre-training, namely CLIP~\cite{radford2021learning}, also works on image text feature fusion. Although it doesn't work on the \ac{vqa} task directly, it demonstration an effective way to fuse image and text features.

Another task called \ac{wsol}~\cite{zhang_weakly_2021} is also related to \ac{vqa}.
Weakly supervised learning refers to a training process when only partial information regarding the task (e.g., class label or bounding box) on a small dataset is available.
WSOL refers to the object detection task with insufficient ground-truth bounding boxes.
Without bounding box labeling, categorical token or language query is helpful for object localization~\cite{gao_ts-cam_2021}.
Token Semantic Coupled Attention Map (TS-CAM)~\cite{gao_ts-cam_2021} first splits an image into a sequence of patch tokens for spatial embedding and then re-allocates category-related semantics for patch tokens, enabling each of them to be aware of object categories.
As a multi-modal task, these language-query-based WSOL problems have a strong connection with the \ac{vqa} tasks.

\subsubsection{Applications}
If we take language-based \ac{wsol} also as \ac{vqa} problems, then it is related to the perception topic.
Otherwise, \ac{vqa} is usually not relevant to perception.
\ac{vqa} tries to understand the image and question at the same time, hence, from this perspective, all the papers mentioned above are related to situation understanding.
Regarding the autonomous driving scenario, an autonomous driving \ac{vqa} dataset called ISVQA \cite{bansal_visual_2020} queries a set of images instead of one single image for answers.
The set of images are taken by six cameras mounted at different positions of a vehicle. 
Traditional \ac{vqa} methods are deployed on this ISVQA dataset, while Transformer architecture achieves better performance.
\ac{vqa} methods provide a tool to extract more knowledge from a given autonomous driving scene.
\section{Knowledge Extraction - Visual Explanations}
\label{ch:visual_extraction}

Visualization is a great way to represent abstract and complex knowledge in impressive form. Therefore, it can also serve as an effective tool to demonstrate knowledge extracted from machine learning models. Unlike text, formulas, or other symbolic interpretations, visualizations can provide humans with an intuitive impression of some complex information. Better understandings and even algorithm improvements can be inspired by the visualization. 

Much research has focused on the visual interpretation of machine learning models. One of them, called \textit{visual analytics}, aims to use the human visual system and human knowledge to identify or hypothesize patterns that are often hidden in large datasets. Different methodologies of visual analytics are discussed in \autoref{sec:visual_analytics}. 
\textit{Saliency map} methods in \autoref{sec:saliency_maps} emphasize how to generate heatmaps on input images for computer vision tasks. 
In \autoref{sec:interpretable_feature_learning}, \textit{interpretable feature learning} demonstrates how visualization results can be used to generate ad-hoc explanations or numerical evidence. All these visual explanation offer semantic level knowledge extraction approach for better interpretability.

\subsection{Visual Analytics}
\label{sec:visual_analytics}
\textit{Authors: Kostadin Cvejoski, Johann Kelsch}

\smallskip
\noindent
\ac{va} is a field of research that defines and investigates processes that use the human visual system and human knowledge for identifying or hypothesizing about usually hidden patterns in big datasets. 
While this encompasses a number of techniques that have long been used in data visualization and therefore applied statistics and data science, the need for thorough examination of complex machine learning models like neural networks has driven the development of \ac{va} methods in this field as well.

The actual deployment of \ac{va} in the \ac{ml} lifecycle is flexible as is the basis and the goal of its investigation.
In the scope of machine learning, \ac{va} investigates and develops both the methodology and organizational processes, i.e., feedback loops among different stages of the \ac{ml} development lifecycle.
For the scope of this concise introduction, we refer to the CRISP-ML Lifecycle and focus on the stages Data Preparation, which includes acquisition and labelling of data, Model Engineering, which includes architecture search and training regimen, and Quality Assurance.
First and most straightforward, \ac{va} can be used as quality check for the Data Preparation stage. 
Investigations focus on comparing data points and checking consistency of related labels in similar or equal situations.
At this point, \ac{va} is also often used to perform dataset analysis, e.g., they investigate and quantify imbalanced distributions and recognize missing cases which is, hence, fed back to the data acquisition process.
Insights can directly be reported and taken into account within Data Preparation stage itself.
Second, by use of a trained model, \ac{va} allows for identifying merits and shortcoming of \ac{ml} models and cluster semantic situations in which they succeed or fail systematically. 
These findings can subsequently be fed back to the Model Engineering stage, specifically to the model definition or model training process.
However, if it becomes apparent that a desired analysis cannot be performed due to lack of input data or insufficient annotations, a corresponding request can be issued to the Data Preparation stage.
For example, \ac{va} can support the labelling process in suggesting meta data for future analysis and straightforward search and selection operations. 
This touches upon the field of active learning (c.f.~\autoref{sec:active_learning}) in which an interplay of a human operator and a trained machine learning models drives the selection of new data points and its labels.
Likewise, intuitive visualization techniques bear overlap with the field of Explainable \ac{ai} (c.f.~\autoref{sec:interpretable_feature_learning}) which as one part aims to derive clear visual representations for hard-to-interpret \ac{ml} models.

Shneiderman \cite{shneiderman_eyes_1996} summarizes the levels of \ac{va} process in his mantra: Overview first, zoom and filter, then details-on-demand. 
This defines the main components of any \ac{va} approach: means for visualization, selection and search.
For this brief overview, we restrict ourselves to forms of \ac{va} that go in line with knowledge extraction or plausibility checking. 
These usually includes data-driven machine learning but we explicitly do not cover visualization methods as they have long been established in data science.
Instead we focus on use cases relevant for automated driving which, by the nature of the field, puts its main emphasis on environment perception, in particular computer vision tasks.

\subsubsection{Surveys of Visual Analytics}

We cite Sacha et al.~\cite{sacha_vis4ml_2019} who provide an overview and a fundamental categorization of known \ac{va} workflows.
Hohman et al.~\cite{hohman_visual_2019} provide an extensive survey over \ac{va} techniques for the most common use case of deep learning.
As one prime example for \ac{va} in the field of safety in automated driving, albeit without a special focus on machine-learning-driven vehicle functionality, we refer to the toolset SafetyLens by Narechania et al.~\cite{narechania_safetylens_2020}.
Chatzimparmpas et al.~\cite{chatzimparmpas_state_2020}, on the other hand, focus on machine learning approaches but open the field of application to safety-critical problems in general. Yuan et al.~\cite{yuan2021survey} provide a survey of \ac{va} Techniques for \ac{ml} where the first-level categories in their proposed taxonomy are: techniques before model building, techniques during model building, and techniques after model building. Further, for each category they provide example analysis task and the most recent works.

\subsubsection{Visual Analytics in Environment Perception}

In particular computer vision but also 3d lidar or radar perception are prime examples for the use of \ac{va} in robotics and automated driving.
Gou et al. \cite{gou_vatld_2021} develop and investigate a \ac{va} system for their sophisticated traffic light classification model.
Computer vision models in automated driving are investigated by Bojarski et al.~\cite{bojarski_visualbackprop_2016} who present a real-time capable visualization method for \acp{cnn} and verify its potential on several well-known datasets from automated driving.
Similarly, Wang et al.~\cite{wang_cnn_2020} present their \ac{cnn} Explainer which aims explicitly at teaching the functionality and stepwise transformations of convolutional neural networks.
Liu et al.~\cite{liu_analyzing_2018} target the susceptibility of deep neural networks to adversarial attacks and allow for visualizing the set of neurons in the network that were fooled by the attack.
In doing so, they enable the human operator to find prevalent adversarial attacks to a network and work out techniques to build up robustness against them.
Ma et al.~\cite{ma_visual_2020} address the problem of domain shift and aim to reveal the knowledge that two models share due to transfer learning.

\subsubsection{Deriving Symbolic Knowledge}
 
Due to their need for straightforward and low-dimensional data representations, \ac{va} is often deployed to derive human-understandable rules in datasets or model behaviour. 
Cao and Brown~\cite{cao_dril_2020} developed a system that combines machine-learning-based rule deduction with a visualization of the corresponding data and the degree of satisfaction of automatically derived rules. 
This enables the human operator to identify promising, semantically meaningful relations without being an expert in machine learning.
Xie et al.~\cite{xie_visual_2020} present a system for the causal analysis of predictions with a small number of semantically meaningful features.
Their system is able to incorporate and convey uncertainty of the decision making process and aims to help decision-makers with multi-causal data. Andrienko et al.~\cite{andrienko2021theoretical} provides theoretical definition for the concept of a pattern as a combination of multiple interrelated elements of two or more data components that can be represented as whole object. This definition raises a range of interactive analytical operation for discovering knowledge. Ge et al.~\cite{ge2021peek} propose the visual reasoning explanation framework (VRX) that gives causal understanding of a model's inference step. The method has the main contributions: (i) understanding to what an \ac{nn} pays attention to using high level concept graphs and their relationships; (ii) to explain the \ac{nn}'s reasoning process they use \ac{gnn}-based graph reasoning network. In their paper they also empirically show that their proposed framework can answer questions like "why" and "why not".

\subsubsection{Applications}

\ac{va} is a field where visual techniques are developed to better understand the data, the building of the models and the decision the models make. In a safety critical domain like autonomous driving, transparency and accountability of the models is of great importance. The increased utilization of \ac{dnn} in the autonomous driving domain requires developing techniques that will give answers to the questions "Why" and "Why not" the model came to certain decision. Using \ac{va} techniques in all stages of the \ac{ml} pipeline can help experts gain better understanding of the strengths and weaknesses of the object detection models they build. \ac{va} systems opens a door for building human-in-the-loop \ac{ai} systems which will help to develop better models for perception. It will give us visual explanation for example "why an object was detected as person" and why not as a dog. By using techniques from \ac{va} we will be able to build better dataset for perception, situation interpretation or planning tasks (\cite{bauerle2020classifier}, \cite{bernard2017comparing}, \cite{chen2020oodanalyzer}, \cite{junior2017analytic}).
\subsection{Saliency Maps}
\label{sec:saliency_maps}
\textit{Authors: Christian Hellert, Franz Motzkus}

\begin{figure}[t]
    \centering
    \includegraphics[width=\columnwidth]{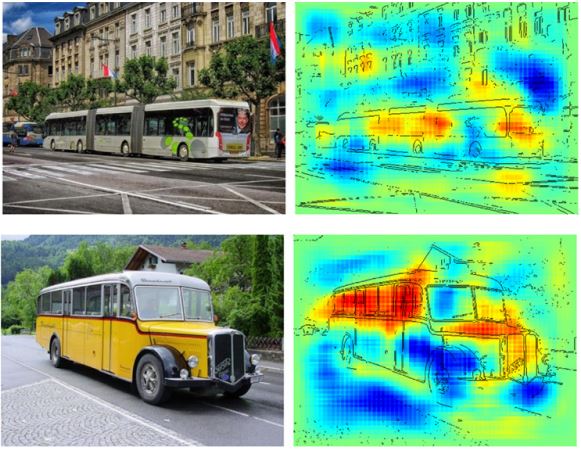}
    \caption{Examples of saliency maps for the task of image classification~\cite{bach_pixel-wise_2015}. Note that reddish and bluish pixels represents positive and negative attribution respectively.}
    \label{fig:example_saliency_maps}
\end{figure}

\smallskip
\noindent
Methods for heat mapping, pixel attribution or saliency maps are methods to obtain interpretable visual representations for individual inputs. These methods are generating heatmaps that indicate the relevance of the input pixels for a certain input sample or a set of input samples towards the output. \autoref{fig:example_saliency_maps} shows two examples of saliency maps. Thereby, these maps can be directly used for manual interpretation or debugging. According to \cite{schwalbe_xai_2021}, there are two kinds of methods regarding locality, which are local and global locality in the context of~\ac{xai}. Local methods explain a \ac{ml} model only on a subset of samples or for a specific behavior on a restricted input data space. Saliency map methods are belonging to the local locality category and often local methods are used on very similar samples to evaluate if a model uses the same clues for a prediction. For example, when classifying pedestrians, specific regions like the head or the body are assumed to be important for the classification result. Global \ac{xai} methods try to explain the behavior of a complete \ac{ml} model. As an example, for the task of image classification, a class is expected to consist of multiple parts. Each of these parts is treated as a concept and the global method will reveal the shared concept between all samples of a class. The concepts themselves can be semantically interpretable and in the case of the class vehicle possible concepts could be tires, windows, or lights. \autoref{sec:interpretable_feature_learning} provides more details on global \ac{xai} methods.

Furthermore, saliency map methods can be model-specific or model-agnostic \cite{molnar_interpretable_2019}. Model-specific methods can only be used for a specific kind of model, e.g., a \ac{cnn}, while model-agnostic methods treat the \ac{ml} model as a blackbox. Typically, model-specific methods use gradient-based and model-agnostic methods use perturbation approaches. \autoref{tab:overview_sailency_map} gives an overview of the saliency map methods along with the type and the task, which will be discussed in the following sections.

\begin{table*}[t]
	\centering
	\caption{Overview of saliency map methods}
	\begin{tabularx}{\textwidth}{|X|X|X|l|}
		\hline
		\textbf{Method} & \textbf{Type} & \textbf{Task} & \textbf{Reference} \\
		\hline
		Image-Specific Class Saliency & specific & Classification & \cite{simonyan_deep_2014} \\ \hline
		DeconvNets                    & specific & Classification & \cite{zeiler_visualizing_2014} \\ \hline
		Guided Backpropagation        & specific & Classification & \cite{springenberg_striving_2014} \\ \hline
		CAM                           & specific & Classification & \cite{zhou_learning_2016} \\ \hline
		Grad-CAM                      & specific & Classification & \cite{selvaraju_grad-cam_2016} \\ \hline
		SIDU                          & specific & Classification & \cite{muddamsetty_sidu_2020} \\ \hline
		Grad-CAM++                    & specific & Classification & \cite{chattopadhay_grad-cam_2018} \\ \hline
		Score-CAM++                   & specific & Classification & \cite{wang_score-cam_2020} \\ \hline
		SmoothGrad                    & specific & Classification & \cite{smilkov_smoothgrad_2017} \\ \hline
		Integrated gradients          & specific & Classification & \cite{sundararajan_axiomatic_2017} \\ \hline
		XRAI                          & specific & Classification & \cite{kapishnikov_xrai_2019} \\ \hline
		DeepLIFT                      & specific & Classification & \cite{shrikumar_learning_2017} \\ \hline
		LRP                           & specific & Detection      & \cite{bach_pixel-wise_2015} \\ \hline
		Deep Taylor Decomposition     & specific & Classification & \cite{montavon_explaining_2017} \\ \hline
		CRP \tablefootnote{Contrastive Relevance Propagation}  & specific & Detection      & \cite{tsunakawa_contrastive_2019} \\ \hline
            CRP \tablefootnote{Concept-wise Relevance Propagation} & specific & Classification & \cite{achtibat_where_2022} \\ \hline
            L-CRP                         & specific & Detection      & \cite{dreyer_revealing_2022} \\ \hline
		LIME                          & agnostic & Classification & \cite{ribeiro_why_2016} \\ \hline
		RISE                          & agnostic & Classification & \cite{petsiuk_rise_2018} \\ \hline
		D-RISE                        & agnostic & Detection      & \cite{petsiuk_black-box_2021} \\ \hline
		Anchors                       & agnostic & Classification & \cite{ribeiro_anchors_2018} \\ \hline
		SHAP                          & agnostic & Classification & \cite{lundberg_unified_2017} \\ \hline
		Explain to Fix                & agnostic & Detection      & \cite{gudovskiy_explain_2018} \\ \hline						
	\end{tabularx}
	\label{tab:overview_sailency_map}
\end{table*}

\subsubsection{Model-specific Methods}

One of the first saliency map methods was introduced by Simonyan~et~al.~\cite{simonyan_deep_2014} and called image-specific class saliency or often called vanilla gradients. The method generates the saliency map for classification tasks by computing the derivative of the class score with respect to the input image. The interpretation is that the method seeks for the input pixels that needs to be changed the least and affects the class score the most, which is equal to the magnitude of the class score derivative. One problem of the image-specific class saliency method is the saturation problem~\cite{shrikumar_learning_2017}, which was solved by~\cite{zeiler_visualizing_2014}, by redefining backpropagation for ReLU layers, so that only positive error signals are backpropagated. In addition, Springenberg~et~al.~\cite{springenberg_striving_2014} combined both approaches, resulting in a method called Guided Backpropagation.

Another method was introduced by Selvaraju~et~al.~\cite{selvaraju_grad-cam_2016} and is called Gradient-weighted Class Activation Map (Grad-CAM) and is able to generate saliency maps for different tasks as long as a convolutional layer is involved in the architecture. Typical tasks involve image classification, visual question answering and image captioning. The method starts by performing a forward step followed by setting the class gradient of interest (layer before soft-max) to one and all others to zero. Then, the gradients are backpropagated till the first occurring convolutional layer. The gradients are then globally averaged pooled and afterwards, a ReLU is applied to obtain a class saliency map. Grad-CAM can also be used for object detection without modification, but for very small objects that are very close nearby, it will not be possible to distinguish between them, since Grad-CAM stops at the last CNN-layer, which has a considerably lower resolution compared to the input data or image. Therefore, some of the spatial information gets lost. Note that Grad-CAM is the generalization of CAM \cite{zhou_learning_2016}. There are also some improved versions of Grad-CAM: Chattopadhay~et~al.~\cite{chattopadhay_grad-cam_2018} propose a method that gives better object localization and improved heatmaps, when multiple object instances occur and in \cite{muddamsetty_sidu_2020} a method with a more precise heatmap is described using a similarity difference mask, which is combined with a heatmap to generate the final heatmap. In addition, Wang~et~al.~\cite{wang_score-cam_2020} introduced a class activation mapping method, that removed the dependency on the gradients by using a linear combination of weights and activation maps.

For all gradient-based methods, an extension is described by Shrikumar~et~al.~\cite{smilkov_smoothgrad_2017}, where noise is added to the input image. The resulting batch of saliency maps is then averaged to obtain the final heatmap. The intuition behind this is that gradients in \acp{dnn} are fluctuating on slight changes in the input. Hence, this method smoothes the gradients and is therefore also called SmoothGrad.

There are also methods to propagate information through the complete model from the output back to the input. One of these methods is called \ac{lrp}~\cite{bach_pixel-wise_2015}. \ac{lrp} performs a decomposition of the classifier output into sums of feature and pixel relevance scores. The relevance scores are propagated from the output (classification results) through the layers (features) to the input, i.e., pixel level. Thereby, \ac{lrp} is defined by a set of constraints resulting in several rules, which can be used solely or in combination for the backpropagation. Another way of obtaining the relevance scores is proposed by Montavon~et~al.~\cite{montavon_explaining_2017} and called Deep Taylor Decomposition, where the outputs of the neurons are rewritten in first-order Taylor expansions. This allows the computation of partial relevance scores for the neurons in previous layers. The Deep Taylor Decomposition is also seen as the theoretical foundation of \ac{lrp}. Multiple extensions exist for adapting \ac{lrp} to object detection. While \cite{tsunakawa_contrastive_2019} introduce contrastive relevance propagation for the special case of a \ac{ssd} network architecture, \cite{karasmanoglou_heatmap-based_2022} extend \ac{lrp} for explaining Yolo \cite{redmon_you_2016} models. An official extension to object detection and segmentation tasks is presented with L-CRP~\cite{dreyer_revealing_2022}.
The approach from \cite{dreyer_revealing_2022} also builds on top of the advancements in \cite{achtibat_where_2022} bringing \ac{lrp} to concept level with concept-wise relevance propagation. Concepts are thereby seen as the encodings of single channels or neurons in the latent space, where a filtering method is applied to the relevance in a specified layer.

Many gradient-based saliency map methods have the problem, that sometimes the resulting heatmap looks like noise. One of the underlying problems is that well-trained models are flat in the vicinity of an input. This problem is addressed in~\cite{sundararajan_axiomatic_2017}, where a method called \ac{ig} is proposed. \ac{ig} introduces a baseline or reference, which can be in terms of an image classification task, a black image. The baseline is a kind of reference point where basically no information exists. From this point along a straight line to the input image, the gradients are calculated and integrated. Practically, the integration is replaced by stepwise approximation. An adaptation of \ac{ig} is described in~\cite{kapishnikov_xrai_2019} and is called XRAI. XRAI uses two baselines, a black and a white image, to compute the attributions for a provided image. In addition, the image is segmented into regions and the final saliency map is obtained by adding regions in dependency of the summed attributions. Another method that uses a reference is DeepLIFT~\cite{shrikumar_learning_2017}. In contrast to \ac{ig} the contribution of a feature towards the output is computed by the difference to the reference using multipliers, which are basically a modified variant of partial derivatives. In addition, some rules are proposed to assign the contribution scores for different layer types.

\subsubsection{Model-agnostic Methods}\label{sec:saliency_modelagnostic}

Apart from gradient-based methods, there are also methods that perturb the input in order to estimate the importance of the input to the output. Ribeiro~et~al.~\cite{ribeiro_why_2016} introduced a method called \ac{lime}, which learns a surrogate model that can explain a blackbox model locally. Thereby, an input sample is perturbed resulting in multiple samples in the proximity of the original input sample. Afterwards, the predictions from the blackbox model are obtained and finally, an interpretable \ac{ml} model (e.g., a decision tree) is learned. \ac{lime} is also applicable for image data, by perturbing the image via superpixels. A superpixel is a region in the image that belongs together by similar color values and can be blinded out by setting the superpixel to a defined color value (e.g., grey or black). Another method that also perturbs the input, but does not learn a surrogate model, is proposed by Petsiuk~et~al.~\cite{petsiuk_rise_2018} and is called \ac{rise}. The method perturbs input images by randomly generated masks, which blind out input pixels, resulting in multiple images, where different pixels are blacked out. Then, the predictions are obtained from the image classification model and a weighted sum multiplied by the respective mask generates the saliency map. This method was also adapted for object detection in~\cite{petsiuk_black-box_2021}. Furthermore, another perturbation-based method was published in \cite{ribeiro_anchors_2018} and is called Anchors. Like \ac{rise}, the input is perturbed, but the predictions are then used to generate rules, the anchors, containing model features along with coverage and precision. The method can be applied for image classification, but unfortunately not for object detection.

An alternative to perturbation-based methods are methods that use Shapley values~\cite{shapley_value_1953}. Shapley values are inspired by game theory and measure the contribution of each feature or feature group to a respective result. The method \ac{shap}~\cite{lundberg_unified_2017} uses Shapley values to let features be present or absent for calculating the contributions, by random sampling. Afterwards, the predictions are obtained from the original model along with the weights and then a linear model is trained to finally obtain coefficients of the linear model. A further extension of \ac{shap} is introduced by Gudovskiy~et~al.~\cite{gudovskiy_explain_2018} for object detection tasks.

\subsubsection{Applications}

Almost all methods listed in \autoref{tab:overview_sailency_map} can be used to gain local explanations for \textit{perception} components. The majority of the methods can be applied to models solving classification tasks, but some of them are already, or can be, adapted for object detection, which is the primary task to solve for perception components in automated vehicles. Often, the local explanations are used to debug or improve the \ac{ml} models´ performance, but the results from those methods can also give hints about the robustness of the perception component when the input is perturbed (e.g., dirt on the camera lens). Of course, there is in general one major disadvantage of all the methods: they can only give local explanations to specific inputs and it is not directly possible to generalize single observations. But there are also attempts to obtain global information from local saliency maps~\cite{lapuschkin_unmasking_2019} and to combine local and global explanations as in \cite{achtibat_where_2022}. Furthermore, especially for perception also the computational costs for the saliency map methods are considerable. Model-agnostic methods, like \ac{rise} or \ac{lime}, require several forward evaluations of the \ac{ml} model, while structure-based or gradient-based methods, e.g., \ac{lrp}, only require one backward pass of the (modified) output gradient. In addition, smoothing approaches as \ac{ig} or SmoothGrad, require again some forward and backward runs to ensure stability.

Saliency maps are not directly applicable for \textit{situation interpretation}. On the one hand, saliency maps can be used to gain interconnections between objects, where certain features contribute to multiple object instances. On the other hand, a heatmap can also serve as an additional source of uncertainty. Of course, the drawbacks are additional computational costs. For \textit{planning} components, saliency map methods are also applicable, if they can handle tabular data, which is true for some of them. Therefore, it is possible to estimate the importance of object instances with respect to the predicted trajectory of the automated vehicle. This can give insides into the behavior of the automated vehicle and other road participants. Formally, it could be possible to perform sanity checks if the predicted trajectory is compliant with a set of rules. Furthermore, saliency map or explainability methods can be used to reveal, which objects influenced the trajectory.
\subsection{Interpretable Feature Learning}
\label{sec:interpretable_feature_learning}
\textit{Authors: Erwin Kraft, Leonie Kreuser}

\smallskip
\noindent
The performance of vision based machine learning models strongly depends on their feature representations. Early works in computer vision focused on the development of hand-crafted features such as \ac{hog} \cite{dalal_histograms_2005}. With the advent of deep learning, feature representations were automatically learned from the training data and did no longer require manual tuning. Automatically learned features have been shown to outperform classical hand-designed representations \cite{felzenszwalb_discriminatively_2008}, \cite{dalal_histograms_2005} by very large margins, especially for image classification \cite{he_deep_2016}, \cite{tan_efficientnet_2019} and object detection tasks \cite{bochkovskiy_yolov4_2020}, \cite{tan_efficientdet_2020}. However, feature representations found in state-of-the-art deep learning models are often difficult to understand and interpret by humans \cite{kim_interpretability_2018}, \cite{wu_towards_2020}. This is usually caused by their huge dimensionality and complexity. Explainability and interpretability are very important safety requirements as one has to be able to understand the limitations of a safety-critical system before deploying it to a real-world application. The opaque black-box nature of deep learning models therefore makes it difficult to meet such demands. This applies in particular for applications such as AI-enabled automated driving, where any prediction error could have severe consequences.

To alleviate the problem, researchers have introduced a variety of methods to explain the feature representations of deep learning models \cite{kim_interpretability_2018}, \cite{wu_towards_2020}, \cite{ghorbani_towards_2019}. According to Wu et al. \cite{wu_towards_2020}, these can be roughly grouped into local explanations, based on individual data samples, and global ones, which try to derive decision patterns for the samples of entire classes. In the following section, we give a brief overview of current state-of-the-art developments with an emphasis on methods that provide global explanations.

\subsubsection{Post-hoc Explanation Methods}

To interpret the internal states of pre-trained deep neural networks, Kim et al. \cite{kim_interpretability_2018} introduced \acp{cav}, which are used to test whether the features of an image classification model are sensitive to a specific visual concept. A high-level concept is defined by a set of images, for example images showing striped patterns. \acp{cav} are derived by comparing the feature activations obtained from concept images to those of random images. To do this, a linear binary classifier is trained to separate the concept activations from the non-concept activations by constructing a hyperplane. A \ac{cav} is then defined as a normalized perpendicular vector to this plane. Using \acp{cav} in combination with directional derivatives, it is possible to measure the presence or absence of a visual concept for a specific category. For example, the method can be used to verify if the visual concept of striped patterns was learned by an image classification model to distinguish zebras from horses. \acp{cav} are also used by Graziani et al. \cite{graziani_concept_2020}, who investigate their application in medical domains. They extend the method to also include non-binary concepts since these are more suitable to explain the model behaviour to physicians.

One of the drawbacks of \acp{cav} is that the visual concepts have to be manually selected. This makes it difficult to deploy the method in an automated setting. To address this limitation, Ghorbani et al. \cite{ghorbani_towards_2019} introduced \ac{ace}, which can be seen as an automated version of \acp{cav}, as it does not need human supervision. \ac{ace} starts with segmenting a given set of images from the same class into groups of pixels. In the next step, these segements are passed through a \ac{cnn} and similar segments are identified by using the Euclidean distance in the activation space of the final layer. After removing some outliers, an importance score is calculated for the extracted concepts, resulting in a set of automatically generated visual concepts.

Another interesting post-hoc analysis method is presented by Wu et al. \cite{wu_towards_2020}, who utilize adversarial patterns to occlude specific feature representations in the network and thus are able to measure their overall importance for classifying images. It has been shown that small image perturbations (adversarial patterns) can be used to fool classification models into producing wrong predictions \cite{biggio_wild_2018}. Wu et al. \cite{wu_towards_2020} use these patterns to build global feature occluders, which supress the activations of specific features in the network for a given category. Thus, they are able to compute feature importance scores. This allows to investigate how important a feature set is for classifying an image.

\subsubsection{Disentangled Representations}
An approach to learn interpretable features in neural networks is based on disentangled representations. The idea of disentangled representations is to train neural networks in a way to identify meaningful input properties that do not influence each other. Thus, if one input property changes, all other properties remain largely unaffected. For example, Zhang et al. \cite{zhang_interpretable_2018} train an interpretable \ac{cnn} by forcing each filter in the network to learn a disentangled representation.  A similar idea is also investigated by Chen et al. \cite{chen_concept_2020} who replace the batch normalization layers of image classification models with concept whitening modules. The purpose of theses modules is to disentangle the latent feature space and align it with pre-defined concepts. This is done by learning a whitening matrix that decorrelates and standardizes the data and maximizes the activations of known concepts along the latent space axes. In general, there exist different ways to achieve disentangled representations \cite{de_mijolla_human-interpretable_2020}. Unsupervised methods extract factors of variation directly from the data while supervised disentanglement uses data with the desired semantic properties. Locatello et al. \cite{locatello_challenging_2019} point out some shortcomings of unsupervised methods and argue that well-disentangled models cannot be learned without supervision.

\subsubsection{Interpretable Model Design}

It is possible to use a set of pre-defined high-level concepts in the network design. Koh et al. \cite{koh_concept_2020} introduced \acp{cbm}. Unlike end-to-end models that go directly from raw input to a prediction, \acp{cbm} first learn a set of concepts and thereafter use these concepts to make a prediction. The final prediction module has only access to these pre-defined concepts (information bottleneck). Therefore, it is possible to assess to what degree a concept contributed to the decision and if the decision made by the model follows human reasoning. It is also possible to create a concept bottleneck using variational autoencoders. For example, Mijolla et al. \cite{de_mijolla_human-interpretable_2020} develop an explainability framework based on an encoder-decoder architecture, where the encoder module aims to reduce raw input features of high-dimensional data to compact, semantically meaningful and thus more interpretable latent features.

Interpretable models are also learned by Chen et al. \cite{chen_this_2019} who introduce prototypical part networks (ProtoPNet). The idea is to classify images by identifying object parts that can be matched to prototype parts of object categories. For example, if the input image shows a bird, the model identifies parts of the bird (e.g., the head) that are matched to the parts of a prototypical bird. A prediction is then made based on the weighted combinations of the part similarity scores. Since the prototypical parts are learned by a clustering method, it is possible to inspect the clusters and assign them to high-level concepts. In this way, a model is created that does achieve comparable results with some of the best-performing deep learning models in terms of accuracy and is also more interpretable. However, some shortcomings of ProtoPNet are demonstrated by Hoffmann et al. \cite{hoffmann_this_2021} who claim that image patches that look similar to a ProtoPNet might not necessarily look similar to a human. In their experiments, they show that the models do not perform well in the presence of compression noise.

\subsubsection{Applications}

Interpretability is a very important prerequisite for object detection tasks in autonomous driving. Detection (2D or 3D) usually resembles the backbone of camera based scene perception. As autonomous driving is a safety critical application, the need for interpretable object detection arises. One of the challenges is that many methods in the literature focus exclusively on image classification tasks. It would be therefore interesting to investigate how these methods could be extended or improved for object detection tasks.

\section{Knowledge Conformity}
\label{ch:conformity}


Autonomous vehicles are safety-critical systems, which means that their malfunction potentially has severe consequences, \eg when a pedestrian is overlooked by the detection system.
We thus must ensure that they operate safe and reliable. In particular, they should conform with existing safety principles and knowledge.
One such principle is the identification and handling of \textit{uncertainties}, \ie factors that potentially cause the system to behave in unpredictable ways. 
We provide an overview over existing concepts and methods for the estimation and assessment of uncertainties in \autoref{sec:uncertainty}.
Another principle is interpretability, \ie humans should ideally be able to understand why a system made a particular decision.
To improve in that regard, the decision making process of a DL system should be more aligned with human decision-making of which \textit{causal reasoning} is a central component. 
Therefore, we discuss methods for infusing causal reasoning into DL systems in \autoref{sec:causal_reasoning}.
Another aspect is conformity with existing knowledge about the environment of the autonomous vehicle.
In particular, the environment is subject to certain \textit{rules}, \eg traffic regulations, physical laws or human common sense. \autoref{sec:rule_conformity} discusses concepts and approaches regarding rule conformity.
The ability to generalize to previously unseen phenomena is another essential requirement for the safe and reliable use of neural networks. Hence, approaches for a \textit{verification} of the intended functionality are discussed in ~\autoref{sec:ai_ver}. 
Another important aspect is the safeguarding of autonomous systems during run-time. \textit{Monitoring} systems offer the possibility of detecting and displaying deviations from predefined rules, standards and procedures. Relevant approaches are presented in the final~\autoref{sec:runtime_network_ver}.

\subsection{Uncertainty Estimation}
\label{sec:uncertainty}
\textit{Authors: Maximilian Alexander Pintz, Christian Wirth, Sebastian Houben}

\smallskip
\noindent
Conventional machine learning algorithms, like neural networks, are only able to predict a single best estimate $y=f(x,\omega)$ (with $\omega$ as the network parameters). 
By default, they do not provide any probability of this single estimate being correct neither do they provide a probability for all of the possible estimates of being correct, i.e., a distribution over the output space. 
Therefore, neither (reliable) probabilities of correctness for classification tasks, nor a measure of spread of the prediction for regression tasks are available. 
However, these pieces of information are highly relevant for safety-critical applications like autonomous driving. 
A measure of correctness or spread allows us to determine if and to what extent it is safe to use a prediction for downstream tasks. 
Additionally, a system might recognize when the input deviates from the \ac{odd} and can therefore likely not be handled correctly. 
Furthermore, depending on the use case, it is possible to use numeric limits to "err on the side of caution" or to "err on the side of being optimistic". 
As an example, when performing visual object detection, we should be conservative about "free parking" signs (e.g., only predict if sure) but should predict pedestrians even if we are not sure. 
Especially in urban scenarios, we also have to factor in the uncertainty over the exact position of a road user and keep some distance, based on the possible localization error. 
When considering trajectory predictions for other road users, we also need to be able to react to highly different possibilities (like turn vs. drive straight), which requires multiple estimates with according probability of correctness.
Furthermore, several advanced machine learning methods, like active learning (c.f.~\autoref{sec:active_learning}) or continuous learning (c.f.~\autoref{sec:continual_learning}), explicitly require some form of confidence or uncertainty estimates.


In the literature \cite{hullermeierAleatoricEpistemicUncertainty2021}, two different types of uncertainty are usually considered: aleatoric and epistemic uncertainty. 
Aleatoric uncertainty relates to the uncertainty that is inherent to the data, meaning it can not be "explained away" with better models or more data (e.g., observation noise or information that is lacking to solve the problem optimally). 
Epistemic uncertainty is induced by the limitations of our machine learning model(s) and their training and can theoretically be eradicated with more data or better approaches.

\subsubsection{General Methods}
\label{sec:uncertainty_general_methods}

In uncertainty estimation, we seek to find distributions that describe the uncertainty in our machine learning models. The distributions are typically in the space of model parameters (for describing epistemic uncertainty) or in the output space (describing aleatoric or combinations of aleatoric and epistemic uncertainty). Uncertainty in the output space is modeled by replacing scalar output values $y=f(x,\omega)$ with a distribution $p(y \vert x, \omega)$ (see likelihood optimization in \autoref{sec:deterministic_inference}). For learning distributions in the parameter space, we usually compute the posterior probabilities of parameter values, given the (training) data by applying Bayes theorem
\begin{equation}
    \label{eq:bayes_posterior}
    p(\omega \vert \mathcal{D}) = \frac{p(\mathcal{D} \vert \omega) p(\omega)}{p(\mathcal{D})},
\end{equation}
with $p(\mathcal{D} \vert \omega)$ being the probability of the (training) data $\mathcal{D}$ given the parameter values (the joint likelihood), $p(\omega)$ is a prior over the parameters and $p(\mathcal{D})$ is the probability of observing the data. The prior $p(\omega)$ captures an initial belief about the model parameters before any training data is observed and is commonly chosen to be uniform or Gaussian.
We can then compute the predictive distribution
\begin{equation}
    \label{eq:bayes_predictive}
    p(y \vert x, \mathcal{D}) = \int_{\omega} p(y \vert x, \omega) p(\omega \vert \mathcal{D}) d\omega,
\end{equation}
to describe the uncertainty for new datapoints $x$. However, computing the posterior or the predictive distribution of neural networks  directly is computationally intractable and approximations are required. It is possible to empirically approximate the posterior (see MCMC-based methods in \autoref{sec:sample_based_methods}), but the number of required samples is usually too high for modern deep learning architectures.

\paragraph*{\textbf{Parametric Approximations}}
\label{sec:UE_parametric}
The requirement for an empirical approximation of $p(\omega \vert \mathcal{D})$ can be circumvented by the assumption that it is possible to approximate the posterior distribution with a parametric distribution $p(\omega \vert \mathcal{D}) \approx q(\omega, \theta)$, like a Gaussian or Laplace distribution. This allows us to only store the distribution parameters $\theta$, instead of the empirical samples and directly sample from the approximation of $p(\omega \vert \mathcal{D})$ for inference (see \autoref{sec:sample_based_methods}). 
This procedure reduces the amount of samples required for a sufficient approximation of the predictive distribution $p(y \vert x, \mathcal{D})$ (Equation~\eqref{eq:bayes_predictive}).
In some cases (see \autoref{sec:deterministic_inference}), it is even possible to compute the predictive distribution analytically, without any sampling. 

Aleatoric uncertainty is commonly modeled via parametric distributions $p(y \vert \theta(x))$ that replace the (deterministic) outputs of a network, e.g.,
\begin{equation}
    \label{eq:parametric_approx_aleatoric}
    \theta = (\mu,\sigma) = f(x,\omega).
\end{equation} 
Similarly, epistemic uncertainty is often modeled by replacing the network parameters $\omega$ with a parametric distribution.
Parametric approximations commonly use loc-and-scale family distributions like Gaussian, Laplace or Cauchy distributions. In these cases, the scalar value we replace is the location $\mu$ of the distribution while the scale $\sigma$ allows us to capture the "range" of possible values (e.g., the displacement error for object detection). In the aleatoric case, it is possible to either also use a model output for defining the scale parameter (as in Equation~\eqref{eq:parametric_approx_aleatoric}), or to consider it a free parameter, which does not depend on the inputs. The first setting is called heteroscedastic uncertainty, whereas the second variant relates to homoscedastic uncertainty. Homoscedastic uncertainty is input-independent observation uncertainty, e.g., identical for every datapoint, whereas heteroscedastic uncertainty captures data dependence.

\paragraph*{\textbf{Multimodality}}
\label{sec:uncertainty_multimodality}
By replacing the weights (epistemic) or outputs (aleatoric) with a unimodal distribution, we can capture the variance but not multimodalities. Within the epistemic setting, this is a less severe problem as it is still possible to obtain multimodal predictions due to the model-inherent nonlinearities. There are efforts to employ multimodal parameter distributions, like Gaussian mixtures (c.f.~\cite{bishop_mixture_1994}), but this is computationally difficult due to mode collapse and the computational cost of high dimensional distributions. Therefore, independent distributions are often used, see Ensembles in \autoref{sec:deterministic_inference}. In the aleatoric case, the computational costs are less severe, as the mixture only scales with the number of outputs, not the number of model parameters. 

\paragraph*{\textbf{Covariance Matrix Approximation}}
\label{sec:uncertainty_matrix_approx}
In the case of parametric approximations of the posterior or multidimensional outputs, fully specifying the multidimensional distribution goes along with storing and learning the full covariance matrix. This can induce scalability issues, especially for networks with many parameters or high dimensional outputs, as the matrix scales quadratically. Therefore, it is often assumed that all distributions are independent, disregarding the covariances. For epistemic uncertainty, this is also known as the mean-field approximation. However, it can be empirically shown that it is highly relevant to also capture the dependence between the distributions \cite{foong_-between_2019,collier_simple_2020}, meaning approximations of the covariance matrix are of relevance. The lower Cholesky triangular approximation reduces the size of the covariance matrix by a factor of 2, exploiting that the co-variance matrix is symmetric positive definite but still scales quadratically. Therefore, many state-of-the-art approaches use a vector decomposition of the covariance matrix as an approximation \cite{wen_batchensemble_2020,collier_simple_2020}. 

\paragraph*{\textbf{Subspace Approximation}}
\label{sec:uncertainty_subspace}
As mentioned, full covariance matrix approximations are usually computationally very costly. 
This is especially problematic for epistemic uncertainty estimation due the large number of parameters encountered in conventional neural networks. 
However, even vector approximations (c.f. previous paragraph) are often not sufficient as approximation \cite{kristiadi_being_2020}. 
As an alternative, one delineates neural networks into two distinct parts: A feature projector which only projects into a latent space and the classifier/regressor itself. 
In case of computer vision approaches, the feature projector is the backbone network whereas the classifier is only the last layers, the head. 
This allows us to capture only the uncertainty in the classifier/regressor head, substantially reducing the number of parameters that need to be approximated by a parametric distribution. 
This approach is in line with conventional machine learning, like Gaussian Processes, which are not using any learned feature projectors. 
Due to the parameter reduction, it is then possible to approximate the full covariance matrix,  which has shown empirically better results than an all-layer vector or mean-field approximation \cite{kristiadi_being_2020}. 
Additionally, this allows us to build on uncertainty estimation methods used in conventional machine learning, like replacing the last layer with a Gaussian Process \cite{liu_simple_2020}. Another area of research tries to use non-predefined sub-spaces, by selecting the model parts where the probabilistic treatment is most important. As example, \cite{daxberger_bayesian_2021} determines relevant model parameters by considering the induced Wasserstein distance to the full probabilistic solution.

\subsubsection{Deterministic Inference}
\label{sec:deterministic_inference}
Deterministic approaches do not require Monte Carlo sampling for inference and provide uncertainty estimates with a single forward pass. These approaches avoid the inherent cost of sampling, but may substantially increase the memory footprint and operation count of a network. Therefore, these are often either computationally costly \cite{huber_bayesian_2020} or only work under certain model assumptions, like specific activation functions \cite{wu_deterministic_2019}. In the following, we present general areas of deterministic inference, that are not restricted by severe model assumptions. 

\paragraph*{\textbf{Likelihood Optimization}}
\label{sec:likelihood_optimization}
Among the simplest deterministic approaches is letting the network directly compute the parameters of a predictive distribution $p(y \vert \theta(x))$ (Equation~\eqref{eq:parametric_approx_aleatoric}) and using maximum likelihood or a-posteriori optimization for learning the parameters.
Such approaches do not require the computation of (costly) posterior distributions $p(\omega \vert \mathcal{D})$ and thus enable training and inference without any need of sampling, but come at the cost of restricting the predictive distribution to a simple parametric form, such as a Gaussian in regression or categorical softmax-based distributions in case of classification. In the latter case, the uncertainty is characterized by the probability of the predicted class or by the entropy of the estimated distribution.
However, softmax-based class probabilities are often poorly calibrated \cite{gast_lightweight_2018} and thus require recalibration \cite{guo_calibration_2017,kuleshov_accurate_2018} or more advanced training procedures besides likelihood optimization.

A natural extension of this approach is to consider the distribution parameter $\theta$ to be uncertain as well.
Evidential deep learning \cite{NEURIPS2020_aab08546,sensoy_evidential_2018} considers a parametric distribution on $\theta$, i.e., $p(\theta \vert \eta(x))$ that takes the form of a Dirichlet distribution (classification) or a Normal-Inverse-Gamma distribution (regression). 
This enables new ways of measuring uncertainty and taking other types of uncertainties into account, such as epistemic uncertainty.
These measures include the variance or entropy of $\theta$ or the mutual information between $y$ and $\theta$.
The distribution parameters are learned by optimizing the marginal likelihood $\sum_i p(y_i \vert \eta(x_i))$, a procedure also known as Type II maximum likelihood optimization \cite{bishop:2006:PRML}.
A related approach are prior networks, with the main difference lying in the optimization procedure. 
Prior networks \cite{10.5555/3327757.3327808} optimize a \ac{kl} divergence between $p(\theta\vert \eta(x))$ and hand-crafted distributions centered on the ground-truth.
Besides the given training inputs, the model is also explicitly trained on chosen out-of-data inputs (e.g., corrupted training inputs) to encourage the variance of $\theta$ to increase for inputs that are atypical relative to the training inputs.

\paragraph*{\textbf{Laplace Approximation}}

The Laplace approximation~\cite{mackay_practical_1992} approximates optimal Gaussian scale parameters given a mean. Meaning, we compute a point estimate $\omega$, minimizing $\mathrm{NLL}(\omega) = -\sum_i \log p(y_i \vert f(x_i, \omega))$, as in conventional deep learning, but with an $L_2$-regularization (or other prior-based terms) added to the loss. The resulting parameters $\omega$ are \ac{mapo} estimates that serve as the mean of a Gaussian approximation to the posterior $p(\omega \vert \mathcal{D}) \approx q(\omega, \theta)$ (c.f. Parametric Approximations in \autoref{sec:uncertainty_general_methods}). The variances can then be estimated by computing the Hessian at the MAP. However, this method is computationally very costly due to the approximation of the Hessian. This applies to the computational cost in terms of operations as it does to memory consumption. 
Deterministic inference can than be performed using probit scaling or the Taylor approximation \cite{bishop:2006:PRML}, using the stored Hessian. Different methods for approximating the Hessian have been suggested \cite{ritter_scalable_2018}. It should also be noted, that it is not required to use deterministic inference approaches, but one can directly sample from the obtained posterior, comparable to the inference step typically employed in \acf{vi} (c.f. variational inference in Subsec.~\autoref{sec:sample_based_methods}).

\paragraph*{\textbf{Ensembles}}
\label{sec:uncertainty_ensembles}

Ensembles subsume all approaches that combine the outputs of several models to obtain several samples that represent a distribution over the output space. 
Usually this involves using techniques for obtaining independent outputs from all models or techniques for properly addressing the dependency among model outputs. 
In this sense, ensembles constitute the frequentist's approach for providing a distribution over the output space. 
Ensembles have been and continue to be deployed for better performance and increased robustness but bear a straightforward solution for uncertainty estimation.
Special cases comprise bagging, i.e.,~training the same model type with different parts of the training data, and boosting \cite{Freund99ashort,hastie_elements_2009}, i.e., selecting and training the next member of the ensemble with the aim of addressing shortcomings of the current ensemble.

Ensembles of neural networks with the same architecture trained with different random initialization are called Deep Ensembles~\cite{lakshminarayanan_simple_2017}. 
Such ensembles capture multimodality using the fact that neural networks usually converge to different solutions, based on the initial parameter weights and the optimizer settings. 
They have been trained and tested with adversarial training which allows for a smoother prediction and a better uncertainty estimation.
Hyper-deep Ensembles \cite{wenzel_hyperparameter_2020} extend the random initialization of weights to a selection of hyperparameters.
However, ensemble techniques are costly in terms of parameters as it is required to store (and train) the parameters for each ensemble member independently. 
Batch Ensembles \cite{wen_batchensemble_2020} try to alleviate this problem by approximating the joint weight matrix over all models by a vector decomposition. 
The weights of each ensemble member are defined by a shared parameter set, multiplied with a row and column vector, specific for each model. 
This method is also applicable for parametric approximations, by replacing the row and column vectors with parametric distributions \cite{dusenberry_efficient_2020}.

\paragraph*{\textbf{Kernel-based Methods}}
\label{sec:uncertainty_kernel}
An alternative way is the use of Kernel methods, capturing the uncertainty in the space of functions. This allows us to build on uncertainty estimation methods used in conventional machine learning, by replacing the last layer (cf.~\autoref{sec:uncertainty_general_methods}) with a kernel function. Known approaches use RBF Layers \cite{van_amersfoort_uncertainty_2020} or Gaussian Processes \cite{liu_simple_2020}. 

However, it should be mentioned that exact kernel methods, are not applicable in this setting due the high number of training datapoints and the resulting size of the kernel matrix. Therefore, it is not possible to use all training datapoints as inducing points. The authors in \cite{van_amersfoort_uncertainty_2020} suggest using a single prototype per class, whereas \cite{liu_simple_2020} uses a random feature approximation of the inducing point space. When using multiple inducing points, it is required to ensure that the deterministic feature projector (everything before the last layer), is preserving the distance between datapoints. Without this constraint, it is possible that the feature projector maps all datapoints of the same class to the same feature vector. This can be achieved by spectral normalization, bounding the Lipschitz constant \cite{liu_simple_2020}. Other commonly considered approximations besides the random feature approximation of the inducing point space, are variational Gaussian processes \cite{tran_variational_2015}.

\subsubsection{Sampling-based methods}
\label{sec:sample_based_methods}

The ability of deterministic approaches to provide uncertainty estimates with a single forward pass typically goes along with restrictions on the uncertainty distribution (e.g., via imposing parametric models).
Sampling-based methods allow to approximate more flexible and complex distributions.
A universal tool for this purpose is \ac{mc} integration. For example, it can be used to approximate an arbitrary complex predictive distribution $p(y | x, \mathcal{D}) \approx 1/N \sum_i p(y | x, \omega^{(i)})$, given $N$ samples $\omega^{(i)}$ from the posterior $p(\omega | \mathcal{D})$. This idea gives rise to multiple approaches that seek to obtain samples from the posterior, that are discussed in the following.
In addition, there are also non-Bayesian approaches that employ \ac{mc} integration, which we discuss at the end of this section.

\paragraph*{\textbf{MCMC-based}} 
\label{sec:uncertainty_mcmc}
A common method for computing an empirical approximation of the posterior distribution is \ac{hmc} \cite{duane_hybrid_1987,neal_mcmc_2010} and its extension \ac{nuts} \cite{hoffman_no-u-turn_2014}. The predictive distribution is then obtained by directly inferring a single prediction for every posterior sample. 
However, these methods are not applicable to batched training and can therefore not be applied to common deep learning tasks. Batch-enabled variants exist, like \ac{sgld} \cite{li_preconditioned_2016} or \ac{sghmc} \cite{chen_stochastic_2014}, but the resulting posterior distributions $p(\omega \vert \mathcal{D})$ are usually very complex and a large amount of samples is required for an sufficiently good approximation. Obtaining these samples, and/or inferring predictions, is usually computationally to costly for modern neural networks. Note that implementations of \ac{hmc}, \ac{nuts} and \ac{sgld} are available in Tensorflow probability and similar libraries.

\paragraph*{\textbf{Variational Inference}} 

Another approach is \acf{vi} \cite{bishop:2006:PRML,10.1145/168304.168306,NIPS2011_7eb3c8be}, that seeks to approximate the parameter posterior using a (simpler) parametric distribution $q(\omega, \theta)$, the so-called variational distribution.
This is done by optimizing the \ac{elbo} 
\begin{equation}
    \hat{V}(\theta) = 1/N \sum_i \log p(Y \vert X, \omega^{(i)}) - \mathrm{KL}[q||p(\omega)],
\end{equation} with $\omega^{(i)} \sim q(\omega, \theta)$, which effectively minimizes the \ac{kl} divergence between the posterior and $q$.
For neural networks, computing gradients of the \ac{elbo} is possible via the reparametrization trick \cite{10.5555/2969442.2969527}, which sets the sampled parameters $\omega^{(i)} = t(\theta, \varepsilon^{(i)})$ to the result of a deterministic function $t$ and a sample $\varepsilon^{(i)}$ from a distribution which does not depend on any parameters. 
A common case is that $q$ is a Gaussian with diagonal covariance, which allows for the reparametrization $\omega = \mu + \mathrm{diag}[\sigma] \varepsilon$ with $\varepsilon \sim \mathcal{N}[0, I]$, where we can compute gradients with respect to the variational parameters $\mu$ and $\sigma$.
Applying a \ac{vi}-based method to an existing neural network requires changing the loss function to the \ac{elbo} and modifying every layer due to the reparameterization.
The methods scale to large \acp{nn} under the choice of small number of samples (typically one chooses $N=1$) and feasible $q$ and priors.
Several variants of this procedure have been proposed that typically aim at reducing computational costs, faster convergence (eg. Flipout \cite{wen_flipout_2018}) and enabling the use of more expressive priors and variational posteriors (e.g., Horseshoe priors \cite{ghosh_model_2019}, MVG \cite{10.5555/3045390.3045571}, normalizing flows \cite{rezende_variational_2015}). 

\paragraph*{\textbf{Dropout}} Another approach to Bayesian \ac{vi} in neural networks is \ac{mc} dropout \cite{gal_dropout_2016}. 
This approach utilizes the dropout mechanism \cite{srivastava_dropout_2014}, which randomly omits units from the neural network in each forward pass.
In practice, this is done by multiplying the activations in each layer with a binary mask, whose entries are sampled from a Bernoulli distribution with "drop probability" $p$.
Gal et. al \cite{gal_dropout_2016} showed that training such a network with a standard $L_2$-regularized \ac{mse} approximately optimizes an \ac{elbo} between a deep Gaussian process model and a variational posterior on network weights, giving further theoretical motivation for the approach.
The mean and variance of the corresponding predictive distribution can be obtained by simply computing the mean and variance (plus a hyperparameter-dependent offset) of multiple dropout forward passes.
\ac{mc} dropout has become a popular approach for uncertainty estimation, especially among applied practitioners, as it requires only a small change in network architecture, has low computational complexity and is directly applicable to several different architectures, such as \acp{cnn} or \acp{rnn}. 
A drawback of \ac{mc} dropout is that it employs a quite restrictive variational posterior, that utilizes heavy independence assumptions. 
In addition, it assumes a fixed-variance Gaussian likelihood, thus allowing it only to describe homoscedastic aleatoric uncertainty.
To mitigate this drawback, the authors in \cite{kendall_what_2017} proposed to combine \ac{mc} dropout with parametric likelihood methods, to incorporate heteroscedastic aleatoric uncertainty as well. 
This method is also the basis for many uncertainty-aware object detection models \cite{harakeh_bayesod_2020}.
Sicking et al.~\cite{sicking_characteristics_2020,sicking_novel_2021} introduced Wasserstein dropout, which directly optimizes the dropout sub-networks to model (heteroscedastic) aleatoric uncertainty.
Several other proposed variants include dropout with trainable drop probabilities $p$ \cite{NIPS2017_84ddfb34}, different distributions besides Bernoulli \cite{10.5555/2969442.2969527} and sampling-free variants \cite{Postels2019SamplingFreeEU}.

\paragraph*{\textbf{Distributional Output}}
The authors in \cite{collier_simple_2020} take a non-Bayesian approach to uncertainty estimation in classification tasks.
Classification networks can generally be thought of as generating a utility value $u_c$ for each class $c$. 
The network assigns class $c$ to a given input if the corresponding utility $u_c$ is larger than the utility for all other classes.
For uncertainty estimation, the framework considers the utility values to be corrupted by random noise $\varepsilon_c$, which allows to assign a probability to each class for having maximum utility.
The probabilities reduce to standard softmax probabilities, if the noise is homoscedastic and follows a standard Gumbel distribution for each class.
However, the framework allows to generalize this to more realistic types of noise, including noise that is not identically distributed across classes or heteroscedastic noise with full covariance.
Due to the higher complexity, the class probabilities cannot be computed in closed form anymore and are thus approximated via \ac{mc} integration.
Regardless, the approach is one of the few highly scalable approaches that incorporate full covariance uncertainty distributions.

\subsubsection{Quality Measures}
Considering ways to express a prediction uncertainty introduces a number of additional quality measures apart from the original performance metrics of the machine learning model, like classification rate, root-of-mean-squared-error, mean average precision and others. 
The quality of uncertainty estimates can be evaluated with proper scoring rules \cite{gneiting_strictly_2007}, which quantify the alignment of the predicted distributions with given ground-truth data points.
They attain their optimum when the predicted distribution coincides with the data distribution.
Commonly used proper scoring rules include the \ac{nll}, brier score (for classification) or the \ac{crps}. 
An alternate way of assessing uncertainty estimates provides the framework of calibration.
In general, calibration measures whether class probability or confidence interval estimates conform to the actual model error.
A calibrated classifier satisfies that the fraction of correct prediction is $p$ among all class predictions with estimated class probability $p$.
This calibration property can be evaluated quantitatively with the \ac{ece}. It should be noted, that \ac{ece} is subject to several pathologies, that should be considered explicitly \cite{nixon_measuring_2019}. A qualitatively assessment is possible via reliability diagrams that compare class probabilities against classification accuracy \cite{guo_calibration_2017}.
Similar notions of calibration also exist for regression models (based on how ground-truth outputs conform with the predicted confidence intervals \cite{kuleshov_accurate_2018}) and for certain application domains such as object detection models  \cite{kuppers_multivariate_2020}.
For a more strict assessment, we might want the calibration property to hold not only across the whole (evaluation) dataset, but also on local regions of the dataset, which is captured by adversarial group calibration \cite{zhao_individual_2020}.
We can also measure the performance of uncertainty estimates in auxiliary tasks, such as out-of-data-detection.
Separating true from false detections or in- from out-of-data inputs is a binary classification task, where the classes are determined by thresholding uncertainty estimates.
Thus, we can employ standard evaluation methods for binary threshold classifiers, including plotting precision-recall or receiver operating curves for qualitative assessment or computing average precision or the area under the \ac{roc} curve.
Another approach is to compare histograms of uncertainty estimates (\eg variance or predictive entropy) on inputs from each class (\eg in-data or out-of-data), as done in \cite{10.5555/3454287.3455541}.  
In technical applications the computational complexity, i.e., the overhead for computing the uncertainty estimate when compared to the single-point-estimate, and the latency, i.e., the part from the computational complexity that cannot be efficiently parallelized or pipelined and, thus, result in additional inference time, are taken into account for assessment as well.


\subsubsection{Applications}

Many of the approaches for uncertainty estimation outlined before apply to a wide range of different models and are used in numerous different application scenarios.
Main challenges of uncertainty estimation in specific application scenarios include a principled treatment of specialized layers, incorporating uncertainty in pre-/post-processing steps, dealing with unusual data structures, leveraging task-specific assumptions and efficiency.
In the following, we discuss several works that employ uncertainty estimation in perception tasks.
A larger-scoped review of applications of uncertainty estimation can be found in the surveys \cite{ABDAR2021243} and \cite{Gawlikowski2021ASO}. 

\paragraph{\textbf{Object detection}}

A wide bandwidth of models employing uncertainty estimation for object detection tasks have been presented in the literature and we refer to \cite{feng_review_2021} for a survey and comparative study with a focus on object detection in autonomous driving. 
The approaches cover both 2d and 3d \cite{Lu_2021_ICCV} localization of objects as well as different input types such as camera images \cite{miller_dropout_2018}, lidar scans \cite{feng_leveraging_2019} or radar \cite{dong_probabilistic_2020}. 
The main techniques employed by such approaches are typically direct optimization of parametric distributions (e.g., \cite{feng_leveraging_2019,he_bounding_2019}), \ac{mc} dropout (e.g., \cite{miller_dropout_2018})  or, based on the work of \cite{kendall_what_2017}, a combination of both (e.g., \cite{10.1109/ITSC.2019.8917494,harakeh_bayesod_2020}). 
Other works consider ensembling \cite{Miller_2019_CVPR_Workshops} or a spatial clustering of redundant detections \cite{8569637}. 
Approaches often differ in the way uncertainty estimates are integrated into non-maximum suppression, a post-processing step to filter redundant detections, or whether anchor-based or anchor-free \cite{lee_localization_2020} methods are employed. 

\paragraph{\textbf{Trajectory Prediction}}
For path planing, it is important to correctly predict the future trajectories of other traffic participants. Naturally this is not possible in a deterministic manner as there are usually multiple options and we can only derive the prediction from the participants past behavior. Therefore, nearly all approaches use multimodal predictions \cite{chai_multipath_2019,phan-minh_covernet_2020,varadarajan_multipath_2021,djuric_multixnet_2021} (c.f. Multimodality in \autoref{sec:uncertainty_general_methods}), usually by reducing the continuous distribution over trajectories (or keypoints) to a categorical distribution over distinct trajectories. These methods may be enhanced by also allowing to predict displacement of the expected trajectory wrt. the predicted trajectory from the categorical distribution. These displacements can be modeled using likelihood optimization (see \autoref{sec:deterministic_inference}), usually using bivariate Gaussians \cite{chai_multipath_2019,varadarajan_multipath_2021,salzmann_trajectron_2020}. This allows to model the aleatoric uncertainty, but not the epistemic uncertainty. Due to the realtime requirements usually involved in trajectory prediction, full Bayesian approaches are uncommon. However, some methods employ \ac{cvae} conditional, variational auto encoder (CVAE) \cite{salzmann_trajectron_2020} for learning a latent space. \acp{cvae} are in fact a kind of variational inference (Variational Inference in Subsec.~\autoref{sec:sample_based_methods} method, but only applied to a very limited subspace of the network (see Subspace Approximation in \autoref{sec:uncertainty_general_methods}).

\paragraph{\textbf{Other}} Parametric distributions and dropout are also utilized in other computer vision-related tasks such as semantic segmentation for single image inputs \cite{mukhoti_evaluating_2018} and video \cite{huang_efficient_2018}, 
as well as camouflaged or salient object detection \cite{li_uncertainty-aware_2021,Yang_2021_ICCV}, which respectively refer to the tasks of segmenting objects that are hidden in the surrounding or stand out for humans on an image.  
Similar uncertainty estimation techniques are also considered for human pose estimation \cite{bertoni_monoloco_2019,Gundavarapu_2019_CVPR_Workshops}. 
To better represent the uncertainty of the rotational part of pose estimates, several works on pose estimation or visual odometry consider employing distributions from directional statistics, such as the von Mises-Fisher \cite{deepdirectstat2018} or Bingham distributions \cite{Deng2020DeepBN,peretroukhin_smooth_2020}.    
Approaches for complete monocular visual odometry also exist that provide uncertainty estimates of trajectories (sequence of position and rotation) based on image sequences. 
The work of \cite{8403264} considers object detection based on a sequence of images from multiple camera positions. 
The approach leverages dropout and Bayesian updating to maintain a posterior class distribution based on multiple images. 
There is also a similar method \cite{8461127} that explicitly incorporates localization information into the estimation. 

The authors in \cite{gast_lightweight_2018} propose a general framework for probabilistic deep learning that employs \ac{adf}, an alternative uncertainty estimation method to pure parametric output distributions or dropout. 
\ac{adf} is an efficient approximate form of expectation propagation, where we seek to find the output distribution of the network that arises from stochastic inputs (that, e.g., follow a Gaussian distribution) when all other network components are deterministic.
The authors apply their framework to large \acp{cnn} to find the optical flow in given video sequences.
Dropout-based uncertainty estimation in optical flow tasks is also considered in \cite{9340963} or \cite{wannenwetsch_probflow_2017}. 
Building on the \ac{adf}-based framework, the authors in \cite{loquercio_general_2020} combine assumed density filtering with dropout to further integrate epistemic uncertainty into the framework. 
They demonstrate their framework on tasks such as future motion prediction and steering angle prediction.

\subsection{\acl{cr}}
\label{sec:causal_reasoning}
\textit{Authors: Tobias Latka, Christian Brunner}

\smallskip
\noindent
\subsubsection{Background}
\label{sec:causal_reasoning_background}
“\textit{Would I have arrived at my destination earlier, had I taken route A instead of route B?}” Such a query looks familiar to humans and is an instance of human reasoning. The above example can be attributed to the notion of \textit{\ac{cr}} \cite{pearl_causality_2013}, a subcategory of \textit{\ac{ci}} \cite{shanmugam_elements_2018} that also deals with \textit{\ac{cd}}, i.e., inferring cause-effect-relationships from data and encoding these in form of a \textit{Causal Model}. As \textit{\acl{cd}} is beyond the scope of this section, we refrain from giving an overview on the topic of \textit{\acl{cd}} here.

There are two common frameworks for the description of \textit{Causal Models} that are mathematically equivalent:

\begin{itemize}
    \item the \textit{Rubin Causal Model} \cite{holland_statistics_1986, rubin_causal_2005} (also known as the \textit{Neyman-Rubin Causal Model} or the \textit{potential outcome framework of Neyman} \cite{neyman_sur_1923} \textit{and Rubin} \cite{rubin_estimating_1974})
    \item the \textit{\acf{scm}}\footnote{"Graphical Models serve as a language for representing what we know about the world, counterfactuals help us to articulate what we want to know, while structural equations serve to tie the two together in a solid semantics." \cite{pearl_seven_2019}} \cite{halpern_causes_2005}.
\end{itemize}

\begin{figure}[t]
\begin{center}
\includegraphics[width=0.8\columnwidth]{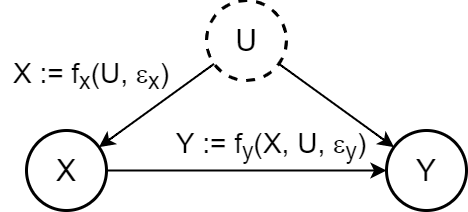}
\caption{An \acs{scm} consists of both a set of causal mechanisms ($f_X$ and $f_Y$) and a \acs{dag} modeling the flow of causation between variables (directed arrows). The shown \acs{scm} is composed of a cause (X) and its effect (Y), which are both confounded by a latent variable (U). Nodes associated with the exogenous noise variables $\epsilon_X$ and $\epsilon_Y$ are omitted for the sake of clarity.}
\label{fig:causal_reasoning:scm_example}
\end{center}
\end{figure}

While the \textit{potential outcome framework of Neyman and Rubin} solely resorts to counterfactual notation, \acsp{scm} make explicit use of a graphical representation in terms of a \textit{\ac{dag}} which counterfactual expressions can be inferred from (see \autoref{fig:causal_reasoning:scm_example} for an illustration of an \acs{scm}). The key characteristic of \acsp{scm} is that they represent each variable as a deterministic function ($f_X$ and $f_Y$ in \autoref{fig:causal_reasoning:scm_example}) of its direct causes together with latent exogenous noise variables ($\epsilon_X$, $\epsilon_Y$ and $U$ in \autoref{fig:causal_reasoning:scm_example}), whose causes lie outside the \acs{scm}.

By definition, \textit{\acl{cr}} is the process of drawing conclusions from a \textit{Causal Model}, similar to the way probability theory reasons about the outcomes of random experiments. However, since \textit{Causal Models} are thought of as the data-generating process, they contain more information than probabilistic ones and are thus more powerful, as they allow to analyze the effects of interventions or distribution changes on target variables \cite{shanmugam_elements_2018}.

Questions posed to the \textit{Causal Model} are phrased as \textit{Causal Queries}. They are at the heart of \textit{\acl{cr}} and can be differentiated into three levels of causation of increasing complexity \cite{pearl_causality_2013}:

\begin{itemize}
    \item \textit{Associations} (correlations),
    \item \textit{Interventions} (actively changing causal mechanisms),
    \item \textit{Counterfactuals} (retrospective reasoning).
\end{itemize}

The just described \textit{Causal Hierarchy} is also often referred to as \textit{Pearl’s Ladder of Causation} (see \autoref{tbl:causal_reasoning:causal_hierarchy}).

\begin{table*}[t]
\caption{The \textit{Causal Hierarchy}. Questions at level $i$ can be answered only if information from level $i$ or higher is available. Reproduced from \cite{pearl_seven_2019}.}
\centering
\begin{tabular}{|l|l|l|l|} 
\hline
\textbf{Level (Symbol)} & \textbf{Typical Activity} & \textbf{Typical Questions} & \textbf{Examples} \\ 
\hline
\makecell{1. Association \\ $p(y|x)$} & \makecell{Seeing} & \makecell{What is? \\ How would seeing $X$\\ change my belief in $Y$?} & \makecell{What does a symptom tell me about a disease? \\ What does a survey tell us about the election results?}\\
\hline
\makecell{2. Intervention \\ $p(y|do(x), z)$} & \makecell{Doing, \\ Intervening} & \makecell{What if? \\ What if I do $X$?} & \makecell{What if I take aspirin, will my headache be cured? \\ What if we ban cigarettes?}\\
\hline
\makecell{3. Counterfactuals \\ $p(y_x|x^{\prime}, y^{\prime})$} & \makecell{Imagining, \\ Retrospection} & \makecell{Why? \\ Was it $X$ that caused $Y$? \\ What if I had acted differently?} & \makecell{Was it the aspirin that stopped my headache? \\ Would Kennedy be alive had Oswald not shot him? \\ What if I had not been smoking the past two years?}\\
\hline
\end{tabular}
\label{tbl:causal_reasoning:causal_hierarchy}
\end{table*}

On top of that, there are different types of \textit{\acl{cr}} \cite{pearl_causality_2013}:

\begin{itemize}
    \item \textit{Prediction} (reasoning forward in time),
    \item \textit{Abduction} (reasoning from evidence to explanation),
    \item \textit{Transduction} (reasoning through common causes),
    \item \textit{Induction} (from experience to causal knowledge).
\end{itemize}

Abduction, for instance, is the first step in the three-step process in counterfactual reasoning (abduction, intervention, prediction), as the following example illustrates:

The introductory counterfactual query can be mapped to the graph in \autoref{fig:causal_reasoning:scm_example} and formulated mathematically as $\mathbb{E}(Y_{X=A}|Y^{\prime}, X^{\prime}=B)$. Here, $\mathbb{E}$ is the expected value, $Y$ is the time of arrival at the desired destination, $X$ stands for the route taken by the driver and $X^{\prime}$ and $Y^{\prime}$ are the evidences. The background factor (e.g., the traffic situation) influencing both the driver's decision and the time of arrival is signified by $U$, which is inferred during the abduction step such that it complies with the evidences. With this information, the expression, $\mathbb{E}(Y_{X=A}|Y^{\prime}, X^{\prime}=B)$ reads: "\textit{I would have arrived at my desired destination at $Y$ instead of $Y^{\prime}$ o'clock, if I had taken route $A$ instead of route $B$.}"

One of the strengths of \textit{\acl{cr}}, that is closely linked to the above example, is its capability to assess the causal effect of one variable on another from observational data alone, without the need of collecting experimental data. A causal query that is (uniquely) evaluable in this way is said to be \textit{identifiable}, but whether it is \textit{identifiable} highly depends on the causal structure used for answering it. There is a rich literature on theoretical work regarding \textit{\acl{cr}}, e.g., Do-calculus \cite{huang_pearls_2006}, c-components \cite{jin_tian_studies_2002} and identification of joint \cite{huang_identifiability_2006, ilya_shpitser_identification_2006, shpitser_complete_2008} and conditional interventional distributions \cite{shpitser_identication_2006} even under partial observability \cite{lee_causal_2020} for the sake of commonsensical decision making and policy evaluation \cite{pearl_causality_2013}.

\subsubsection{Connection to Machine Learning Insufficiencies}
\label{sec:causal_reasoning_ML_connection}
Although \textit{\ac{ci}} and \textit{\ac{ml}} arose separately, there is, now, an increasing interest in both fields to benefit from the advances of the other \cite{scholkopf_toward_2021}. The intersection of both fields gives rise to Causal Machine Learning (CausalML). For a detailed review on the state of this nascent field and open problems see \cite{kaddour2022causal}. Despite, many \acs{ml}-methodologies still do not consider causality, but keep modeling correlations between variables, although the maxim “\textit{Correlation does not imply causation}” proves itself true on every occasion. Indeed, the open problems of today's \acs{ml}-systems like insufficient \textit{robustness}, \textit{generalizability} and \textit{explainability} are closely related to the lack of causality considerations. See \cite{scholkopf_toward_2021} for an excellent review on this topic. Further discussions from various domains on why causality is linked to \acs{ml}-deficiencies and on how to equip \acs{ml} with \textit{\acl{ci}} can be found in \cite{prosperi_causal_2020, castro_causality_2020, scholkopf_causality_2019, richens_improving_2020, pearl_seven_2019, pearl_theoretical_2018}.

We will next discuss already existing applications of \textit{\acl{cr}} in the field of \textit{\acl{ml}}. For this purpose, we highlight a couple of publications from various domains that take aspects of \textit{\acl{cr}} explicitly into account for the sake of overcoming today's \acs{ml}-deficiencies. 
Eventually, we get on to its potential applications in the context of \textit{\ac{ad}}.

\subsubsection{Applications}
\label{sec:causal_reasoning_applications}
\paragraph{\textbf{Policy Search / Decision Making}}
\label{sec:causal_reasoning_policy_search}

As for policy search in \textit{\ac{mbrl}} (see \autoref{Knowledge_Integration/reinforcement_learning} for more information on \ac{rl}), \citeauthor{buesing_woulda_2018} \cite{buesing_woulda_2018} present an algorithm called \textit{\ac{cfgps}} that utilizes counterfactual reasoning explicitly for improving an agent’s policy. Thus, it effectively combines \acs{rl}-concepts with those from \acl{ci} to enhance the sampling efficiency on required experiences. Moreover, \acs{cfgps} is thought of as an extension of \textit{\ac{mbps}} algorithms. In \acs{cfgps}, the environment is modeled as a \textit{\ac{pomdp}}, but cast into the \acs{scm}-framework, where its exogenous variables summarize all aspects of the environment that cannot be influenced by the agent and where its causal mechanisms connect state-action-pairs to new states. The central idea of \acs{cfgps} is that, instead of running an agent on scenarios sampled from scratch from a model, one infers these scenarios in hindsight from given off-policy data (real experience, see \autoref{Knowledge_Integration/reinforcement_learning} for more information on off-policy \acs{rl}). Then, the agent’s policy is evaluated and improved on model predictions of alternate outcomes consistent with both the inferred scenarios and the given or learned causal mechanisms, but under counterfactual actions (i.e., actions that had not actually been taken), while keeping everything else the same. Concretely, while \acs{mbps} algorithms usually sample from a prior trajectory distribution for model rollout, \acs{cfgps} samples trajectories from a posterior distribution under a certain model that is anchored in off-policy data from another policy.

Based on these ideas, \citeauthor{hart_counterfactual_2020} \cite{hart_counterfactual_2020} propose a \textit{\ac{cpe}} algorithm with which the authors assess the safety of an ego vehicle’s policy prior to its execution in the actual world using counterfactual worlds. In a counterfactual world, the policy of at least one vehicle near the ego vehicle is replaced to estimate the impact another vehicle’s policy may have on the ego vehicle’s decision. To validate their method, the authors learn a policy using a \ac{sac} \cite{haarnoja_backprop_2016} \acs{rl}-approach in a lane merging scenario. In the application-phase, the ego vehicle’s policy will only be executed after \acs{cpe} had been performed and if the policy is found to be safe. Here, safety means that the expected collision rate of the current policy does not exceed a predefined threshold. Otherwise, the ego vehicle follows a simple lane-following policy.

Unlike \citeauthor{buesing_woulda_2018} \cite{buesing_woulda_2018}, who use counterfactuals to approximate draws from the interventional distribution, Oberst \& Sontag \cite{oberst_counterfactual_2019} treat the counterfactual distribution as the primary object of interest. The authors point out that evaluating an \acs{rl}-policy on observational (off-policy) trajectories is challenging, as it is often prone to issues such as confounding and a lack of introspection. To overcome the latter, they propose to pair observed with counterfactual trajectories, so that a domain-expert can “sanity-check” a proposed policy, especially in high-risk settings, e.g., healthcare. In particular, \citeauthor{oberst_counterfactual_2019} introduce an off-policy evaluation procedure based on counterfactual distributions in \acsp{pomdp} with discrete states and actions and stochastic state transitions. There, they try to answer counterfactual questions such as: “\textit{How would a specific trajectory have changed, had the policy been different and had all other variables (including noise) been the same?}” The main idea behind their approach is that if the counterfactual trajectory is unreasonable given an observed trajectory (for instance, an individual patient’s full health record), the policy might be flawed.
The authors finally demonstrate their method on a synthetic environment of sepsis management. From comparison of divergent counterfactual and observed trajectories, they reveal dangerous failure modes of the learned policy, even when off-policy evaluation is overly optimistic.

\citeauthor{richens_improving_2020} \cite{richens_improving_2020} compare the accuracy of associative diagnostic algorithms to their counterfactual counterparts for the purpose of clinical decision making. The authors argue that the inability of associative algorithms to disentangle correlation from causation can result in sub-optimal or even dangerous diagnoses. Reformulating diagnosis as a counterfactual inference task, they show that counterfactual analysis, in contrast to standard associational approaches, yields expert clinical accuracy on electronic health records, especially for rare and very-rare diseases. In their analysis, they compare the diagnostic accuracy of ranking diseases using traditional associative posterior probabilities to the results obtained from counterfactual inference. To be more precise, the authors set up a disease model in terms of a three-layer Bayesian Network with a Noisy-Or-operation connecting diseases to symptoms. A team of doctors and epidemiologists parameterize this model using prior knowledge, while prior and conditional probabilities of disease and risk factors are attained from multiple independent medical sources and doctors.

\citeauthor{he2022causal}~\cite{he2022causal} present \textit{CausalCF}, a causal \acs{rl} approach using associations, interventions and counterfactuals for training an agent applied to robotic manipulation tasks. The authors use elements from \textit{\ac{cophy}} \cite{baradel2019cophy}. \acs{cophy} is concerned with the problem of counterfactual learning of object mechanics. Having observed a mechanical experiment, e.g., a tower of building blocks falling over, the task is to predict how the outcome would have changed assuming different initial conditions, such as when displacing or removing one object. The altered future is predicted given the altered past and a latent representation of confounders (e.g., the masses or coefficients of friction of the objects) learned by the model in an end-to-end fashion, without supervision of the confounders. CausalCF adapts the CoPhy architecture to \acs{rl} tasks and uses the causal latent representation for training an agent. The training is split into two stages. In the first stage, the model producing the latent representation is learned by making counterfactual predictions about future states of the environment, whereby actions are generated by a pre-trained agent. In the second stage, the causal latent representation from counterfactual training is concatenated to all the observations for the training of the \acs{rl} agent.

\paragraph{\textbf{Explainability / Interpretability}}
\label{sec:causal_reasoning_explainability}

In an attempt to find human-understandable explanations for the incentives driving the behavior of \textit{\ac{mfrl}}-agents, \citeauthor{deletang_causal_2021} \cite{deletang_causal_2021} illustrate a methodology for investigating the underlying causal mechanisms. In particular, they show that each question cannot be addressed by pure observation (Level 1 in the \textit{Causal Hierarchy}) alone, but instead requires conducting experiments with systematically chosen interventions (Level 2 in the \textit{Causal Hierarchy}) in order to get the correct causal evidences. For this purpose, an analyst formulates a causal hypothesis (\acs{scm}), conducts experiments with carefully chosen interventions and confirms the predictions made by the resulting causal model. Similarly, \cite{carey_incentives_2021, everitt_understanding_2019} use structural causal incentive diagrams, a hybrid of influence diagrams and \acsp{scm}, to reason about the causal pathways of decisions in the service of maximizing certain utility functions. 

In another approach, \citeauthor{madumal_explainable_2020} \cite{madumal_explainable_2020} use an \acs{scm} to derive causal explanations of the behavior of \acl{mfrl}-agents. In particular, they propose a method that learns an \acs{scm} during \acs{rl}, where the causal structure is given, but where the \acs{scm}’s structural equations are learned as multivariate regression models. Subsequently, the learned \acs{scm} generates explanations of behavior from counterfactual analysis.

A more direct approach of using \acsp{scm} in \ac{dl} is presented by \citeauthor{pawlowski_deep_2020} \cite{pawlowski_deep_2020}. The authors propose a framework capable of learning \acsp{scm} with \textit{\acp{dnn}} as causal mechanisms. In particular, they focus on Markovian causal models with mutually independent exogenous variables and with no latent confounders. Additionally, the authors present an approach capable of inferring exogenous noise (potentially up to approximation) using a tractable evaluation of the abduction step via (amortized) variational inference or normalizing (gradient) flows. To that end, Pawlowski et al. assume that each structural function in the \acs{scm} can be expressed as a functional composition of an invertible (low-level) function (e.g., a cumulative distribution function as used in the reparameterization trick) and a non-invertible (high-level) branch (e.g., a \acs{cnn} or a probabilistic decoder). The proposed method shows good performance on both real (brain \acs{mri} scans) and simulated data (Morpho-MNIST).

In contrast to the formal approaches, \citeauthor{dasgupta_causal_2019} \cite{dasgupta_causal_2019} recently showed that \textit{\acl{cr}} can also emerge via Meta-Reinforcement Learning (see \autoref{sec:meta_learning} for more information on Meta-Learning) when no knowledge of causality is explicitly provided to an agent. The agent is trained with \acl{mfrl} to solve a range of problems backed by (latent) causal structures, which themselves are represented as \textit{\acp{cbn}}. They split the experimental setup into an \textit{information phase} and a \textit{quiz phase}. During the \textit{information phase} the agent is allowed to collect information by observing or interacting with the environment (\acs{cbn}) to acquire causal knowledge of the environment. In the subsequent \textit{quiz phase}, the agent is asked to answer causal queries along all three rungs of the \textit{Ladder of Causation}. The authors conclude that the trained agent is able to perform \textit{\acl{cr}} implicitly from observational data, even in novel situations, and can even make counterfactual predictions without having access to a full causal inference machinery.

For a more comprehensive survey of both traditional and frontier methods in learning causality and relations along with the connections between causality and \acs{ml}, see \cite{guo_survey_2020}. Another survey \cite{moraffah_causal_2020} discusses interpretable / explainable models from a causal perspective and contrasts these with traditional interpretability approaches in \acs{ml}.

\paragraph{\textbf{\acl{ad}}}
\label{sec:causal_reasoning_autonomous_driving}
As advertised, we next examine the potential applicability of \textit{\acl{cr}} in the domain of \ac{ad}. In particular, we will explore the three main subtasks in \acs{ad}: \textit{Planning}, \textit{Situation Interpretation} and \textit{Perception}.

According to \citeauthor{pearl_causality_2013} \cite{pearl_causality_2013}, a \textit{plan} is an ordered sequence of value assignments to the control variables (actions). \textit{Planning} itself is an optimization procedure of a sequence of concurrent or sequential actions, where each action may be influenced by its predecessors in the plan, with a view of achieving a specified goal (like arriving at a desired destination or preventing accidents). Therefore, \textit{Planning} is closely connected to causality. \textit{\acl{cr}} can be employed not only for planning the ego vehicle's trajectory, but also for predicting physically reasonable trajectories of other vehicles in its immediate vicinity. A prerequisite for this approach is that existing knowledge about the vehicles' motion (governed by physical laws) is modeled by an \acs{scm} in both diagrammatic and algebraic form. As the resulting \acs{scm} serves as a lightweight simulator of the vehicle dynamics, it can be leveraged to generate physically reasonable trajectories by repeatedly calling the \acs{scm} on sequential state-action tuples, one for each time step. As for planning, actions in the plan can be either dictated by the ego vehicle's learned policy (a mapping from state to action) or enforced by hard interventions on the \acs{scm} (setting the action to a fixed value irrespective of the policy's output). The obvious advantage of using an \acs{scm} modeling the vehicle dynamics is that an agent does not have to interact with the environment at all, but can rather \textit{act} (in the sense of \textit{interventions}) in an imagined space that is created by the \acs{scm}'s predictions. For instance, the \acs{cfgps} method presented in \cite{buesing_learning_2018} could potentially be transferred to the \acs{ad} domain, but must first be adapted for application in continuous action spaces.

Besides \textit{Planning}, \textit{\acl{cr}} naturally lends itself to \textit{Situation Interpretation}, insofar as that a causal analysis of a particular plan assesses an action's impact on the trajectory's subsequent evolution. There are basically two forms of causal analyses that matter here: The \textit{counterfactual} and the \textit{interventional} analysis. For instance, the \textit{counterfactual} analysis of an existing trajectory, that is recorded under a certain policy, can unveil explanations / causes (like specific actions taken at a certain time step) for events / effects (such as colliding vehicles after an elapsed time interval). The latter only works as a \textit{post hoc} method, \textit{after} an event has occurred, while the \textit{interventional} analysis (an \textit{ex ante} approach) does not make use of a recorded trajectory. As opposed to the counterfactual analysis, the interventional analysis assesses the importance of certain actions for future events in specific situations and is solely based on the assumptions encoded in the \acs{scm}. On the whole, \textit{\acl{cr}} can give answers to questions like "\textit{Which actions have led or might lead to an accident in a certain situation?"}

As for perception, things are not that easy, since \textit{\acl{cr}} struggles to derive high-level concepts (like objects in images) from low-level features (e.g., image pixels). However, applications of \textit{\acl{cr}} to vision tasks can be found in, e.g., \cite{wang_visual_2020, qi_two_2020, yang_deconfounded_2020}. In \citeauthor{wang_visual_2020} \cite{wang_visual_2020}, for instance, high-level features are extracted by an \acs{rcnn}, while the high-level features are post-processed and used to make these plausible by tools of \textit{\acl{cr}}, which are also part of the learning objective. 
\subsection{Rule Conformity}
\label{sec:rule_conformity}
\textit{Authors: Abhishek Vivekanandan, Etienne B\"{u}hrle, Hendrik K\"{o}nigshof}

\smallskip
\noindent
\subsubsection{What is Rule conformity?}

According to ISO 9000:2000 norms, conformity is defined as fulfilling certain requirements. Rules are a set of formal codes derived from local laws and socio-behavioural structures while effectively considering the \acf{odd} of the algorithm. Conforming to rules is a crucial factor while attempting to categorize the risk an action could lead to, making the \acfp{av} safe not only for the passengers but also for other traffic participants like pedestrians and cyclists etc.

Formalizing a subset of traffic rules and associating priorities to the vehicles not only achieves compliant behavior but also maintains the vehicle in a safe state, as can be seen from the works \cite{rizaldi_formalising_2015}. 

\subsubsection{Rules and Safety}

From the literature works \cite{mobileye_implementing_2021, nister_introduction_2019}, one can define mathematical rules to formalize common-sense notions/behaviors such as maintaining a safety gap between vehicles, safe behavior to merge and as such to avoid maneuvers which could lead to possible collisions. Depending upon the \ac{odd} one could broadly categorize the behaviors based on formal rules; which are stated by law and could also be derived from the environment, these include stop signs, appropriate speed markings, solid double or dotted lane markings. Each of these briefly define laws which demand an integration into the learning process of the model. In contrast to rules which are explicitly stated by law, informal rules, on the other hand, are difficult to formalize but expected to be followed to avoid collision or (and) increase the comfortability of the passengers. For example, to leave way for an emergency vehicle, it is acceptable to cross the solid line or pass a red light at an intersection paving a gap for the ambulance to pass through. In this situation if there exists a maneuver where breaking the rule does not endanger both the passenger and the \acp{vru} then under defined priority this maneuver could be executed with strict adherence to safety.

\subsubsection{Always be compliant, but have common sense} 

When to follow or to relax the rules is a complex decision process for \acp{av} to comply with, since recognizing the situation itself which lead to such behavior requires \emph{General Intelligence} which could be seen in humans due to the implicit notion of common sense. Let us consider the example of achieving a collision free maneuver in case of an unanticipated merge by an adjacent vehicle driving on the right onto our lane. To generate a compliant behavior, the algorithm should respect the properties iterated below. This hierarchy is defined from the perspective of Ego Vehicle, following a simple bicycle model.

\begin{enumerate}
    \item Detects and tracks the surrounding objects via forecasting networks, which considers the uncertainties based on the object class. For example, a cyclist could emit different states when compared with pedestrians whose degrees of movement are higher. 
    
    \item Calculate the safe lateral and longitudinal gaps defined via Formal rules.
    
    \item Based on 2; a continuous trajectory path which complies not only with the traffic rules but provides safety and comfortability assurances for the passengers and \acp{vru} involved. 
    \begin{enumerate}
        \item If no such valid paths exist, priority should be given to the next best plan, which could be switched at optimal cost. 
        
        \item Backup plans could be pre-calculated (via multi-modal trajectory forecasting) by ensuring that a delicate balance is maintained between relaxation of rule compliance versus safety of the passengers. For example, if it means that lives could be saved by crossing over a double solid lane, then it should be taken.
        
        \item Bringing the vehicle to a \ac{mrc} should also be evaluated to bring the vehicle to a safe state.
    \end{enumerate}
\end{enumerate}
    
Within the current scope of this work, we do not consider the effects of uncertainty and the ethics involved about human lives. Normally, these plans should be calculated under uncertainty to reduce the risk posed by the real world. 

\subsubsection{Constraints and Conformity}

 Embedding constraints on the learning functions are a direct way to ensure conformance of a neural network to a certain objective by defining upper and lower bounds to the output space. Modern \acp{av} utilize machine learning model which are trained with algorithms based on Stochastic Gradient Descent via regularizers and optimizers; placing \emph{soft constraints} in the subspace. Here, regularizers are a set of rules to fine-tune the expected behavior. On the other hand, when hard constraints are a part of the neural network structure \cite{karpatne_physics-guided_2018, mohan_embedding_2020} the networks should be attentive to the physics involved, and their predictions should not deviate from the said constraints which are imposed. With which we expect the outputs from the networks to be restricted to the physical subspace, ensuring a compliance to the bounds placed.  When compared with their Neural Network counterparts trained via soft constraints, physics guided nets respect the bounds implicitly by never deviating from them. 

Conformance to rules could be rephrased as a constraint satisfaction problem by reasoning about the temporal properties of the system. Ensuring the safety of the vehicles during planning operation is a crucial task such that possible collisions could be mitigated. As a result, expressing rules and satisfying those specifications in the form of temporal logics \cite{fainekos_temporal_2005} and verifying the system properties using \acf{stl} \cite{roehm_stl_2016}, where rules are checked and verified for their formal conformance. Although, works from \citeauthor{censi_liability_2019} take a different approach by defining rules in the form of relational hierarchical structures represented via \acf{ltl}. This allows one to have the flexibility to describe the relative priorities between rules which result in a control command to the planner deemed compliant according to the specification. These scalable “rule book” could explicitly capture the common-sense and informal rules to some extent in the context of automated driving.  
\subsection{Artificial Intelligence Verification}
\label{sec:ai_ver}
\textit{Author: Tino Werner}

Deep \ac{ai} models are well-known for their often very good prediction accuracy. However, the authors in \cite{katz_reluplex_2017} point out that \ac{ai} often lack of generalization ability in the sense that there is no guarantee that it behaves correctly on unseen data points. Hence, \ac{ai} verification is a necessary step in order to prove that a particular \ac{ai} model works correctly. A major difficulty is the non-linearity of most activation functions so that any kind of \ac{ai} verification leads to non-convex problems. Moreover, due to a \ac{nn} being composed of non-linear and transcendental real-valued functions, it is in general undecidable whether a particular property holds \cite{franzle_efficient_2006}. Rational approximations of real numbers are generally too expensive \cite{pulina_never_2011}. 

Standard approaches in order to perform \ac{ai} verification are over-approximation of non-linear functions by intervals or sets that cover the true function values with certainty, while for time-dependent models, the reachable sets are over-approximated. Usually, the constraints that the \ac{ai} has to satisfy are negated so that one checks whether there exists points for which the negated constraints are satisfied (i.e., the \ac{ai} fails on these points) using approaches like \acf{sat} (e.g., \cite{franzle_efficient_2006}, \cite{franzle_hysat_2007}, \cite{scheibler_recent_2013}, \cite{scheibler_accurate_2016}), \acf{smt} (e.g., \cite{nieuwenhuis_solving_2006}, \cite{de_moura_z3_2008}), stochastic \ac{smt} (e.g., \cite{franzle_stochastic_2008}, \cite{teige_stochastic_2008}, \cite{teige_constraint-based_2011}, \cite{gao_solving_2015}, \cite{gao_csisat_2016}, \cite{gao_verification_2017}), \ac{milp} or \acf{smc} (e.g., \cite{nuzzo_calcs_2010}, \cite{shoukry_smc_2018}). 

Although many verification algorithms are motivated by safety and security, there are other highly relevant aspects of trustworthy \ac{ai} (see \cite{wing_trustworthy_2021} for a critical overview) apart from safety and security for which there are already verification algorithms, e.g., for privacy \cite{zhang_lightdp_2017} or fairness \cite{albarghouthi_fairsquare_2017}. As for conformity checking, the verification techniques can be used as tools in order to verify whether the knowledge constraints imposed on the \ac{ai} models are satisfied, lifting the concept to the whole field of informed \ac{ml}, including applications in autonomous driving.

The authors in \cite{huang_survey_2020} provide a thorough survey on safety verification and trustworthiness (certification combined with explanation as defined in their work). They categorize the methods for safety verification into constraint solving, over-approximation, global optimization and search-based methods. Most popular \ac{ai} verification approaches are those based on constraint solving and over-approximation. Constraint solving indicates that the requirements can be encoded as a set of constraints. Then, transforming the verification problem into constraint verification provides deterministic guarantees and can be achieved using \ac{smt}, SAT or \ac{milp} solvers. Over-approximation is a technique that intends to cover the whole output range of a neuron by simple constraints. One well-known example (see \cite{ehlers_formal_2017}) is to over-approximate a ReLU node by a linear constraint. If one knows that the input $c$ for the ReLU node is contained in some interval $[l,u]$, maybe by having applied interval arithmetic, the output $z$ of the ReLU node is contained in the set determined by the three linear constraints $z \ge 0$, $z \ge c$ and $z \le u(c-l)/(u-l)$. The advantage is that this over-approximation avoids to distinguish different cases (for ReLU, whether the input is non-negative or negative) and covers a non-linear input-output relationship with linear constraints, therefore allowing for linear solvers. As a drawback, the set corresponding to the over-approximation contains non-feasible points, but this issue is encountered in literature with more tight over-approximations. Over-approximation is also used for reachability analysis, in particular for time-dependent \acp{nn}, in order to over-approximate the states that can be attained by the system, given some input space. Further categorizations include adversarial robustness verification and testing. Other surveys include \cite{leofante_automated_2018} who provide an overview of \ac{dnn} verification algorithms and outline that one may use model reduction by pruning or by distillation, i.e., training a smaller \ac{nn} with only insignificant accuracy loss. Alternatively, one may decompose the \acp{nn} so that one only verifies critical parts of the models as in \cite{huang_safety_2017}. A thorough survey on the algorithms for verification of \acp{dnn} themselves is given by the authors in \cite{liu_algorithms_2021} who also show code snippets.  See \cite{xiang_reachability_2018} for another survey. The authors in \cite{urban_review_2021} provide a survey of the application of formal methods in machine learning and distinguish between complete and incomplete (scalable, but prone to false positives) methods.

\subsubsection{Adversarial robustness} 

There are discussions in literature in order to distinguish \ac{ai} verification from checking adversarial robustness (e.g., \cite{bastani_measuring_2016}) since the latter is achieved if there are no adversarial samples around the training points which only covers a limited radius around these points, leaving a large part of the input space unexplored. At most, only statistical safety guarantees can be shown or, as spelled out by the authors in \cite{katz_reluplex_2017}, only an "approximation of the desired property" can be verified. The authors in \cite{xiao_training_2018} point out that \ac{ai} certification is based on over-approximating the adversarial polytope so that the certification output can be interpreted as a relaxed verification task where safe input points may be flagged as unsafe, see also \cite{everett_certified_2020} for the terminology. Algorithms for verifying adversarial robustness of neural networks are usually based on over-approximation, typically by over-approximating the activation functions and therefore the reachable activations. 

The authors in \cite{weng_towards_2018} propose the FastLin algorithm that over-approximates ReLU activations linearly, and also the FastLip algorithm that bounds the local Lipschitz constant of a ReLU network. Both algorithms do not require optimization algorithms but compute the bounds explicitly layer per layer. The authors in \cite{wong_provable_2018} (ConvDual) propose a convex outer approximation of the set of reachable ReLU activations in feed-forfard \acp{nn}.  The authors in \cite{singla_second-order_2020} consider feed-forward networks with differentiable activations and use global curvature bounds while the authors in \cite{zhang_efficient_2018} (CROWN) and \cite{lyu_fastened_2020} (FROWN) consider feed-forward networks with general non-linear activations and bound them with linear and quadratic functions. In \cite{fazlyab_safety_2020}, quadratic abstractions of nonlinear activation functions are considered. \cite{singh_boosting_2019} (RefineZono) and \cite{singh_abstract_2019} (DeepPoly) cover ReLU, maxpool and sigmoid-type activations and can also handle \acp{cnn} using \ac{milp} resp. over-approximation. Reachability analysis for \acp{cnn} has been proposed by the authors in \cite{tran_verification_2020} who use the ImageStar set representation which can both perform exact and over-approximated set-based analysis of \acp{cnn} using linear programming which is then used for adversarial robustness certification. 

There are also approaches that do not directly over-approximate activations. For example, the authors in \cite{bastani_measuring_2016} propose an adversarial robustness metric and show, for ReLU activations, how to approximate the constraint system corresponding to the robust set by a linear program. The authors in \cite{xiang_output_2018} over-approximate the reachable set for the output layer which, for monotonic activation functions, can be done using convex optimization. The authors in \cite{cheng_maximum_2017} consider feed-forward ReLU and softmax networks and compute maximum perturbation bounds using \ac{milp}. \cite{gopinath_deepsafe_2018} (DeepSafe) propose an approach for classification networks by combining clustering and Reluplex which encodes ReLU activations via the Simplex algorithm (see \cite{katz_reluplex_2017}). The authors in \cite{tjeng_evaluating_2017} (MIPVerify) compute an over-approximation of the set of tolerable perturbations for piece-wise linear feed-forward ReLU networks using \ac{milp}. The authors in \cite{ruan_global_2018} compute the adversarial robustness of \acp{cnn} w.r.t. the $l_0$-norm.

Some works focus on better over-approximation of ReLU activations. A speed-up technique for ReLU networks has been proposed by the authors in \cite{xiao_training_2018} who point out that a major issue of ReLU network verification is that ReLU nodes branch in the sense that one has to respect both cases, i.e., active or inactive nodes, i.e., the output is non-zero or zero. The authors in \cite{xiao_training_2018} concentrate on minimizing the number of such splits by maximizing the number of stable ReLU nodes which stay active or inactive, therefore calling their approach ReLU stability.
As for time-dependent \acp{nn}, the authors in \cite{everett_certified_2020} (CARRL) consider deep \ac{rl} (\ref{Knowledge_Integration/reinforcement_learning}) and compute lower bounds of the $Q$-values. 
As for the adversarial robustness of \acp{binnn}, an \ac{milp} approach has been proposed by \cite{narodytska_verifying_2018}. The certification of quantized \acp{nn} which imposes the additional difficulty of quantization errors is considered by the authors in \cite{sena_verifying_2021} who adapt software model checking techniques to this setting. 

From the perspective of informed \ac{ml}, such adversarial robustness verification/certification approaches are tailored to safety and therefore not directly applicable in order to verify the satisfaction of knowledge constraints.

\subsubsection{Verifying non-recurrent \acp{nn}} 

\ac{ai} verification itself is indeed tailored to some set of constraints, for example, given by inequalities, equalities or temporal logic. One of the first algorithms of this kind was NEVER from \cite{pulina_challenging_2012} where \acp{mlp} (or \acp{ffnn}) are verified according to some \ac{smt} formula. The basic idea is to over-approximate the activation function, leading to an "abstraction" or "abstract activation function" given in terms of intervals. This abstraction is propagated, leading to abstractions of the outputs. The satisfaction check is then done using HYSAT \cite{franzle_hysat_2007} w.r.t. the negated constraints so that either the constraint set is unsatisfiable which is equivalent to the result that there is no point that violates any of the original constraints, so the \ac{ai} is safe. In this case, a refinement step is executed in order to find a tighter approximation of the activation function. Otherwise, the algorithm finds an input interval so that the abstract output violates the constraints, i.e., the input interval serves as abstract counterexample which itself is then checked for feasibility.
Note that NEVER is not restricted to ReLU activations but can, in principle, handle general non-linear activation functions. Many works on non-recurrent \ac{nn} verification however focus mainly on ReLU activations.

Since the NEVER algorithm, as stated by the authors in \cite{katz_reluplex_2017}, is only feasible for networks with at most 20 hidden nodes, \cite{katz_reluplex_2017} proposed the Reluplex algorithm which adapts the Simplex algorithm to ReLU constraints. Their algorithm inherits the soundness (the decision of satisfiability resp. unsatisfiability is correct) and completeness (any starting configuration leads to a decision) properties from the Simplex algorithm. Reluplex was extended by the Marabou platform provided by \cite{katz_marabou_2019} which speeds up Reluplex by a lazy search strategy so that even \acp{cnn} with piece-wise linear activation functions can be handled. However, as pointed out by the author in \cite{ehlers_formal_2017}, Reluplex is not feasible for max-pooling activations since replacing them with multiple ReLU layers would be too cumbersome. The author in \cite{ehlers_formal_2017} himself proposes the algorithm Planet where general nonlinear activation functions in \acp{ffnn} are approximated in a piece-wise linear fashion and linear programming is used in order to approximate the \ac{nn} behavior and discard infeasible parts of the search space in advance. 

Apart from the works that focus on ReLU over-approximation in the context of adversarial robustness, there are additional works on ReLU over-approximation in the context of \ac{ai} verification. A symbolic linear relaxation of ReLU units has been proposed by the authors in \cite{wang_efficient_2018} who introduce the Neurify algorithm which can handle much larger \acp{nn}. In \cite{henriksen_deepsplit_2021}, the indirect effect of splitting a ReLU node on the subsequent nodes is estimated, leading to the DEEPSPLIT algorithm. ReLU over-approximation has also been considered by the authors in \cite{rossig_advances_2021} who combine \ac{mip} and branching. 

Apart from the verification of ReLU networks, there are also approaches that consider other activation functions. Lipschitzian activation functions (which cover convolutional and fully-connected ReLU layers as well as max-pooling and contrast-normalization layers as shown in \cite{szegedy_intriguing_2013}) have been considered by the authors in \cite{ruan_reachability_2018} who show that softmax, hyperbolic tangent and sigmoid activation layers are also Lipschitz and who derive linear constraints for an output reachability problem for \acp{ffnn} with such activation functions. They call their software tool DeepGO. In \cite{dutta_output_2018}, feed-forward ReLU networks are considered for which output range analysis is performed with the goal to determine a tight interval that covers the network output for the whole input space. Their algorithm Sherlock is based on \ac{milp}. In \cite{meyer_reachability_2021},  the output of a \ac{ffnn} is over-approximated using mixed monotonicity. The authors in \cite{lomuscio_approach_2017} provide an \ac{milp}-type variant of Reluplex for feed-forward ReLU networks. In \cite{guidotti_pynever_2021}, star sets are considered in their pyNeVer algorithm. 

The authors in \cite{bunel_unified_2018} consider piece-wise linear neural networks and inequality verification tasks in the sense that, e.g., the output of the network has to be non-negative. They propose a general branch-and-bound algorithm that splits the input domain into sub-domains where sub-domains are pruned away if the lower bound of the network is too high there, i.e., if the desired property cannot be falsified there. They identify Reluplex and Planet as special instances of their algorithm. Their ReluVal algorithm is much faster than Reluplex on small networks but it cannot cope with large \acp{nn}. A speed-up strategy based on symbolic interval arithmetic in order to avoid \ac{smt} solvers has been proposed by \cite{wang_formal_2018}. 

Pruning approaches where several parts of a deep \ac{nn} are pruned in order to make verification algorithms feasible have been proposed in \cite{gokulanathan_simplifying_2020} and \cite{guidotti_verification_2020}. The verification of \acp{binnn} has been considered in several works. In \cite[Thm. 1]{cheng_verification_2018}, it is proven that verifying a \ac{binnn} is NP-complete. They show how to encode \acp{binnn} as SAT instances and propose an inter-neuron factoring approach for \ac{binnn} verification, see also \cite{jia_efficient_2020}. In \cite{amir_smt-based_2021}, the Reluplex algorithm is adapted to \ac{binnn} by extending it to sign activation functions which allow for strict binarity using sign constraints and new splitting rules that can handle them.

\subsubsection{Verifying time-dependent \acp{nn}} 

The techniques listed so far consider static models in the sense that they do not invoke a time component. For time-dependent models, there are already many techniques in literature that can handle these cases. We distinguish between hybrid automata, controller systems where the \ac{nn} maps states into actions and \acp{rnn}. A survey on stochastic hybrid systems, including verification, is given by \cite{lavaei_automated_2021}.

\paragraph{\textbf{Verifying hybrid automata}} 

As for the verification of hybrid automata, popular approaches include a reachability analysis where one aims at guaranteeing that the system never can attain unsafe states, often in a weakened fashion, for example for a finite horizon or in terms of probabilistic guarantees. 

The authors in \cite{wang_sreach_2014} propose SReach for safety verification of stochastic hybrid systems. The idea is to verify a weakened version of the requirements over a $k$-step horizon. Random variables are sampled and a $\delta$-complete analyzer and dReach from \cite{gao_delta-complete_2014} is applied, i.e., inequality constraints like $a>0$ are replaced by $a>-\delta$ so that the algorithm either returns non-satisfaction or $\delta$-satisfaction. However, as they restrict themselves to $k$-step reachability, they cannot provide guarantees for infinite horizons. The authors in \cite{shmarov_probreach_2015} propose Probreach, an algorithm that computes the probability that a hybrid system reaches an unsafe region of the state space. They more precisely consider probabilistic $\delta$-reachability and partition the state space into iteratively finer intervals and check the safety on these intervals. According to the authors in \cite{huang_probabilistic_2017}, this method is computationally very expensive. In \cite{ivanov_verisig_2019}, the Verisig algorithm for safety verification of closed-loop \ac{nn}-controlled systems is proposed. They translate the \ac{dnn} into a hybrid system with sigmoid activation and apply reachability analysis. Their method also covers actor-critic \ac{rl} methods. 

The authors in \cite{xue_probably_2019} instead propose a probabilistic safety verification of dynamic hybrid systems for infinite time horizon by computing PAC (probably approximately correct) barrier certificates (differentiable functions that map each initial state to a real number where unsafe states always get a positive number and whose gradients satisfy a regularity condition) using linear programming which amounts to chance-constrained optimization of some convex cost function. These chance constraints are circumvented by randomly drawing control inputs and enforcing the solution to satisfy the corresponding hard constraints for each realization.

\paragraph{\textbf{Verifying \ac{nn} controllers}} 

There are many reachability approaches for \ac{nn} controllers based on an over-approximation/abstraction of reachable states.
Reachability approaches for the verification of \ac{nn} controllers are given by the authors in \cite{julian_reachability_2019} who over-approximate reachable states for \ac{nn}-controlled agents, see also \cite{xiang_reachability_2018}, \cite{xiang_reachable_2020}. The authors in \cite{tran_safety_2019} propose star sets instead of polyhedra for reachability analysis of ReLU-\ac{ffnn}-based systems by over-approximation while \cite{schilling_verification_2021} combine Taylor models and zonotype set representations for \ac{nn} controllers for a more precise reachability analysis, and \cite{everett_efficient_2021} consider reachability analysis for closed-loop \ac{nn}-controlled systems and for a discrete-time linear time-varying system and derive explicit bounds for the next state by respecting convex constraints and input constraints. The authors in \cite{ivanov_verisig_2021} propose Verisig2.0 which approximates \ac{nn} controllers by Taylor models. Quadratic abstraction of nonlinear activation functions for reachability analysis is considered by the authors in \cite{hu_reach-sdp_2020}. In \cite{sun_formal_2019}, finite state abstraction for reachability analysis for discrete-time continuous linear dynamics and a ReLU controller network are considered. They propose to partition the state space into a set of polytopes according to the laser angles of the LIDAR perception in their context. They encode the problem as a \ac{smc} optimization problem \cite{shoukry_smc_2018} in order to detect the set of safe resp. unsafe states (which can result in an unsafe trajectory). In \cite{huang_reachnn_2019}, ReachRNN is proposed which approximates a Lipschitz-continuous \ac{nn} (including ReLU, hyperbolic tangent and sigmoid activations which have shown to be Lipschitz in \cite{ruan_reachability_2018}) by Bernstein polynomials and estimates the approximation error bound. Their algorithm is parallelized by the authors in \cite{fan_reachnn_2020}. Bernstein polynomials as a tool for over-approximating node activations in \ac{nn} controllers have also been used in the POLAR algorithm from \cite{huang_polar_2021}. In \cite{sidrane_overt_2021}, piecewise linear approximations of ReLU activations for \ac{nn} controllers in their OVERT algorithm are considered.

Nevertheless, there are also algorithms that aim at finding probabilistic guarantees that the system does not reach unsafe states. 
In \cite{bacci_probabilistic_2020}, the MOSAIC algorithm for probabilistic guarantees for \ac{rl} based on iterative state abstractions via \acp{mdp} is introduced, assuming that the policies are memoryless and deterministic. They consider $k$-step reachability and the goal is to under-approximate the safe set or to over-approximate the worst-case probability of failure. Having state abstractions $\hat s$, an environment abstraction is a mapping $\hat T: \hat S \times A \rightarrow \hat S$ which can be constructed using interval arithmetic, leading to abstracted \acp{mdp}, where the state abstractions are refined by splitting the corresponding regions. The authors in \cite{kazak_verifying_2019} propose Verily for the verification of deep \ac{rl}-controlled systems. The tool provides a concrete scenario where the safety requirements are violated or where the liveness requirements cannot be met. The algorithm searches for a physically reasonable sequence of states (i.e., every state can be reached from the previous state) and that end up in a bad state. They only consider paths of limited length and use Marabou for solving the verification problem. 

The authors in \cite{katz_generating_2021} over-approximate deep policies for finding probabilistic safety guarantees. They partition the input space into cells and perform the verification on each cell for over-approximated policies that map such whole cells into the action space. They propose an adaptive strategy how to find a suitable partition.

The authors in \cite{tuncali_reasoning_2018} compute barrier certificates for \ac{ffnn}-based cyber-physical systems, enhanced with simulation. The idea is to generate a set of simulation traces by randomly sampling initial states and by solving a linear program. More precisely, the validity of the barrier certificate conditions for each candidate are checked using dReal \cite{gao_dreal_2013}. The authors in \cite{huang_probabilistic_2017} also apply barrier certificates for probabilistic safety verification of hybrid systems and assume that the behavior of the system is defined by polynomial constraints but that the initial states are stochastic, so the goal is to bound the probability of facing initial states that lead into an unsafe region. 

Other approaches include \cite{akintunde_formal_2020} where a \ac{ffnn}-based agent interacting with a non-deterministic environment is studied and where the verification problem is re-written as an \ac{milp} with recursive encoding, extending their former work \cite{akintunde_reachability_2018} which considers a static environment. Their algorithm is called NSVerify. In \cite{carr_verifiable_2020}, verification of \ac{rnn}-based policies for POMDP based on discretization techniques is proposed. The authors in \cite{stahl_online_2020} study a general online verification concept using an online verification monitor (supervisor) aiming for full safety coverage of a black-box model. In \cite{xiong_scalable_2021}, it is proposed to decouple training and verification for better scalability and invoke verified linear controllers (safety shields) to guarantee the safety of the training procedure and based on the verified properties, \ac{rl} is used to learn other properties.  The authors in \cite{pek_online_2019} provide an online verification framework for motion planning in order to guarantee that a safe plan is computed, i.e., collision-free given all possible future plans of other agents/dynamic obstacles.  For each traffic participant, a mathematical model for the feasible future behaviors is computed in an over-approximative way. The fail-safe trajectories are computed by solving a quadratic optimization problem with a convex cost function and kinematic constraints.

\paragraph{\textbf{Verifying \acp{rnn}}} 

As for the certification of the adversarial robustness of \acp{rnn}, \cite{ko_popqorn_2019} propose POPQORN which computes linear bounding planes for nonlinear activation functions and especially cross-linearity (products of activations) for \acp{rnn}, GRUs and \acp{lstm}. The relaxation of all non-linear operations of \acp{lstm}, in particular, products of sigmoid and tanh or sigmoid and identity operations, with linear convex operations so that bounds can be derived is considered in the algorithm R2 from \cite{ryou_fast_2020}. 

An approach that really aims at \ac{ai} verification for time-independent models available has been proposed by the authors in \cite{jacoby_verifying_2020} who propose invariant inference to reduce the \ac{rnn} verification problem to an \ac{ffnn} verification problem which has been introduced by \cite{akintunde_verification_2019}. The idea is to over-approximate the \ac{rnn} by the \ac{ffnn} of the same size by encoding time-invariant properties of the \ac{rnn} into the \ac{ffnn}. \cite{jacoby_verifying_2020} criticize that this unrolling approach produces a very large \ac{ffnn}, hence it is infeasible because an \ac{rnn} with $K$ memory units produces an \ac{ffnn} with $K(t-1)$ new nodes if $t$ is the number of time steps corresponding to the property that is to be verified. The \ac{ffnn} is constructed by replacing each memory unit by a standard neuron, connected with the original neurons and weights. Then, by bounding the values from each new node and properly adjusting these bounds so that they are neither too tight nor too weak, standard \ac{ffnn} verification techniques are applicable.

\subsubsection{Verifying \ac{nn} differences} 

There are some works that have lifted the verification concept to the analysis of differences of neural networks, for example, in order to verify the accuracy of a compressed network. For the original (deep) neural network $f$ and the compressed counterpart $f'$, the verification amounts to showing that $|f(x)-f'(x)|<\epsilon$ for all $x$ in the input space. The authors in \cite{paulsen_reludiff_2020} use symbolic interval analysis for analyzing differences in the values of neurons and gradients and call their algorithm ReluDiff which over-approximates the respective differences and checks in a forward pass if the resulting output difference exceeds $\epsilon$. According to the authors in \cite{paulsen_neurodiff_2020}, ReluDiff is too conservative, so the authors provide the extension NeuroDiff which is based on fine-grained convex approximations of the difference intervals. The whole concept has been lifted from feed-forward networks to \acp{rnn} by the authors in \cite{mohammadinejad_diffrnn_2021} who propose the DiffRNN algorithm in order to verify the equivalence of structurally similar \acp{rnn}. As ReluDiff cannot handle non-linear activation functions, including the products that occur in \acp{lstm}, they apply the \ac{smt}-solver dReal \cite{gao_dreal_2013} that can handle non-linear real-valued functions. In fact, they rewrite the activation differences $\sigma(x')-\sigma(x)$ by $\sigma(x+\delta_x)-\sigma(x)$ for $\delta_x=x'-x$ and compute tight bounding boxes corresponding to the ranges of $x$ and $\delta_x$ using dReal. The authors in \cite{lahav_pruning_2021} aim at finding neurons that can be removed from the \ac{dnn} (not only neurons with zero weight but also redundant ones), so they aim at verifying that the pruned network is equivalent to the original one. The idea is domain slicing so that the original \ac{nn} $N$ associated to domain $D$ is represented by a family of \acp{nn} where $NN_i$ is associated do domain slice $D_i$. For each $NN_i$, they apply \ac{milp} in order to check whether neurons are redundant.

\subsubsection{Repair of \acp{nn}} 

One question that the \ac{ai} verification algorithms do not answer is how to proceed if the models do not satisfy the requirements (cf. \cite{dong_towards_2020}). If an \ac{ai} verification algorithm concludes that the \ac{nn} does not satisfy the given requirements, i.e., the \ac{nn} is falsified, a na\"{i}ve approach could be to retrain the \ac{nn} and to hope that the retrained one satisfies the constraints. The idea of repairing a \ac{nn} however is to avoid expensive retraining by using the falsified \ac{nn} as warm start. As for the notion, the term "repair" has already been used in \cite{pulina_challenging_2012}, however, as pointed out in \cite{guidotti_never_2020}, their technique is essentially adversarial re-training based on counterexamples that NEVER detected. The authors in \cite{sohn_search_2019} propose Arachne which however, as pointed out in \cite{dong_towards_2020}, modifies the weights of neurons which are relevant for failures using a particle swarm optimization algorithm but it does not guarantee the correctness of the repaired model. The authors in \cite{yu_deeprepair_2020} propose DeepRepair, but it is essentially based on style-guided data augmentation using clustering so that unknown failure patterns can be learned, so it follows ideas like adversarial testing or DeepXplore. In \cite{usman_nn_2021}, NNRepair is proposed which modifies one intermediate layer (neuron values) or the output layer (decision boundaries) for classification \acp{dnn}. However, their approach is based on detecting adversarial examples for repair and on positive examples to keep the performance there.

As for true repair, recently, \cite{goldberger_minimal_2020} propose provably minimal modifications of \acp{dnn} in order to make them satisfy the given requirements. In order to quantify the modification, they propose a distance measure between two \acp{dnn} of the same architecture. The idea is to solve a constrained optimization problem where a modified \ac{nn} is computed so that the distance of the \acp{nn} is minimized while the modified \ac{nn} satisfies all requirements which enter as hard constraints. Due to non-convexity and high-dimensionality, they propose to solve layer-wise verification problems by computing the layer change matrix for the layer where a modification should happen so that all previous layers remain unchanged. The authors in \cite{dong_towards_2020} point out that the approach in \cite{goldberger_minimal_2020} aims at finding a repaired network that behaves correctly on multiple specific inputs while \cite{dong_towards_2020} want to have a verified repaired network that always behaves correctly. The modified layer can be the output layer itself which, according to \cite{goldberger_minimal_2020}, has the advantage to make the verification problem linear. The authors in \cite{dong_towards_2020} however argue that modifying only the output layer is often too late. The idea of \cite{dong_towards_2020} is based on a minimal modification by detecting the neurons which are most relevant for the failure and performing gradient descent w.r.t. these neurons. The goal is to find a repaired \ac{nn} that satisfies the required properties on each erroneous partition (on which the original \ac{nn} violated them). One starts with an initial input space partition and checks if a partition is possible into disjoint constraints. If a partition is possible, one checks for each partition if the \ac{nn} is verified. If a counterexample is detected, they aim at finding a modification of the \ac{nn} with the same architecture such that the violation loss is minimized. As for the repair itself, they order the neurons according to the magnitude of their gradients of the loss and fix a budget of neurons that are allowed to be modified. If after modifying the most relevant neuron the \ac{nn} is repaired, the procedure ends, otherwise, it is repeated for the second-most relevant neuron and so forth. 
Their empirical results show that the overall task performance of the repaired network does not significantly decrease compared to the original one. They admit that their method may become infeasible for large \acp{nn}. The authors in \cite{sotoudeh_provable_2021} describe the method of \cite{goldberger_minimal_2020} as layer-wise repairs whose computational time grows exponentially (becoming infeasible except for linear activations). Their repair algorithm allows for point-wise (for specifications concerning finite points) and polytope repair (infinitely many points) and can handle general activation functions. They point out that directly encoding the constraints and the \ac{dnn} into an \ac{smt} is infeasible due to coupled weights. They introduce decoupled \acp{dnn} (DDNNs) where, by a first-order expansion of the activation function, the activated values and the activations are decoupled. For point-wise repair, the DDNN is then repaired by checking whether modifying a single value channel layer leads to satisfaction using a LP which can be done in polynomial time. They also provide an algorithm for polytope repair for piece-wise linear activations. The extension to repairing multiple layers is given in \cite{refaeli_minimal_2021} whose idea is to split the \ac{dnn} into sub-networks consisting of subsequent layers of the large \ac{dnn} and to repair one layer of each sub-network. The authors in \cite{majd_local_2021} use the task loss function for optimization instead of weight differences and propose a mixed integer quadratic programming approach. A repair on the output layer is considered in \cite{leino_self-repairing_2021}. The authors in \cite{fu_sound_2021} do not repair neurons but propose to use the piece-wise linear nature of ReLU networks and to construct patch networks for each linear region which violates the constraints. They point out that modifications of neurons or the architecture result in global output changes so that there is no guarantee that formerly safe input points stay safe after the modification. Their patch functions are linear functions that are added to the original \ac{nn} in the respective region and whose absolute maximum should be as small as possible. The corresponding optimization problem is solved using linear programming. They theoretically and experimentally prove the efficiency of their so-called REASSURE algorithm.

The authors in \cite{xie_neuro-symbolic_2022} point out that complex properties, for example, that an agent always stops in front of a stop sign, are very hard to verify. Therefore, they propose a neuro-symbolic (\ref{sec:neuralsymbolic_integration}) approach where the \acp{nn} take the role of proxies for semantic properties, e.g., whether the perception \ac{nn} detects the stop sign and whether the controller \ac{nn} decides for deceleration. The resulting neuro-symbolic property is translated into a verification condition which can be checked using existing \ac{ai} verification tools.

\subsubsection{Applications}

A quite large number of \ac{ai} verification techniques has been outlined above which are all relevant for autonomous driving due to the heterogeneous \ac{ai} applications in this area. Roughly speaking, verification techniques for non-recurrent \acp{nn}, especially \acp{cnn}, correspond to perception or image segmentation modules in order to check whether there are situations where these models become unrealiable. As for planning, techniques for \ac{rnn} verification become more relevant. 
\subsection{Run-time Network Verification}
\label{sec:runtime_network_ver}
\textit{Author: Mert Keser, Youssef Shoeb, Gesina Schwalbe}

Run-time network monitoring plays a crucial role in safeguarding the operations of autonomous vehicle systems, ensuring they continuously align with established knowledge bases and safety tenets. As these vehicles navigate their environments, they must constantly validate their actions and decisions against a vast repository of rules, regulations, and best practices. Run-time monitoring serves as the system's real-time audit mechanism, instantly identifying any deviations from predefined norms or emerging unpredictable behaviors. By doing so, it provides an immediate feedback loop, facilitating timely interventions and adjustments. This proactive approach not only ensures that the vehicle's behavior consistently adheres to traffic regulations, physical laws, and general common sense, but also guarantees a state of knowledge conformity throughout its operation. In essence, run-time network monitoring is the sentinel that ensures the integrity and reliability of autonomous systems, keeping them in sync with the ever-evolving landscape of knowledge and safety standards.

The deployment of DNN-based methods in autonomous driving systems is limited by our ability to guarantee their correct behavior.
The study of convergence guarantees and safety certification for complex software systems at the design stage is a well-established area of research (see \autoref{sec:ai_ver}).
However, the complexity of learning-based perception and the unknowable future deployment conditions of an autonomous vehicle makes the generalization of design-time verification to run-time problematic.
Researchers have studied network monitoring techniques for detecting and predicting run-time errors in response to this issue. 
Network monitoring, also known as online monitoring \cite{samann2020strategy,delseny2021white}, online verification, run-time monitoring \cite{watanabe2018runtime,schwalbe2020survey}, run-time verification \cite{watanabe2018runtime}, or operation-time monitoring \cite{burton2022safety}, refers to architectural measures used to observe the performance of a deep neural network during run-time.
No ground-truth labels are available during operation, so a network monitor is used to assess how well a model is doing by detecting and predicting errors during run-time.
Network monitoring is an essential component that is critical in ensuring the safety and reliability of a DNN-based component in an autonomous vehicle with two main applications of the error indication:
(1) It may serve as a trigger mechanism for transferring control to a human operator or a less capable but safer system when necessary.
Additionally, (2) network monitoring may determine uncertainties in the trustworthiness of DNN outputs that can be propagated from one component to another in the processing pipeline.
This propagation may be used to adjust towards more conservative decisions, such as driving at reduced speeds in scenarios where multiple uncertainties accumulate, thereby enhancing the overall safety and reliability of the system. 
 
In the following, we categorize the different methods for run-time network monitoring based on the types of failures the network monitor seeks to detect during run-time. In \autoref{ood_errors}, we cover network monitoring techniques that attempt to identify failures caused by inputs belonging to classes not represented in the training data. These are referred to \textit{out-of-distribution errors}. In ~\autoref{in-erros}, we cover the remainder of possible failures. These are caused by inherent limitations of the model; we refer to those as \textit{in-distribution errors}. 


\subsubsection{Out-of-Distribution Errors}
\label{ood_errors}
The success of supervised learning methods relies on the implicit assumption that the training and test data are generated from the same data distribution. However, a DNN may encounter scenarios that violate this assumption in an autonomous driving use case. Errors due to this violation are called \acf{ood} errors.

A rich line of research has been developed to empirically \cite{salehi2021unified} and theoretically \cite{fang2022out} address the problem of \ac{ood} detection. \ac{ood} detectors aim to identify inputs not belonging to the training set’s distribution. However, the term \ac{ood} is often fuzzy throughout the literature, and there is no clear definition of whether a data point is considered \ac{ood}. Therefore, we start this section by defining what we consider \ac{ood} data points and then follow with examples that use ideas related to \ac{ood} detection for online network monitoring.


\vspace{1em}
\noindent\textbf{Definition: }
In a supervised learning setting, a DNN model $f_{\theta}$ with parameters $\theta$ is trained using a finite dataset $D_{\text{train}} \subset \mathcal{X}\times\mathcal{Y}$ consisting of $n$ data points with labels from an oracle $\Omega$:
\begin{gather*}
D_{\text {train }}=\left\{\left(x_i, \Omega(x_i)\right) \mid i\in\{1,\dotsc,n\}\right\} \subset \mathcal{X}\times \mathcal{Y}
\end{gather*}
where vectors in $\mathcal{X}$ are the (non-semantic) input features, and $\mathcal{Y}$ holds the (semantic) label features.
The density of the points from $D_{\text{train}}$ in $\mathcal{X}\times\mathcal{Y}$
allows to approximate a probability distribution $p_{\text{ID}}$ on $\mathcal{X}\times\mathcal{Y}$, the in-training-data distribution or short in-distribution.
An \acf{ood} data point $(x,y)$ is any data point that has a low or zero probability of occurrence $p_\text{ID}((x,y))<\epsilon$, i.e., occurs in regions with low- or zero-density of the training samples. Two main types of \ac{ood} are: (1) \emph{covariate shift} \cite{keser2021content} is caused by input features $x_i$ that are rarely seen in the training data, e.g., other styles; and (2) \emph{semantic shift} is caused by label features $y_i\in\mathcal{Y}$ that are unseen or seldom in the training data, e.g., novel classes. Note that semantic shift usually also requires seldom or novel input features.
%
The task of an out-of-distribution network monitor $m$ is to identify these \ac{ood} data points. An ideal \ac{ood} monitor $m^*$ perfectly approximates $p_\text{ID}$, in particular:
\begin{gather*}
\forall x \in \mathcal{X}\colon \begin{cases}m^*(x)\approx 0 & \text {if $x$ \ac{ood}} \\ m^*(x) \gg 0 & \text {if $x$ in-distribution.}\end{cases}
\end{gather*}

The goal of identifying \ac{ood} samples is frequently seen in the literature under several different problems; these include outlier detection \cite{aggarwal2001outlier,hodge2004survey,wang2019progress}, anomaly detection \cite{ruff2021unifying,pang2021deep,bulusu2020anomalous,chalapathy2019deep}, novelty detection \cite{pimentel2014review,miljkovic2010review,markou2003novelty,markou2003novelty-2}, and open set recognition \cite{boult2019learning,geng2020recent,mahdavi2021survey,vaze2021open}. While the goals behind these problems are similar, there are slight variations between the problems in terms of how the boundaries of in- versus out-of-distribution are defined. For a comprehensive understanding of the relationship between the different problems, we refer the reader to \cite{yang2021generalized}. 

Throughout this section, we refer to \ac{ood} samples as only the samples with semantic shift, more precisely, such that do not have overlapping labels with respect to the labels of the training dataset. 

\vspace{1em}
\noindent\textbf{Run-Time \ac{ood} Network Monitors: }
Approaches for \ac{ood} detection during run-time can be differentiated by the type of distribution shift they detect (semantic or covariate) and the \ac{ood} problem that is to be solved.
Other practical considerations are whether the monitor can be applied post-hoc or the DNN architecture and training needs to be adapted; whether it is unsupervised or a dataset of \ac{ood} samples is needed; and what is used to identify \ac{ood} samples: DNN inputs, the neural activation patterns in intermediate or final outputs, or additional self-evaluations of the DNN trustworthiness like uncertainty estimates. In the following, we will give examples along the latter axes, namely what information used to identify \ac{ood} samples.

\textbf{Uncertainty Estimation }A DNN deployed on an autonomous vehicle operating in an open-world setting is vulnerable to making over-confidently wrong predictions on \ac{ood} data samples \cite{ovadia2019can}. 
A vast field of methods that can be utilized for \ac{ood} detection is that of uncertainty estimation methods \cite{feng2022review}. The main goal of uncertainty estimation is to obtain along each DNN output a score that represents the probability of correctness of the output. This can be achieved by, e.g., means of model ensembling \cite{henne2020benchmarking,kendall2017what}, direct modeling of parameters as distributions as in Bayesian DNNs \cite{gast2018lightweight}, and post-hoc calibration of the DNN confidence outputs \cite{guo2017calibration}.
For details on uncertainty estimation in AD the reader is referred to ~\autoref{sec:uncertainty} and \cite{feng2022review}.
It should be noted, however, that it is hard to differentiate between different causes for low uncertainties, such as noisy data, decision boundaries, and \ac{ood}. 

\textbf{\ac{ood} on Inputs: Autoencoder }To determine whether an input is \ac{ood}, a common approach is to use the reconstruction error of an autoencoder trained on the training data as an \ac{ood}-score.
Richter and Roy~\cite{richter2017safe} proposed using an autoencoder as an \ac{ood} network monitor in a collision-avoidance system. The autoencoder is used to recognize whether an input is novel; in case of a novel input, the system reverts to a safe setting where it can collect additional training data to continually improve the learning procedure.
For detecting \ac{ood} optical flow during runtime, \cite{feng2021improving} proposed a lightweight variational autoencoder-based framework that uses input space feature abstraction and exploits the spatio-temporal correlation of motion in videos to detect environmental motion that was not present in the training set during run-time.
While the previous examples use post-hoc attached autoencoders, \cite{yoshihashi2019classification} proposed a deep hierarchical reconstruction net architecture that is trained for joint classification and reconstruction of input samples. This approach allows the model to learn a representation that preserves valuable information about what is considered \ac{ood} and what was seen during training in an ante-hoc manner.
%

\textbf{\ac{ood} on Inputs: Comparable Representations }Autoencoders only allow indirect comparison of samples against a training set. Some methods leverage compression techniques to find a representation of samples that allows direct comparison between (all) training samples and new samples without having to store all training samples during run-time.
Such an approach for run-time \ac{ood} detection was proposed by \cite{cai2020real}. They use inductive conformal prediction and anomaly detection to obtain an adaptive \ac{ood}-score threshold and get a well-calibrated false positive rate. Two approaches are suggested to calculate the non-conformality score that is used as \ac{ood}-score: Again the reconstruction error of a variational autoencoder; and training a Deep Support Vector Data Description model to find a mapping of the training data into a hypersphere, where the distance of newly mapped samples to the sphere center serves as \ac{ood}-score.

\textbf{Neural Activation Patterns }More recent work by \cite{hashemi2023runtime} proposes a real-time network monitor for \ac{ood} detection in 2D object detection systems. The monitoring algorithm extends on the Gaussian-based monitoring proposed in \cite{hashemi2021gaussian}. Their work quantitatively measures the difference between the training set and new samples by monitoring neuron activation patterns and integrating it into the inductive conformal anomaly detection framework \cite{laxhammar2015inductive} to detect \ac{ood} samples during runtime.
An uncertainty-based \ac{ood} detection method that leverages additional training and post-hoc statistics without requiring an external \ac{ood} dataset was proposed in \cite{NEURIPS2018_abdeb6f5} and \cite{nitsch2021out}. The proposed approach compares the neural activation pattern of a new sample, both in intermediate and the final layers, against each the activation distribution typical for one class. Samples without a similar class counterpart are identified as \ac{ood}. For better calibration, a Generative Adversarial Network (GAN) is used to drive the object classifier to assign untypical confidences to \ac{ood} data.
Similarly, but using only the confidence outputs of a classifier, \cite{jafarzadeh2021automatic} formalized the \ac{ood} detection problem and presented multiple automatic reliability assessment policies. All do a comparison of the confidence value distribution of a sample against a typical one. The proposed \ac{ood} detection performs well compared to a baseline method in closed-set and open-world contexts.

\subsubsection{In-Distribution Errors}
\label{in-erros}
In-distribution errors are caused by inherent limitations in the trained model.
These limitations stem from inadequate representations learned from the data. 

%
In this section, we cover previous work on run-time network monitoring techniques for detecting failures in a neural network \cite{rahman2021run}. We here subdivide the literature according to their necessity for additional outputs and constraint formulations into the categories: additional \emph{self-evaluation outputs}, plausibilization of the available model behavior against \emph{given} constraints, and \emph{learned} meta-classification of the model output quality.

\vspace{1em}
\noindent\textbf{Self-evaluation: }
A common approach towards error retrieval during runtime is to utilize trustworthiness scores that are directly part of the model output. This usually requires
a modification of the model to provide these outputs and/or properly calibrate
them. Two reknowned examples are uncertainty outputs as error indicators and
classification with included reject option, both discussed below.

\textbf{Uncertainty Estimation }The uncertainty estimation techniques mentioned in Run-Time \ac{ood} Network Monitors section are also applicable for identifying samples that are close to a decision boundary, or generally should be treated with care due to high model uncertainty~\cite{kendall2017what}.
The obtained uncertainty score can be used as an indicator for high error probability. However, as mentioned earlier, it is hard to discriminate different sources of uncertainty.

\textbf{Selective Classification }Selective classification generally refers to classification with a reject option, which can be calculated jointly with the model prediction or by a separate meta-classifier. This method improves the model performance but at the cost of test coverage.
In the following, some examples of jointly trained reject options for self-evaluation are given.
Geifman and El-Yaniv~\cite{geifman2017selective} propose a simple implementation for the selective classification of DNNs, introducing a reject function that enables control over risk based on the softmax output.
Continuing on this, \cite{geifman2019selectivenet} introduces SelectiveNet, a two-headed network that jointly trains classification and rejection functions.
\cite{hecker2018failure} proposed an approach to augment any network with a failure head to learn to predict the occurrence of model failures using a failure score.

\vspace{1em}
\noindent\textbf{Plausibilization: }
Under plausibilization we here summarize approaches to error indication that check fulfillment of constraints on the DNN behavior during run-time. Typical constraint types are stability constraints, semantic constraints, and consistency between alternative model predictions.

\textbf{Semantic constraints }can be applied to semantic inputs, and the intermediate and final outputs as well as processing characteristics like feature saliency. Examples are locality of feature saliency for object detection, or conformity with class and attribute relations such as semantic class hierarchies like \enquote{Cars are vehicles} \cite{roychowdhury2018image}, spatial class hierarchies like \enquote{Arms typically belong to a person} \cite{schwalbe2022enabling}, or 
other attribute relations like \enquote{Only traffic lights can give green signal} \cite{giunchiglia2022roadr}. In particular \cite{schwalbe2022enabling,giunchiglia2022roadr} interpret outputs as fuzzy truth values, leveraging fuzzy logic to evaluate consistency with rules that relate the outputs. 


\textbf{Stability and Model Consistency }Examples of stability constraints are temporal consistency \cite{varghese2020unsupervised} and local robustness against slight input changes.
Consistency between model outputs is classically known from redundancy and voting safety architectures \cite[7.4.12]{iso/tc22/sc322018isoa}, for DNNs better known as ensembles (all same input) and late fusion (different input sensor modalities).
Consistency can also be checked using additional control outputs, e.g., for object detection comparing the predictions of an additional depth estimation head and a LiDAR sensor.
And lastly, one may compare against a custom oracle model that is only queried for outputs of interest, e.g., potentially safety-relevant ones.
For example, in a recent study \cite{keser2023interpretable}, the decision of 2D object detectors was evaluated using interpretable domain-invariant concept bottleneck models to verify the interaction of these detectors with the environment model in real-time. The surrogate model can be pre-trained on any relevant dataset in the target application of interest, and only a small portion of the user's dataset is needed for fine-tuning.


\textbf{Meta-classification }Run-time network monitoring can be done using a surrogate model that uses past examples of successes and failures of a base network to detect these types of errors. The task of the base network can be any arbitrary task: object detection, image segmentation, or image classification. The surrogate model is trained using positive and negative samples where the base network performs its task. The auxiliary model functions alongside the base network throughout the deployment phase and forecasts whether the base network will succeed or fail for the given input. This is often also referred to as meta-classification, and approaches can differ by the input to the surrogate model: either just the model inputs, the model outputs, or any further insights into the model representations and behavior like saliency, or combinations thereof.


\textbf{Input-based Meta-clasifiers} Some examples of earlier work that use surrogate models to predict the success and failures of a DNN output were presented in \cite{scheirer2008predicting,wang2007modeling,aggarwal2011predicting,jammalamadaka2012has}. These works use \textit{evaluator algorithms} to predict the algorithm’s failures by assessing the similarity scores between the input samples and the training set. 
\cite{wang2007modeling} also utilizes the similarity score within template images, and \cite{jammalamadaka2012has} takes it one step further and explicitly considers other factors obtained from the output of the base model. These methods adopt the notion of self-evaluation and present it as a binary classification problem. Using examples of failure from the training set of the base model, a surrogate binary classifier is trained to serve as the evaluator. The base model's performance is assessed using a threshold on the evaluator's quality during inference.
\cite{daftry2016introspective,saxena2017learning} applied approaches similar to the \textit{Alert} framework to predict failures in autonomous navigation tasks. Their work uses spatiotemporal convolutional neural networks to extract features and a support vector machine to extract situations where navigational failure may occur.
More recent work for online monitoring of object detectors \cite{rahman2021online} use a cascaded neural network to monitor the performance of the object detector by predicting the quality of its mean average precision on a sliding window of the input images.
%

\textbf{Output-based Meta-classifiers} Following the work of \cite{jammalamadaka2012has}, \cite{zhang2014predicting} proposed a generalized failure detection framework called \textit{Alert} to detect the failures of any vision system. \textit{Alert} uses generic image features for various applications. Experimental results show that this framework effectively predicts failure in various tasks, such as image segmentation, 3D layout estimation, camera parameter estimation, and object detection.
In \cite{shao2020increasing}, the surrogate model takes as input the deployed DNN’s softmax probability output and directly indicates if the DNN’s prediction result is correct or not, thus leading to an estimate of the accurate inference accuracy.
To detect false positives in object detectors, \cite{schubert2021metadetect} use meta-classification to distinguish between true-positive and false-positive. This approach uses meta-regression to predict the intersection-over-union score without using any ground-truth labels during deployment.
\cite{rottmann2020prediction} follows a similar approach but is applied to semantic segmentation models. They predict the quality of a segmented region of one class as the mean entropy of the pixel confidences.

\textbf{Representation-based Meta-classifiers} Following the approach of using surrogate models for network monitoring, \cite{mohseni2019predicting} train a surrogate model using the saliency maps generated from input images to forecast the base model's failures. The network monitor is trained using errors in the steering angle predictions of the base model from frames in the training set; their evaluations show that saliency maps outperform raw images in predicting model failure.
%
In a similar approach, \cite{rahman2021per} use the internal features of the object detector to predict if the per-frame mean average precision drops below a predefined threshold.

\subsubsection{Applications}
\label{sec:runtimemonitoring-applications}

Run-time network monitoring has been employed across various application domains. A significant portion of research in this area emphasizes the use of probabilistic frameworks to enhance predictions based on existing models, especially those applied for classification \cite{gast2018lightweight, guo2017calibration, henne2020benchmarking} and object detection problems \cite{feng2022review, schubert2021metadetect}. One notable application in this domain is real-time \ac{ood} detection. Numerous methods have been devised to detect out-of-distribution samples in real-time for visual navigation \cite{richter2017safe}, classification \cite{hashemi2021gaussian, NEURIPS2018_abdeb6f5}, object detection \cite{hashemi2023runtime, nitsch2021out}, and trajectory prediction \cite{laxhammar2015inductive}. Ensuring model performance in run-time is of paramount importance, and the established methods have found utility in various tasks such as image classification \cite{geifman2017selective, geifman2019selectivenet, zhang2014predicting, shao2020increasing}, end-to-end driving \cite{hecker2018failure}, object detection \cite{schwalbe2022enabling, keser2023interpretable, rahman2021online,schubert2021metadetect, rahman2021per}, semantic segmentation \cite{varghese2020unsupervised, rottmann2020prediction}, face recognition \cite{wang2007modeling, scheirer2008predicting, aggarwal2011predicting}, human pose estimation \cite{jammalamadaka2012has}, and autonomous vision-based flight systems \cite{daftry2016introspective, saxena2017learning}.

\newpage
\section*{List of Abbreviations}
\addcontentsline{toc}{section}{List of Abbreviations}
\begin{acronym}[SGHMC]
    \acro{a3c}[A3C]{Asynchronous Advantage Actor-Critic}
    \acro{ace}[ACE]{Automated Concept-based Explanation}
    \acro{ad}[AD]{Autonomous Driving}
    \acro{adf}[ADF]{Assumed Density Filtering}
    \acro{admm}[ADMM]{Alternating Direction Method of Multipliers}
    \acro{ae}[AE]{Auto Encoder}
    \acro{agem}[A-GEM]{Averaged Gradient Episodic Memory}
    \acro{ai}[AI]{Artificial Intelligence}
    \acro{al}[AL]{Active Learning}
    \acro{alm}[ALM]{Augmented Lagrangian Method}
    \acro{av}[AV]{Autonomous Vehicle}
    \acro{bev}[BEV]{Bird's-Eye-View}
    \acro{bp}[BP]{Backpropagation}
    \acro{bnn}[BNN]{Bayesian Neural Network}
    \acro{binnn}[BinNN]{Binarized Neural Network}
    \acro{cav}[CAV]{Concept Activation Vector}
    \acro{cbm}[CBM]{Concept Bottleneck Model}
    \acro{cbn}[CBN]{Causal Bayesian Network}
	\acro{cd}[CD]{Causal Discovery}
	\acro{cfgps}[CF-GPS]{Counterfactually-Guided Policy Search}
    \acro{ci}[CI]{Causal Inference}
    \acro{cnn}[CNN]{Convolutional Neural Network}
    \acro{cophy}[CoPhy]{Counterfactual Learning of Physical Dynamics}
    \acro{cpe}[CPE]{Counterfactual Policy Evaluation}
    \acro{cr}[CR]{Causal Reasoning}
    \acro{crps}[CRPS]{Continuous Ranked Probability Score}
    \acro{ct}[CT]{Computerised Tomography}
    \acro{cv}[CV]{Computer Vision}
    \acro{cvae}[CVAE]{Conditional Variational Auto Encoder}
    \acro{dag}[DAG]{Directed Acyclic Graph}
    \acro{ddpg}[DDPG]{Deep Deterministic Policy Gradient}
    \acro{dl}[DL]{Deep Learning}
    \acro{dll}[DLL]{Differentiable Logic Layer}
    \acro{dnn}[DNN]{Deep Neural Network}
    \acro{dot}[DOT]{U.S. Department of Transportation}
    \acro{dpg}[DPG]{Deterministic Policy Gradient}
    \acro{dsl}[DSL]{Domain-Specific Language}
    \acro{dqn}[DQN]{Deep Q Learning}
    \acro{ece}[ECE]{Expected Calibration Error}
	\acro{elbo}[ELBO]{(empirical) Evidence Lower Bound}
    \acro{erm}[ERM]{Empirical Risk Minimization}
    \acro{fc}[FC]{full-covariance}
    \acro{fcn}[FCN]{Fully Convolutional Network}
    \acro{ffnn}[FFNN]{Feed-Forward Neural Network}
    \acro{fol}[FOL]{First order Logic}
    \acro{fpn}[FPN]{Feature Pyramid Network}
    \acro{gaal}[GAAL]{Generative Adversarial Active Learning}
    \acro{gail}[GAIL]{Generative Adversarial Imitation Learning}
    \acro{gan}[GAN]{Generative Adversarial Network}
    \acrodefplural{gan}[GANs]{Generative Adversarial Networks}
    \acro{gem}[GEM]{Gradient Episodic Memory}
    \acro{gis}[GIS]{Geographical Information System}
    \acro{gmm}[GMM]{Gaussian Mixture Model}
    \acro{gnn}[GNN]{Graph Neural Network}
    \acro{gpm}[GPM]{Gradient Projection Memory}
    \acro{hd}[HD]{High-Definition}
    \acro{hmc}[HMC]{Hamiltonian Monte Carlo}
    \acro{hog}[HOG]{Histograms of Oriented Gradients}
    \acro{ig}[IG]{Integrated Gradients}
    \acro{ilp}[ILP]{Inductive Logic Programming}
    \acro{irl}[IRL]{Inverse Reinforcement Learning}
    \acro{kf}[KF]{Kronecker Factorization}
    \acro{kg}[KG]{Knowledge Graph}
	\acrodefplural{kg}[KGs]{Knowledge Graphs}
	\acro{kge}[KGE]{Knowledge Graph Embedding}
    \acro{kkt}[KKT]{Karush Kuhn Tucker}
    \acro{kl}[KL]{Kullback-Leibler}
    \acro{krl}[KRL]{Knowledge Representation Learning}
    \acro{lime}[LIME]{Local interpretable model-agnostic explanations}
    \acro{lrp}[LRP]{Layer-wise Relevance Propagation}
    \acro{lstm}[LSTM]{Long Short-Term Memory}
    \acro{ltl}[LTL]{Linear Temporal Logic}
    \acro{ltn}[LTN]{Logic Tensor Network}
    \acro{maml}[MAML]{Model-Agnostic Meta-Learning}
    \acro{map}[mAP]{mean Average Precision}
    \acro{mapo}[MAP]{Maximum-a-posteriori}
    \acro{marl}[MARL]{Multi-Agent Reinforcement Learning}
    \acro{mbps}[MB-PS]{Model-based Policy Search}
    \acro{mbrl}[MBRL]{Model-based Reinforcement Learning}
    \acro{mc}[MC]{Monte Carlo}
    \acro{mcts}[MCTS]{Monte-Carlo Tree Search}
    \acro{mdp}[MDP]{Markov Decision Process}
    \acro{mf}[MF]{mean-field}
    \acro{mfrl}[MFRL]{Model-free Reinforcement Learning}
    \acro{milp}[MILP]{Mixed-Integer Linear Programming}
    \acro{mip}[MIP]{Mixed-Integer Programming}
    \acro{ml}[ML]{Machine Learning}
    \acro{mlp}[MLP]{Multi-Layer Perceptron}
    \acro{moba}[MOBA]{Multiplayer Online Battle Arena}
    \acro{mpc}[MPC]{Model-predictive Control}
    \acro{mrc}[MRC]{Minimal Risk Condition}
    \acro{mrf}[MRF]{Markov Random Field}
    \acro{mri}[MRI]{Magnetic Resonance Imaging}
    \acro{mse}[MSE]{Mean Squared Error}
    \acro{nas}[NAS]{Neural Architecture Search}
    \acro{nll}[NLL]{Negative Log-Likelihood}
    \acro{nlm}[NLM]{Neural Logic Machines}
    \acro{nlp}[NLP]{Natural Language Processing}
    \acro{nlu}[NLU]{Natural Language Understanding}
    \acro{nmt}[NMT]{Neural Machine Translation}
    \acro{nn}[NN]{Neural Network}
    \acrodefplural{nn}[NNs]{Neural Network}
    \acro{nsbp}[NSBP]{Neural-Symbolic Behavior Program}
    \acro{nsdp}[NSDP]{Neuro-Symbolic Decision Program}
    \acro{nsp}[NSP]{Neuro-Symbolic Program}
	\acro{nsps}[NSPS]{Neuro-Symbolic Program Search}
	\acro{nuts}[NUTS]{No-U-Turn Sampling}
    \acro{oedr}[OEDR]{Object and Event Detection and Response}
    \acro{odd}[ODD]{Operational Design Domain}
    \acro{ood}[OoD]{Out-of-Distribution}
    \acro{path}[PATH]{California Partners for Advanced Transportation Technology}
    \acro{pde}[PDE]{Partial Differential Equation}
    \acro{pomdp}[POMDP]{Partially Observable Markov Decision Process}
    \acro{ppo}[PPO]{Proximal Policy Optimization}
	\acro{qa}[QA]{Question Answering}
	\acro{rcnn}[R-CNN]{Region-based Convolutional Neural Network}
	\acro{r-cnn}[R-CNN]{Regions with CNN}
	\acro{rgb}[RGB]{Red Green Blue}
	\acro{rise}[RISE]{Randomized Input Sampling for Explanation}
	\acro{rl}[RL]{Reinforcement Learning}
	\acro{rpn}[RPN]{Region Proposal Network}
	\acro{rnn}[RNN]{Recurrent Neural Network}
	\acro{roc}[ROC]{Receiver Operating Characteristic}
	\acro{roi}[ROI]{Region of Interest}
    \acro{sa}[SA]{Situation Awareness}
    \acro{sac}[SAC]{Soft Actor Critic}
    \acro{sat}[SAT]{Satisfiability Solvers}
    \acro{sbr}[SBR]{Semantic Based Regularization}
    \acro{scm}[SCM]{Structural Causal Model}
    \acro{sdv}[SDV]{Self-Driving Vehicle}
    \acro{sgd}[SGD]{Stochastic Gradient Descent}
    \acro{sghmc}[SGHMC]{Stochastic Gradient Hamiltonian Monte Carlo}
    \acro{sgld}[SGLD]{Stochastic Gradient Langevin Dynamics}
    \acro{shap}[SHAP]{SHapley Additive exPlanations}
    \acro{si}[SI]{Synaptic Intelligence}
    \acro{smc}[SMC]{Satisfiability Modulo Convex Programming} 
    \acro{smt}[SMT]{Satisfiability Modulo Theory}
    \acro{srm}[SRM]{Structural Risk Minimization}
	\acro{ssd}[SSD]{Single Shot MultiBox Detector}
	\acro{ssm}[SSM]{State Space Model}
	\acro{stl}[STL]{Signal Temporal Logic}
	\acro{stvo}[StVO]{Straßenverkehrsordnung}
	\acro{sqp}[SQM]{Sequential Quadratic Program}
	\acro{svm}[SVM]{Support Vector Machine}
	\acro{trpo}[TRPO]{Trust Region Policy Optimization}
	\acro{ttc}[TTC]{Time-To-Collision}
	\acro{uav}[UAV]{Unmanned Aerial Vehicles}
	\acro{va}[VA]{Visual Analytics}
	\acro{vae}[VAE]{Variational Auto Encoder}
	\acro{vi}[VI]{Variational Inference}
	\acro{vqa}[VQA]{Visual Question Answering}
	\acro{vru}[VRU]{Vulnerable Road User}
	\acro{wsol}[WSOL]{Weakly Supervised Object Localization}
	\acro{xai}[XAI]{Explainable Artificial Intelligence}
\end{acronym}

\newpage
\printbibliography[heading=bibintoc]

\end{document}